\documentclass[acmtog, nonacm]{acmart}

\usepackage{booktabs} 

\usepackage[ruled]{algorithm2e} 
\usepackage{tabularray}
\usepackage{enumitem}

\SetAlFnt{\small}
\SetAlCapFnt{\small}
\SetAlCapNameFnt{\small}
\SetAlCapHSkip{0pt}

\acmJournal{TOG}

\usepackage{titletoc}
\usepackage{graphicx}
\usepackage{amsmath}
\usepackage{booktabs} 

\usepackage{booktabs}
\usepackage{multirow}
\usepackage{multicol}
\usepackage{bbm}
\usepackage{color}
\usepackage{colortbl}
\usepackage{dcolumn}
\usepackage{tablefootnote}
\usepackage{subcaption}

\usepackage{array}

\definecolor{pinkl}{HTML}{FB2B97}

\newcommand{\rev}[1]{#1}
\newcommand{\revv}[1]{#1}

\newcommand{\icosphere}{icosphere }

\usepackage{pifont}
\definecolor{limegreen}{HTML}{32CD32}
\newcommand{\cmark}{\textcolor{limegreen}{\ding{51}}}%
\newcommand{\xmark}{\textcolor{red}{\ding{55}}}%
\newcommand{\hmark}{\textcolor{orange}{\large \checkmark\kern-1.1ex\raisebox{.7ex}{\rotatebox[origin=c]{125}{--}}}}%

\begin{document}

\title{Mesh Neural Cellular Automata}

\author{Ehsan Pajouheshgar}
\authornote{Both authors contributed equally to this research.}
\email{ehsan.pajouheshgar@epfl.ch}

\author{Yitao Xu}
\authornotemark[1]
\email{yitao.xu@epfl.ch}
\affiliation{%
  \institution{EPFL}
  \country{Switzerland}
}

\author{Alexander Mordvintsev}
\affiliation{%
  \institution{Google Research}
  \country{Switzerland}}
\email{moralex@google.com}

\author{Eyvind Niklasson}
\affiliation{%
  \institution{Google Research}
  \country{Switzerland}}
\email{eyvind@google.com}

\author{Tong Zhang}
\affiliation{%
  \institution{EPFL}
  \country{Switzerland}
}
\email{tong.zhang@epfl.ch}

\author{Sabine Süsstrunk}
\affiliation{%
  \institution{EPFL}
  \country{Switzerland}
}
\email{sabine.susstrunk@epfl.ch}

\renewcommand{\shortauthors}{Pajouheshgar and Xu, et al.}

\begin{abstract}
    Texture modeling and synthesis are essential for enhancing the realism of virtual environments. Methods that directly synthesize textures in 3D offer distinct advantages to the UV-mapping-based methods as they can create seamless textures and align more closely with the ways textures form in nature. We propose \textbf{Mesh} \textbf{N}eural \textbf{C}ellular \textbf{A}utomata (MeshNCA), a method that directly synthesizes dynamic textures on 3D meshes without requiring any UV maps. MeshNCA is a generalized type of cellular automata that can operate on a set of cells arranged on non-grid structures such as the vertices of a 3D mesh. MeshNCA accommodates multi-modal supervision and can be trained using different targets such as images, text prompts, and motion vector fields. Only trained on an Icosphere mesh, MeshNCA shows remarkable test-time generalization and can synthesize textures on unseen meshes in real time. We conduct qualitative and quantitative comparisons to demonstrate that MeshNCA outperforms other 3D texture synthesis methods in terms of generalization and producing high-quality textures. Moreover, we introduce a way of grafting trained MeshNCA instances, enabling interpolation between textures. MeshNCA allows several user interactions including texture density/orientation controls, grafting/regenerate brushes, and motion speed/direction controls. Finally, we implement the forward pass of our MeshNCA model using the WebGL shading language and showcase our trained models in an online interactive demo, which is accessible on personal computers and smartphones and is available at \href{https://meshnca.github.io}{https://meshnca.github.io/}. 
\end{abstract}




\begin{teaserfigure}
  \centering
    \captionsetup{type=figure}
    \includegraphics[width=\linewidth,trim={50 100 50 100},clip]{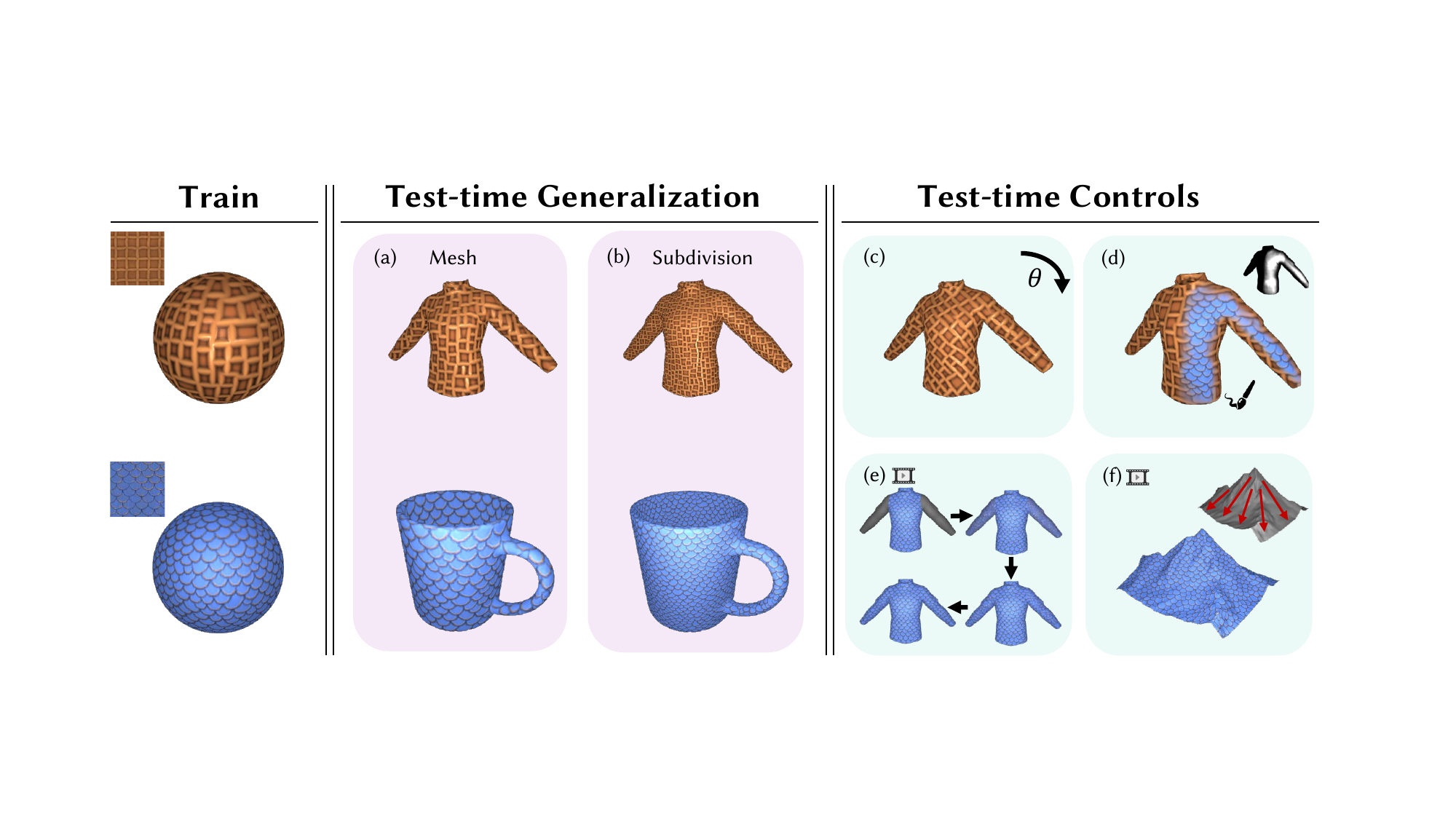}
    \vspace{-19pt}
    \captionof{figure}{ 
    Mesh Neural Cellular Automata (MeshNCA) synthesizes 3D surface textures in real time and exhibits various unique test-time properties that facilitate user interaction. \textbf{Train:} MeshNCA is trained only on an \icosphere  guided by the target appearance. \textbf{Test-time Generalization:} (a) MeshNCA generalizes to unseen meshes during the test time and (b) can operate at different levels of mesh subdivision, allowing users to control texture density. \textbf{Test-time Controls: } In test time, users can interactively control the behavior of MeshNCA in real time, including (c) texture orientation, (d) texture grafting, (e) self-regeneration, and (f) motion. See our online demo at \textcolor{pinkl}{\href{https://meshnca.github.io/}{\textbf{https://meshnca.github.io/}}}. 
    }
    \vspace{12pt}
    \label{fig:teaser}
\end{teaserfigure}

\newcolumntype{G}{>{\centering\arraybackslash} m{75pt} }  
\newcolumntype{F}{>{\centering\arraybackslash} m{80pt} }  

\newcommand{\arrow}{\textcolor{black}{\Longrightarrow}}%
\newcommand{\ofarrow}{\hspace*{-8pt}$ \overset{OF}{\arrow}$}

\newif\ifimgprefix
\imgprefixtrue   


\maketitle

\section{Introduction}
\label{sec:intro}
The appearance of objects surrounding us depends on their textures. The modeling and synthesizing of textures are key components for enhancing the realism and visual appeal of our virtual environments. By enriching the surface-level details of 3D models, textures add depth, complexity, and authenticity to our renderings and breathe life into the digital worlds we create. 


Various methods have been developed to synthesize textures for 3D objects, most of which fall into two main categories. 
1) Methods that synthesize textures directly within a 3D space, and 2) Methods that synthesize textures in a 2D domain which are then mapped onto a 3D surface via UV mapping.
While UV-mapping-based methods are the most common approach due to their simplicity and compatibility with existing graphics pipelines \rev{\cite{procedure-first-novelframework, procedure-mat-generative-text, matformer, MATch, appearance_space, tchapmi2022generating, richardson2023texture-texture}}, they come with an inherent limitation: The process of creating accurate UV maps for intricate 3D objects often leads to visible artifacts such as overlapping regions, distortions, and visible seams \rev{\cite{yuksel2019rethinking}}.

In this paper, we focus on methods that directly synthesize textures in 3D without using UV mapping. These methods offer a set of inherent advantages. Primarily, they can naturally produce seamless textures. 
Moreover, they tend to align more closely with natural texture formation processes, including phenomena such as growth, erosion, and deposition \cite{erosion-and-deposition-texture}, all of which occur in 3D space.
We classify these methods into three groups: \textit{Solid Textures}, \textit{Surface Textures}, and  \textit{Cellular Automata (CA)}.



\textit{Solid Texture} synthesis methods assign colors to the points inside a 3D volume. Although this kind of approach is suitable for modeling textures carved out of a solid block, such as wood or marble, it is not apt for modeling texture-related processes that take place on the surface of an object, such as motion or erosion. Instead, \textit{Surface Texture} synthesis approaches directly assign a color to each point on the surface of an object. Although these methods can be aware of the underlying 3D surface, they require either separate training for each 3D mesh or a large dataset of textured meshes for training. Lastly, \textit{Cellular Automata} models utilize local update rules to iteratively generate textures, mirroring the dynamic behaviors seen in natural systems such as Reaction-Diffusion \cite{turing-pattern}. These methods offer many test-time controls and allow users to interact with the synthesized texture. However, they require manual design for each texture and cannot be trained to synthesize a target texture. 


In this paper, we focus on and improve upon the Cellular-Automata-based 3D texture modeling as this approach offers more user interaction compared to the other two categories, which is of paramount importance for design, art, and creative applications. We build upon Neural Cellular Automata (NCA\footnote{In this paper, we use NCA to refer to both Neural Cellular \textbf{Automata} and Neural Cellular \textbf{Automaton}.}), a differentiable counterpart of conventional CA \cite{mordvintsev2020growing, niklasson2021self-sothtml}. NCA models are parameterized by very small neural networks and have proven to be effective for texture synthesis due to various advantages they possess, such as parameter efficiency \cite{mordvintsev2021mu-micronca}, inherent parallelizability \cite{niklasson2021self-sothtml, dynca}, a capability to operate at arbitrary resolutions \cite{niklasson2021self-sothtml}, real-time user interactivity \cite{niklasson2021self-sothtml}, and their ability to control texture motion \cite{dynca}.

One major limitation of current NCA models used for texture synthesis \cite{niklasson2021self-sothtml, dynca} is that the cells are aligned in a regular grid, like the pixels of an image or a hexagonal lattice, limiting their applicability to more general structures such as meshes. We generalize these models beyond the regular grid structure and allow them to operate on an arbitrary arrangement of cells given by a mesh. We call our model \textbf{Mesh} \textbf{N}eural \textbf{C}ellular \textbf{A}utomata (\textbf{MeshNCA}). MeshNCA enables real-time and interactive texture synthesis on 3D meshes while entailing all the remarkable properties of 2D NCA. Figure \ref{fig:teaser} highlights the test-time properties of our MeshNCA model, including generalization to unseen meshes, texture density control, texture orientation control, self-organization, and texture motion control. Our development ensures minimal computational overhead, preserving the efficiency inherent to the 2D NCA model \cite{dynca, niklasson2021self-sothtml}. This optimization facilitates the smooth operation of MeshNCA even on edge devices. Additionally, we propose a way to graft two NCA models in test time to produce smooth transitions from one texture to another.  


MeshNCA works by replacing the 2D convolution kernels of the NCA model with a more general message-passing scheme based on spherical harmonics. Mesh vertices constitute the cells in our model and the edges of the mesh define the neighborhood for each cell. We train our model to synthesize a target exemplar texture, described either by an image or a text prompt. While MeshNCA is only trained on an \icosphere  mesh, remarkably, it can generalize to any mesh in test time, as it only relies on local interactions between the cells. 
Finally, we show the superiority of MeshNCA over existing 3D texture synthesis methods through qualitative and quantitative comparisons.
Our contributions summarized are as follows:

\begin{itemize}[leftmargin=0pt]
    \item We introduce MeshNCA, a novel model for real-time and interactive texture synthesis on 3D meshes. Our model possesses remarkable generalization capabilities along with diverse test-time properties.
    \item MeshNCA can be trained under image or text guidance, facilitating multimodal texture synthesis.
    \item We perform 3D dynamic texture synthesis by guiding the motion of the synthesized 3D texture through projected 2D motion supervision. MeshNCA can seamlessly adapt and synthesize the correct motion patterns on unseen meshes. 
    \item We propose MeshNCA grafting, where two MeshNCA models cooperate to synthesize novel hybrid textures during the test time.
    \item We implement the forward pass of our MeshNCA model using the WebGL shading language to showcase various user interactions and numerous test-time properties of MeshNCA. Our demo runs in real time on low-end edge devices, demonstrating the efficiency of MeshNCA. It is available at \textcolor{pinkl}{\href{https://meshnca.github.io}{https://meshnca.github.io/}}.
\end{itemize}

\section{Related Works}

In the following section, we discuss existing 3D texture synthesis methods. For comparison, Table~\ref{tab:method-compare} summarizes the strengths (\cmark) and shortcomings (\xmark) of these methods.  


\label{sec:related_works}

\newcommand\RotText[1]{\rotatebox{90}{\parbox{2cm}{\centering#1}}}




\begin{table}[t]
\caption{Comparison of 3D texture synthesis methods. 
\textbf{(\#$\mathbf{\theta}$)} The number of parameters required to synthesize new textures \protect\footnotemark.
\textbf{(A)} Can synthesize a diverse set of textures 
\textbf{(B)} Does not need high poly meshes 
\textbf{(C)} Does not require manual parameter tuning (Can be optimized)
\textbf{(D)} Can produce dynamic textures on 3D meshes 
\textbf{(E)} Does not need re-training for a new mesh
\textbf{(F)} Is aware of mesh semantics
\textbf{(G)} Can synthesize new texture instances in real-time
\textbf{(H)} Allows re-orienting the texture on the 3D mesh
\textbf{(I)} Allows texture grafting. 
\cmark$^\dagger$ Cannot generalize beyond the mesh categories in the training dataset. \cmark$^\ddagger$ The motion is random and cannot be controlled.
\label{tab:method-compare}
}
\resizebox{\linewidth}{!}{%
\begin{tabular}{cc|cccccccccc}
\toprule
\textbf{Type} & 
\textbf{Method} & 
\textbf{\#$\mathbf{\theta}$} & A & B & C & D  & E & F & G & H & I \\
\midrule
\multirow{6}{*}{\rotatebox{90}{\parbox{1.2cm}{\centering \textbf{\hspace{0pt} \small{Solid Textures}}}}}

& \boxed{\large 1} & \cellcolor[HTML]{FFFFFF}{$10^1$} & 
\xmark & \cmark & \xmark & \xmark & \cmark & 
\xmark & \cmark & \cmark & \xmark   \\
 
& Kopf et al. \shortcite{Solid_from_2d} & \cellcolor[HTML]{FFFFFF}{$10^7$} & 
\cmark & \cmark & \cmark & \xmark & \cmark & 
\xmark & \xmark & \cmark & \xmark   \\

& Oechsle et al. \shortcite{texture-fields} & \cellcolor[HTML]{FFFFFF}{$10^8$} & 
\xmark & \cmark & \cmark & \xmark & \cmark$^\dagger$ & 
\cmark & \cmark & \xmark & \xmark   \\

& \boxed{2} & \cellcolor[HTML]{FFFFFF}{$10^5$} & 
\cmark & \cmark & \cmark & \xmark & \cmark & 
\xmark & \cmark & \cmark & \xmark   \\

& Michel et al. \shortcite{text2mesh} & \cellcolor[HTML]{FFFFFF}{$10^6$} &
\cmark & \xmark & \cmark & \xmark & \xmark & 
\cmark & \cmark & \xmark & \xmark   \\

& Ma et al. \shortcite{xmesh} & \cellcolor[HTML]{FFFFFF}{$10^7$} &
\cmark & \xmark & \cmark & \xmark & \xmark & 
\cmark & \cmark & \xmark & \xmark   \\

\midrule

\multirow{3}{*}{\rotatebox{90}{\parbox{1.2cm}{\centering \textbf{\hspace{0pt} \small{Surface Textures}}}}} 

& \boxed{3} & \cellcolor[HTML]{FFFFFF}{$10^5$} & 
\cmark & \xmark & \cmark & \xmark & \xmark & 
\xmark & \xmark & \xmark & \xmark   \\

& Han et al. \shortcite{fast-example-2dto3d-texture} & \cellcolor[HTML]{FFFFFF}{$10^5$} & 
\cmark & \xmark & \cmark & \cmark & \xmark & 
\xmark & \xmark & \xmark & \xmark   \\

& \boxed{4} & \cellcolor[HTML]{FFFFFF}{$10^8$} & 
\xmark & \xmark & \cmark & \xmark & \cmark$^\dagger$ & 
\cmark & \cmark & \xmark & \xmark   \\

\midrule

\multirow{5}{*}{\rotatebox{90}{\parbox{1.5cm}{\centering \textbf{\hspace{0pt} \small{Cellular Automata}}}}}

& Turk \shortcite{turk-rd-texture} & \cellcolor[HTML]{FFFFFF}{$10^1$} & 
\xmark & \xmark & \xmark & \cmark$^\ddagger$ & \cmark & 
\xmark & \cmark & \cmark & \xmark   \\

& Fleischer et al. \shortcite{cellular-texture-generation} & \cellcolor[HTML]{FFFFFF}{$10^2$} & 
\xmark & \xmark & \xmark & \cmark & \xmark & 
\xmark & \cmark & \xmark & \xmark   \\

& Gobron et al. \shortcite{3d-surface-ca} & \cellcolor[HTML]{FFFFFF}{$10^1$} & 
\xmark & \cmark & \xmark & \cmark$^\ddagger$ & \cmark & 
\xmark & \cmark & \xmark & \xmark   \\

& Mordvintsev et al. \shortcite{diff-program-rdsystem} & \cellcolor[HTML]{FFFFFF}{$10^4$} &
\cmark & \xmark & \cmark & \cmark$^\ddagger$ & \cmark & 
\xmark & \cmark & \xmark & \xmark   \\

& \textbf{MeshNCA (Ours)} & \cellcolor[HTML]{FFFFFF}{$10^4$} & 
\cmark & \xmark & \cmark & \cmark & \cmark & 
\xmark & \cmark & \cmark & \cmark   \\

\bottomrule \\[-8pt]
\multicolumn{12}{l}{\boxed{1} \citet{perlin_noise} - \citet{solid_textures}} \\

\multicolumn{12}{l}{\boxed{2} \citet{on-demand-solid-texture} - \citet{neural-texture-space} - \citet{gramgan}} \\

\multicolumn{12}{l}{\boxed{3} \citet{turk2001texture} -  \citet{wei2001texture-neighbor-template} - \citet{ying2001texture-uvmapping} - \citet{progressively-variant-textures}} \\

\multicolumn{12}{l}{\boxed{4}\label{gp:group4} \citet{mesh2tex} - \citet{face_graph_texture} - \citet{texturify}} \\

\vspace{2pt}

\end{tabular}
}
\end{table}


\subsection{Solid Texture Synthesis}
The seminal work by Peachey \shortcite{solid_textures} and \citet{perlin_noise}, introduce the concept of \textit{Solid Textures} and the idea of using a \textit{space function} to map 3D coordinates to colors. Solid textures eliminate the need to design UV maps to texture a mesh. However, their approach cannot produce a diverse set of textures and requires a lot of creative thinking and manual design processes to obtain a specific texture.
To enable synthesizing a more diverse set of textures, \citet{Solid_from_2d} propose an optimization-based method that creates an explicit solid texture volume from a 2D exemplar texture.  




Extending the idea of solid texturing into the deep learning era, \citet{neural-texture-space} leverage \textbf{M}ulti \textbf{L}ayered \textbf{P}erceptron (MLP) to model the coordinate-to-color mapping function. Following the work of \citet{perlin_noise}, the input coordinates to the MLP are first transformed into noise at multiple frequency levels. Portenier et al. \shortcite{gramgan} propose to inject noise into the hidden layers of MLP. \citet{on-demand-solid-texture} use a 3D convolutional neural network to transform the input noise into a solid texture volume. For training, all of these methods use a form of Gram loss \cite{gatys2015texture} which is evaluated on 2D slices of the synthesized solid texture.


Instead of slice-based training, \citet{text2mesh} utilize a differentiable renderer to directly texture a given mesh. They apply positional encoding on the vertex coordinates and use an MLP to output color and displacement values for each vertex. Improvements have been made towards dynamically incorporating the text guidance information \cite{xmesh} or fitting to more vertex attributes \cite{tango}. \citet{texture-fields} present solid textures using a residual MLP which is conditioned on the shape code of the target object and the appearance code of the intended texture.  While \textit{Solid Texturing} methods can craft high-fidelity textures on 3D meshes, they are fundamentally incapable of resembling texture-related processes that happen on the surface of objects such as motion.

\subsection{Surface Texture Synthesis}

\footnotetext{The number of parameters for \citet{fast-example-2dto3d-texture} and the methods in group \boxed{3} is considered to be 3 times the number of pixels in the RGB exemplar texture since these methods need to store the exemplar to synthesize new textures.}

Surface texture synthesis methods assign a color for each point on the surface of a mesh. Early approaches rely on the \textbf{M}arkov \textbf{R}andom \textbf{F}ield (MRF) model for texture synthesis, in which the vertex color depends on its neighborhood \cite{efros1999texture-nonpa1, freeman2011mrf-nonpa4}.
Their neighborhood construction schemes include surface sweeping \cite{turk2001texture}, template matching \cite{wei2001texture-neighbor-template}, and neighborhood mapping from 2D to 3D \cite{ying2001texture-uvmapping}. After obtaining the neighborhood for a vertex, the algorithm searches for the best matching patch from the 2D exemplar texture and assigns the central pixel's color to the corresponding vertex. \citet{fast-example-2dto3d-texture} utilize discrete optimization techniques to improve the synthesis performance. Additional user-provided inputs such as texton masks \cite{progressively-variant-textures} or vector fields \cite{fast-example-2dto3d-texture}, can be incorporated into training to synthesize progressive-variant textures and dynamic textures on the surface, respectively. These methods allow modeling surface-related texture processes such as motion. Nevertheless, they require from-scratch optimization for each new mesh.

To alleviate the need for re-optimizing for new meshes and textures, the recent surface texture synthesis methods have shifted their focus from exemplar-based training to a dataset-based training schema \cite{face_graph_texture, mesh2tex, texturify}. 
\citet{texturify} propose a GAN training scheme in which the generator predicts target textures based on surface-feature-conditioned latent code. Building on this, \citet{mesh2tex} incorporate an MLP to achieve more detailed face textures. Using a different architecture and a similar training strategy, \citet{face_graph_texture} propose using graph neural networks operating on the faces of a mesh to synthesize textures without UV maps. These methods use large neural networks and train their model on a large dataset of textured meshes such as ShapeNet \cite{shapenet2015}. As a result, these methods are limited to the family of meshes and textures in the training dataset, restricting their applicability to real-world meshes and more diverse textures. Moreover, none of the surface texturing methods allow real-time user interaction to control the texture synthesis in test-time.

\subsection{Cellular Automata For Texture Synthesis}
\textit{Cellular Automata} (CA) \cite{vonneumann-introca} models are comprised of a set of cells with three main components: cell state, neighborhood, and update rule. The cell state stores all the information of a cell such as its color. The update rule defines how the cell states change over time, depending on the state of the cell and its neighbors. Inspired by Turing's seminal work \rev{\cite{turing-pattern}} proposing the Reaction-Diffusion (RD) model \footnote{RD models are a form of cellular automata as they use a discrete grid for simulations.}, \citet{turk-rd-texture} extend the reaction-diffusion simulation to arbitrary surfaces using Voronoi region-based neighborhood definition. \citet{3d-surface-ca} extend 2D cellular automata \cite{vonneumann-introca} to the 3D domain by discretizing mesh polygons with regular grids and subsequently performing 2D CA updates for each polygon. \citet{cellular-texture-generation} devise an elaborate cell simulation, in which solving a set of partial differential equations (PDEs) defined on cell attributes yields various textures. These CA-based models have the potential to generate 3D textures in real time and can generalize to a new mesh without retraining. However, all of these models rely on a manually crafted set of rules which makes them incapable of synthesizing a diverse set of textures. 

To address this issue, \citet{mordvintsev2020growing} propose Neural Cellular Automata (NCA) as a trainable counterpart of the traditional CA and reaction-diffusion systems. \citet{niklasson2021self-sothtml, mordvintsev2021mu-micronca} train NCA models to synthesize 2D textures. \citet{diff-program-rdsystem} resort to the Laplacian operator which allows them to transpose a 2D reaction-diffusion model onto the surface of 3D meshes. However, due to the isotropic nature of the Laplacian operator, their model is unaware of directional information and thus has limited capability of texture learning.
Our proposed model, MeshNCA, leverages non-parametric spherical-harmonics-based filters to incorporate directional information for all cells while preserving the purely local communication scheme. Hence, MeshNCA acts as a natural extension of the 2D NCA models into the 3D domain, while retaining all its remarkable properties such as efficiency, robustness, and controllability.

\section{Method}
\label{sec:meshnca_arc}
In the following sections, we discuss NCA models and their limitations. Following this, we explain the details of our MeshNCA model, which serves as a more generalized version of the vanilla NCA. Finally, we describe a scheme for grafting two NCA models to create interactive texture interpolation.


\subsection{Preliminaries}

\begin{figure}[t] 
	\centering
	\includegraphics[width=\linewidth, trim={0.0cm 24.90cm 17.53cm 0.0cm},clip]{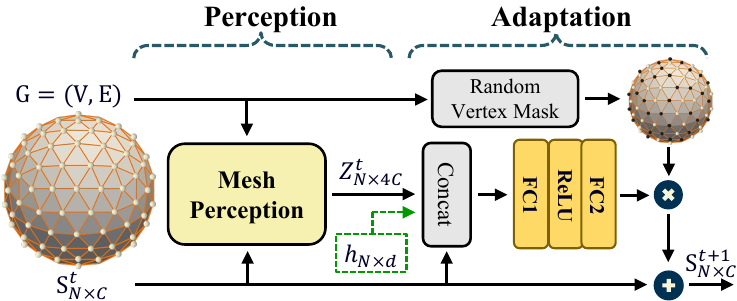}
	\vspace{-10pt}
	\caption{
        A single update step of MeshNCA. The underlying graph topology of the mesh is given by $G$. At each time step $t$, the cell states $\mathbf{S}^t_{N \times C}$ first pass through the \textbf{Mesh Perception} where each cell collects information from its neighborhood. The obtained perception tensor $\mathbf{Z}^t_{N \times 4C}$ is concatenated to the cell states and an optional per-cell conditional vector $h_{N \times d}$. Each cell then updates its state via the same \textbf{Adaptation} rule, parameterized by a two-layer MLP. The MLP outputs the residual update for cell adaptation, and all cells perform a stochastic update by multiplying a random vertex mask. Consequently, the cell states are determined at the time step $t + \Delta t$. We set $\Delta t = 1$ in our experiments. }
	\label{fig:architecture}
	\vspace{-15pt}
\end{figure}

\textit{Cell states}, \textit{Neighborhood Function}, and \textit{Local Update Rule} are the core components of any Cellular Automata model. The \textit{cell states} at time $t$ are represented by $\mathbf{S}^t \in \mathbb{R}^{N \times C}$, where $N$ is the total number of cells and $C$ is the dimensionality of a single cell state. A \textit{neighborhood function} $\mathcal{N}$ represents the neighboring relations between the cells such that $\mathcal{N}(i) = \{j \;| \; j \textup{ is a neighbor of } i \}$, where $i,j$ are cell indices. All cells in an CA model follow the same \textit{local update rule}. This update rule determines the next state of each cell based on the cell's state and its neighboring cell states. 

Neural Cellular Automata models are a subset of CA models in which the update rule is parameterized by a neural network. We decompose the \textit{local update rule} in NCA models into two parts: \textit{Perception} stage and \textit{Adaptation} stage. In the \textit{Perception} stage, each cell receives information from its local neighborhood, resulting in the perception vector $Z \in \mathbb{R}^{N \times C'}$. This newly gathered information subsequently directs the update of the cell's state in the \textit{Adaptation} stage. Notice that the perception stage is the only part of the update rule that allows the cells to receive information from their neighboring cells. 

The existing NCA models use the Moore neighborhood and operate on a regular grid of cells where the state can be represented as $\mathbf{S}^t \in \mathbb{R}^{H \times W \times C}$. They utilize frozen\footnote{The kernels are not optimized during training.} convolution kernels, including Sobel and Laplacian, to perform the perception operation \cite{mordvintsev2020growing, niklasson2021self-sothtml, mordvintsev2021mu-micronca, dynca}. Using Sobel and Laplacian kernels offers the advantage of enabling interactive directional controls in testing time \cite{niklasson2021self-sothtml, dynca}. However, this approach constrains the applicability of existing NCA models to grid-like, regular cell structures like pixel images, preventing their use with more flexible cell structures such as 3D meshes.

Our MeshNCA model extends the vanilla NCA, enabling it to work with arbitrarily positioned sets of cells and eliminating the need for a grid cell structure. We model the \textit{perception stage} of MeshNCA using the message-passing scheme in Graph Neural networks \cite{message-passing, gnca} and generalize the convolution filters of vanilla NCA to any arbitrary cell structure by utilizing Spherical Harmonics. In the following section, we elaborate on the MeshNCA architecture.

\subsection{MeshNCA Architecture}

\label{sec:architecture}

We use a graph to represent the arrangement of the cells in our model. Given a mesh, we construct a graph $\mathbf{G} = (V, E)$ where $V \in \mathbb{R}^{N\times3}$ denotes the vertices of the mesh and their 3D positions and $E$ denotes the edges of the mesh. The nodes of this graph represent the cells and the edges of the graph determine the neighborhood relations between the cells. In this work, we use undirected edges to have a symmetric neighborhood relationship between the cells. MeshNCA interprets each vertex of the mesh as a cell and updates the cell states over time, starting from a constant zero tensor $\mathbf{S}^0$ as the initial state. Figure~\ref{fig:architecture} illustrates one step of the MeshNCA update rule. We decompose the update rule into two main stages, \textit{Perception}, and \textit{Adaptation}, which will be further explained in the following sections.






\subsubsection{\textbf{Perception Stage}}

\begin{figure}[t] 
	\centering
	\includegraphics[width=\linewidth, trim={0.0cm 14.0cm 16.05cm 0.0cm},clip]{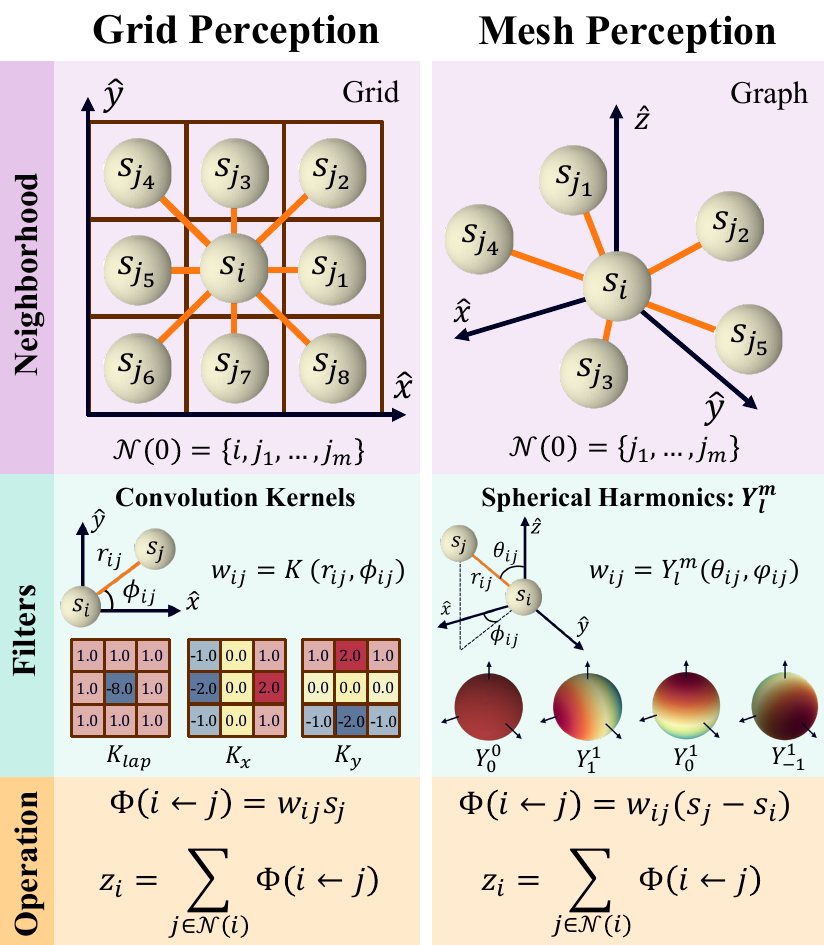}
	\vspace{-10pt}
        
	\caption{
        A comparison between regular \textbf{Grid Perception} and our \textbf{Mesh Perception}. \textbf{Neighborhood:} In grid perception, each cell has a fixed number of neighbors with fixed relative directions and distances. In mesh perception, each cell can have a different number of neighbors with arbitrary relative positions to the center cell. \textbf{Filters:} The convolution filters in grid perception can be rewritten as a function of the distance and direction between the center node and its neighbor. Extending this idea, we use spherical harmonics as the basis of our mesh perception filters, acting as a continuous generalization of discrete convolution filters to non-grid-like structures.
        \textbf{Operation:} Both perception methods share similar information-gathering operations to form the perception vector $z_i$ for each cell.}
 \label{fig:perception}
	
\end{figure}

Figure~\ref{fig:perception} shows an overview of our \textit{Mesh Perception} stage. We use the message-passing scheme of Graph Neural networks \cite{message-passing} as the basis of our \textit{Mesh Perception}. During message-passing, each cell $i$ receives a message $\Phi (i\leftarrow j)$ from the cell $j$. 
The perception vector for each cell is simply the summation of all the messages received by that cell: 
\begin{equation}
    z_i = \sum_{j \in \mathcal{N}(i)} \Phi(i \leftarrow j)
    \label{eq:message_passing}
\end{equation}

The main idea underlying our \textit{Mesh Perception} layer is to generalize the 2D convolution kernels of the vanilla NCA model to non-grid structures such as meshes by reformulating the convolution operation using a graph-based message-passing scheme. Our key insight is that any 2D convolution filter is a discrete lookup table that can be realized using a continuous function $filter(\phi_{ij}, r_{ij})$ that maps the distance $r_{ij}$ and the polar angle $\phi_{ij}$ between two pixels $i$ and $j$ to a coefficient $w_{ij}$ as shown in Figure~\ref{fig:perception}. The message that the center cell $i$ receives from its neighbor cell $j$ is then defined as $\Phi (i\leftarrow j) = w_{ij} s_j$ where $s_j$ is the state of the neighbor cell $j$.
The vanilla NCA architecture uses 3 different convolution filters in the perception stage, including two Sobel filters $K_x, K_y$ and one Laplacian filter $K_{lap}$. For example, the weights of the Sobel filter $w_{ij} = K_x(\phi_{ij}, r_{ij})$ can be realized using the following function:
$$K_x(\phi_{ij}, r_{ij}) = \left\{\begin{matrix}
0 & \textup{if} \; r_{ij} = 0\\ 
2 \; \textup{sgn}(\cos \phi_{ij} ) \cos^2 (\phi_{ij})  & \textup{otherwise}
\end{matrix}\right.$$
where $\textup{sgn}$ is the Sign function. 


Interpreting \textit{convolution filters as functions} facilitates the extension to non-grid structures such as meshes. For two arbitrarily located cells $i$ and $j$ in 3D space, a filter function can be written as $f(r_{ij}, \theta_{ij}, \phi_{ij})$ where $[r_{ij}, \theta_{ij}, \phi_{ij}]$ are the spherical coordinates of the vector connecting the cell $i$ to the cell $j$. We find that removing the dependence on distance $r_{ij}$ is not harmful to the expressivity of our model for texture synthesis. 
Therefore, we only consider functions of the form $f(\theta_{ij}, \phi_{ij})$, and choose \textit{Spherical Harmonics} $Y^m_l(\theta, \phi)$ as the basis of our Mesh Perception filters, as shown in Figure~\ref{fig:perception}. We use the first-order spherical harmonics, $l \leq 1$, which lead to four basis functions as our filters. For each filter $Y^m_l$, we evaluate the message coefficient $w_{ij} = Y^m_l(\theta_{ij}, \phi_{ij})$ and use the following equation
\begin{equation}
    \Phi (i\leftarrow j) = w_{ij} (s_j - s_i)
    \label{eq:perception}
\end{equation} 
to evaluate the message passed from neighbor cell $j$ to the center cell $i$. With our proposed message passing scheme introduced in Equation~\ref{eq:perception}, the first four spherical harmonics become analogous to the Laplacian and Sobel filters, where $Y^0_0$ corresponds to the Laplacian filter $K_{lap}$ and $Y^1_1, Y^1_0, Y^1_{-1}$ correspond to the Sobel filters $K_x, K_z, K_y$, respectively. Notice that our mesh perception is completely non-parametric and is not optimized during the training. We refer the reader to our supplementary material for an ablation study on parametric versus non-parametric perception. 
As shown in Figure~\ref{fig:architecture}, the output of the perception stage is $Z^t_{N\times 4C}$, where $z^{t}_i$ is the perception vector of the cell with index $i$ at time step $t$. The 4-time expansion in the number of channels is due to the fact that we use the first four spherical harmonics bases in our perception filters. 
We refer the reader to our supplementary material for an ablation study on the degree of the spherical harmonics.

\revv{Spherical harmonics show up in many different places in geometric computing and computer graphics. From the \textit{Manifold Harmonics} perspective, spherical harmonics are solutions of the Laplace equation on a sphere manifold \cite{vallet2008spectral}. They have also been used to create more sophisticated lighting sources for rendering applications \cite{lighting-spherical} and directionally-varying colors for neural rendering \cite{ReluField}. In this work, we use spherical harmonics to encode, for a given cell, the anisotropic influence of its neighboring cells as a function of their relative position. 
}


\subsubsection{\textbf{Adaptation Stage}}

In the \textit{Adaptation} stage, each cell $i$ independently updates its own state based on the information available to it. This information constitutes $z_i^t \in \mathbb{R}^{4C}$, $s_i^t \in \mathbb{R}^{C}$ and $h_i \in \mathbb{R}^d$ where $z_i^t, s_i^t$ are the perception vector and the cell state at the current step, respectively. The vector $h_i$ stores the optional per-cell conditional information\footnote{In the text-guided and image-guided experiments (Sect.~\ref{sec:exp-image-synthesis} and \ref{sec:exp-text-synthesis}), we omit the condition vector. In the dynamic texture experiment (Sect.~\ref{sec:exp-motion}), we use the projected motion vector field as the condition vector. }. We concatenate $z_i^t, s_i^t$, and $h_i$ in the channel dimension and pass the resulting $5C + d$ dimensional vector to a \textbf{M}ulti-\textbf{L}ayered-\textbf{P}erceptron (MLP) with two layers and a ReLU activation function. The output of the MLP is then masked by a random variable $M$ having Bernoulli distribution with $p=0.5$ and added to the current state as shown in the following equation:

\begin{equation}
s^{t+1}_{i} = \left\{\begin{matrix}
s^{t}_{i} & M = 0\\ 
s^{t}_{i} + \textup{MLP} (s^{t}_{i}, z^{t}_i,h_i) & M = 1
\end{matrix}\right.
    \label{eq:adaptation}
\end{equation}

Stochastic random-vertex masking fosters asynchronicity in the update rule and enables the NCA to generate new textures over time. Note that all operations in the \textit{adaptation} stage are vertex-wise and can be executed in parallel for all cells, allowing for an efficient implementation.

\subsection{Grafting}

\label{sec:grafting}

We introduce a useful property of MeshNCA that we call "grafting", named after the practice of physically joining tissues of different plants within the same species. Grafting is a test-time property of MeshNCA that allows spatially/temporally interpolating between two different textures. Grafting requires both a particular method of test-time interpolation and a certain training scheme. In the following section, we first introduce the concept of compatible MeshNCA instances and then propose our method for test-time grafting of MeshNCA instances. The training scheme for acquiring graftable MeshNCA instances is detailed in Section~\ref{sec:grafting-training}.
We refer the reader to the supplementary for a discussion on the challenges of texture interpolation and the importance of grafting.

\subsubsection{Compatible MeshNCA Instances} 

We demonstrate that MeshNCA naturally lends itself to solving the texture interpolation problem. When one MeshNCA $S$ is trained on a specific texture, and a second MeshNCA $T$ is trained on another texture, initialized with the converged weights $W_S$ from $S$, we empirically observe that cells (vertices) from both $S$ and $T$ communicate, cooperate and coexist in a constructive manner. On the contrary, when both $S$ and $T$ are independently trained from scratch, such constructive collaboration between cells disappears. This is reminiscent of the difficulty in biological grafting of two plants of different species.  
We refer to MeshNCA models that are from the same lineage (being trained from a common ancestor) as "compatible", akin to the concept of \textit{genetic ties}\footnote{A \textit{genetic tie} signifies a biological link that is passed down through genes and contributes to the similarities and resemblances observed among related individuals.} in biological systems.

\subsubsection{Test-time Grafting Scheme}

We demonstrate one simple and effective scheme to effectively exploit "compatible" MeshNCA instances for texture interpolation in our online demo, implemented as the \textit{Graft} paintbrush tool. Note that if compatible cells are simply placed next to each other along a boundary, they tend to try and form their specific pattern on each side of this boundary, and only exactly on the boundary do they perform limited coordination of visual features between the two textures. 
If instead we proportionally interpolate the updates computed by each of two compatible MeshNCA, a smooth transition between the textures is produced. Suppose that we want to interpolate between the corresponding textures of $S$ and $T$, with trained weights $W_S$ and $W_T$, respectively, where $W_T$ is trained by initializing MeshNCA using $W_S$.
For each vertex on the mesh, we then choose a value $0.0 \leq \alpha \leq 1.0$ according to a soft grafting mask, which denotes the interpolation weight between two textures. The grafting mask is updated when the user paints on the mesh using the grafting brush tool. 
During the test time, at each time step, we compute both $u_T = \text{MeshNCA}_{W_T}(x)$ and $u_S = \text{MeshNCA}_{W_S}(x)$, and interpolate the updates according to $u = \alpha * u_T + (1-\alpha) * u_S$. This results in a natural-looking transition between the two textures, with macroscopic features from each texture appearing partially "inside" the other at times. Alternatively, they might exhibit a strong alignment with analogous features in the other texture. We refer the readers to the supplementary material for an ablation study of our grafting scheme.





\section{MeshNCA Training}

\begin{figure}[t] 
	\centering
	\includegraphics[width=\linewidth, trim={0.21cm 22.02cm 15.52cm 0.15cm},clip]{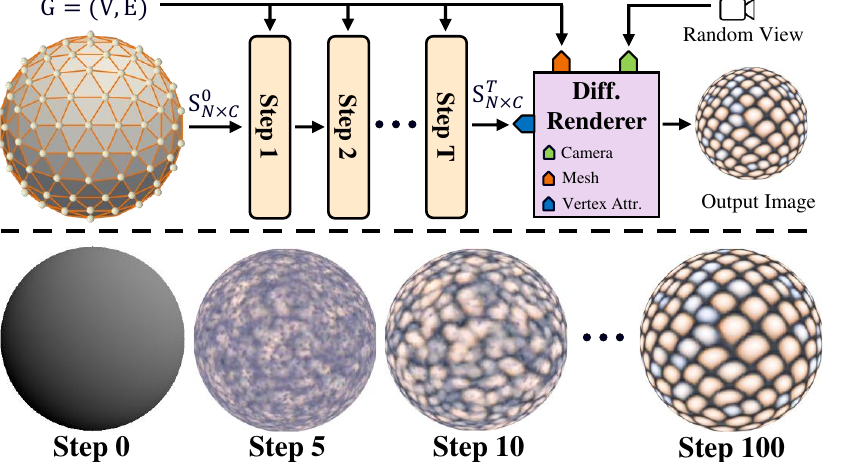}
	\vspace{-10pt}
	\caption{\rev{\textbf{Top row:} Overview of the texture synthesis process of MeshNCA. Starting with seed cell states $\mathbf{S}^0_{N \times C}$, MeshNCA iteratively updates the cell states by alternating the perception and adaptation stages. The underlying graph topology remains the same across all time steps. The last cell state $\mathbf{S}^T_{N \times C}$ and the mesh are then passed to a differentiable renderer and rendered using a perspective camera with a random view to get the output image. \textbf{Bottom row:} Synthesized textures rendered from cell states at different time steps.}    
    }
	\label{fig:meshnca_seq}
\end{figure}

MeshNCA acts as a PDE defined on the mesh graph. It evolves over time and produces an ordered series of cell states $\mathbf{S}^g=\{{\mathbf{S}^{g_1}, \mathbf{S}^{g_2}, \cdots}\}$. By extracting different channels from the cell states, we modify the vertex attributes and alter the appearance of the mesh accordingly, forming a sequence of textured mesh $\mathcal{M}^g=\{\mathcal{M}^{g_1}, \mathcal{M}^{g_2}, \cdots\}$. Utilizing an interpolation-based differentiable renderer $R$ \cite{kaolin}, we render the textured mesh with a perspective camera to obtain the output image as shown in Figure~\ref{fig:meshnca_seq}. The rendering of the entire mesh sequence results in an array of images $\mathcal{I}^g=\{{\mathcal{I}^{g_1}, \mathcal{I}^{g_2}, \cdots}\}$. The rendered images are then evaluated against different target modalities, that is, images or texts, to compute the corresponding appearance losses $\mathcal{L}_{appr-im}$ and $\mathcal{L}_{appr-tx}$. Additionally, we can also impose 2D-only supervision on the emergent motion during training to synthesize the desired 3D dynamic textures. The motion loss term is denoted as $\mathcal{L}_{mot}$. We elaborate on the loss functions in the following sections.

\subsection{Image-guided Training}
\label{sec:image-guided-training}

Leveraging a differentiable renderer, we backpropagate the gradient flow from the rendered images to the MeshNCA parameters. Hence, 2D texture synthesis schemes can be adopted, which typically work by aligning the statistics of deep features of the generated images to those of the target image. Following \cite{dynca}, we utilize a pre-trained VGG16 network \cite{vgg} to extract deep features, using the rendered images as input. We denote VGG by $\mathcal{F}_{VGG}$ and the feature map drawn from the layer $l$ as $\mathcal{F}^{l}_{VGG}$. 
Considering that the synthesized 3D texture should be visually analogous to the 2D exemplar image from any specific view, we adopt the multi-view rendering procedure similar to the one in \cite{dip}. To ensure the stability of training and the quality of the results, we implement the style loss proposed by \cite{kolkin2019style-otloss}. 
It consists of a relaxed Wasserstein distance and a moment-matching term. The premise behind this approach is to analytically solve a relaxed optimal transport problem between the extracted features and use this distance as a metric for evaluating style similarity. First, the algorithm converts the deep feature map of size $C \times H \times W$ into a deep feature set by flattening the feature map along the spatial dimensions. This results in the feature sets $A, B$ derived from the synthesized and target images, respectively. The relaxed Wasserstein distance $\mathcal{L}_{W}$ and the moment-matching term $\mathcal{L}_{M}$ are then defined as: 


\begin{equation}
\begin{aligned}
    W(A, B) =  \frac{1}{|A|} \sum_{i} \mathop{min}\limits_{j} \left ( D_{cos} (A_i, B_j) \right. 
    \left. + \mathbbm{1}_{RGB} D_{L2} (A_i, B_j) \right )\\
    \mathcal{L}_{W}(A,B) = \mathop{max} \left ( W(A, B), W(B, A) \right ) \quad \quad \quad \quad \\
\end{aligned}
\label{eq:ot-structure}
\end{equation}

\vspace{0pt}
\begin{equation}
\begin{aligned}
   \mathcal{L}_{M}(A,B)=\frac{1}{C} D_{L1}(\boldsymbol{m}_A,  \boldsymbol{m}_B)  + \frac{1}{C^2} D_{L1}(\boldsymbol{T}_A, \boldsymbol{T}_B)
\end{aligned}
\label{eq:ot-moment}
\end{equation}
Here, $i, j$ are element indices; $D_{cos} $, $D_{L1}$, and $D_{L2} $ are the cosine, L1, and L2 distances; and $\boldsymbol{m}$, $\boldsymbol{T}$ are the mean and covariance matrices of the corresponding sets, respectively. $\mathbbm{1}_{RGB}$ is the indicator function that returns $1$ if $A, B$ are the input RGB images and $0$ if they are features extracted from VGG. 
The final loss of image-guided training $\mathcal{L}_{appr-im}$ involves matching multi-layer features. Let $A^l_k$ and $B^l$ be VGG activations extracted by $\mathcal{F}^{l}_{VGG}$ or the input images, where $l$ and $k$ stand for the VGG layer index and the view index, respectively. The $l=0$ stands for image input. $\mathcal{L}_{appr-im}$ is then defined as:
\begin{equation}
    \mathcal{L}_{appr-im}=\frac{1}{K}\sum^{K}_{k=1} \sum^L_{l=0} \left (\mathcal{L}_{W} (A^l_k, B^l) +  \mathcal{L}_{M} (A^l_k, B^l)\right )
\label{eq:appr-im-loss}
\end{equation}
We choose the activations after each convolution layer as well as each pooling layer.  

A unique property of NCA is that all the channels of the cell states have aligned structures without imposing external supervision \cite{niklasson2021self-sothtml}. This advantage allows MeshNCA to simultaneously fit multiple texture maps such as albedo, surface normal, roughness, height, and ambient occlusion. We treat each of these texture maps as target texture images and use different channels of the cell states to fit each texture map. We assign 3 channels of the cell states $S_{N\times C}$ to the albedo, 3 channels to the surface normal, 1 channel to the height map, 1 channel to the roughness map, and 1 channel to the ambient occlusion map. For single channel texture maps such as the height map, we repeat them 3 times along the channel dimension to get an RGB image before evaluating our appearance loss $\mathcal{L}_{appr-im}$. 
The design of MeshNCA naturally fulfills the requirement of homologous structure between different texture maps, thus allowing efficient and simultaneous synthesis of different texture maps for physically-based rendering (PBR) applications.

\subsection{Text-guided Training}
\label{sec:text-guided-training}

The multi-modal embedding space provided by the CLIP model \cite{clip} renders text-guided image generation and editing feasible. In such a training scheme, a text prompt $\mathcal{T}$ is used to describe the desired outcome. We use the following prompt template: "\textit{An image of a/an <Object> made of <Prompt>}", where "\textit{<Object>}" is the mesh name, such as \textit{bunny, vase, chair}, and "<Prompt>" describes the target texture, such as \textit{feathers, stained glass, colorful crochet}.
Contrary to existing methods in 3D texture synthesis \cite{tango, text2mesh, xmesh}, our approach eliminates the need to add specialized branches to fit the additional geometric vertex attributes. Upon the iteration of MeshNCA, we extract the first 3 channels of the cell states as vertex colors and the fourth channel as the vertex displacement along the surface normal, namely the geometric channel. We constrain the range of the displacement to maintain the overall shape of the mesh. Inspired by \cite{text2mesh}, we render two images from the same textured mesh, the first using both color and geometric channels, denoted as $\mathcal{I}^{g-color}$, and the second using only the geometric channel by assigning a gray color to the vertices, denoted as $\mathcal{I}^{g-geo}$. 
To avoid potential degeneration of the result, we employ a global augmentation $Aug_{glo}$ and a local augmentation $Aug_{loc}$ to $\mathcal{I}^{g-color}$ and $\mathcal{I}^{g-geo}$ to obtain multiple variations of the synthesized texture. We use the ViT-B/32 model of CLIP to encode all rendered images to 1D latent codes. An average of these latent codes, taken across all rendering views, provides us with a robust representation of the textured mesh in the CLIP's latent space:

\begin{equation}
    \begin{aligned}
        & \mathcal{R}_{glo} = \frac{1}{K} \sum^{K}_{k=1} CLIP(Aug_{glo}(\mathcal{I}^{g-color}_k)) \\
        & \mathcal{R}_{loc} = \frac{1}{K} \sum^{K}_{k=1} CLIP(Aug_{loc}(\mathcal{I}^{g-color}_k)) \\
        & \mathcal{R}_{loc-geo} = \frac{1}{K} \sum^{K}_{k=1} CLIP(Aug_{loc}(\mathcal{I}^{g-geo}_k))
    \end{aligned}
    \label{eq:clip-image-encoding}
\end{equation}
where $K$ is the number of rendering views. To further avoid corrupted results, we extend the directional CLIP loss\cite{clip-direction-loss, cliptexture-clip-direction-loss} to the 3D domain. Unlike the 2D case, the negative components still preserve certain semantic information. Specifically, we define the negative mesh $\mathcal{M}^{neg}$ as the original mesh with a gray color and without any geometric change. The negative images $\mathcal{I}^{neg}$ are rendered from $\mathcal{M}^{neg}$. The negative prompt, $\mathcal{T}^{neg}$, is defined to be the name of the mesh object. The negative images are encoded into latent codes following Equation \ref{eq:clip-image-encoding} to obtain $\mathcal{R}^{neg}$. We compute the direction vectors between $\mathcal{R}$ and $\mathcal{R}^{neg}$ as well as $\mathcal{T}$ and $ \mathcal{T}^{neg}$ using Equation~\ref{eq:clip-direction-component}.

\begin{equation}
\begin{aligned}
    & \Delta \mathcal{R} = \mathcal{R} - \mathcal{R}^{neg}, \\
    & \Delta \mathcal{C} = CLIP(\mathcal{T}) - CLIP(\mathcal{T}^{neg})
\end{aligned}
\label{eq:clip-direction-component}
\end{equation}
The 3D directional CLIP loss $\mathcal{L}_{appr-tx}$ is defined accordingly:
\begin{equation}
\begin{aligned}
    \mathcal{L}_{appr-tx} = \sum_{i} D_{cos}(\Delta \mathcal{R}_i, \Delta \mathcal{C})
\end{aligned}
\label{eq:clip-directional-loss}
\end{equation}
where $i \in \{glo, loc, loc-geo\}$.  $D_{cos}$ is the cosine distance between two 1D vectors. We refer the readers to our supplementary material for an ablation study of the proposed 3D directional CLIP loss.

\subsection{Dynamic Texture Synthesis}
\label{sec:motion-training}

MeshNCA demonstrates many unique emergent properties after training, one of which is spontaneous motion, a characteristic also observed in the NCA model \cite{niklasson2021self-sothtml,dynca}. \citet{dynca} succeed in controlling the motion generated by NCA by introducing 2D motion training signals. However, guiding the motion on a 3D surface remains a challenging task. First, there is no pre-trained model for estimating the 3D motion for each vertex. Moreover, unlike the 2D motion in the vanilla NCA that can happen in any direction, the motion on a 3D mesh is limited to the local tangent plane. To solve these problems, we propose to project the user-defined 3D target vector field into 2D to obtain the target optical flow from the camera perspective. Then, we use this projected 2D optic flow to supervise the MeshNCA motion.  

Let $U: \mathbb{R}^{3} \rightarrow \mathbb{R}^{3}$ be the user-defined global motion vector field. We evaluate this 3D motion vector field on the mesh vertices to get per vertex motion $U^{v} \in \mathbb{R}^{N \times 3}$. Note that this target per vertex motion is not necessarily in the local tangent plane and needs to be projected onto the mesh surface. Given a 3D mesh with vertex normal set $N=\{\hat{n}_1, \hat{n}_2, \cdots\}$, we project $U^{v}$ onto the local tangent space of the mesh to obtain the tangent motion vector field for each vertex $U^{t}$. 
\begin{equation}
    U^t_{i} = U^{v}_i - (U^{v}_i \cdot \hat{n}_i) \hat{n}_i,
\end{equation}
where $i$ is the vertex index. To make the cells aware of the target motion direction, we set the per-vertex conditional vector, shown in Figure~\ref{fig:architecture} to $h_{N\times d} = U^t$, resulting in \textbf{Motion Positional Encoding} (MPE). Our ablation study in the supplementary shows the necessity of MPE for successfully training MeshNCA to synthesize dynamic textures. 

Given a rendering view with a camera transformation matrix $CT_{k}$, we project the per-vertex tangent motion vector field $U^t$ into the camera-centered coordinate via linear projection $U^{p}_k = U^t \cdot CT_{k}$. Finally, by using the vertex-to-pixel correspondence, we can compute the 2D target motion vector field $U_{k}^{2D} \in \mathbb{R}^{H \times W \times 2}$ for the given view, where $H \times W $ is the render resolution. By keeping the camera parameters unchanged before and after the MeshNCA iterations, a pre-trained 2D optical flow estimation network $\mathcal{F}_{OF}$ can predict the view-dependent optic flow for 2D consecutive synthesized images $U^{gen} = \mathcal{F}_{OF}(\mathcal{I}^g_{t_1}, \mathcal{I}^g_{t_2})$. Here, $t_1, t_2$ are two different time steps along the evolution of MeshNCA, with $t_1 < t_2$. We draw inspiration from \cite{dynca} to supervise the direction and strength of the generated motion via two losses $\mathcal{L}_{dir}$ and $\mathcal{L}_{str}$, as described in Equations \ref{eq:dir_loss} and \ref{eq:strength_loss}, respectively. Our training also incorporates a multi-view strategy to ensure motion consistency on the mesh surface. 

\begin{equation}
\mathcal{L}_{dir} = \frac{1}{K} \sum^{K}_{k=1} \frac{1}{HW} \sum_{i, j} D_{cos} (U^{gen}_{ijk}, U^{2D}_{ijk}) ,
\label{eq:dir_loss}
\end{equation}

\begin{equation}
\mathcal{L}_{str} = \frac{1}{K} \sum^{K}_{k=1} \frac{1}{HW} \sum_{i, j}  D_{L1} 
\left (
    \frac{T \left \| U^{gen}_{ijk} \right \|_2}{t_2 - t_1}, \left \| U^{2D}_{ijk} \right \|_2 
\right ).
\label{eq:strength_loss}
\end{equation}
Where $i, j$ are spatial indices and $k$ is the camera view index. Here, $D_{cos}$ and $D_{L1}$ are the cosine and $L1$ distances, respectively. The constant $T$ regulates the correspondence between the number of MeshNCA steps and the desired motion strength of a single frame.
To prioritize the direction matching loss, we formulate the final loss $\mathcal{L}_{mot}$ as:

\begin{equation}
\mathcal{L}_{mot} = \left \{
\begin{array}{lc}
    \mathcal{L}_{dir}, & \mathcal{L}_{dir} \geq 1.0 \\
    \left ( 1.0 - \mathcal{L}_{dir} \right ) \mathcal{L}_{str} + \gamma \mathcal{L}_{dir}, & \mathcal{L}_{dir} < 1.0 
\end{array}
\right.
\label{eq:motion-loss}
\end{equation}
where $\gamma$ is a constant hyperparameter. This approach enables the model to learn to synthesize the correct direction first, resulting in more coherent motions. The motion loss term, $\mathcal{L}_{mot}$, serves as an additional component in MeshNCA training and can be combined with both image and text-guided training. The final loss term $\mathcal{L}_{dyn}$ in dynamic texture synthesis becomes:

\begin{equation}
\mathcal{L}_{dyn} = \mathcal{L}_{appr} + \lambda \mathcal{L}_{mot}.
\label{eq:dynamic-final-loss}
\end{equation}
$\mathcal{L}_{appr}$ may take the form of $\mathcal{L}_{appr-im}$ in Section \ref{sec:image-guided-training} or $\mathcal{L}_{appr-tx}$ in Section \ref{sec:text-guided-training}.


\subsection{Graftable models Training }
\label{sec:grafting-training}

MeshNCA models exhibit the property of different trained models being able to communicate and form coherent transitions between textures. This property arises only when the models are trained in a specific fashion - a method that resembles the process of "fine-tuning" as is done with large language and large image models. We consider two NCA models to be "compatible" if neighboring cells running one of the two rules, on the same substrate (whether a mesh as in MeshNCA or a 2D grid as in the original NCA), exhibit visual coherence in the pattern formed across the boundary between them. We refer the readers to Section \ref{sec:grafting} for more details on how to graft trained MeshNCA instances.

We find that one way to encourage two MeshNCA models to be compatible is to first train a single model on a texture, and then use the resulting converged weights as the initial weights when training the second MeshNCA on a different texture. Note that the naive approach to train/fine-tune separate MeshNCA models for each pair of textures is not practical, as the training time grows quadratically with the number of textures. We propose a different training scheme that eliminates the need for pairwise training. We choose one texture as the parent, train one MeshNCA model on the parent texture, and then train all subsequent models using the weights of the parent model for initialization. In this manner, all trained MeshNCA models are from the same lineage, thereby ensuring compatibility. We observe that with our proposed scheme, training converges faster compared to training with random initialization while achieving higher or on-par visual quality. This suggests that the underlying functionality the model learns to produce one texture is at least in part applicable to forming other textures. We refer the readers to the supplementary for an ablation study of our training strategy for creating graftable models.





\section{Experiments}
\label{sec:exp}




We conduct experiments on the 3D texture synthesis capabilities of MeshNCA, including \textbf{Image-guided} and \textbf{Text-guided} synthesis, in Section \ref{sec:exp-image-synthesis} and \ref{sec:exp-text-synthesis}, respectively. In all texture synthesis experiments, MeshNCA shares the same configuration: cell state dimensionality $N_C=16$, and the output dimensionality of the first layer of MLP being $N_{FC}=128$. This configuration results in a very parameter-efficient model with 12432 trainable parameters. \rev{We perform ablation studies on $N_C$ and $N_{FC}$, showing that this configuration achieves an optimal balance between model size and synthesis quality. We refer the readers to the supplementary material for details of the ablation studies.}
Moreover, we demonstrate that by only imposing motion supervision on the albedo output channels, MeshNCA exhibits the capacity to generate diverse dynamic textures while preserving the motion coherence between albedo and other texture maps. With the additional motion targets, we set $N_C=32$, resulting in 25120 parameters. We use Nvidia Kaolin \cite{kaolin} as our differentiable renderer. All experiments are performed on an Nvidia-A100 GPU and use Adam optimizer with an initial learning rate of 0.001. Notice that MeshNCA is only trained on an \icosphere  mesh in all of our experiments.


We refer the readers to the supplementary for our ablation studies and more qualitative results. Results in video format are available in our online demo under the supplementary section \textcolor{pinkl}{\href{https://meshnca.github.io/supplementary}{meshnca.github.io/supplementary}}.






\begin{table}[]
\caption{The rendering parameters used in \textbf{Image-guided}, \textbf{Text-guided} synthesis, and corresponding dynamic texture synthesis experiments. For \textbf{Text-guided} experiments, we train each text prompt using 3 different camera distances and pick the one with the lowest loss. $K$: The number of views in Equation \ref{eq:appr-im-loss}. $C$: The dimensionality of the cell state. $h$: per-cell conditional vector. None means without conditioning. MPE is the proposed Motion Positional Encoding in Section \ref{sec:motion-training}.}
\resizebox{\linewidth}{!}{
\begin{tabular}{ccccccc}
\toprule
\textbf{Guidance} & \parbox[c]{2cm}{\centering\hspace{-5pt} \textbf{Render Resolution}} & \parbox[c]{2cm}{\centering\hspace{-5pt} \textbf{Camera Distance}} & \textbf{$K$} & \textbf{$C$} & \textbf{$h$} &  \textbf{\# Params} \\
 \midrule
\; Image & $320\times320$ & 2.5 & 6 & 16 & None& 12k \\
+ Motion & $128\times128$ & 2.0 & 6 & 32 & MPE & 25k \\
\midrule
\; Text & $224\times224$ & \{2.0, 3.0, 4.0\} & 8 & 16 & None & 12k \ \\
+ Motion & $224\times224$ & 2.0 & 8 & 16 & MPE & 13k \\
\bottomrule
\end{tabular}
}
\label{tab:image-text-param}
\end{table}

\subsection{Image-guided Synthesis}
\label{sec:exp-image-synthesis}
We collect 72 PBR textures from \cite{3dtexture-web}, all available under the CCO license. Each of these textures comes accompanied by albedo, roughness, height, ambient occlusion, and normal maps.
For fitting the albedo and normal maps, we extract three distinct channels for each map from the MeshNCA cell states. For each of the other texture maps that are typically represented by a gray-scale image, we repeat one distinct channel of the cell states three times to respect the single-channel nature of the target texture maps. We apply our appearance loss function, defined in Equation~\ref{eq:appr-im-loss}, separately on each of the texture maps. We only use ambient lighting to train our image-guided models since the values of these texture maps reflect physical properties and their values should not be influenced by lighting during training. The rendering parameters are given in Table~\ref{tab:image-text-param}. The results are shown in Figure \ref{fig:image-synthesis-result}.
We provide our trained models in an online demo available at \textcolor{pinkl}{\href{https://meshnca.github.io/}{https://meshnca.github.io/}} that incorporates a physically-based rendering shader to combine all of the MeshNCA synthesized textures into a single rendering. Notice that all of these models are trained to be graftable as described in Section~\ref{sec:grafting-training} and the users can experiment with the interactive grafting brush in our demo to create texture interpolations. For graftable model training, we first use the ``\textit{Wall\_Shells\_001}'' texture to train a MeshNCA model and subsequently train all other models initialized by the weights of that pre-trained model.


\newcommand{\imgimg}[1]{%
  \ifimgprefix
    \includegraphics[height=60pt]{figures/Experiments/image-synthesis/#1}%
  \else
    \includegraphics[height=60pt]{figures/Experiments/image-synthesis/L_#1}%
  \fi
}

\begin{table*}[]
\resizebox{\textwidth}{!}{%
\begin{tabular}{cc||cc||ccc}
 \multicolumn{2}{c||}{\textbf{Target Texture Maps}}  &  \multicolumn{2}{c||}{\textbf{Icosphere Train}}  &  \multicolumn{3}{c}{\textbf{Test-time Generalization}}  \\
 Albedo & Attributes &  Albedo & Attributes &  Albedo & Attributes & Rendered \\
 \midrule
\imgimg{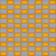}  &  \imgimg{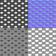}  &  \imgimg{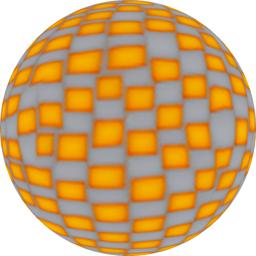}  &  \imgimg{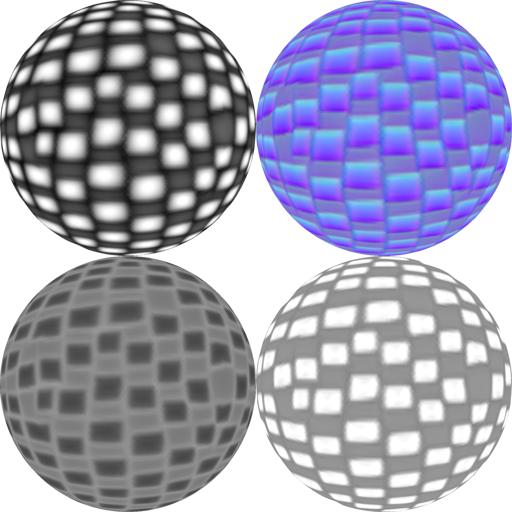}  &  \imgimg{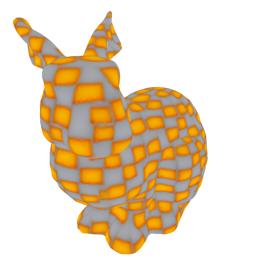}  &  \imgimg{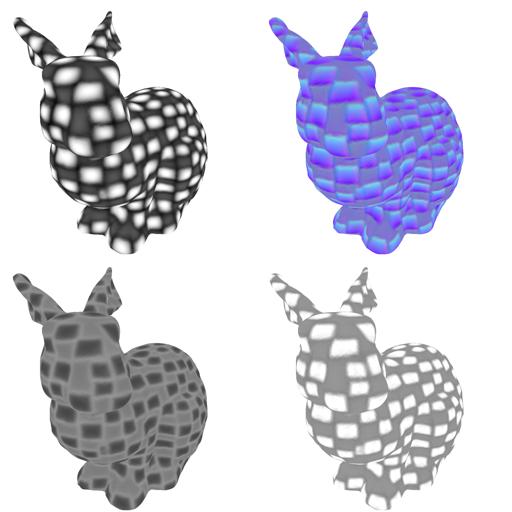}  &  \imgimg{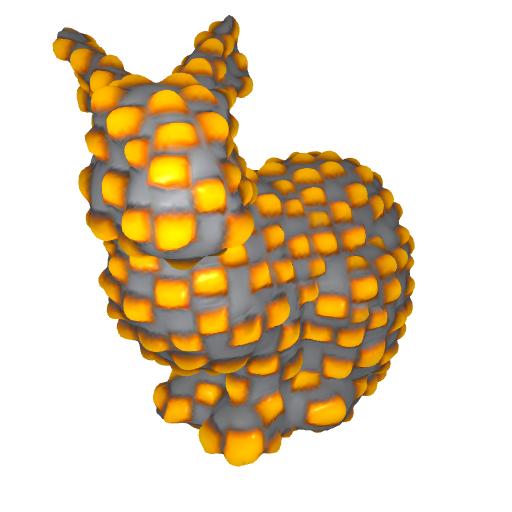} \\
\imgimg{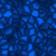}  &  \imgimg{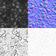}  &  \imgimg{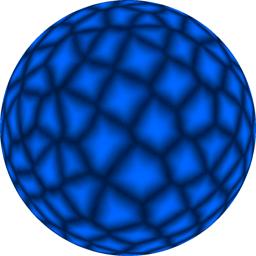}  &  \imgimg{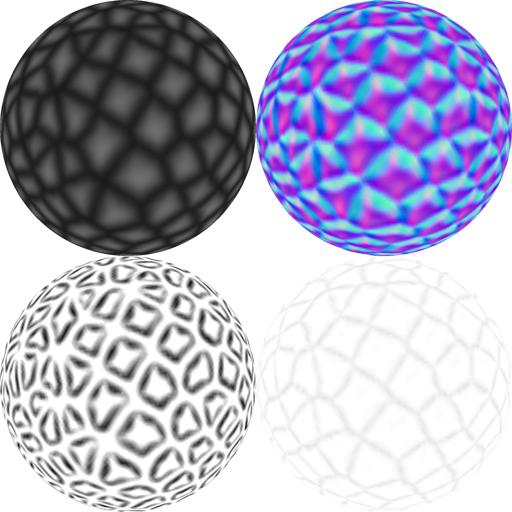}  &  \imgimg{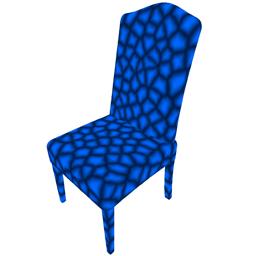}  &  \imgimg{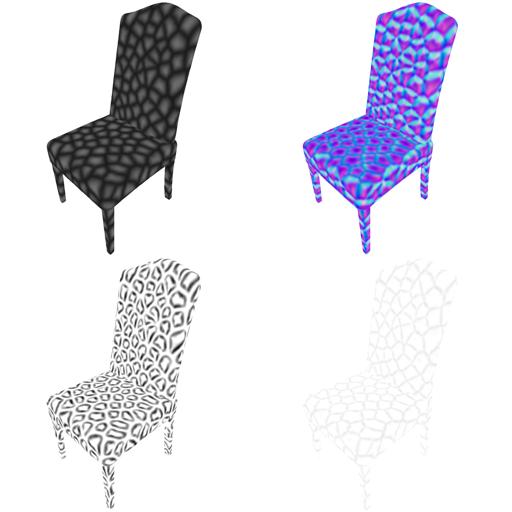}  &  \imgimg{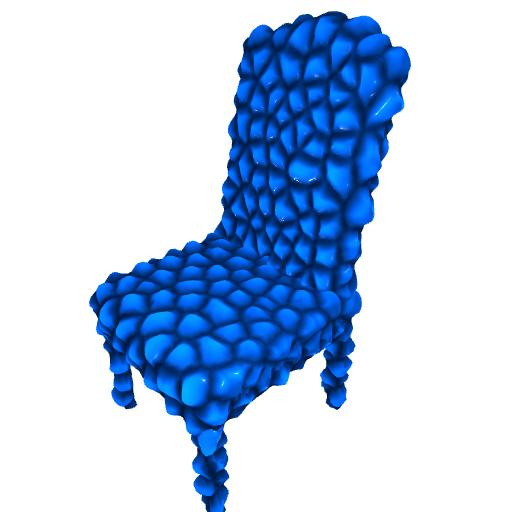} \\
\imgimg{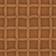}  &  \imgimg{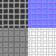}  &  \imgimg{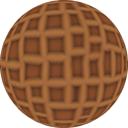}  &  \imgimg{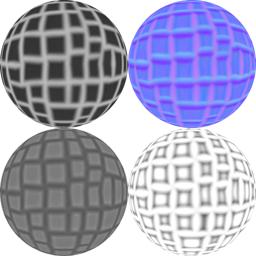}  &  \imgimg{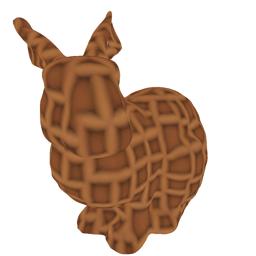}  &  \imgimg{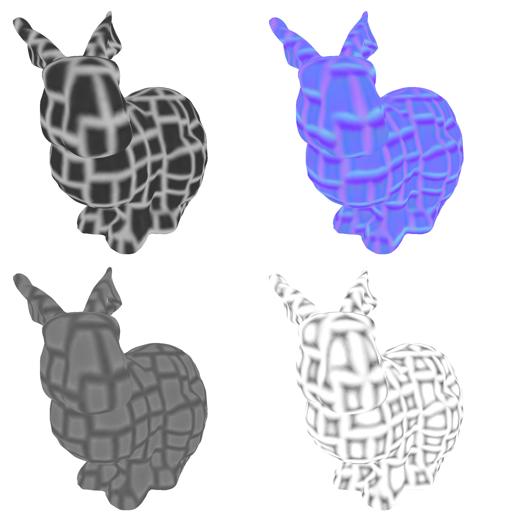}  &  \imgimg{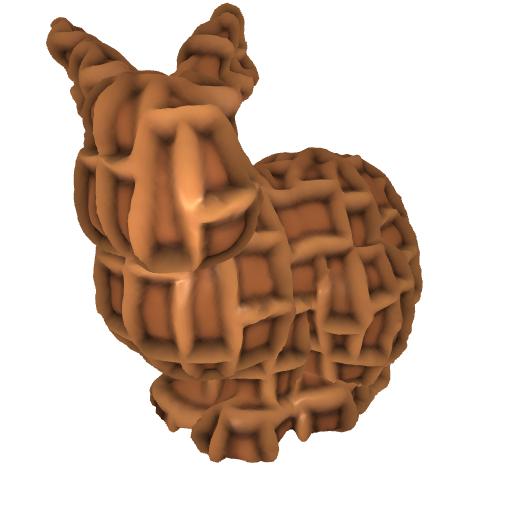} \\
\imgimg{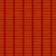}  &  \imgimg{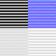}  &  \imgimg{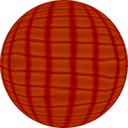}  &  \imgimg{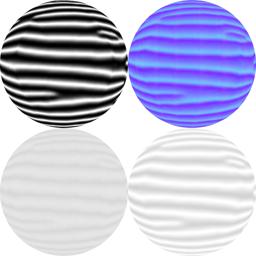}  &  \imgimg{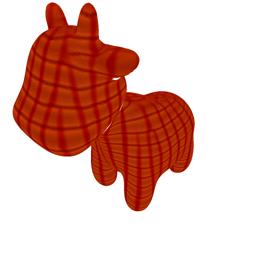}  &  \imgimg{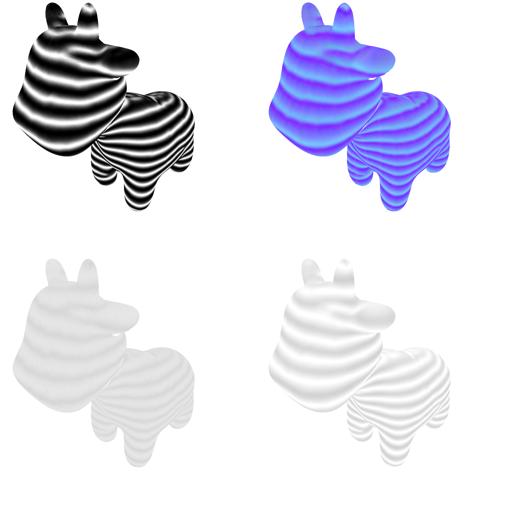}  &  \imgimg{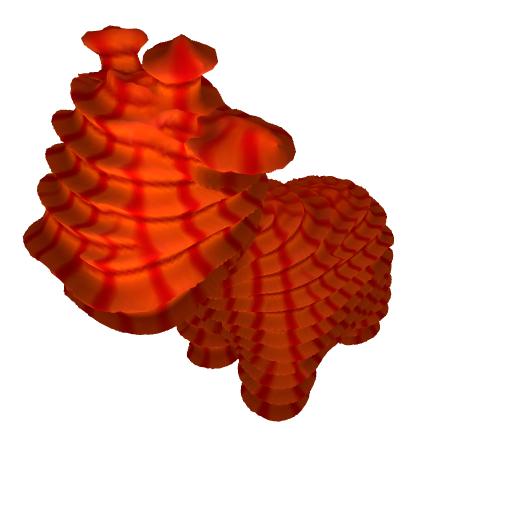}
\end{tabular}
}
\captionof{figure}{\rev{Results of \textbf{Image-guided} 3D texture synthesis. The attributes are height, normal, roughness, and ambient occlusion maps, from top to bottom, left to right. MeshNCA is trained only on the \icosphere mesh. At test time, it generalizes to unseen meshes with aligned texture maps. After rendering, the synthesized texture maps show correct shading effects.}}
\label{fig:image-synthesis-result}

\end{table*}

\subsection{Text-guided Synthesis}
\label{sec:exp-text-synthesis}
We use the ViT-B/32 CLIP model for supervising MeshNCA in the text-guided experiments. Unlike image-guided training, we use view-independent spherical harmonic lighting in our text-guided training and synthesis. This is because rendering a mesh with only ambient lighting hinders the CLIP model from understanding the 3D structure of the scene and evaluating it against a prompt that involves the object name. The rendered images are fed into the CLIP model to obtain the latent codes $\mathcal{R}$ in Equation~\ref{eq:clip-image-encoding}. The negative components are constructed following the description in Section~\ref{sec:text-guided-training}. All rendering parameters are shown in Table~\ref{tab:image-text-param}. For the text-guided experiments, we assign 3 channels of the cell states to the RGB color of the vertex and 1 channel to the geometric displacement of the vertex along the surface normal.
Despite not encountering any mesh other than \icosphere  during training, the synthesized geometric changes also generalize well to other meshes, confirming that MeshNCA learns the desired geometric texture instead of overfitting to the \icosphere  mesh.  Figure~\ref{fig:text-synthesis-result} shows the synthesized textures in our text-guided experiments. We refer the readers to our supplementary material and our demo for more results on text-guided texture synthesis, grafting, and videos. To enable grafting, we first train a MeshNCA using the prompt ``\textit{An image of a sphere made of colorful crochet}'' and then use the obtained model weights to initialize the training of all other models.

\newcolumntype{H}{>{\centering\arraybackslash} m{25pt} }

\newcommand{\imgtext}[1]{%
  \ifimgprefix
    \includegraphics[height=50pt]{figures/Experiments/text-synthesis/#1}%
  \else
    \includegraphics[height=50pt]{figures/Experiments/text-synthesis/L_#1}%
  \fi
}

\begin{table}[]
\resizebox{\linewidth}{!}{
\begin{tabular}{c||c||cc}
\multirow{2}{*}{\textbf{Prompts}} & \multirow{2}{*}{\textbf{Train}} & \multicolumn{2}{c}{\textbf{Test}}  \\
 & & \rev{Geometry} & \rev{+Color} \\
 \midrule
\rotatebox{90}{\parbox[c]{2cm}{\centering\hspace{0pt} \textbf{Feathers}}}  &  \imgtext{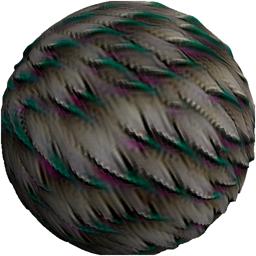}  &  \imgtext{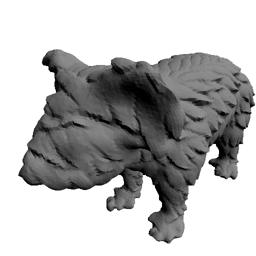} & \imgtext{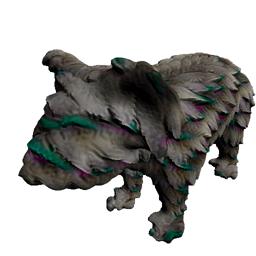} \\
\rotatebox{90}{\parbox[c]{2cm}{\centering\hspace{0pt} \textbf{Patchwork Leather}}}  &  \imgtext{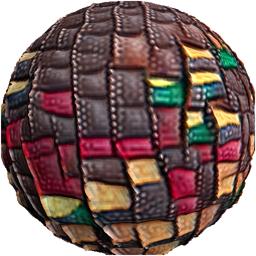}  &  \imgtext{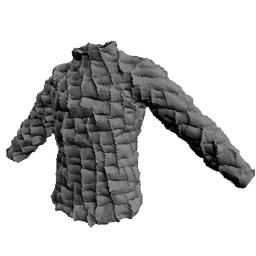} 
 & \imgtext{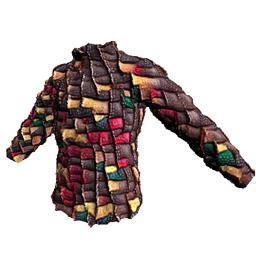} \\
\rotatebox{90}{\parbox[c]{2cm}{\centering\hspace{0pt} \textbf{Stained Glass}}}  &  \imgtext{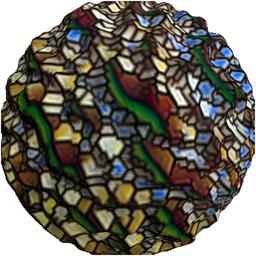}  & \imgtext{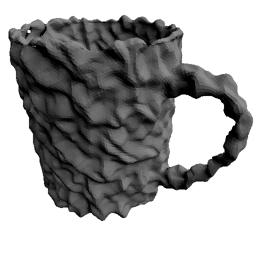}  & \imgtext{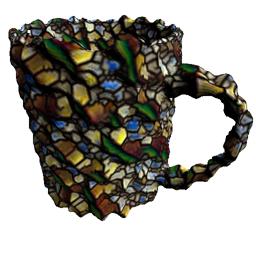}    \\
\rotatebox{90}{\parbox[c]{2cm}{\centering\hspace{0pt} \textbf{Bark}}}  &  \imgtext{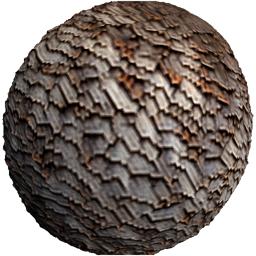}  &  \imgtext{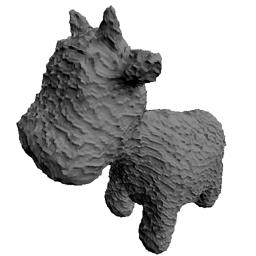} & \imgtext{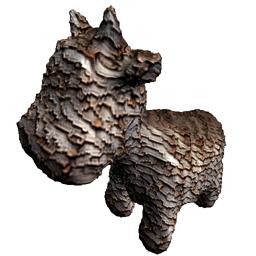} 
\end{tabular}
}
\captionof{figure}{\rev{Results of \textbf{CLIP-guided} 3D texture synthesis. MeshNCA is trained only on the \icosphere mesh. During test time, it generalizes to new meshes both in terms of color and geometric textures.}}
\label{fig:text-synthesis-result}
\end{table}

\subsection{3D Dynamic Texture Synthesis}
\label{sec:exp-motion}

\begin{table}[]
\caption{Hyperparameters used specifically in 3D dynamic texture synthesis experiments. $T, \gamma,  \lambda$ are described in Equation~\ref{eq:strength_loss}, \ref{eq:motion-loss}, and \ref{eq:dynamic-final-loss}, respectively.}
\centering
\begin{tabular}{cccc}
\textbf{Guidance} & $T$ & $\gamma$ & $\lambda$ \\
 \midrule
Image + Motion & 24.0 & 1.5 & 0.67 \\
Text + Motion & 10.0 & 0.15 & 0.67 \\
\end{tabular}
\label{tab:dyts-param}
\end{table}

We train MeshNCA to learn a user-given motion defined by a 3D vector field. In our experiments, we use eight manually designed vector fields as target motions, covering motions with different directions, sink and source motion patterns, and motions with spatially varying strength. The mathematical definition of these 8 motion vector fields is given in the supplementary material. The pre-trained optic-flow estimation network from \cite{two_stream, dynca} is used to predict the motion information given two consecutive frames. In each step of the training, We feed the network with two rendered images before and after MeshNCA iterations to estimate the optic flow on the surface. Additional parameters used for dynamic texture synthesis experiments are given in Table~\ref{tab:dyts-param}. Figure~\ref{fig:image-motion-result} and \ref{fig:text-motion-result} show the optical flow estimations to visualize the generated motions and their corresponding target motions, demonstrating the consistency of the motion on 3D surfaces. The 2D optical flow visualizations are obtained by placing the mesh center at the origin and placing the camera on the x-axis pointing towards the origin. The flow visualization follows \citet{flow-visualization}. The results in video format are available in our \textcolor{pinkl}{\href{https://meshnca.github.io/supplementary/}{demo supplementary}}.

\newcommand{\motionimg}[1]{%
  \ifimgprefix
    \includegraphics[height=60pt]{figures/Experiments/image-motion/#1}%
  \else
    \includegraphics[height=60pt]{figures/Experiments/image-motion/L_#1}%
  \fi
}

\begin{table*}[]
\centering
\resizebox{\textwidth}{!}{
\begin{tabular}{cc||c||c|ccc}
 \multicolumn{2}{c||}{\textbf{Target Dynamics}}  &  \textbf{Train}  &  \multicolumn{4}{c}{\textbf{Test-time Generalization}}  \\
 
 Vector Field & 2D Projections & Albedo OF & Attribute OF & Albedo &  Albedo OF & Attribute OF \\
 \midrule
\motionimg{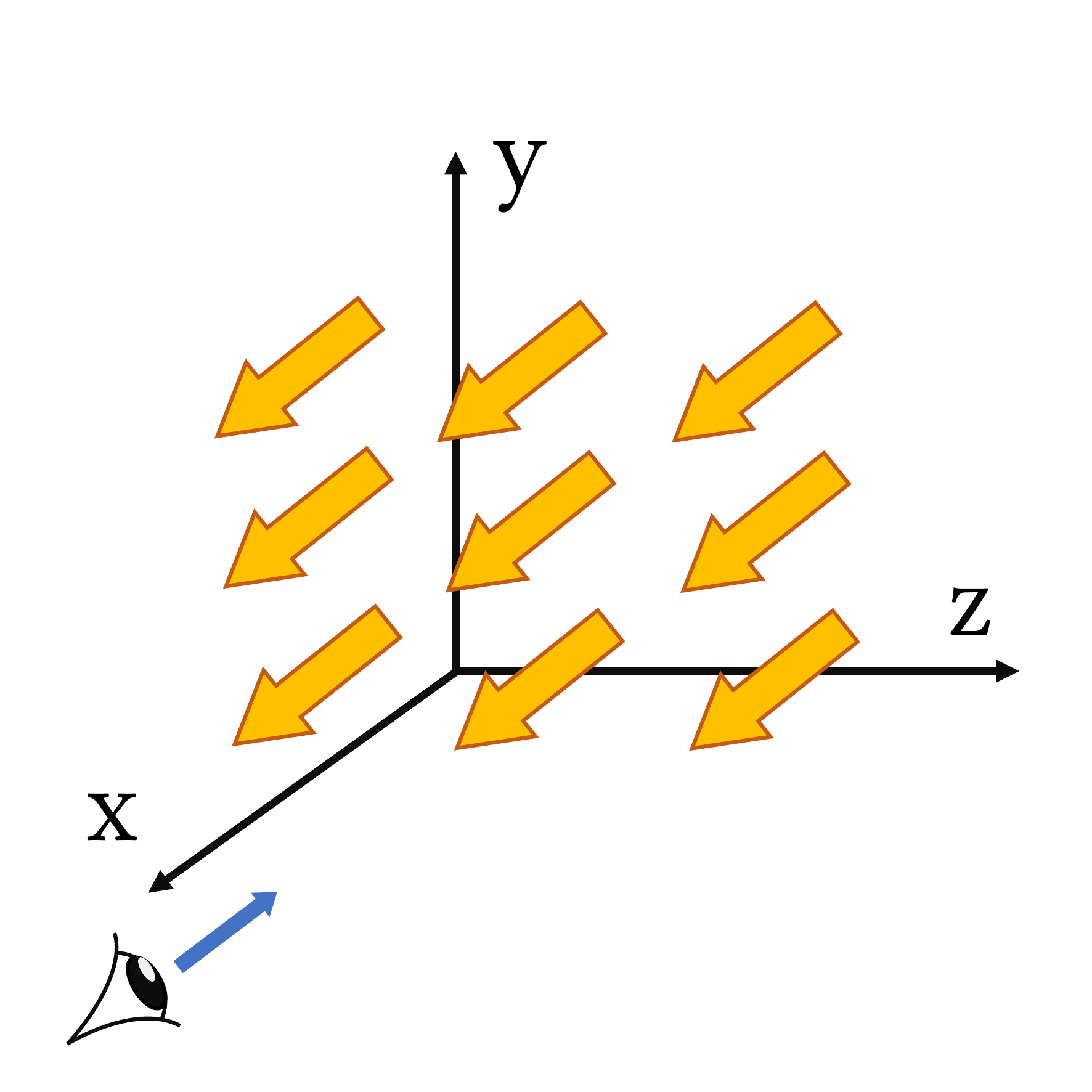}  &  \motionimg{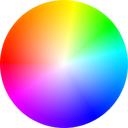}  &  \motionimg{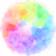}  &  \motionimg{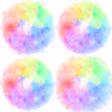}  &  \motionimg{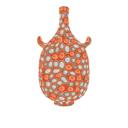}  &  \motionimg{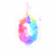}  &  \motionimg{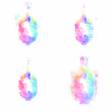} \\
\motionimg{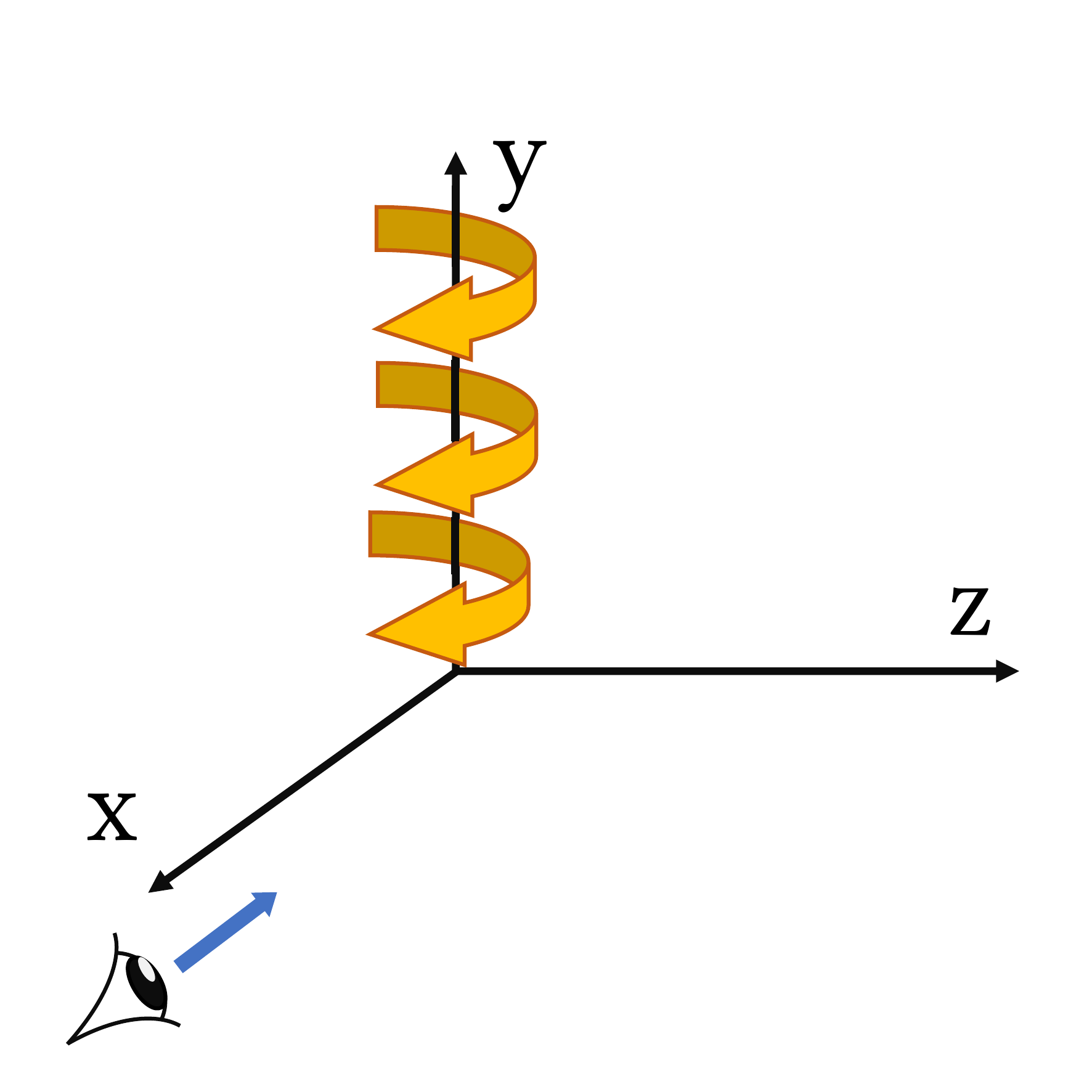}  &  \motionimg{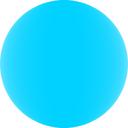}  &  \motionimg{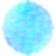}  &  \motionimg{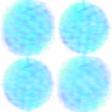}  &  \motionimg{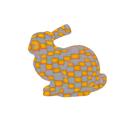}  &  \motionimg{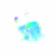}  &  \motionimg{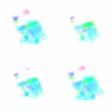} \\
\motionimg{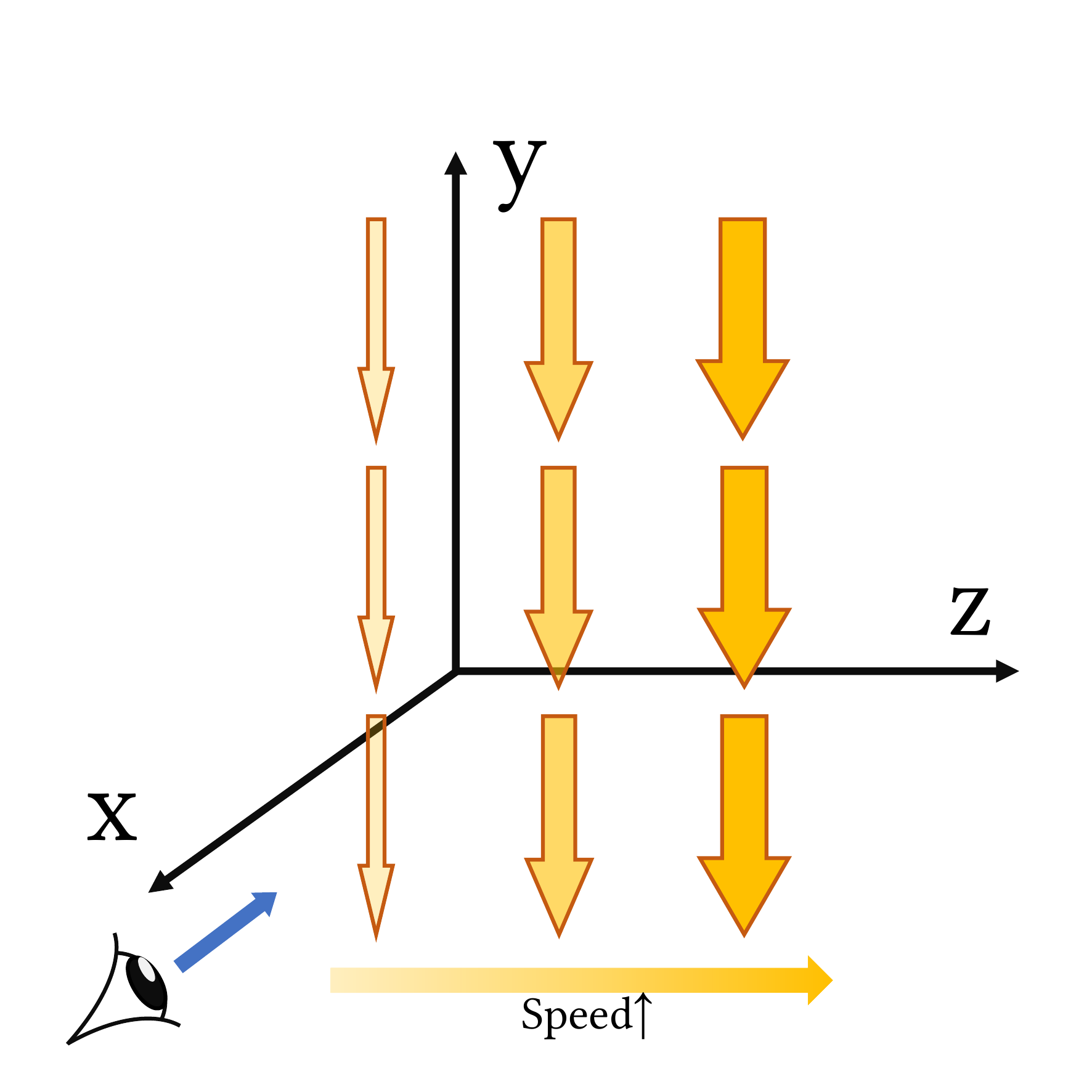}  &  \motionimg{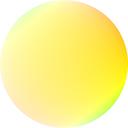}  &  \motionimg{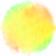}  &  \motionimg{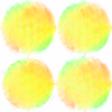}  &  \motionimg{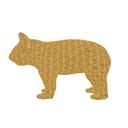}  &  \motionimg{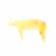}  &  \motionimg{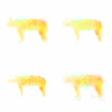}
\end{tabular}
}
\captionof{figure}{\rev{Results of dynamic texture synthesis with image targets. In all visualizations, the camera is fixed to looking at the negative $x$ direction from the positive $x$. The 2D projections of the 3D vector fields onto the camera plane are generated accordingly. During training, motion supervision is only imposed on the albedo channels. Since the channels are implicitly aligned in MeshNCA, all other attribute channels follow the same dynamic pattern during test time. Moreover, the learned motion on the \icosphere can also be generalized to unseen meshes. We highly encourage the readers to watch the corresponding videos available at the \textcolor{pinkl}{\href{https://meshnca.github.io/supplementary/Image+Motion/}{demo supplementary}} to see the dynamic textures.}}
\label{fig:image-motion-result}
\end{table*}

\newcommand{\motiontext}[1]{%
  \ifimgprefix
    \includegraphics[height=50pt]{figures/Experiments/text-motion/#1}%
  \else
    \includegraphics[height=50pt]{figures/Experiments/text-motion/L_#1}%
  \fi
}

\begin{table*}[]
\centering
\resizebox{\textwidth}{!}{
\begin{tabular}{c||cc||c||c|ccc}
 \multirow{2}{*}{\textbf{Prompts}} & \multicolumn{2}{c||}{\textbf{Target Dynamics}}  &  \textbf{Train}  &  \multicolumn{4}{c}{\textbf{Test-time Generalization}}  \\
 
 & 3D Vector Field & 2D Projections & Color OF & Geometry OF & Color  & Color OF & Geometry OF \\
 \midrule
\rotatebox{90}{\parbox[c]{2cm}{\centering\hspace{0pt} \textbf{Jelly Beans}}}  &  \motiontext{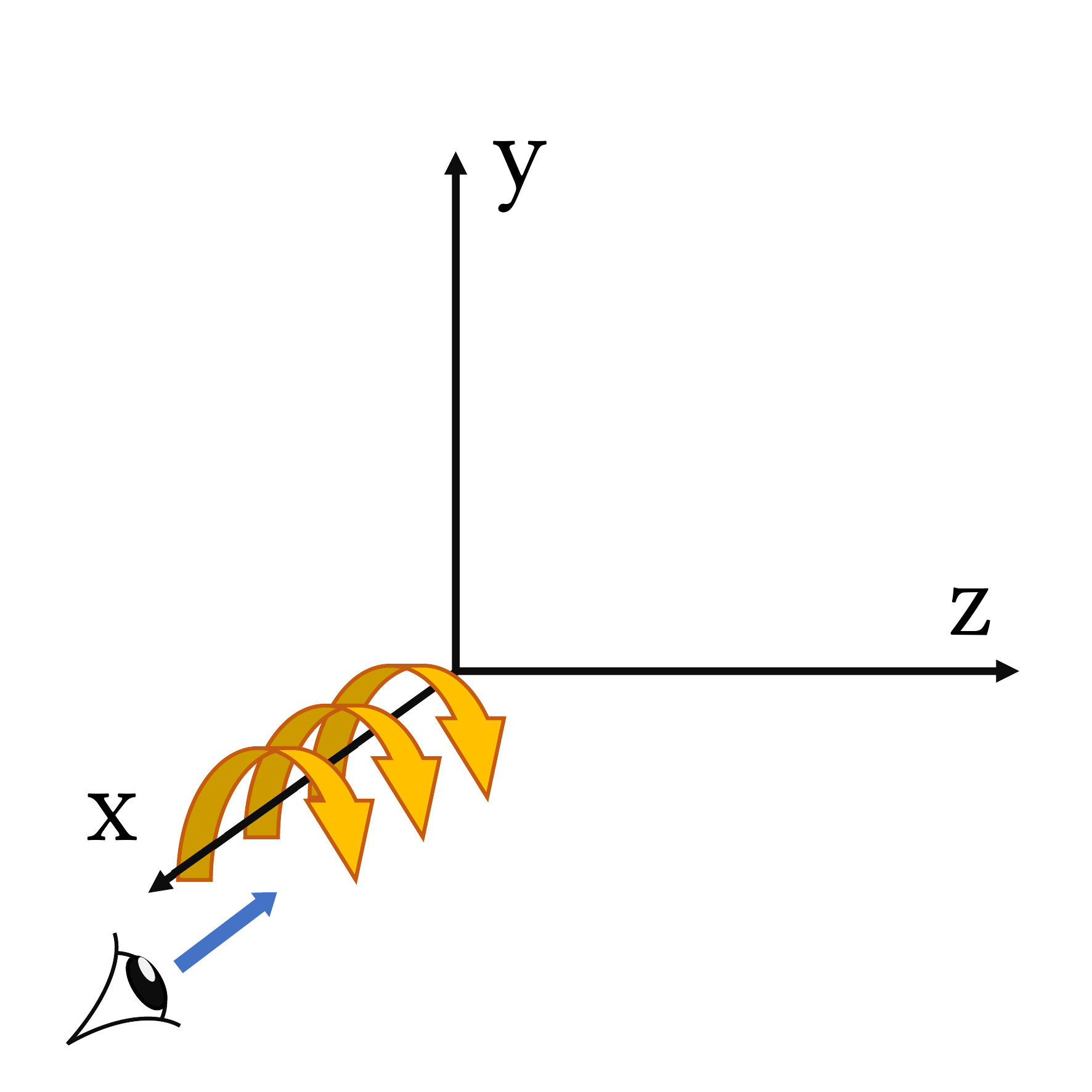}  &  \motiontext{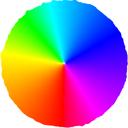}  &  \motiontext{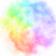}  &  \motiontext{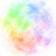}  &  \motiontext{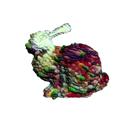}  &  \motiontext{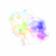}  &  \motiontext{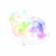} \\
\rotatebox{90}{\parbox[c]{2cm}{\centering\hspace{0pt} \textbf{Animal Fur}}}  &  \motiontext{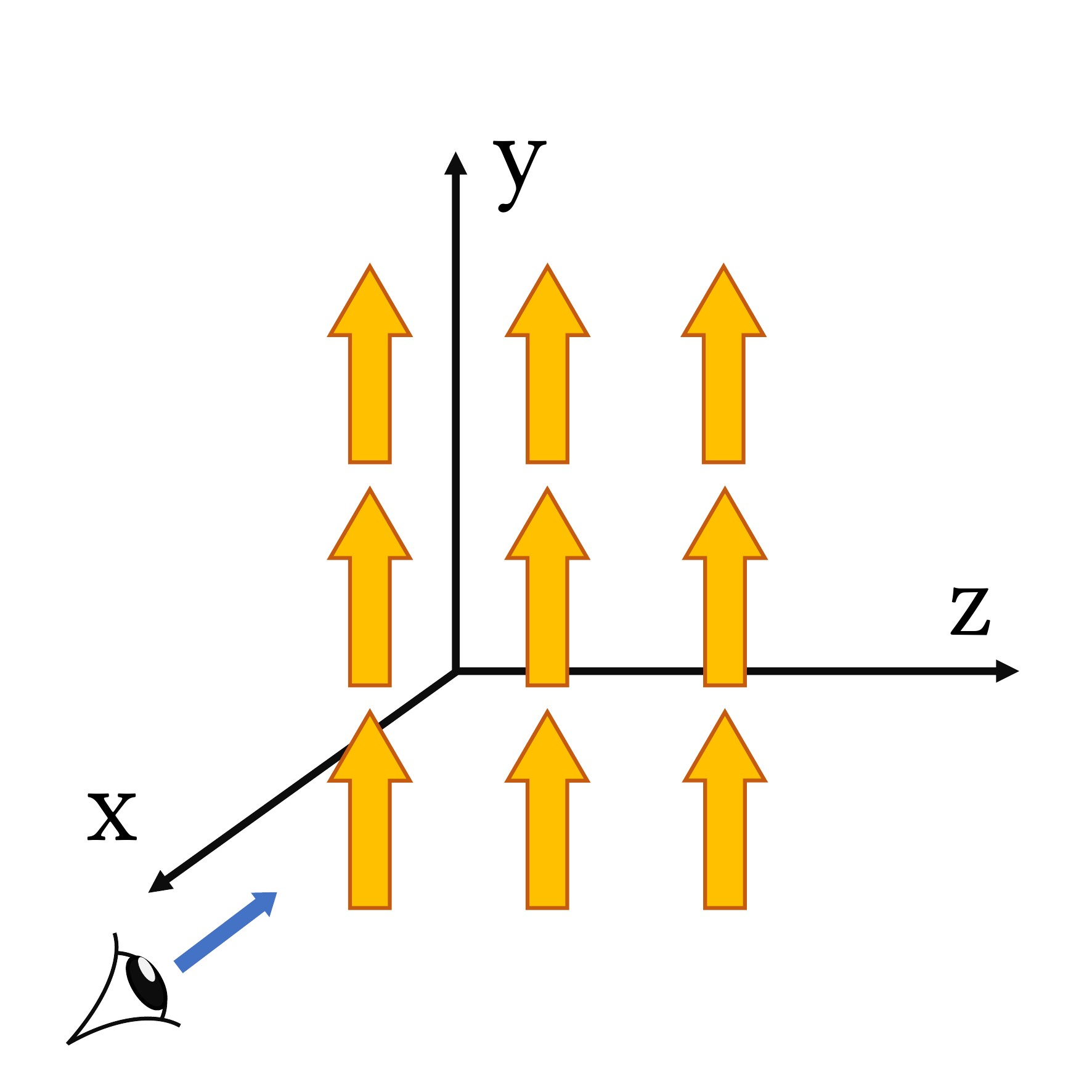}  &  \motiontext{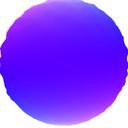}  &  \motiontext{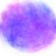}  &  \motiontext{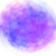}  &  \motiontext{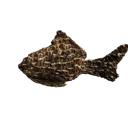}  &  \motiontext{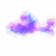}  &  \motiontext{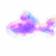} \\
\rotatebox{90}{\parbox[c]{2cm}{\centering\hspace{0pt} \textbf{Colorful Crochet}}}  &  \motiontext{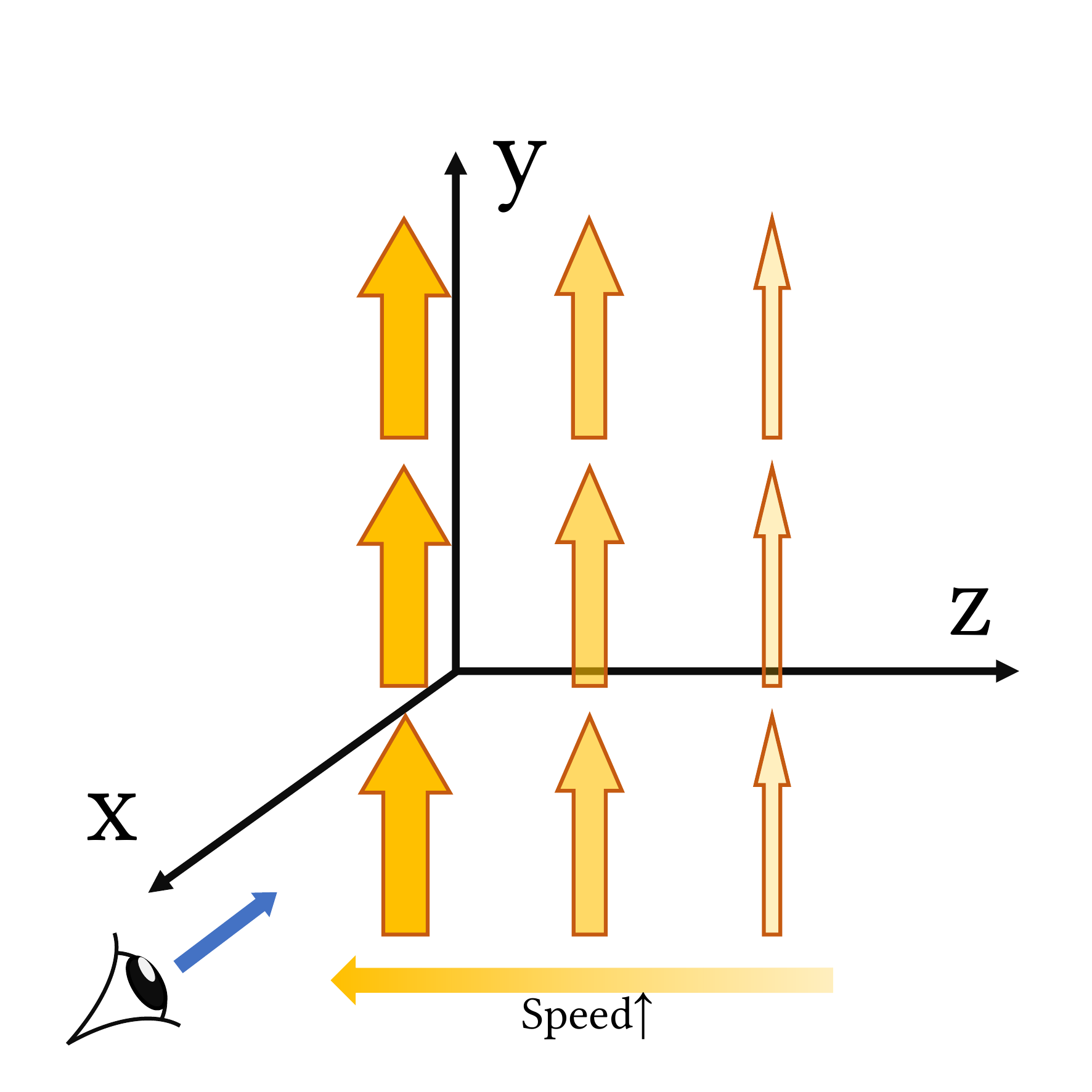}  &  \motiontext{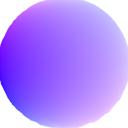}  &  \motiontext{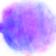}  &  \motiontext{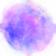}  &  \motiontext{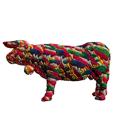}  &  \motiontext{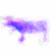}  &  \motiontext{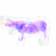}
\end{tabular}
}
\captionof{figure}{Results of dynamic texture synthesis under text guidance. In all visualizations, the camera always looks at the negative $x$ direction from the positive $x$. The seemingly noisy 2D projections are due to the change in mesh geometry. The geometric textures show consistent motion with the color textures, while only the color texture is supervised towards the target motion. The synthesized dynamic textures also seamlessly generalize to new meshes without any fine-tuning. We highly encourage readers to watch the videos released at the \textcolor{pinkl}{\href{https://meshnca.github.io/supplementary/Text+Motion/}{demo supplementary}} to see the dynamic textures.}
\label{fig:text-motion-result}
\end{table*}


\subsection{Grafting}
MeshNCA models trained using the method proposed in Section \ref{sec:grafting-training} can synthesize hybrid textures during test time. Figure \ref{fig:graft-result} shows the grafting results after image-guided and text-guided training. The hybrid texture is different from the linear interpolation of the two synthesized textures, as demonstrated in our ablation study on the grafting scheme in the supplementary material.

\newcommand{\graftimg}[1]{%
  \ifimgprefix
    \includegraphics[height=65pt]{figures/Experiments/graft-abl/#1}%
  \else
    \includegraphics[height=65pt]{figures/Experiments/graft-abl/#1}%
  \fi
}

\newcommand{\imgtextgraftmain}[1]{%
  \ifimgprefix
    \includegraphics[height=65pt]{figures/Experiments/supp/text-graft/#1}%
  \else
    \includegraphics[height=65pt]{figures/Experiments/supp/text-graft/#1}%
  \fi
}

\newcommand{\centeredtxtmain}[1]{
\begin{tabular}{l}
\parbox{2.0cm}{\vspace{-50pt} \centering \large #1}
\end{tabular}
}

\newcolumntype{H}{>{\centering\arraybackslash} m{1pt} }

\begin{table}[htbp]
\centering
\resizebox{\linewidth}{!}{
\begin{tabular}{Hc|c||c} 
& \textbf{Targets} & \textbf{Texture Mask} & \textbf{Results} \\
 \midrule
\multirow{2}{*}{\rotatebox{90}{\parbox[c]{0.77cm}{\hspace{0pt} \textbf{Albedo}}}}
 & \graftimg{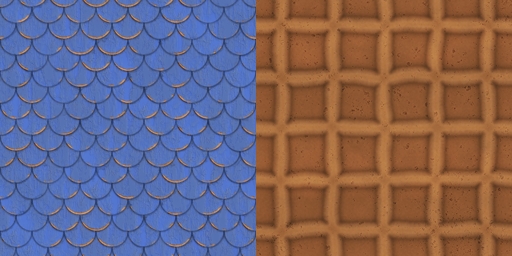} & \graftimg{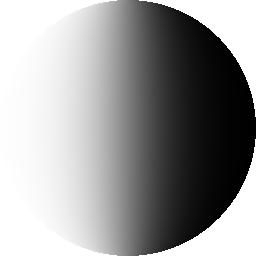}  & \graftimg{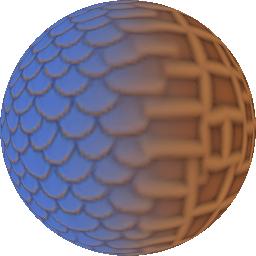} \\
& \graftimg{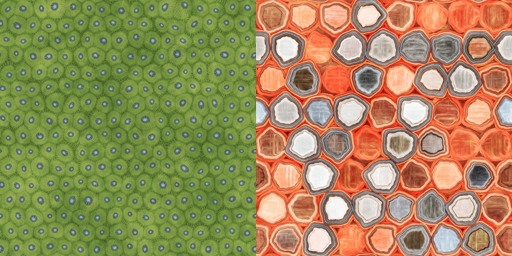} & \graftimg{mask.jpg}  & \graftimg{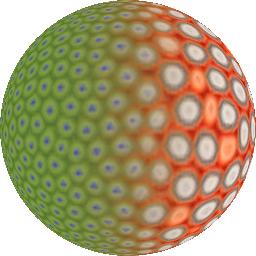} \\
\midrule
\multirow{2}{*}{\rotatebox{90}{\parbox[c]{1cm}{\centering\hspace{0pt} \textbf{Prompt}}}} & \centeredtxtmain{\textbf{Jelly Beans}} \hspace{-5pt}\centeredtxtmain{\textbf{Colorful Crochet}}  &  \graftimg{mask.jpg}  & \imgtextgraftmain{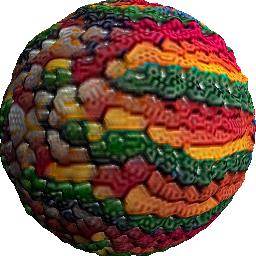} \\
& \centeredtxtmain{\textbf{Moss}}   \hspace{-5pt}\centeredtxtmain{\textbf{Cactus}}  &  \graftimg{mask.jpg}  & \imgtextgraftmain{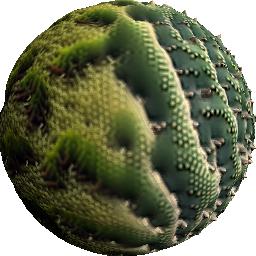} 
\end{tabular}
}
\captionof{figure}{Results of MeshNCA grafting. MeshNCA models are separately trained under the guidance of left and right targets. During testing, the two MeshNCA models cooperate to generate hybrid textures, which we term grafting.}
\label{fig:graft-result}
\end{table}


\section{Comparisons}
In this section, we perform qualitative and quantitative comparisons between MeshNCA and methods from different categories, outlined in Table~\ref{tab:method-compare}. Our comparative analysis comprises three aspects.  First, we compare MeshNCA with other methods in terms of the quality of the synthesized texture in the training domain. This comparison aims to compare the expressivity and diversity of the texture synthesized by different methods. Our second comparison targets the generalization capability of different methods and demonstrates MeshNCA's inherent generalizability to unseen meshes. Finally, we conduct user studies to quantitatively assess the quality and fidelity of the textures synthesized by different methods.

We exclude the methods of \cite{mesh2tex, face_graph_texture, texturify, texture-fields} from our comparisons, as the focus of MeshNCA is exemplar-based 3D texture synthesis, while these methods focus on dataset-based learning for texture synthesis and require a large dataset of textured meshes for training. We also exclude the methods that require manual parameter tuning, \cite{perlin_noise, solid_textures, cellular-texture-generation, 3d-surface-ca}, from our comparison as they cannot be trained to synthesize textures from image exemplars. 
Among the exemplar-based 3D texture synthesis methods, we could not find any publicly available code for the methods of \cite{turk2001texture, wei2001texture-neighbor-template, ying2001texture-uvmapping, progressively-variant-textures, Solid_from_2d}. 

From the remaining methods in Table~\ref{tab:method-compare}, we chose \cite{on-demand-solid-texture}  \footnote{as the representative of group \boxed{2} in Table~
\ref{tab:method-compare}, due to their higher quality results on 3D meshes}, \cite{diff-program-rdsystem}, and \cite{text2mesh} for comparing with image-guided 3D texture synthesis methods. Among the chosen methods, \cite{text2mesh} is not designed for image-guided texture synthesis. To ensure its competitiveness, we disable the geometry head in their model and use the texture loss in Section~\ref{sec:image-guided-training} to train it. \footnote{Although there is an image-guided mesh stylization method proposed in \cite{text2mesh}, it fails on most textures in our dataset, often resulting in outputs that are visually inconsistent with the intended texture quality and detail.} Figure~\ref{fig:image-comparison-train} shows an overview of the methods in our image-guided texture synthesis comparison. 
For text-guided texture synthesis, we chose \cite{text2mesh} and \cite{xmesh} for comparison. Figure~\ref{fig:text-comparison-train} shows an overview of the methods in our text-guided texture synthesis comparison. We refer the readers to our supplementary for more comparative results.

\begin{table}[]
\caption{\rev{Comparison of image-guided 3D texture synthesis methods.}}
\resizebox{\linewidth}{!}{
\begin{tabular}{c|ccccc}
\toprule
\multirow{ 2}{*}{\textbf{Method}}    &  \multirow{ 2}{*}{\textbf{\# Params}}  & \multicolumn{2}{c}{\textbf{Training}} & \textbf{Synthesis} & \multirow{2}{*}{\textbf{User Study}} \\
& & \textbf{Mem.} &\textbf{Time} & \textbf{Time} & \\
\midrule
\rev{\small{\citet{on-demand-solid-texture}}} & 85k & 1.5GB & 28min & 150ms & $10.9\% \pm 0.6\%$ \\ 
\rev{\small{\citet{diff-program-rdsystem}}} & 8k & 19GB & 27min & 280ms & $15.3\% \pm 0.7\%$ \\ 
\rev{\small{\citet{text2mesh}}} & 659k & 4GB & 4min & 130ms & $2.8\% \pm 0.3\%$ \\  
\textbf{MeshNCA (Ours)}   & 12k & 9GB & 16min & 180ms & $71.1\% \pm 0.9\%$ \\   
\bottomrule
\end{tabular}
}
\label{tab:image-comparison}
\end{table}

\begin{table}[]
\caption{\rev{Comparison of text-guided 3D texture synthesis methods.}}
\resizebox{\linewidth}{!}{
\begin{tabular}{c|ccccc}
\toprule
\multirow{ 2}{*}{\textbf{Method}}    &  \multirow{ 2}{*}{\textbf{\# Params}}  & \multicolumn{2}{c}{\textbf{Training}} & \textbf{Synthesis} &  \multirow{2}{*}{\textbf{User Study}} \\
& & \textbf{Mem.} & \textbf{Time} & \textbf{Time} &  \\
\midrule
\rev{\small{\citet{text2mesh}}} & 0.66M & 4GB & 13min/mesh & 150ms & $17.3\% \pm 0.7\%$ \\
\rev{\small{\citet{xmesh}}} & 9.61M & 5GB & 11min/mesh & 150ms & $18.7\% \pm 0.7\%$\\ 
\textbf{MeshNCA (Ours)} & 0.01M & 8GB & 105min & 180ms & $64.0\% \pm 0.9\%$ \\   
\bottomrule
\end{tabular}
}
\label{tab:text-comparison}
\end{table}

\subsection{Synthesis in the training domain}
\label{sec:exp-direct-compare}

In this section, we compare the quality of synthesized textures by each method in their corresponding training domain. MeshNCA's training domain is a sphere mesh. The method of \cite{diff-program-rdsystem} trains a reaction-diffusion model in the 2D domain, which is then lifted into 3D. For Text2Mesh \cite{text2mesh} and X-Mesh \cite{xmesh}, we choose the same sphere mesh in MeshNCA as their training domain.
Lastly, the solid texturing method in \cite{on-demand-solid-texture} uses a cube as its training domain. 

\subsubsection{\textbf{Image-guided Texture Synthesis}}

Figure ~\ref{fig:image-comparison-train} compares the MeshNCA results with \cite{diff-program-rdsystem, text2mesh, on-demand-solid-texture}, trained on six different target textures. All methods solely utilize the albedo map as their training target. As shown in Figure~\ref{fig:image-comparison-train}, our synthesized textures are more faithful to the target texture in terms of both color and structure. The other methods synthesize textures with either the wrong structure or color. Moreover, some of their results contain undesired high-frequency noise. This proves that MeshNCA, despite having very few parameters, is an expressive model and can synthesize a diverse set of textures.
We refer the reader to the supplementary for comparison results on more textures. 

\subsubsection{\textbf{Text-guided Texture Synthesis}}

Figure ~\ref{fig:text-comparison-train} compares our results with Text2Mesh \cite{text2mesh} and X-Mesh \cite{xmesh} on 6 different target prompts. We refer the reader to the supplementary for comparison results on more prompts. Our results appear to be more coherent and less noisy than the other two methods. Moreover, the other two methods contain spike geometry or non-texture artifacts, while our results adhere to the texture described by the prompt. This result signifies the importance of the inductive bias introduced by our local cell communication scheme and the advantage of directional CLIP loss in text-guided 3D texture synthesis. The necessity of using the directional CLIP loss is shown in the ablation study in our supplementary material.





\newcommand{\imagedir}[1]{%
  \ifimgprefix
    \includegraphics[height=60pt]{figures/Experiments/comparison/image-direct/#1}%
  \else
    \includegraphics[height=60pt]{figures/Experiments/comparison/image-direct/#1}%
  \fi
}

\newcommand{\imagegen}[1]{%
  \ifimgprefix
    \includegraphics[height=60pt]{figures/Experiments/comparison/image-gen/#1}%
  \else
    \includegraphics[height=60pt]{figures/Experiments/comparison/image-gen/#1}%
  \fi
}

\newcommand{\textdir}[1]{%
  \ifimgprefix
    \includegraphics[height=50pt]{figures/Experiments/comparison/text-direct/#1}%
  \else
    \includegraphics[height=50pt]{figures/Experiments/comparison/text-direct/#1}%
  \fi
}

\newcommand{\textgen}[1]{%
  \ifimgprefix
    \includegraphics[height=50pt]{figures/Experiments/comparison/text-gen/#1}%
  \else
    \includegraphics[height=50pt]{figures/Experiments/comparison/text-gen/#1}%
  \fi
}

\newcommand{\rowdircomp}[1]{
\imagedir{#1.jpg} & 
\imagedir{meshnca_#1_sphere_6.jpg} & 
\imagedir{text2mesh_#1_sphere_6.jpg} & 
\imagedir{diffrd_#1_train.jpg} 
& \imagedir{ondemand_#1_cube.jpg}
}

\newcommand{\rowgencomp}[2]{
\imagedir{#1.jpg} & 
\imagegen{meshnca_#1_#2.jpg} & 
\imagegen{text2mesh_#1_#2.jpg} & 
\imagegen{diffrd_#1_#2.jpg} &
\imagegen{ondemand_#1_#2.jpg}
}

\begin{table*}[]
\resizebox{\textwidth}{!}{%
\begin{tabular}{c||cccc}
 \multirow{2}{*}{\textbf{Target}}  &  \multicolumn{4}{c}{\textbf{Methods}}  \\
 
 & Ours & \rev{\small{\citet{text2mesh}}} & \rev{\small{\citet{diff-program-rdsystem}}} & \rev{\small{\citet{on-demand-solid-texture}}} \\
 \midrule

\rowdircomp{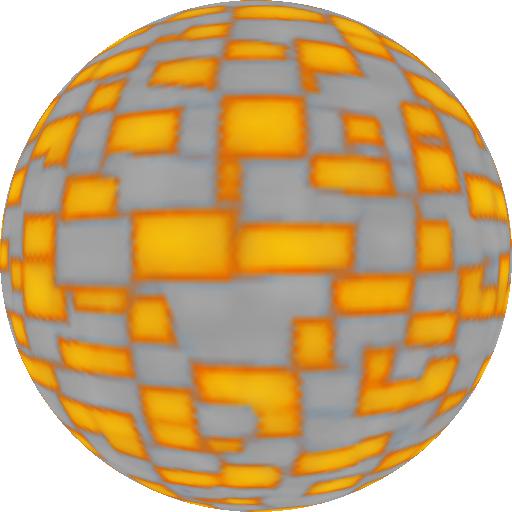} \\ 
\rowgencomp{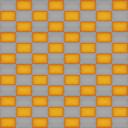}{bunny} \\ 

\rowdircomp{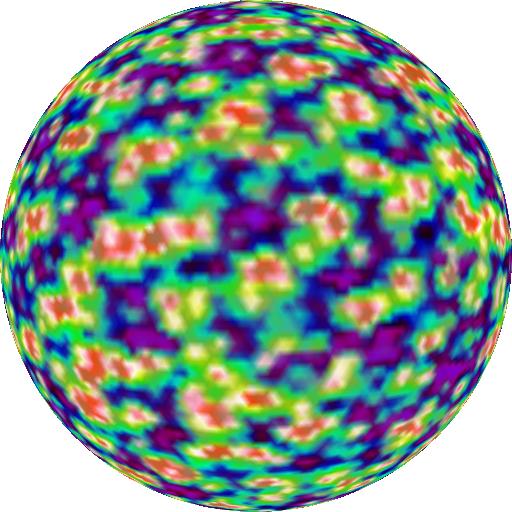} \\ 
\rowgencomp{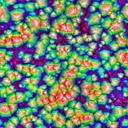}{armadillo} \\ 

\rowdircomp{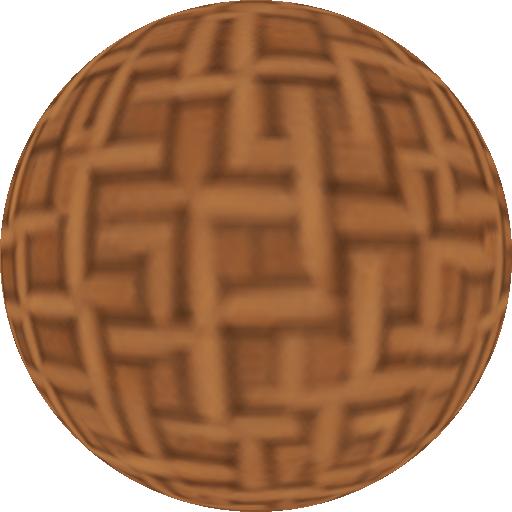} \\ 
\rowgencomp{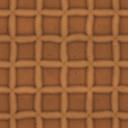}{springer} \\ 

\end{tabular}
}
\captionof{figure}{\rev{Comparison on texture quality in \textbf{Image-guided} synthesis between MeshNCA (Ours), \citet{text2mesh}, \citet{diff-program-rdsystem}, and \citet{on-demand-solid-texture} in the training domain and generalization. The method of \citet{text2mesh} fails to preserve patterns during generalization. The method of \citet{diff-program-rdsystem}, due to its isotropic nature, is not capable of synthesizing textures with correct structure in most cases. The method of \citet{on-demand-solid-texture} does not capture the essence of most textures and generates textures either with the wrong color or structure. In comparison, our method synthesizes the most faithful textures to the target ones.}}
\label{fig:image-comparison-train}
\end{table*}

\newcommand{\rowtextdir}[1]{
\textdir{meshnca_#1.jpg} & \textdir{text2mesh_#1.jpg} & \textdir{xmesh_#1.jpg} 
}

\newcommand{\rowtextgencomp}[2]{
\textgen{meshnca_#1_#2.jpg} & 
\textgen{text2mesh_#1_#2.jpg} & 
\textgen{xmesh_#1_#2.jpg} 
}

\newcommand{\textgenspec}[1]{%
  \ifimgprefix
    \includegraphics[height=50pt]{figures/Experiments/comparison/text-specific/#1}%
  \else
    \includegraphics[height=50pt]{figures/Experiments/comparison/text-specific/#1}%
  \fi
}

\newcommand{\rowtextgencompspec}[2]{
\textgen{meshnca_#1_#2.jpg} & 
\textgenspec{text2mesh_#1_#2.jpg} & 
\textgenspec{xmesh_#1_#2.jpg} 
}

 






\begin{table*}[]
\resizebox{\linewidth}{!}{%
\begin{tabular}{c|ccc||ccc}

\multirow{2}{*}{\textbf{Prompts}} & \multicolumn{3}{c||}{\textbf{Training Domain}} & \multicolumn{3}{c}{\textbf{Generalization}} \\
& Ours & \rev{\small{\citet{text2mesh}}} & \rev{\small{\citet{xmesh}}} & Ours & \rev{\small{\citet{text2mesh}}} & \rev{\small{\citet{xmesh}}}  \\
\midrule
\rotatebox{90}{\parbox[c]{2cm}{\centering\hspace{0pt} \textbf{Moss}}} & \rowtextdir{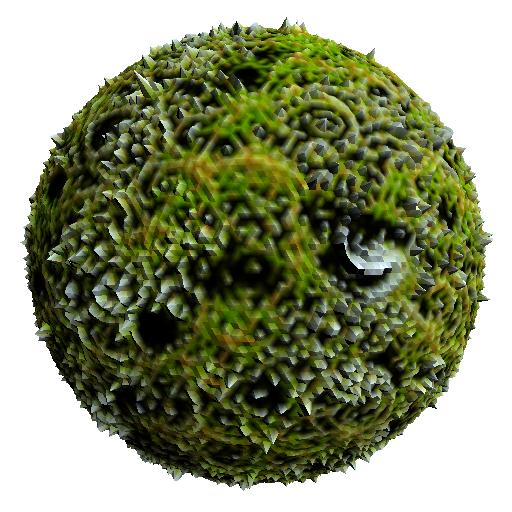} &  \rowtextgencomp{moss}{mug} \\
\rotatebox{90}{\parbox[c]{2cm}{\centering\hspace{0pt} \textbf{Colorful Crochet}}} & \rowtextdir{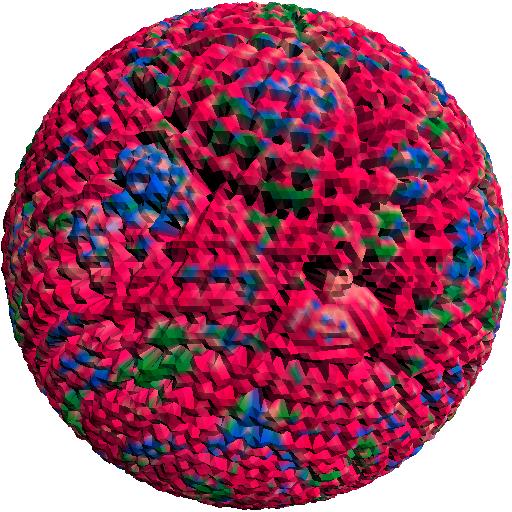} &  \rowtextgencomp{colorful_crochet}{chair} \\
\rotatebox{90}{\parbox[c]{2cm}{\centering\hspace{0pt} \textbf{Jelly Beans}}} & \rowtextdir{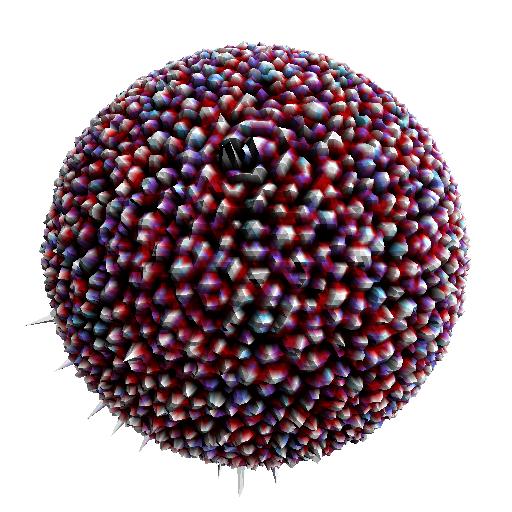} & \rowtextgencomp{jelly_beans}{bunny}  \\
\end{tabular}
}
\captionof{figure}{\rev{Comparison of texture quality in \textbf{Text-guided} synthesis between MeshNCA (Ours), Text2Mesh \cite{text2mesh}, and X-Mesh \cite{xmesh}. Both methods by \citet{text2mesh} and \citet{xmesh} suffer from high-frequency noisy patterns. Text2Mesh contains spike geometry noise, while X-Mesh includes non-texture artifacts, such as the human head in \textit{jelly beans} prompt. On the contrary, our results are more coherent and less noisy than them. Moreover, Text2Mesh and X-Mesh possess little generalization ability, while MeshNCA faithfully preserves the pattern generated in the training domain.}}
\label{fig:text-comparison-train}
\end{table*}

\subsection{Generalization to unseen meshes}
In this section, we conduct experiments to evaluate the generalizability of different methods to unseen meshes. All the methods are trained as described in Section \ref{sec:exp-direct-compare}. For the solid texturing method in \cite{on-demand-solid-texture}, we generate a cuboid texture volume using the pre-trained model and then use trilinear interpolation to query the color from this volume to render the textured mesh. For all other methods, we load the pre-trained models and perform the test on unseen meshes. 

\subsubsection{\textbf{Image-guided Texture Synthesis}}

Figure \ref{fig:image-comparison-train} compares the generalization results of MeshNCA with \cite{diff-program-rdsystem, text2mesh, on-demand-solid-texture} on the same textures used in Section \ref{sec:exp-direct-compare} to unseen meshes. The method of \cite{text2mesh} can only preserve the rough color of the target texture and fails to capture the structure of the patterns. The differentiable reaction-diffusion system \cite{diff-program-rdsystem} generates the same pattern as in its training domain. However, due to its isotropic nature, it cannot synthesize the desired texture in most cases. The solid texture synthesis method \cite{on-demand-solid-texture} inherits all patterns from the cuboid volume, but fails to synthesize high-fidelity textures. MeshNCA, on the contrary, generalizes to unseen meshes seamlessly without compromising the quality of the synthesized textures. 

\subsubsection{\textbf{Text-guided Texture Synthesis}}

For the evaluation of generalization capabilities in text-guided synthesis, we present all the results on unseen meshes in Figure \ref{fig:text-comparison-train}. Analogous to the previously discussed analysis of the generalization ability of \cite{text2mesh}, both methods exhibit limited adaptability to new meshes, often resulting in unwanted color blobs or degenerated noisy patterns.

\subsection{User Study}
\rev{Our user study aims to quantify and compare the quality and fidelity of the synthesized textures by different methods.
We use the Amazon Mechanical Turk (AMT) platform and conduct two user studies to quantitatively evaluate the image-guided and text-guided texture synthesis capabilities of the aforementioned methods. We set a minimum requirement on the worker qualifications to avoid inexperienced and untrustworthy participants on AMT. We provide samples of our questionnaire forms at the \textcolor{pinkl}{\href{https://meshnca.github.io/supplementary/UserStudy/}{demo supplementary}}. The results are provided in the last column of Table~\ref{tab:image-comparison} and \ref{tab:text-comparison}. Both results show the percentage of times each method was chosen by the participants, demonstrating the higher performance of MeshNCA compared to the baseline methods in generalizability to unseen meshes and texture quality. It is worth noticing that for \textbf{Text-guided} synthesis, MeshNCA does not even use mesh-prompt tailored training but merely generalizes to unseen meshes. Nevertheless, it still outperforms current state-of-the-art methods. This result further demonstrates a drawback of the previously proposed metric in \cite{xmesh} to quantify the quality of text-guided 3D texture synthesis. We refer the readers to our supplementary material for details of the metric and a full description of the user study.}





\section{Limitations}
MeshNCA suffers from certain limitations. Although the different channels of cell states share similar structures, misalignment between different texture maps can occur on a few textures, leading to wrong shading effects. This failure is likely due to the implicit supervision of the structural alignment of different texture maps. 
Moreover, since we assign texture attributes to the vertices of the mesh, the quality of the synthesized 3D texture depends on the input mesh quality and the number of vertices. To achieve high quality, our model requires high poly meshes with uniformly distributed vertices on the surface with an average valence close to 6.0.  
Furthermore, MeshNCA is sensitive to certain rendering parameters, and variations in aspects such as rendering resolution and camera distance can influence the training outcome to some extent. Consequently, the framework might yield sub-optimal results if these parameters are not properly adjusted. Finally, MeshNCA struggles to synthesize semantic textures as it is agnostic to the underlying semantic structures of the input 3D mesh, e.g., the distinction between head and arms in a human body mesh. We refer the reader to our supplementary material for a visual demonstration of these limitations.

\section{Future Work}

Expanding MeshNCA to operate on mesh faces instead of vertices can reduce the need for high-poly meshes. Moreover, our training scheme can be adjusted in various ways to enhance the quality and fidelity of synthesized textures. For instance, using an orthographic projection, as opposed to a perspective projection, can facilitate training and enhance the quality of the results by allowing the synthesized patterns to be passed to the loss function without distortion. Additionally, alternative training methods, such as the axis-aligned plane scheme discussed in \cite{Solid_from_2d} or the slice-based scheme from \cite{on-demand-solid-texture}, could be considered to potentially enhance training speed and convergence. Furthermore, incorporating a physically-based renderer in the training, similar to \cite{single-svbrdf} and \cite{tango}, could enhance the quality and fidelity of the synthesized textures. Finally, a subsequent development could involve creating NCA models capable of operating on particle systems to enable real-time point cloud and fluid stylization.

\section{Conclusion}
\label{sec:conclusion}

We propose MeshNCA, a Cellular-Automata-based model for real-time and interactive 3D texture synthesis on meshes. Our model can generalize to any mesh at inference time while being trained only on a \icosphere mesh. By employing a message-passing scheme, our proposed spherical-harmonics-based filters extend the convolution-based perception filters of conventional NCA to accommodate more general non-grid structures such as meshes. We conceptualize a procedure to seamlessly graft multiple MeshNCA models, enabling the synthesis of hybrid textures that exhibit characteristics of two texture instances. Our model can be trained with both image targets and text prompt guidance to synthesize the desired pattern, broadening its range of applications. Moreover, we demonstrate that MeshNCA, equipped with our proposed Motion Positional Encoding, can be trained with motion supervision to synthesize dynamic 3D textures that generalize across meshes. Finally, MeshNCA exhibits several unique test-time properties, including texture direction/orientation control and texture grafting/regeneration, all of which can be seamlessly executed in real time on consumer devices through our online interactive demo.  

\begin{acks} 
This work was supported in part by the Swiss National Science Foundation via the Sinergia grant CRSII5-180359.
\end{acks}





\clearpage
\clearpage

\appendix

\noindent\Huge{\textbf{Supplementary Material}}

\normalsize

\section*{Table of Contents}
\startcontents[sections]
\printcontents[sections]{}{1}{}

\section{Online Demo}
\label{sec:demo}

All cells (vertices) follow the same local update rule at each step of MeshNCA's forward pass. Therefore, the update rule for all cells can be executed in parallel, and the forward pass of MeshNCA can be implemented efficiently in any rasterization pipeline to benefit from high-performance computing on GPUs. We choose WebGL for our implementation because of its cross-platform compatibility. We deploy the pre-trained models from the image-guided experiments in our online demo which is accessible at \textcolor{pinkl}{\href{https://meshnca.github.io/}{\textbf{https://meshnca.github.io/}}}. 

In the following sections, we elaborate on the user interface (UI) and the implementation details of our demo.

\subsection{User Interface}

\begin{figure*}
    \centering
    \includegraphics[width=\textwidth, trim ={0 425 205 0},clip]{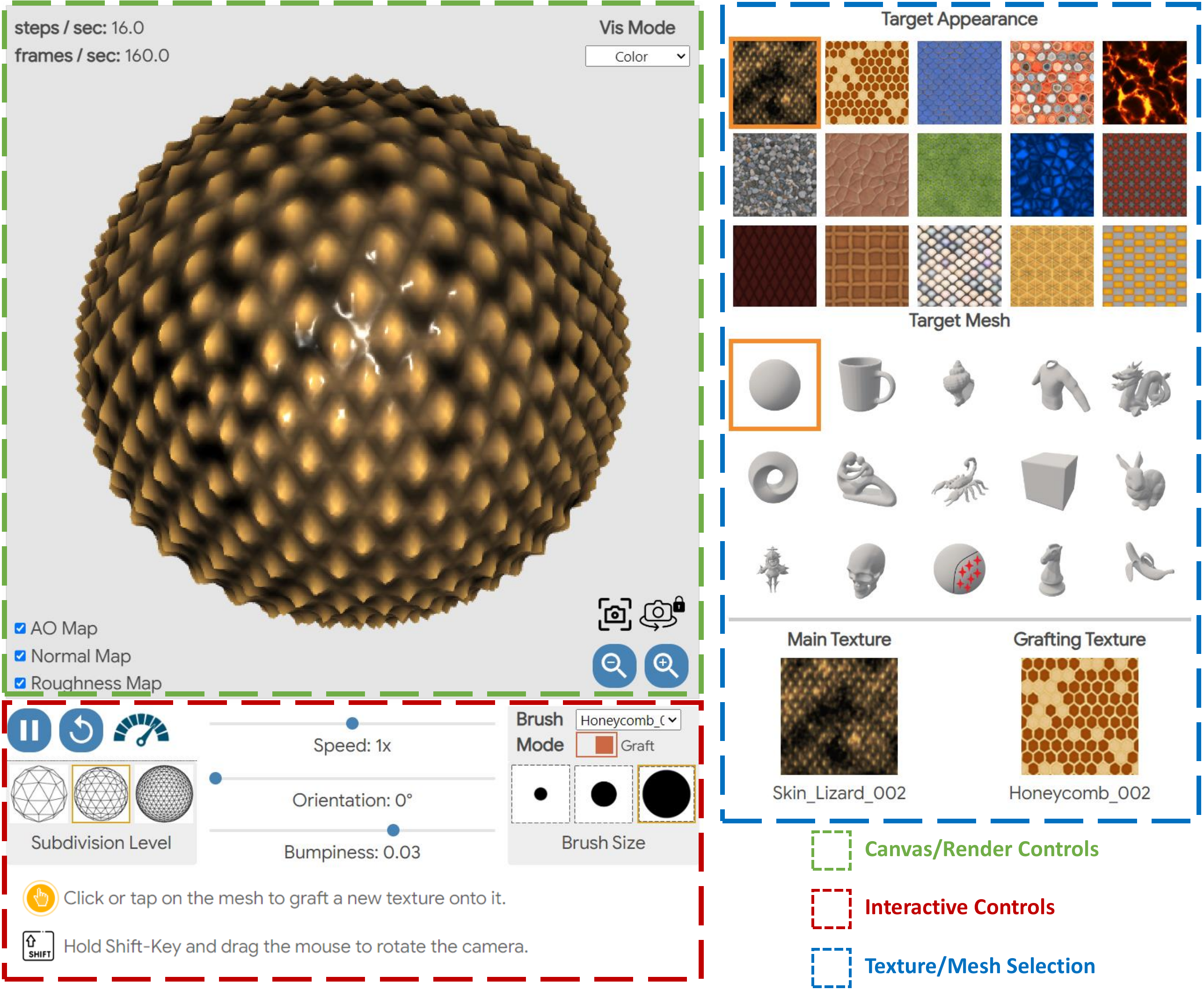}
    \caption{User interface (UI) of MeshNCA demo.  \textbf{\textcolor[HTML]{70AD47}{Canvas and Render Controls}} section displays the stylized mesh and provides basic shading controls. \textbf{\textcolor[HTML]{C00000}{Interactive Controls}} panel allows the users to interact with and control the synthesized texture in different ways \textbf{\textcolor[HTML]{0070C0}{Texture and Mesh Selection}} panels allow the users to choose between 72 different textures and 47 different meshes. 
    }
    \label{fig:demo_ui}
\end{figure*}

Figure~\ref{fig:demo_ui} shows the user interface of our online demo. Here, we describe the role of each element in our UI. There are 3 main parts in our user interface: 1- Canvas and Render Controls, 2- MeshNCA Interactive Controls, and 3- Texture and Mesh Selection.  


\subsubsection{\textbf{Canvas and Render Controls}}

The canvas displays the synthesized texture on the target mesh. We also provide a few options for the users to control the rendering shader.
\begin{itemize}[leftmargin=*]
    \item On the upper left part of the canvas, we show the number of MeshNCA steps per second \textbf{steps / sec}, controlled using the speed slider, and the number of rendered frames per second (fps), \textbf{frames / sec}. Ideally, the fps should match the maximum fps of your display. In case of lag, try decreasing the speed using the speed slider or decreasing the subdivision level.
    \item The \textbf{Vis Mode} selection box is shown on the top right part of the canvas. The available options are \textit{Color}, \textit{Albedo}, \textit{Normal}, \textit{Height}, \textit{Roughness}, \textit{AO (Ambient Occlusion)} and \textit{Graft weights}. The Albedo, Normal, Height, Roughness, and AO options allow the users to visualize the corresponding synthesized texture maps by the MeshNCA model. The Color visualization mode combines all the texture maps and renders the mesh using a physically based rendering (PBR) shader. Choosing the \textit{Graft Weights} option allows the users to visualize the graft weights used for interpolating between two textures. The graft weights change when the user paints on the canvas using the grafting brush.
    \item The checkboxes on the bottom left part of the canvas allow the users to disable/enable the corresponding texture maps in the shading process. When enabled, we use the texture maps (\textbf{AO, Normal, Roughness}) synthesized by MeshNCA for shading, and when disabled, we use a constant map for shading resulting in a more uniform appearance. 
    \item On the bottom right part of the canvas we provide some utilities to visualize the stylized mesh including \textbf{screenshot}, \textbf{camera lock/unlock}, and \textbf{zoom in/out buttons}. The screenshot button will save the current content of the canvas as a PNG image. The camera lock/unlock button toggles the click events between camera-rotation/paint-brush. When the camera is locked, a click/drag on canvas will be interpreted as the use of the brush tool (either in Regenerate or Graft mode). When the camera is unlocked, a click/drag on the canvas will result in changing the viewpoint by rotating the camera around the origin. The zoom-in/out buttons simply bring the camera closer/further to the object. Notice that on personal computers, the camera can be controlled without these buttons and using shortcuts (shift key and mouse scroll wheel).

\end{itemize}

\subsubsection{\textbf{Interactive Controls}} 

As discussed in the main paper, MeshNCA enables several real-time interactive controls for the users to manipulate the synthesized 3D texture. Here, we describe and summarize a list of these controls.
\begin{itemize}[leftmargin=*]

\item \textbf{Play/pause, Reset and Benchmark} buttons. The play/pause toggle button allows the users to stop and resume the iterations of the MeshNCA update rule. The reset button sets all the cell states with the seed state and resets the graft weights to zero. The benchmark button allows the users to benchmark the performance of MeshNCA on their device. 
\item From the \textbf{Subdivison Level} section users can select the mesh subdivision level and control the synthesized texture's density. Notice that changing the subdivision level restarts the MeshNCA from the seed state.
\item \textbf{Speed Slider} allows the users to change the number of MeshNCA update steps per second. More steps per second will result in faster-evolving textures with more salient motion. 
\item \textbf{Orientation Slider} allows the users to control the orientation of the synthesized texture. The angle given by this slider is used in the perception stage to rotate the spherical harmonics basis for each vertex around the surface normal. This operation rotates the orientation of the synthesized texture.
\item \textbf{Bumpiness Slider} The state of this slider together with the synthesized height map, is used to change the underlying geometry of the mesh by simply displacing the vertices along the surface normal by the given bumpiness value.
\item \textbf{Brush Mode} The brush mode toggle allows the users to switch between \textbf{Graft/Regenerate} brushes. When the brush is in the graft mode, the users can also select the desired grafting texture from the drop-down texture list. 
\item The \textbf{Brush Size} section allows the users to manage the size of the painting brush used for grafting or regeneration purposes.

\end{itemize}

\subsubsection{\textbf{Texture and Mesh Selection}}

The two right panels of the user interface allow the users to select between 72 different textures and 47 different meshes. Below the mesh selection panel, we show the main and the grafting textures. The weights of our pre-trained MeshNCA models for all 72 target textures are stored in a single JSON file of 8MB in size. We download this file once when the web page is loading.

\subsection{Shader Prerequisites}

In this section, we describe the common elements used in our WebGL implementation such as the underlying data structures, Framebuffers, and tricks to improve the performance.

\textbf{Storing information in  gl.texImage2D: }
We need the following information to perform one step of MeshNCA update: vertex positions, mesh neighborhood information, per-vertex surface normal, and the weights of pre-trained MeshNCA models. We store all of this information as 2D images in gl.texImage2D, the built-in texture data structure of WebGL shading language.
Vertex positions, vertex normals, and pre-trained weights are each re-arranged and stored in a near-square RGBA 32-bit floating point 2D image. To store the mesh neighborhood information, we first find the neighboring vertex indices list for each vertex and then pad these lists with "-1.0" so that all lists have the same length. These per-vertex neighbor indices lists are then stored in the same way as the vertex positions in floating point textures. Lastly, we re-arrange and store the weights of our pre-trained models in a floating point 2D image. The resulting textures are \textit{vertexPos}, \textit{vertexNormal}, \textit{vertexNeighborhood}, and \textit{pretrainedWeights}.

\textbf{Framebuffers: }
To run one step of MeshNCA update, we need to store and update several tensors containing information such as cell states, graft weights, etc. We re-arrange all of these tensors, store them in Framebuffers, and use a mapping that determines the relationship between a Framebuffer pixel and its corresponding tensor element. 

For synthesizing a single texture (with no grafting) we need 5 different Framebuffers: \textit{cellStates}, \textit{perceptionLayer}, \textit{FCLayer1}, \textit{FCLayer2}, and \textit{updatedCellStates}. When grafting is enabled, we need 4 more Framebuffers: \textit{graftWeights}, \textit{updatedGraftWeights}, \textit{graft\_FCLayer1}, and \textit{graft\_FCLayer2}. These Framebuffers are used as input and outputs for our MeshNCA shaders.

\textbf{Shuffle and unShuffle maps: } Each cell (vertex) state has a 50\% probability to be updated at each step. A naive approach to implement this stochasticity would be to discard the cells based on the values from a pseudo-random generator. However, this implementation is sub-optimal 
since even the non-updated cells are passed through the shader. We propose to use a Shuffle-unShuffle data structure that solves the issues of the vanilla approach and yields nearly 2x performance boost. First, in the pre-processing stage, we create a static data structure called \textit{shuffle and unShuffle maps}. The shuffle map is a sorted list comprised of exactly $\frac{N}{2}$ numbers selected randomly from $0$ to $N - 1$ without replacement where $N$ is the number of vertices (cells) for a given mesh. The shuffle map allows mapping the Framebuffer pixels to their corresponding cells before applying the shuffling operation. The \textit{unShuffle map} stores the inverse of the shuffle map alongside an auxiliary flag for each cell that indicates if a cell was in the shuffle map list. The unShuffle map allows us to update the correct cell given the residual update values from the second fully connected layer. We store each of the shuffle and unShuffle maps in an RGBA 32-bit floating point texture. At each step of the computation, we use a single random index between $0$ and $N - 1$ to offset the shuffle and unShuffle maps that allows us to create a pseudo-random mapping. Notice that the randomness created by this approach differs from the randomness we use in the training. In our WebGL randomness paradigm, exactly 50\% of the cells are updated at each step, while in our training we update each cell with a probability of 50\% and the total number of updated cells at each step can be varied. We find that MeshNCA can generalize to new randomness schemes in test-time without compromising the quality. 

The shuffle mask allows us to reduce the size of some Framebuffers by a factor of 2.0 and gain an almost 2x performance boost. The following Framebuffers utilize the shuffle mask: \textit{perceptionLayer}, \textit{FCLayer1}, \textit{FCLayer2}, \textit{graft\_FCLayer1}, and \textit{graft\_FCLayer2}. 

\subsection{MeshNCA Shaders}
There are two main types of shaders in our WebGL backend: (1) Shaders for computing one step of MeshNCA update to synthesize the texture maps, and (2) A physically based rendering shader to render the mesh with the synthesized texture maps. In the following sections, we delve into the details of these shaders. Moreover, we elaborate on how the interactive controls from our UI interact with these shaders. 

Overall, our implementation contains five main Fragment shaders: \textbf{MeshPerception}, \textbf{FullyConnected}, \textbf{StateUpdate}, \textbf{PaintBrush}, and \textbf{Render}, where the first four are used to run our MeshNCA model and the last one is used to render the output to the canvas.

\textbf{MeshPerception Shader: } This shader uses the \textit{cellStates} Framebuffer and \textit{vertexPos}, \textit{vertexNormal}, \textit{vertexNeighborhood}, and \textit{shuffleMap} textures as uniform inputs. The output of this shader is written to the \textit{perceptionLayer} Framebuffer. This shader contains a for loop that iterates over the neighboring vertex indices extracted from the \textit{vertexNeighborhood} and applies the spherical-harmonics-based perception scheme outlined in the main paper. This works by (1) querying the position of the source and the neighbor vertices from the \textit{vertexPos}, (2) rotating the connecting edge direction by the orientation angle given from the UI along the vertex normal extracted from \textit{vertexNormal}, and (3) passing the rotated edge direction to the spherical harmonics basis. This shader also uses the \textit{shuffleMap} texture to find the actual vertex indices. 

\textbf{FullyConnected Shader: } The goal of this shader is to perform the computation of the Multi-Layered-Perceptron (MLP) in our MeshNCA architecture. The output of this shader is written to either \textit{FCLayer1}, or \textit{FCLayer2} Framebuffers depending on the layer being computed. This shader uses either \textit{perceptionLayer} or \textit{FCLayer1} as a uniform input. The other uniform inputs include \textit{shuffleMap} and \textit{pretrainedWeights}. This shader is comprised of a for loop that performs the matrix multiplication given by the fully connected layer's weights.

This shader is also used with the pretrained weights of the grafting model to compute the grafting update values needed for our grafting scheme. In this case, the output is written to one of the \textit{graft\_FCLayer1}, or \textit{graft\_FCLayer2} Framebuffers.

\textbf{StateUpdate Shader: } This shader applies the residual update given by the output of the FullyConnected shader to finish one step of the MeshNCA update. This shader uses the \textit{cellStates}, \textit{FCLayer2}, \textit{graft\_FCLayer2}, and \textit{graftWeights} Frambuffers alongside \textit{unshuffleMap} texture as uniform inputs and writes the output to the \textit{updatedCellStates} Framebuffer. The \textit{unShuffleMap} texture is used to determine if the cell at hand is going to be updated or not and to find the shuffled index of the cell needed to fetch the necessary information from the other Framebuffers. The \textit{graftWeights} buffer determines the interpolation weight used to mix the residual output from \textit{FCLayer2} and \textit{graft\_FCLayer2} buffers. 
After this shader is called, the \textit{cellStates} and \textit{updatedCellStates} buffers are swapped so that the \textit{cellStates} Frambuffer stores the updated states.

\textbf{PaintBrush Shader: } This shader is used to manipulate either the \textit{cellState} or the \textit{graftWeights} buffers depending on the painting brush mode. This shader uses \textit{vertexPos} and \textit{vertexNormal} as uniform inputs.
To avoid expensive ray-triangle intersection computation, we use an effective but not quite correct hack to find the vertices affected by the brush. For each vertex, we first calculate its 2D viewpoint location by applying the camera transformation. A vertex is then affected by the brush if (1) its view coordinate is in the brush radius of the click/touch position and (2) the vertex normal is facing towards the camera. 

When the brush is in the \textit{Regenerate} mode, the state of the cells (vertices) that are in the proximity of the click location will be set to zero (the seed state). When the brush is used for \textit{Grafting} the graft weights for the vertices around the click point are increased slightly so that the grafting texture receives more weight. In this case, the \textit{graftWeights} and \textit{updatedGraftWeights} buffers are used as double buffers and swapped after the shader is executed.

\textbf{Render Shaders}
Given the synthesized texture map values for each vertex, the goal of the rendering shader is to combine these values using a physically-based-rendering (PBR) algorithm to output the stylized mesh. Our renderer is inspired by the PBR algorithm described in \cite{pbrshader}. The vertex shader displaces each vertex along the surface normal by $d = h^2 \times b$ where $h$ is the synthesized height texture value and $b$ is the bumpiness value from the UI. The rest of the texture maps including albedo, normal, roughness, and ambient occlusion are sent to our PBR fragment shader using the \textit{varying} qualifier so that they are interpolated using the barycentric coordinates.

For lighting, we use a single point light source on the line connecting the camera to the origin at $d=2.0$ from the origin. The intensity of the point light source is adjusted for each mesh to give a more natural experience when looking at different meshes. When the $Vis Mode$ is anything other than \textit{Color}, we disable physically-based rendering and display the synthesized texture values directly. In case the user disables a texture map using the render controls on the canvas, we use the default values for the corresponding texture map. The default values for Ambient Occlusion (AO), Normal, and Roughness are 1.0, vec3(0.0, 0.0, 1.0), and 0.5, respectively.

\subsection{MeshNCA Properties}
\label{sec:meshnca-property}
MeshNCA exhibits various unique test-time properties that make it suitable for real-time and interactive applications. Here, we discuss nine of them and show the corresponding videos for each property at \href{https://meshnca.github.io/#MeshNCAProp}{\textcolor{pinkl}{https://meshnca.github.io/\#MeshNCAProp}}. For a detailed discussion of our WebGL demo, we refer the readers to our supplementary material.
\begin{itemize}[leftmargin=*]
    \item \textbf{Generalization to Unseen Meshes}. MeshNCA demonstrates remarkable generalization ability to various mesh structures that it has not encountered during training. This is primarily attributed to its localized cell communication scheme and update rule. Such a mechanism allows it to effectively handle meshes with distinct topological features, including those with holes or meshes with sharp edges, which differ significantly from the training \icosphere  mesh.
    \item \textbf{Generalization to Animated Meshes}. Our proposed cell communication scheme, based on spherical harmonics, is naturally invariant to scale and translation. This property is particularly beneficial when dealing with animated meshes. MeshNCA is robust against moving vertices and can seamlessly texture animated meshes, despite being trained on a static mesh.
    \item \textbf{Self Organization}. MeshNCA operates on the basis of asynchronous self-updating cells that rely on local information. Such a decentralized update rule promotes robust self-organization within the model, enhancing its resilience to external perturbations. That is, MeshNCA can self-regenerate from external perturbations.
    \item \textbf{Grafting}. Similar to biological grafting, two compatible MeshNCA models can be grafted together and produce a coherent transition between the underlying two textures.
    \item \textbf{Texture Density Control}. The scale-invariant nature of MeshNCA allows for the manipulation of texture density based on the mesh's vertex density. Users can adjust the texture density by subdividing the mesh, providing another degree of control over the final output.
    \item \textbf{Texture Orientation Control}. In MeshNCA, cells perceive their surrounding environment in a direction-aware manner using spherical harmonics. By rotating the spherical harmonics basis around the surface normal for each cell, MeshNCA allows the users to re-orient the synthesized texture in real time.
    \item \textbf{Spontaneous Motion}. Without any video or motion supervision, MeshNCA spontaneously generates dynamic textures with random motion. Additionally, the motion can be supervised using our motion training scheme. 
    \item \textbf{Motion Direction Control}. Similar to \textit{Texture Orientation Control}, MeshNCA allows the users to control the motion direction by rotating the local spherical harmonics basis, linking macro motion to the micro-movements around each cell.
    \item \textbf{Coherent multi-channel alignment for fitting PBR textures}. MeshNCA effectively harnesses structural similarities found in PBR textures, such as \textit{Albedo}, \textit{Normal}, \textit{Height}, \textit{Roughness}, and \textit{Ambient Occlusion} maps. It can learn to  simultaneously synthesize multiple texture maps using different channels of the cell states without imposing structural constraints between these maps.
\end{itemize}

\section{Mesh Collection}
\label{sec:mesh_collection}
MeshNCA textures a mesh by assigning texture attributes, such as albedo and roughness, to the vertices of the mesh. 
The quality of the synthesized texture relies on both the mesh's quality and the distribution of vertices across the mesh surface. The two main factors for ensuring the texture quality are: 
\begin{itemize}
    \item Uniform vertex distribution and equal edge lengths
    \item Valence\footnote{the number of incident edges for a vertex} close to 6
\end{itemize}
Both of these conditions are met for the \icosphere mesh which we use for training our model.

For testing, we have created a dataset of 47 meshes while trying to ensure the aforementioned properties for each mesh.  Most of our meshes come from online repositories and we use PyMeshLab \cite{pymeshlab} to remesh the meshes in our dataset. 
Our mesh dataset is available at \href{https://meshnca.github.io/data/meshes/lionstatue/}{meshnca.github.io/data/meshes/}. We also provide the license of each mesh in the \textit{info.md} file under each mesh folder.  Table~\ref{tab:mesh_dataset} provides an overview of the meshes in our dataset.

\newcommand{\meshthumb}[1]{%
  \includegraphics[width=50pt]{figures/mesh_thumbnails/#1.jpg}
}
\newcommand{\cntrtxtmesh}[1]{
\begin{tabular}{l}
\parbox{1.0cm}{\vspace{-50pt} \centering #1}
\end{tabular}
}

\newcommand{\cntrdigit}[1]{
\begin{tabular}{l}
\parbox{8pt}{\vspace{-50pt} \centering #1}
\end{tabular}
}

\newcommand{\cntrnumber}[1]{
\begin{tabular}{l}
\parbox{0.5cm}{\vspace{-50pt} \centering #1}
\end{tabular}
}

\begin{table*}[]
\caption{An overview of our mesh dataset. The columns $\#F$ and $\#V$ represent the number of faces and vertices, respectively. Notice that the average valence for all the meshes is close to 6.0. }
\resizebox{0.95\textwidth}{!}{
\begin{tabular}{ccccccc||ccccccc}
\toprule
\multirow{2}{*}{\textbf{Name}} & \multirow{2}{*}{\textbf{Mesh}} & \multirow{2}{*}{\#\textbf{F}} & \multirow{2}{*}{\#\textbf{V}} &\multicolumn{3}{c||}{\textbf{Valence}} & \multirow{2}{*}{\textbf{Name}} & \multirow{2}{*}{\textbf{Mesh}} & \multirow{2}{*}{\#\textbf{F}} & \multirow{2}{*}{\#\textbf{V}} & \multicolumn{3}{c}{\textbf{Valence}} \\
&&&& Max & Min & Mean &&&&& Max & Min & Mean \\ 
\midrule
\cntrtxtmesh{sphere} & \meshthumb{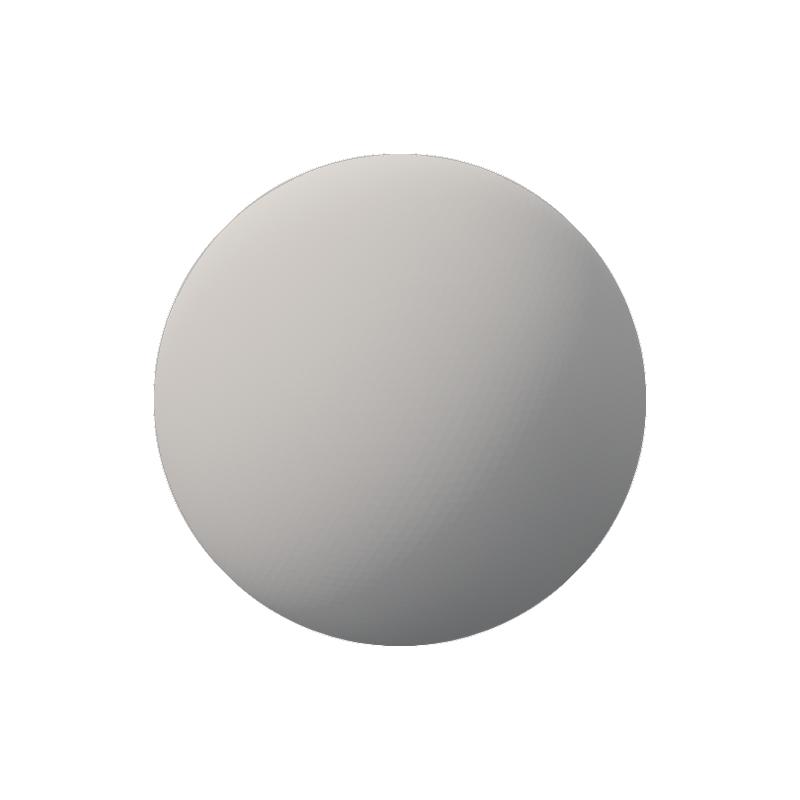} & \cntrnumber{20480} & \cntrnumber{10242} & \cntrdigit{5} & \cntrdigit{6} & \cntrdigit{6.0} &
\cntrtxtmesh{mug} & \meshthumb{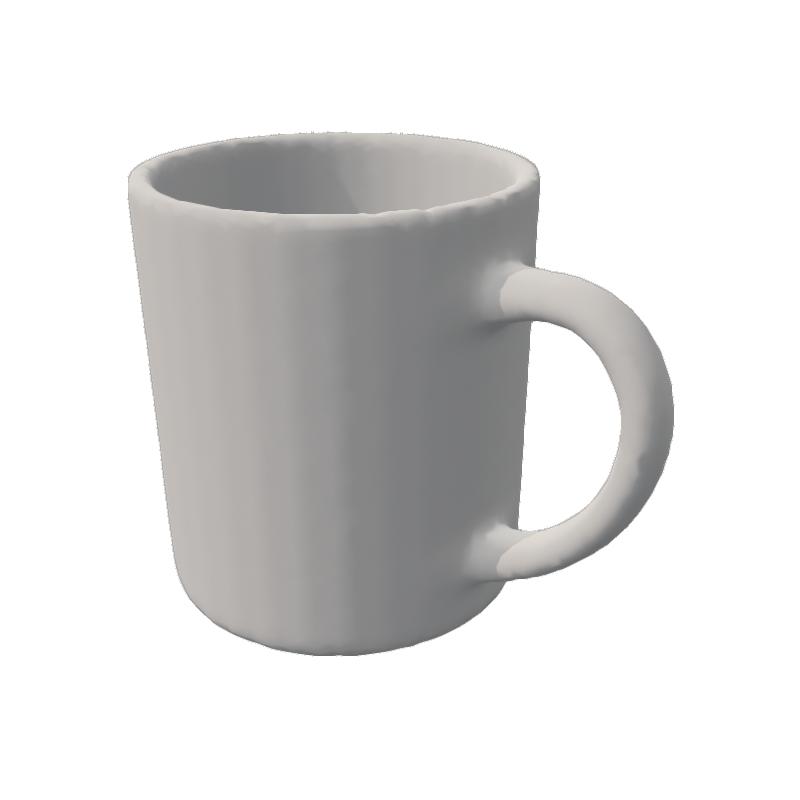} & \cntrnumber{21440} & \cntrnumber{10720} & \cntrdigit{4} & \cntrdigit{8} & \cntrdigit{6.0} \\

\cntrtxtmesh{seashell} & \meshthumb{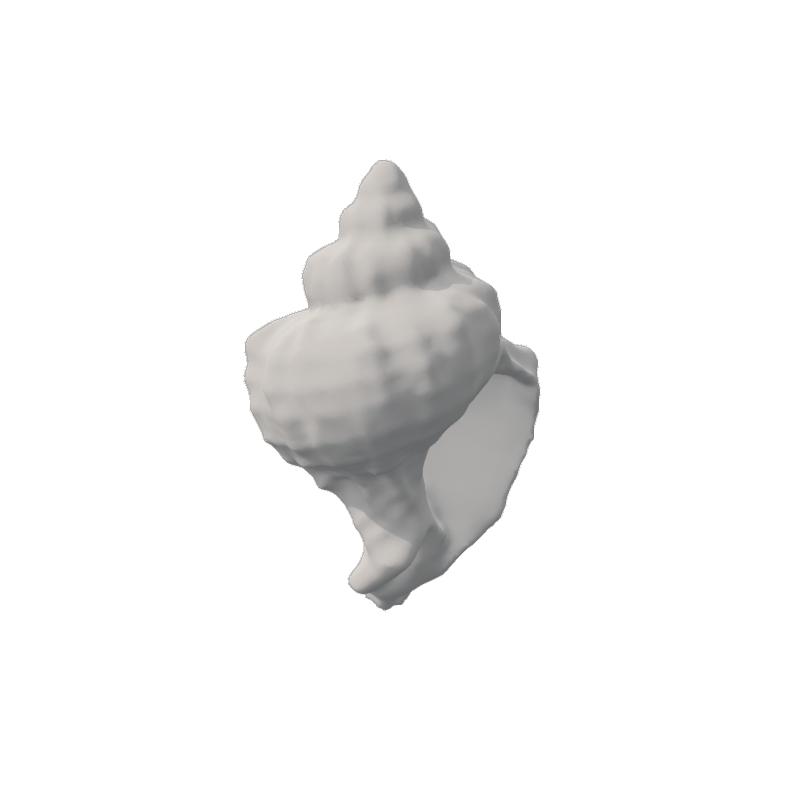} & \cntrnumber{22088} & \cntrnumber{11046} & \cntrdigit{3} & \cntrdigit{9} & \cntrdigit{6.0} &
\cntrtxtmesh{armor} & \meshthumb{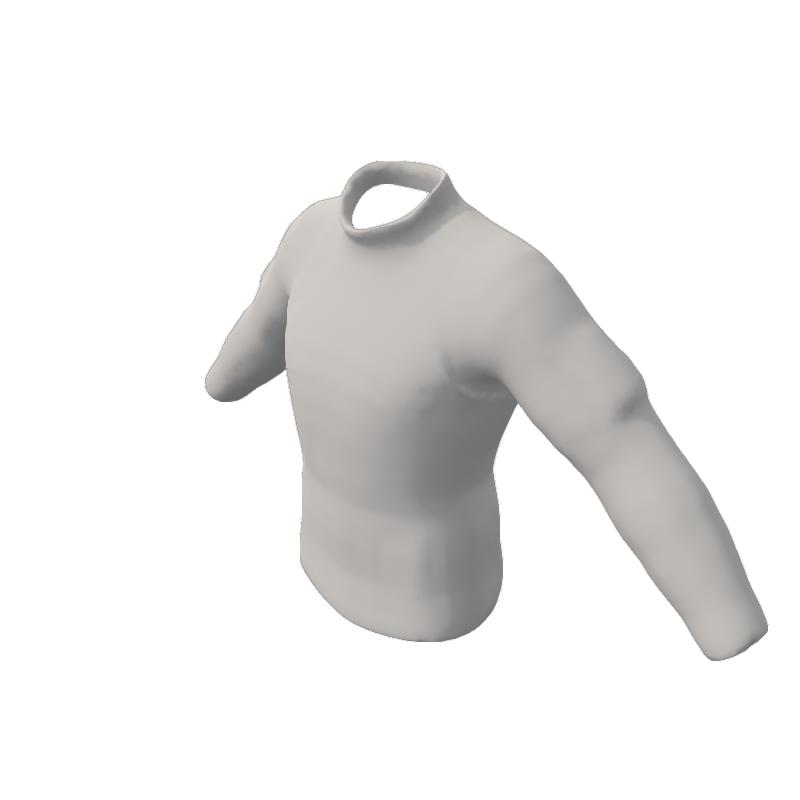} & \cntrnumber{22053} & \cntrnumber{11104} & \cntrdigit{3} & \cntrdigit{9} & \cntrdigit{5.97} \\

\cntrtxtmesh{dragon} & \meshthumb{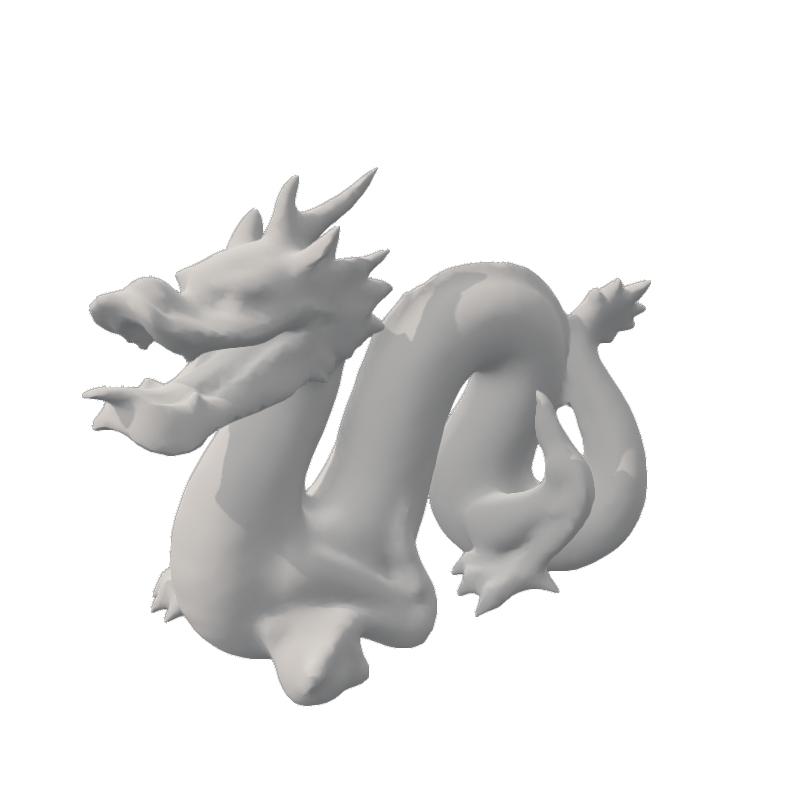} & \cntrnumber{22282} & \cntrnumber{11141} & \cntrdigit{3} & \cntrdigit{9} & \cntrdigit{6.0} &
\cntrtxtmesh{mobius} & \meshthumb{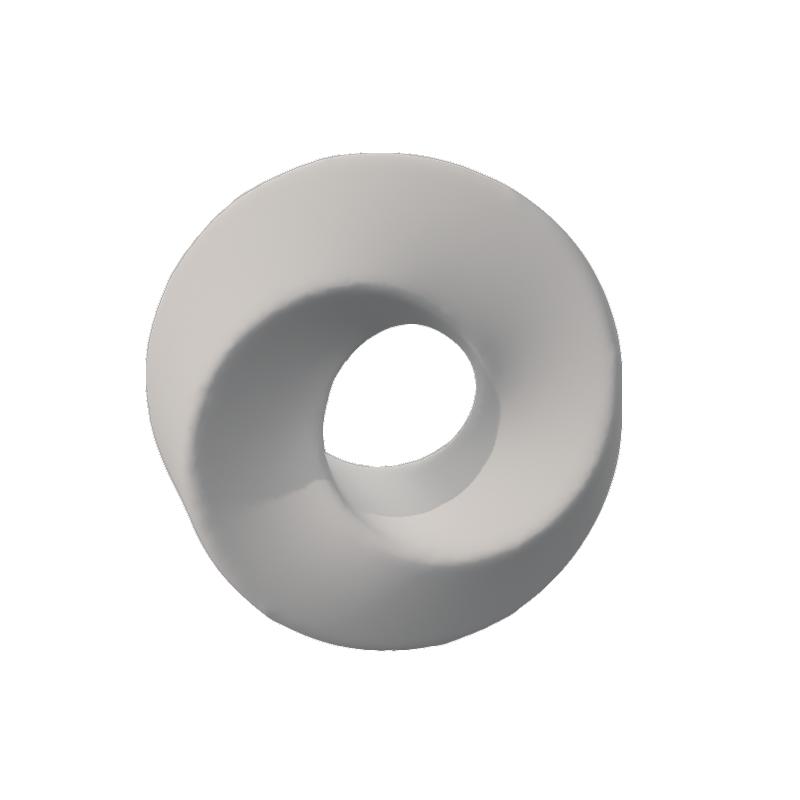} & \cntrnumber{21794} & \cntrnumber{10897} & \cntrdigit{5} & \cntrdigit{8} & \cntrdigit{6.0} \\

\cntrtxtmesh{fertility} & \meshthumb{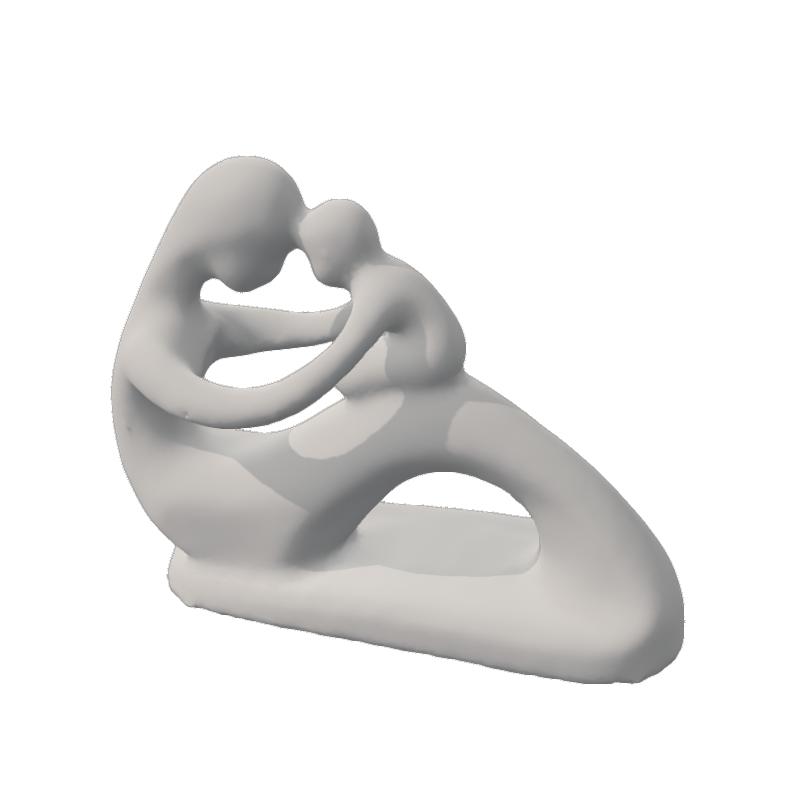} & \cntrnumber{20412} & \cntrnumber{10200} & \cntrdigit{3} & \cntrdigit{8} & \cntrdigit{6.0} &
\cntrtxtmesh{scorpion} & \meshthumb{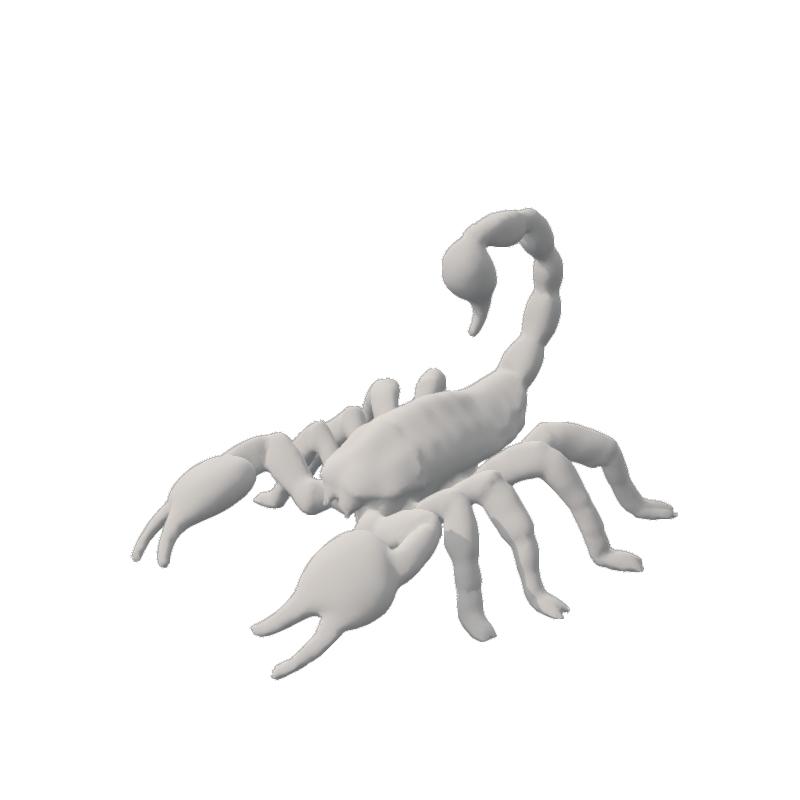} & \cntrnumber{22130} & \cntrnumber{11067} & \cntrdigit{3} & \cntrdigit{10} & \cntrdigit{6.0} \\

\cntrtxtmesh{cube} & \meshthumb{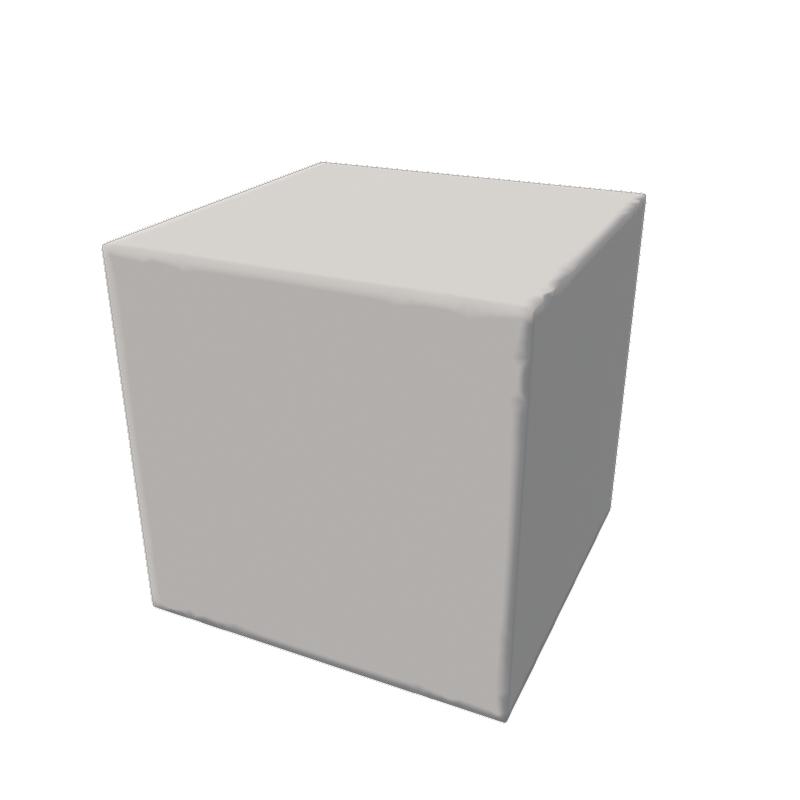} & \cntrnumber{15152} & \cntrnumber{7578} & \cntrdigit{5} & \cntrdigit{7} & \cntrdigit{6.0} &
\cntrtxtmesh{bunny} & \meshthumb{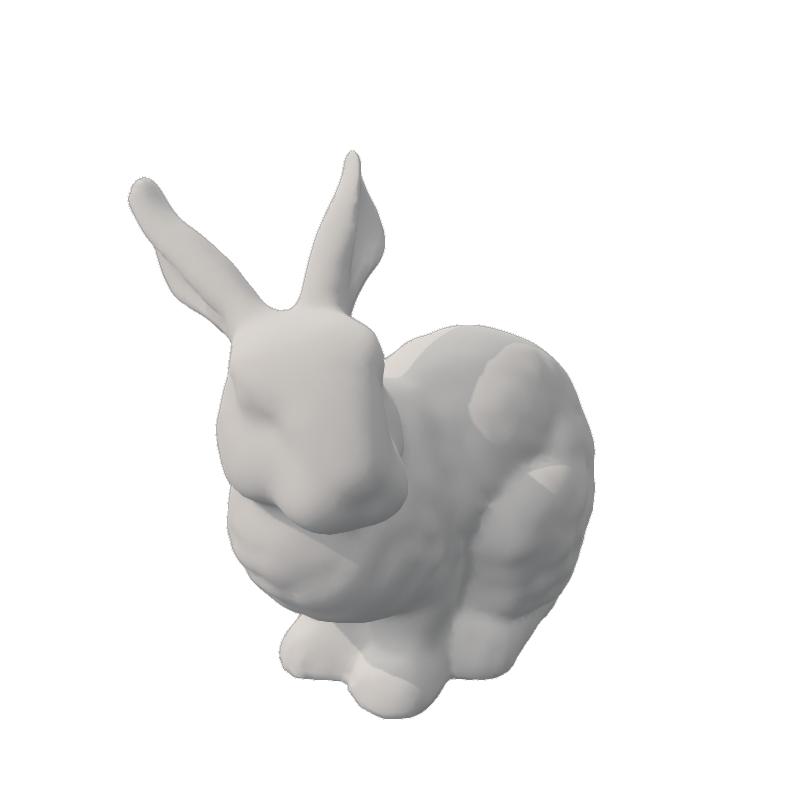} & \cntrnumber{19508} & \cntrnumber{9756} & \cntrdigit{4} & \cntrdigit{8} & \cntrdigit{6.0} \\

\cntrtxtmesh{paimon} & \meshthumb{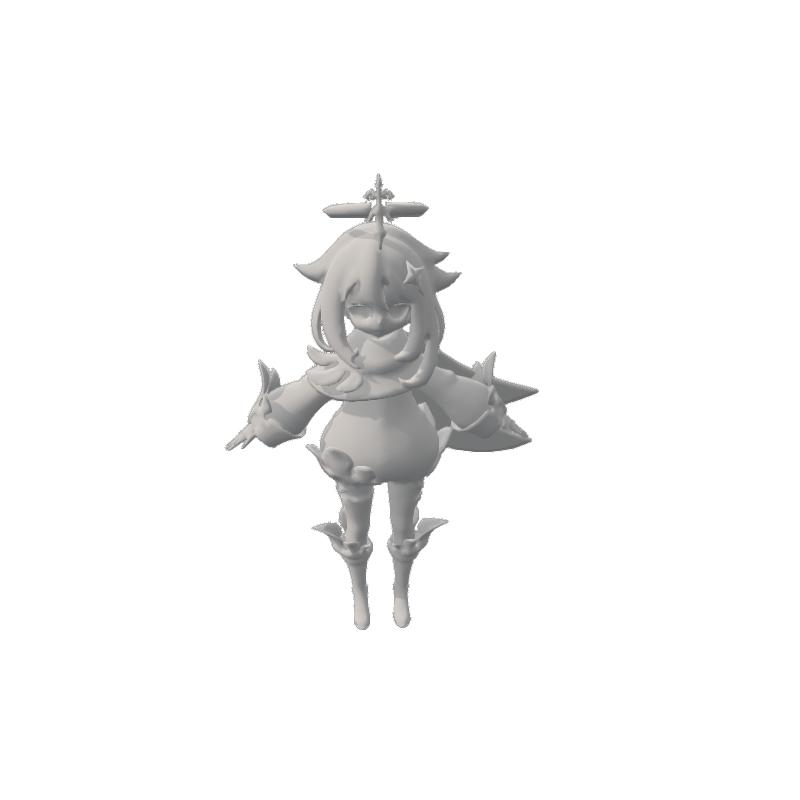} & \cntrnumber{20394} & \cntrnumber{10340} & \cntrdigit{2} & \cntrdigit{13} & \cntrdigit{5.95} &
\cntrtxtmesh{skull} & \meshthumb{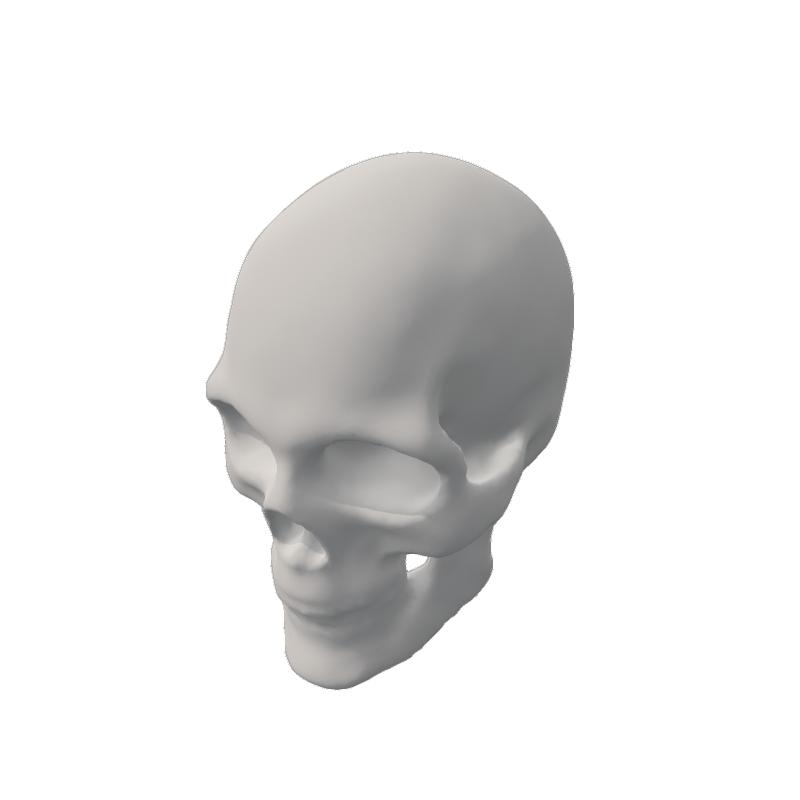} & \cntrnumber{26062} & \cntrnumber{13029} & \cntrdigit{3} & \cntrdigit{10} & \cntrdigit{6.0} \\

\cntrtxtmesh{sphere N.H.} & \meshthumb{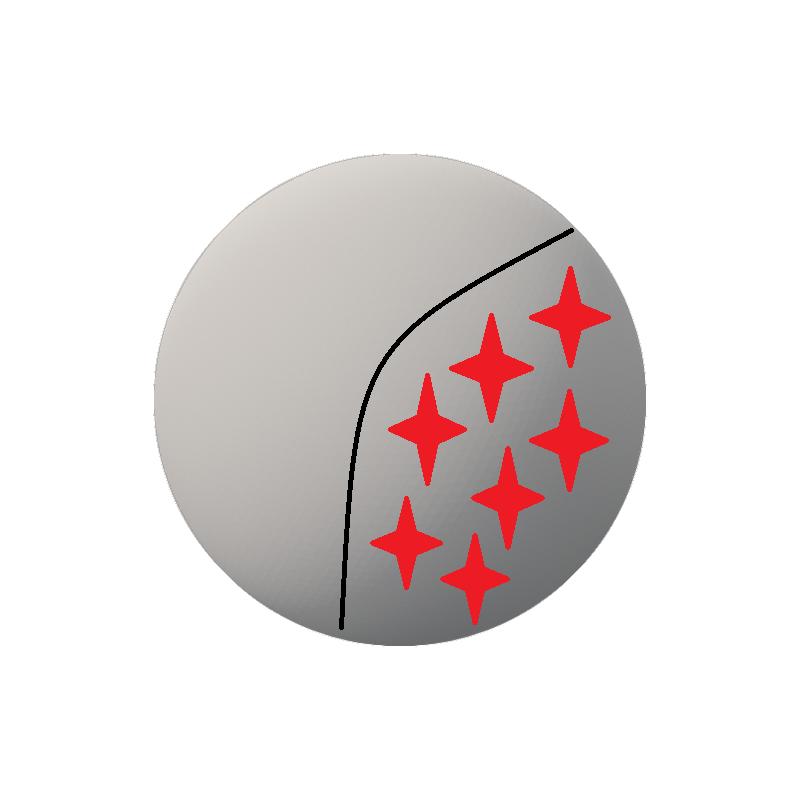} & \cntrnumber{38368} & \cntrnumber{19186} & \cntrdigit{5} & \cntrdigit{8} & \cntrdigit{6.0} &
\cntrtxtmesh{springer} & \meshthumb{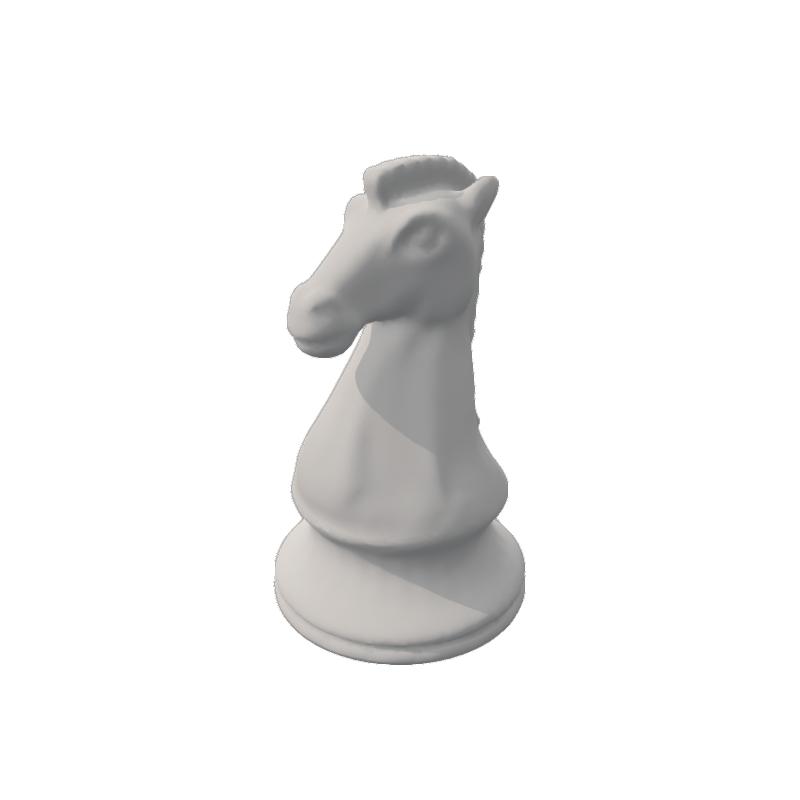} & \cntrnumber{20770} & \cntrnumber{10387} & \cntrdigit{3} & \cntrdigit{8} & \cntrdigit{6.0} \\

\cntrtxtmesh{banana} & \meshthumb{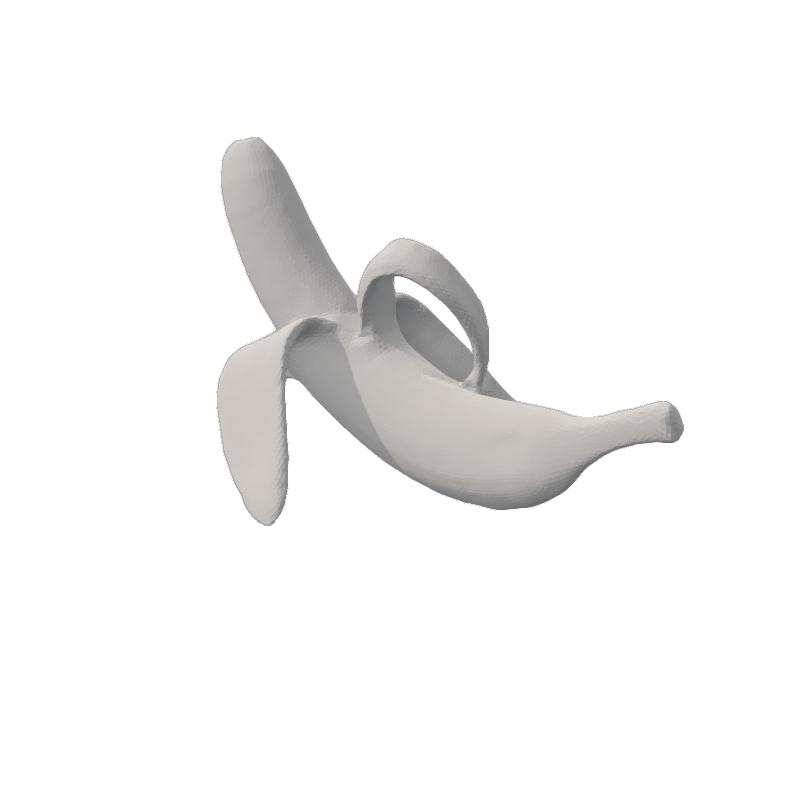} & \cntrnumber{30060} & \cntrnumber{15030} & \cntrdigit{4} & \cntrdigit{9} & \cntrdigit{6.0} &
\cntrtxtmesh{heptoroid} & \meshthumb{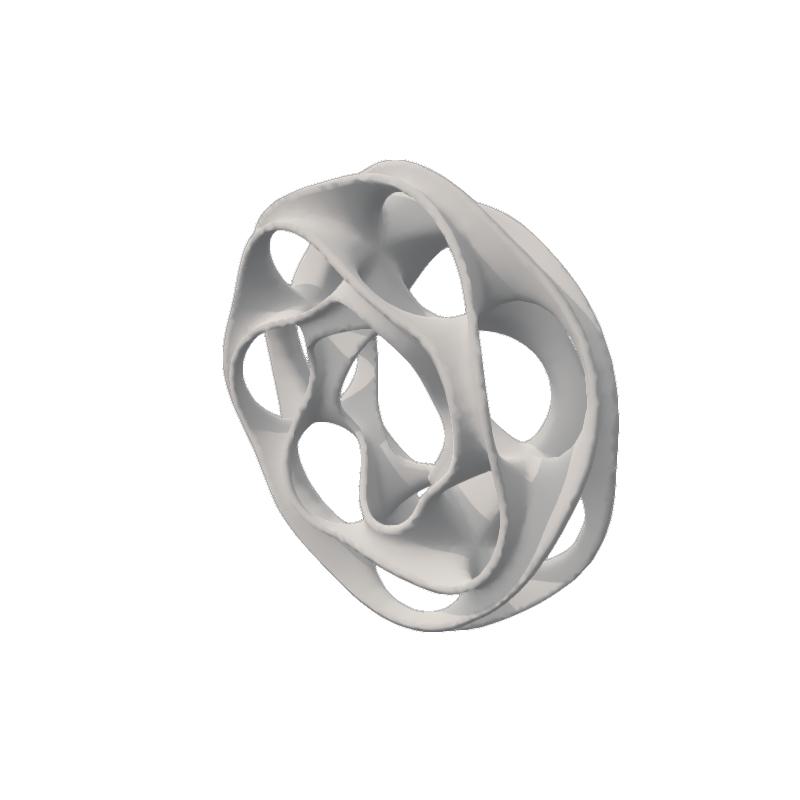} & \cntrnumber{30430} & \cntrnumber{15173} & \cntrdigit{5} & \cntrdigit{9} & \cntrdigit{6.02} \\

\cntrtxtmesh{airplane} & \meshthumb{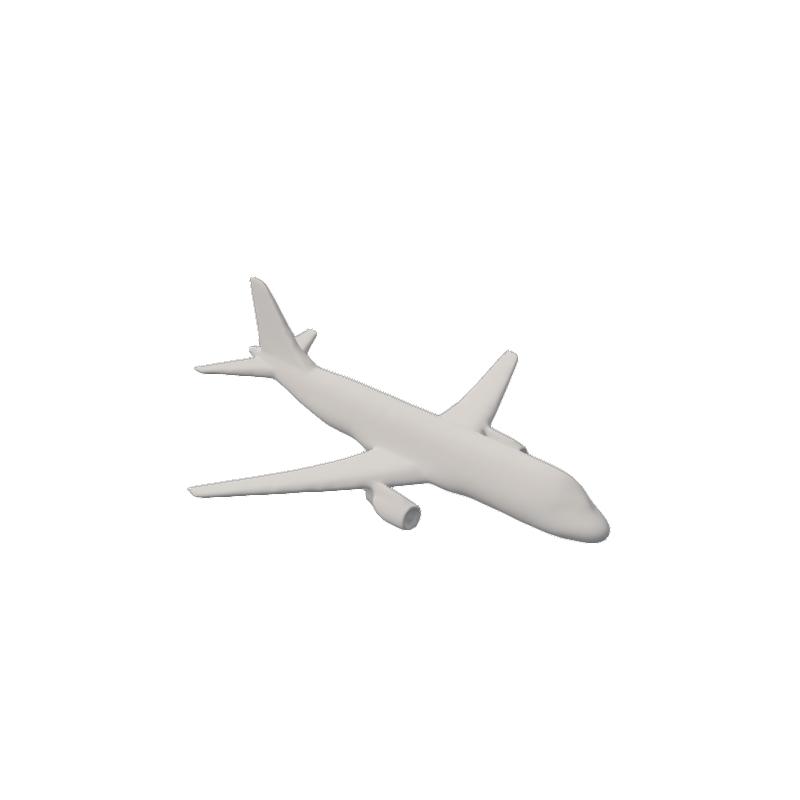} & \cntrnumber{27472} & \cntrnumber{13738} & \cntrdigit{3} & \cntrdigit{9} & \cntrdigit{6.0} &
\cntrtxtmesh{alien} & \meshthumb{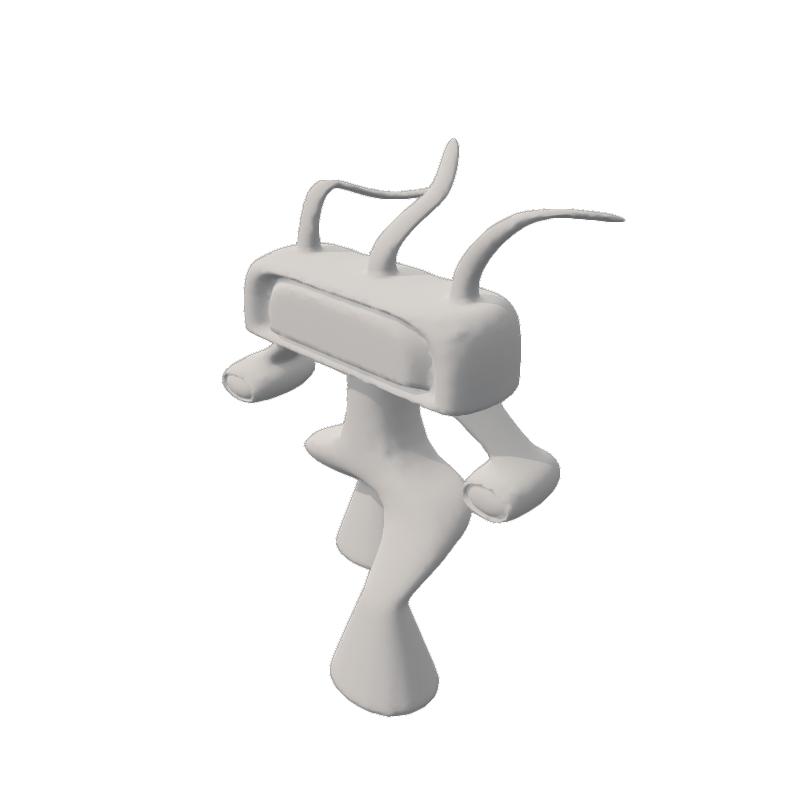} & \cntrnumber{22428} & \cntrnumber{11216} & \cntrdigit{3} & \cntrdigit{9} & \cntrdigit{6.0} \\

\cntrtxtmesh{armadillo} & \meshthumb{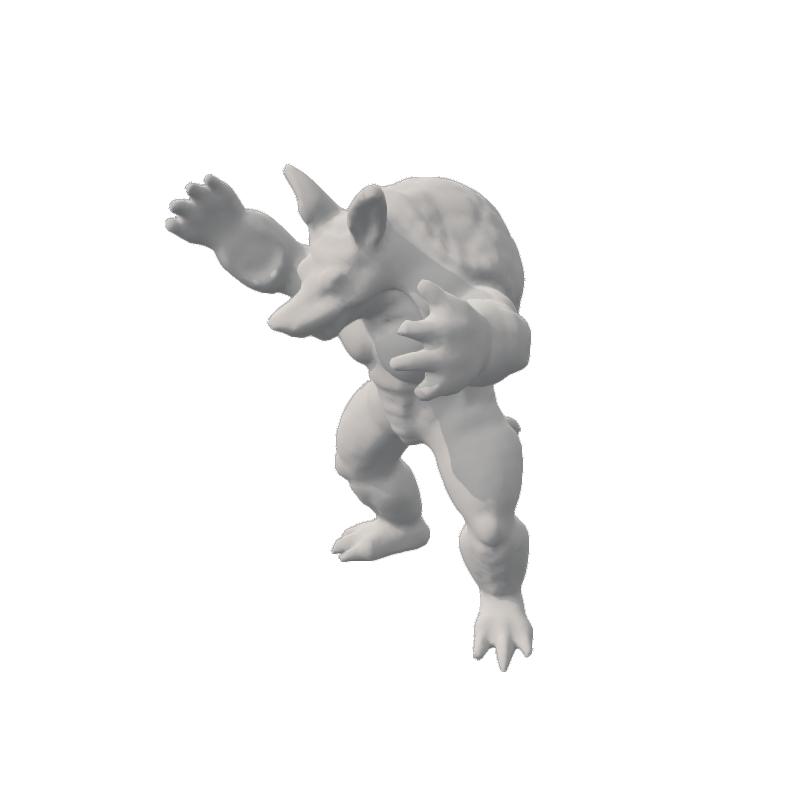} & \cntrnumber{21164} & \cntrnumber{10584} & \cntrdigit{4} & \cntrdigit{10} & \cntrdigit{6.0} &
\cntrtxtmesh{boat} & \meshthumb{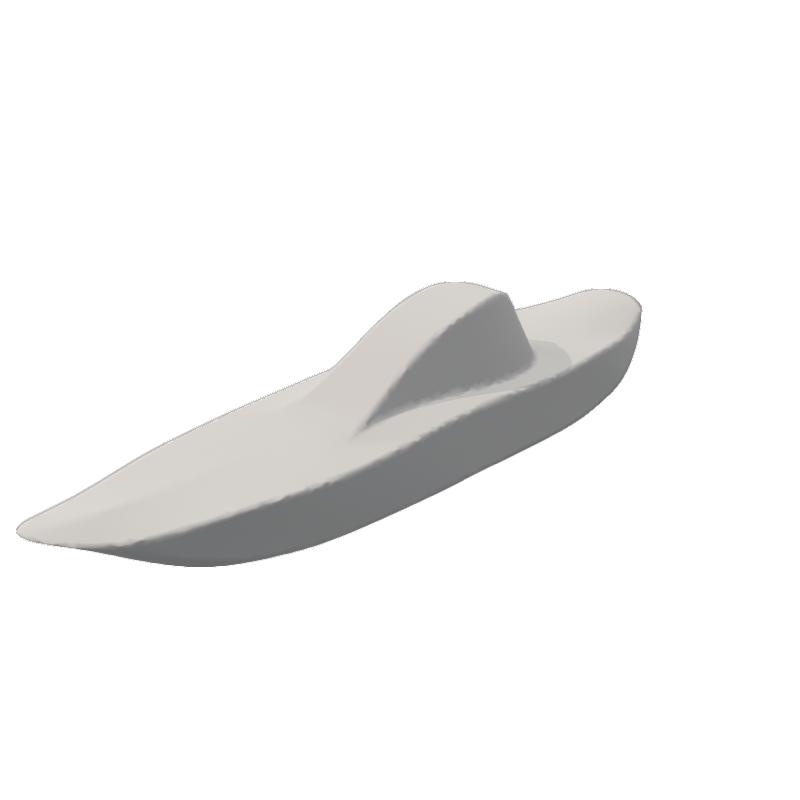} & \cntrnumber{18244} & \cntrnumber{9124} & \cntrdigit{5} & \cntrdigit{8} & \cntrdigit{6.0} \\

\cntrtxtmesh{boot} & \meshthumb{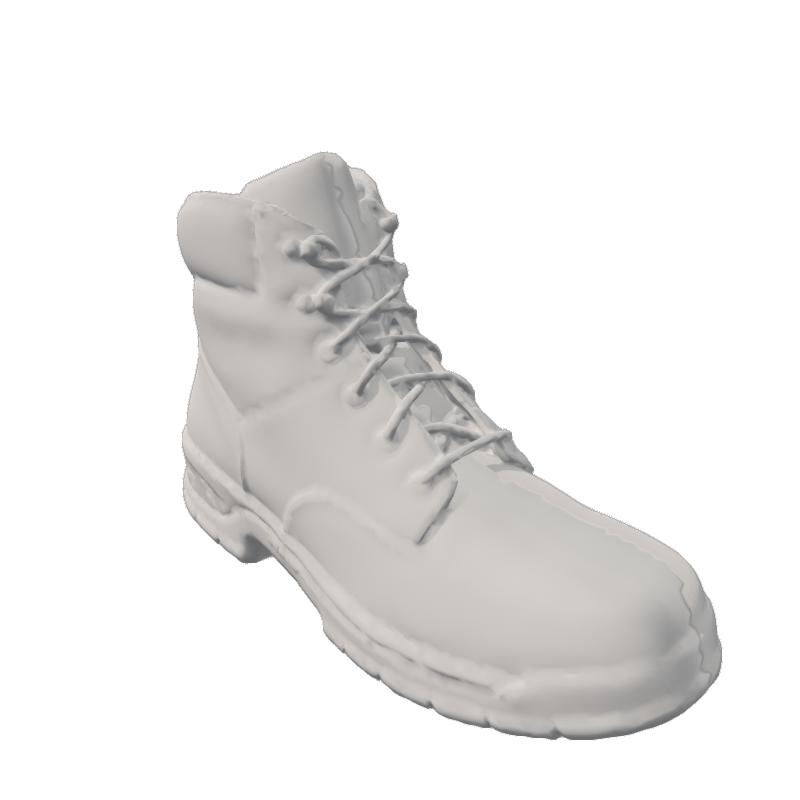} & \cntrnumber{28128} & \cntrnumber{14040} & \cntrdigit{3} & \cntrdigit{12} & \cntrdigit{6.01} &
\cntrtxtmesh{brucewick} & \meshthumb{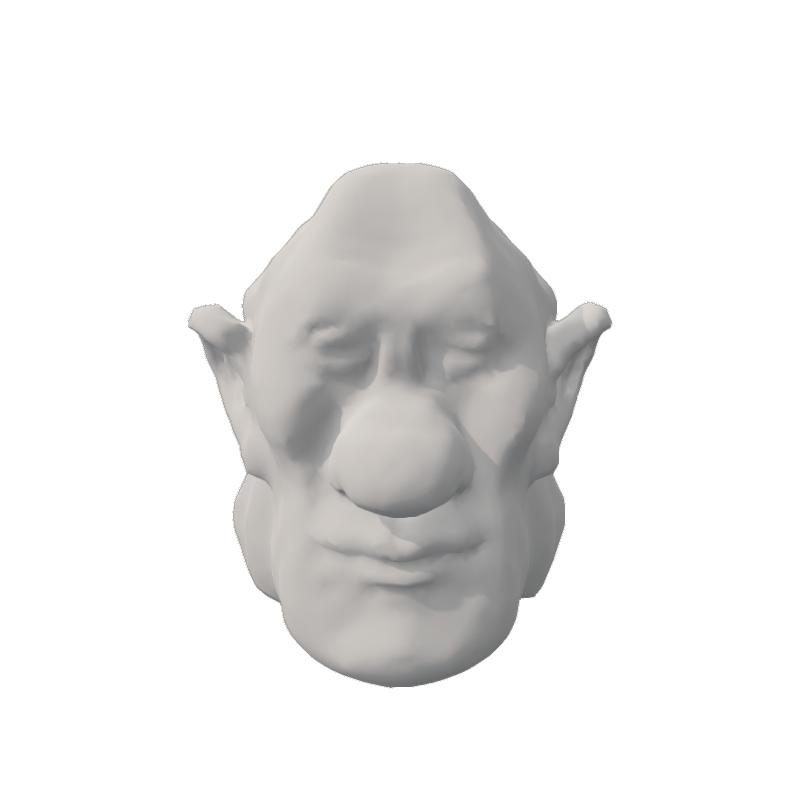} & \cntrnumber{23392} & \cntrnumber{11698} & \cntrdigit{3} & \cntrdigit{9} & \cntrdigit{6.0} \\

\cntrtxtmesh{buddha} & \meshthumb{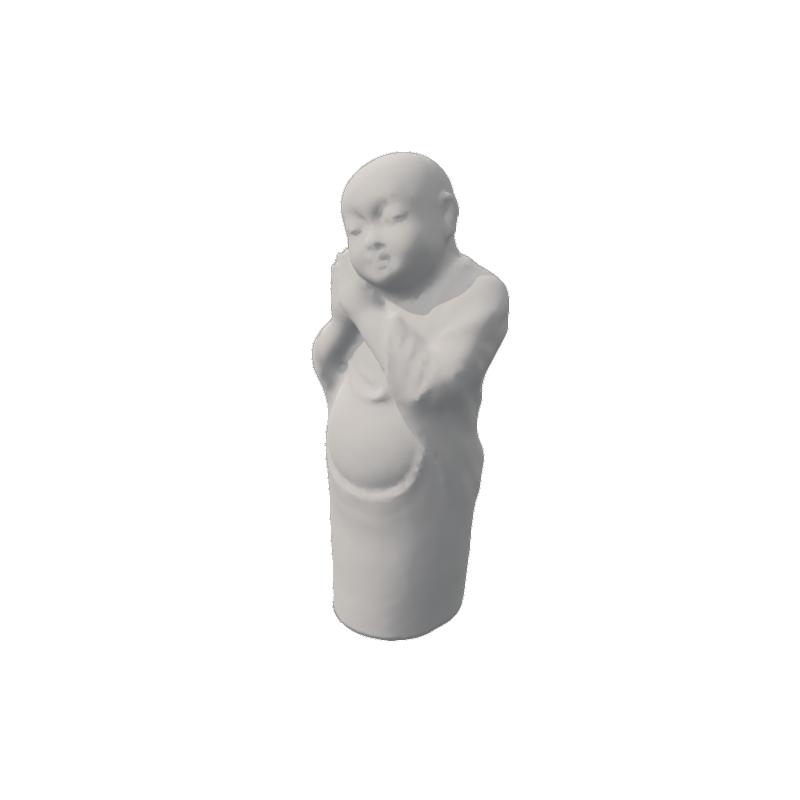} & \cntrnumber{19254} & \cntrnumber{9629} & \cntrdigit{3} & \cntrdigit{10} & \cntrdigit{6.0} &
\cntrtxtmesh{torus} & \meshthumb{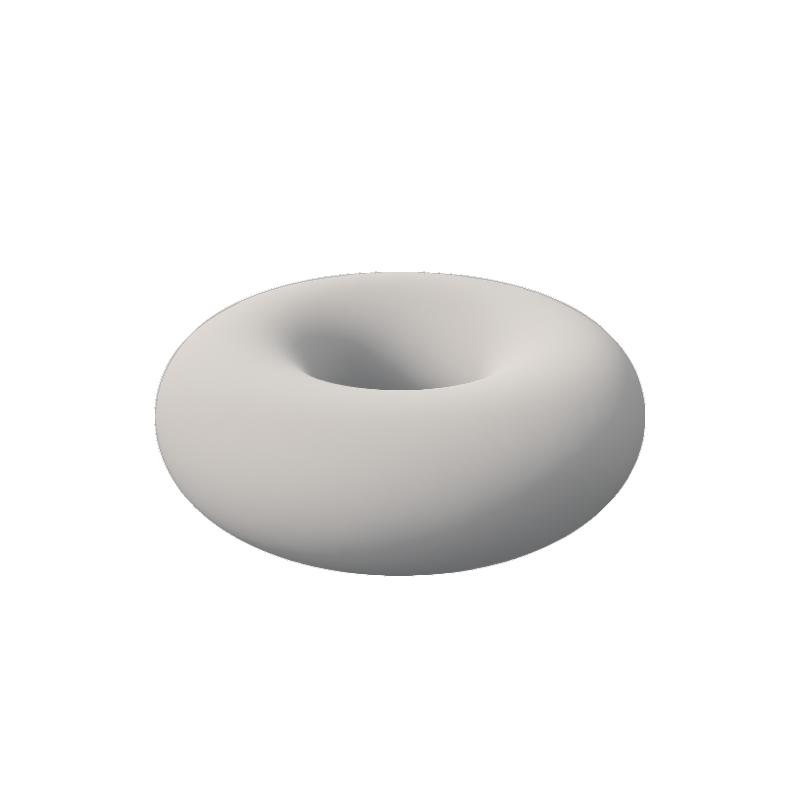} & \cntrnumber{22652} & \cntrnumber{11326} & \cntrdigit{4} & \cntrdigit{8} & \cntrdigit{6.0} \\

\bottomrule
\end{tabular}
}
\vspace{5pt}
\label{tab:mesh_dataset}
\end{table*}

\begin{table*}[]
\caption{An overview of our mesh dataset. The columns $\#F$ and $\#V$ represent the number of faces and vertices, respectively. Notice that the average valence for all the meshes is close to 6.0. }
\resizebox{0.95\textwidth}{!}{
\begin{tabular}{ccccccc||ccccccc}
\toprule
\multirow{2}{*}{\textbf{Name}} & \multirow{2}{*}{\textbf{Mesh}} & \multirow{2}{*}{\#\textbf{F}} & \multirow{2}{*}{\#\textbf{V}} &\multicolumn{3}{c||}{\textbf{Valence}} & \multirow{2}{*}{\textbf{Name}} & \multirow{2}{*}{\textbf{Mesh}} & \multirow{2}{*}{\#\textbf{F}} & \multirow{2}{*}{\#\textbf{V}} & \multicolumn{3}{c}{\textbf{Valence}} \\
&&&& Max & Min & Mean &&&&& Max & Min & Mean \\ 
\midrule

\cntrtxtmesh{flatland} & \meshthumb{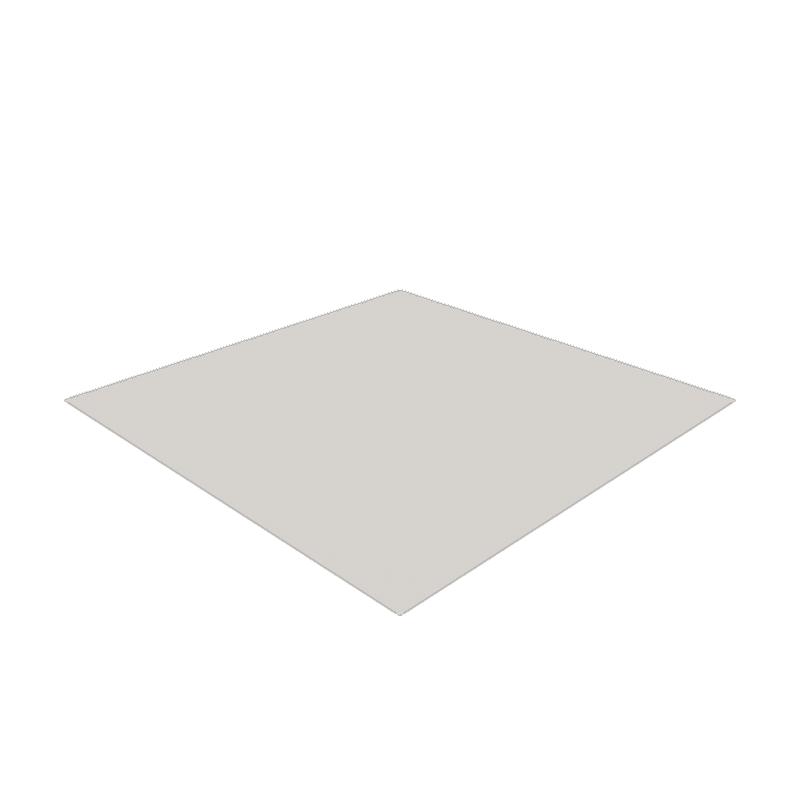} & \cntrnumber{31004} & \cntrnumber{15738} & \cntrdigit{2} & \cntrdigit{8} & \cntrdigit{5.94} &
\cntrtxtmesh{cat} & \meshthumb{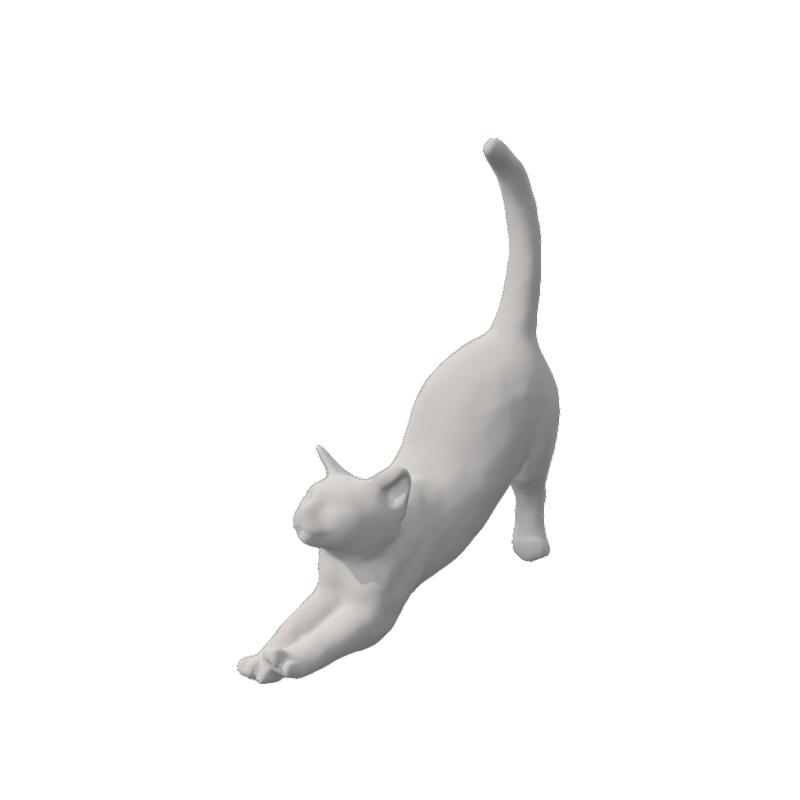} & \cntrnumber{15010} & \cntrnumber{7507} & \cntrdigit{3} & \cntrdigit{8} & \cntrdigit{6.0} \\

\cntrtxtmesh{cow} & \meshthumb{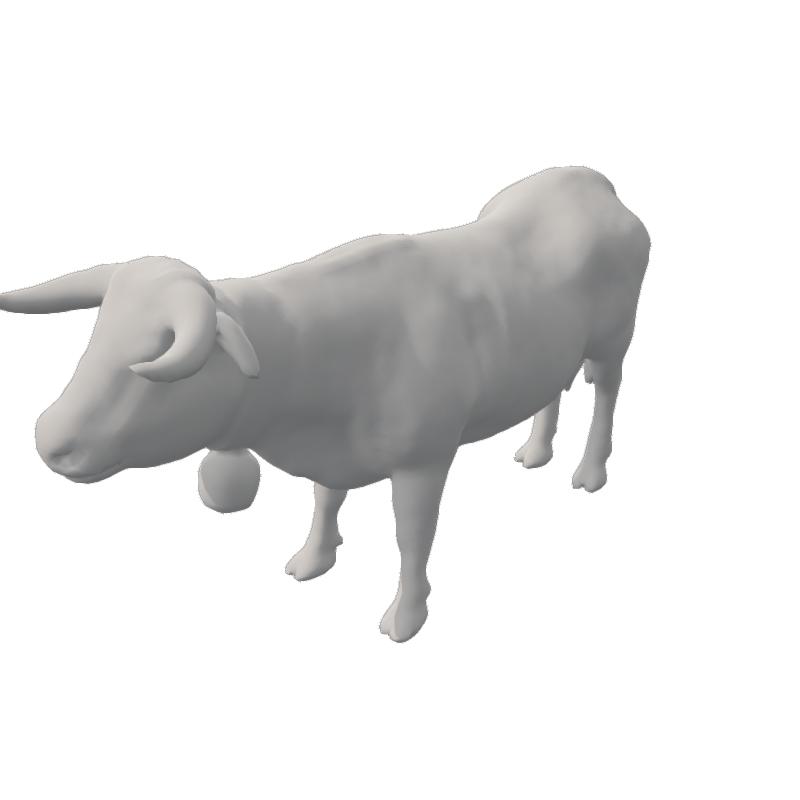} & \cntrnumber{19778} & \cntrnumber{9889} & \cntrdigit{3} & \cntrdigit{12} & \cntrdigit{6.0} &
\cntrtxtmesh{demosthenes} & \meshthumb{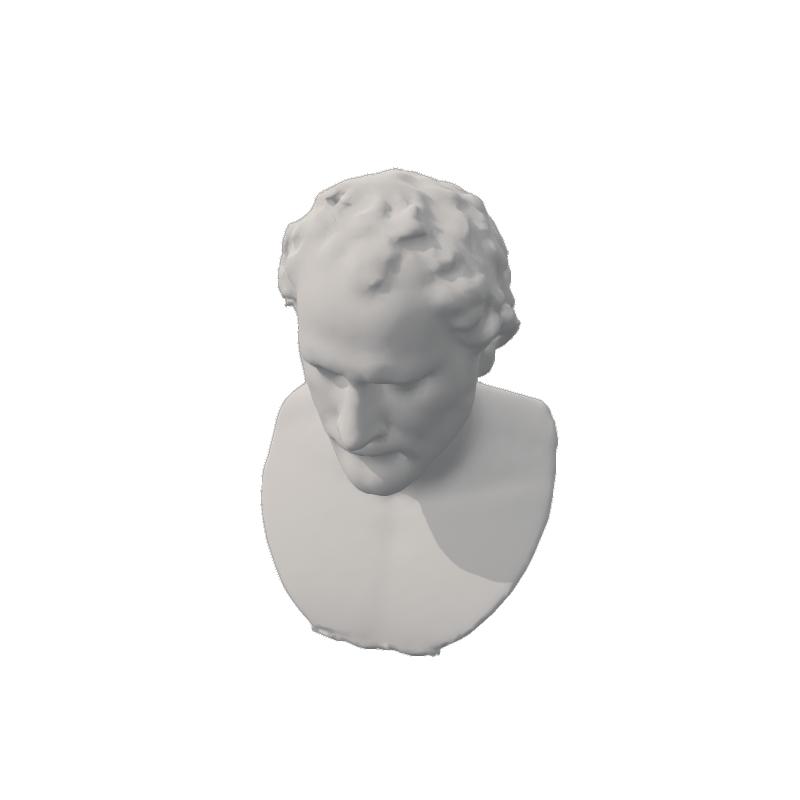} & \cntrnumber{21469} & \cntrnumber{10773} & \cntrdigit{3} & \cntrdigit{10} & \cntrdigit{5.99} \\

\cntrtxtmesh{falcon statue} & \meshthumb{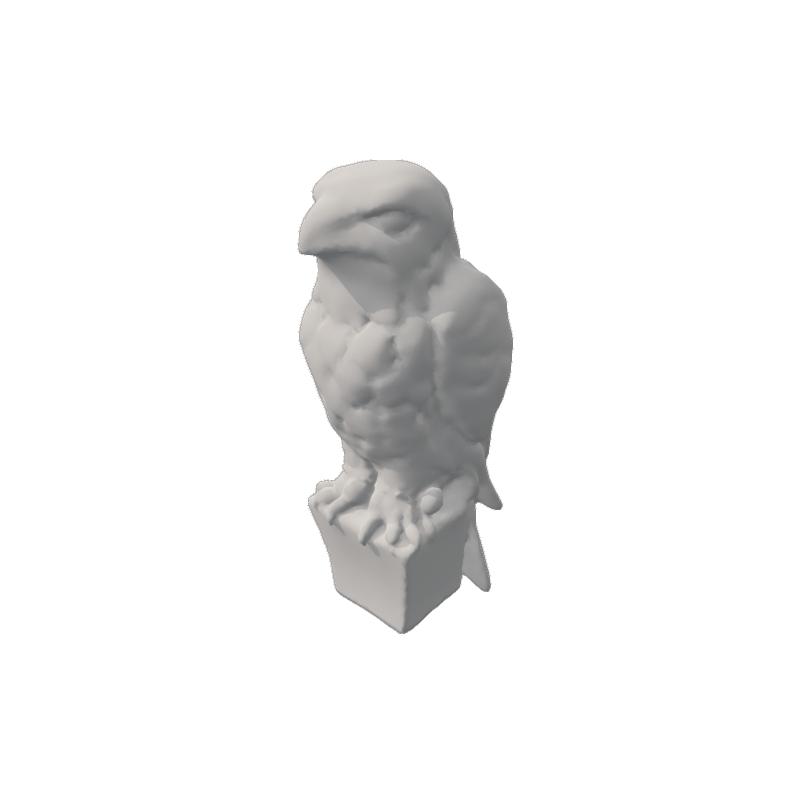} & \cntrnumber{21744} & \cntrnumber{10874} & \cntrdigit{3} & \cntrdigit{11} & \cntrdigit{6.0} &
\cntrtxtmesh{fish} & \meshthumb{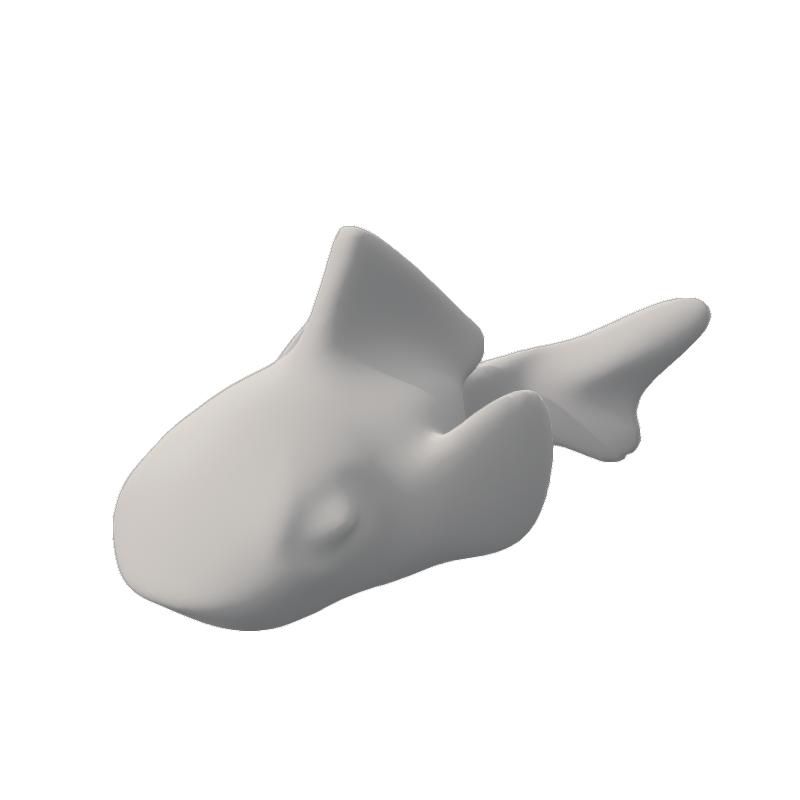} & \cntrnumber{21484} & \cntrnumber{10744} & \cntrdigit{4} & \cntrdigit{9} & \cntrdigit{6.0} \\

\cntrtxtmesh{goathead} & \meshthumb{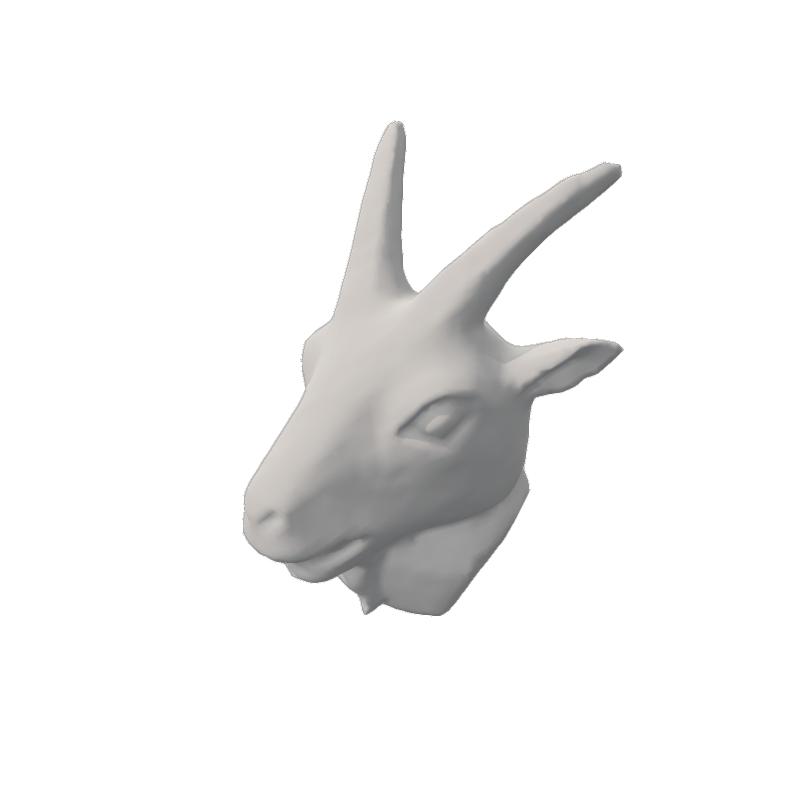} & \cntrnumber{18854} & \cntrnumber{9429} & \cntrdigit{4} & \cntrdigit{8} & \cntrdigit{6.0} &
\cntrtxtmesh{hand} & \meshthumb{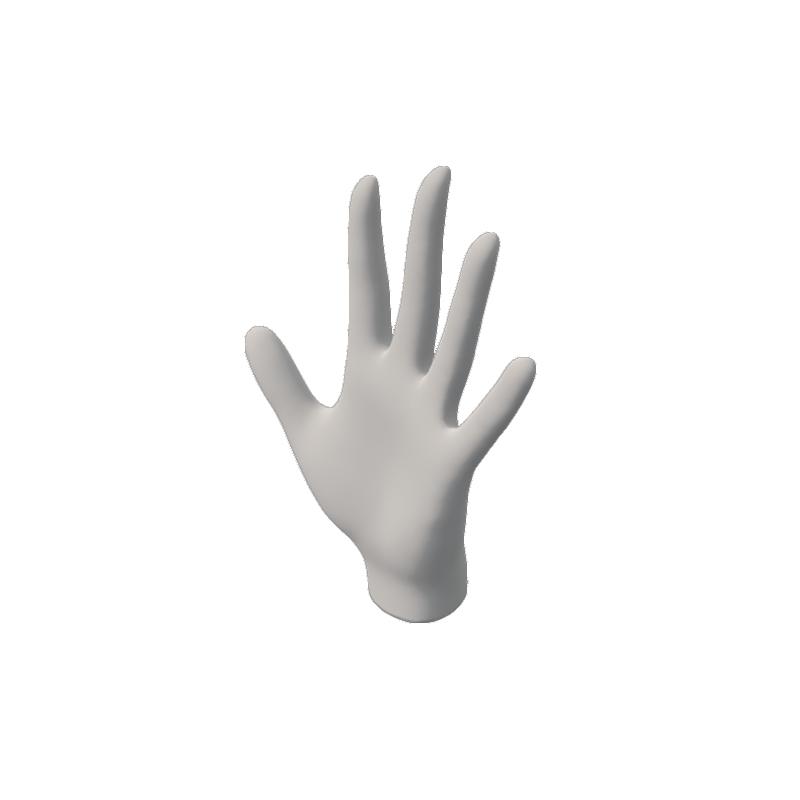} & \cntrnumber{21558} & \cntrnumber{10781} & \cntrdigit{5} & \cntrdigit{8} & \cntrdigit{6.0} \\

\cntrtxtmesh{human man} & \meshthumb{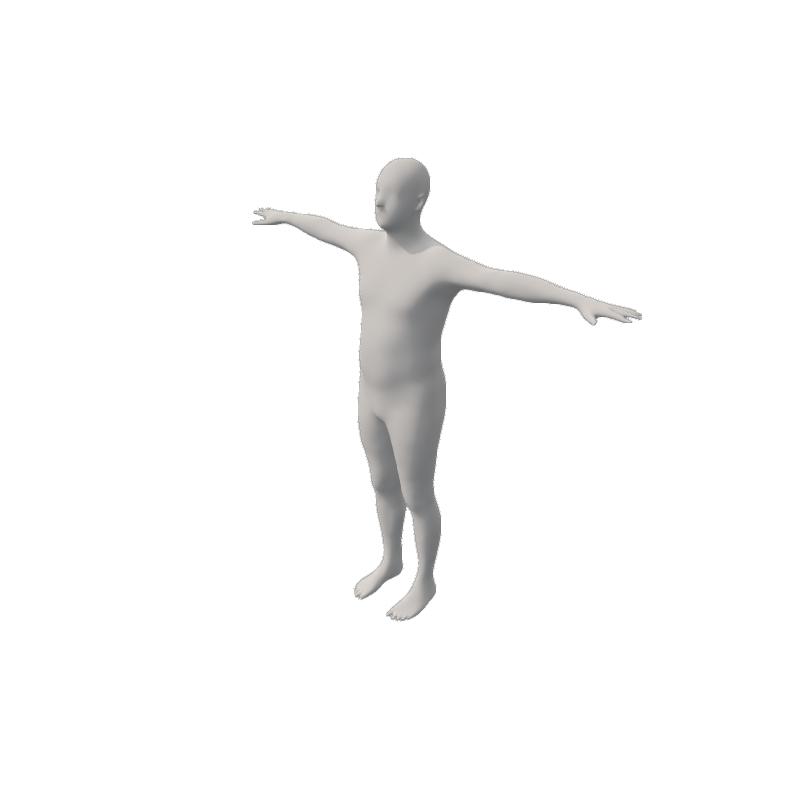} & \cntrnumber{22784} & \cntrnumber{11394} & \cntrdigit{3} & \cntrdigit{11} & \cntrdigit{6.0} &
\cntrtxtmesh{human woman} & \meshthumb{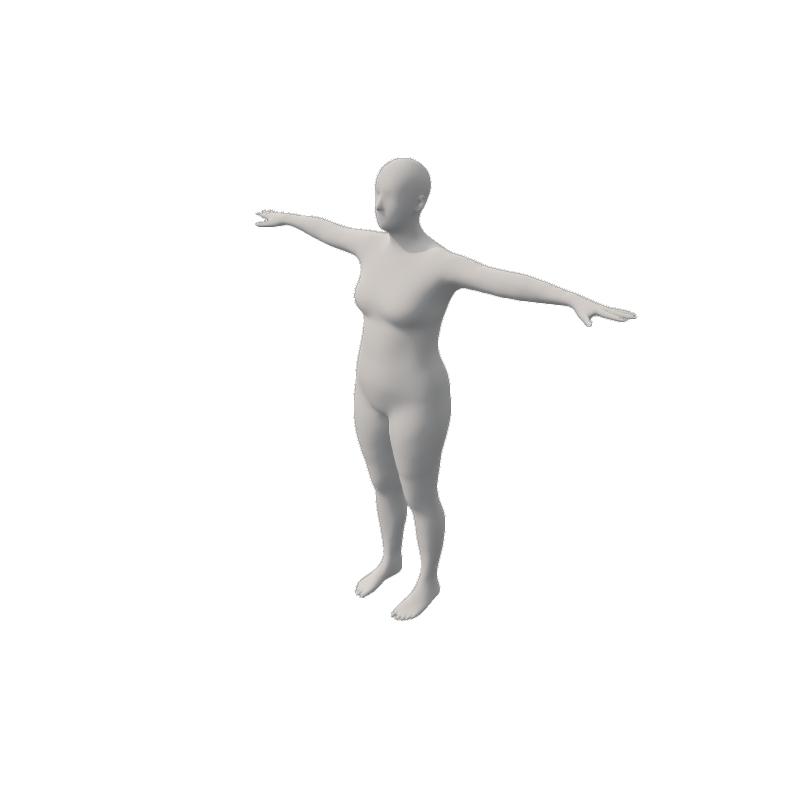} & \cntrnumber{23808} & \cntrnumber{11906} & \cntrdigit{3} & \cntrdigit{11} & \cntrdigit{6.0} \\

\cntrtxtmesh{koala} & \meshthumb{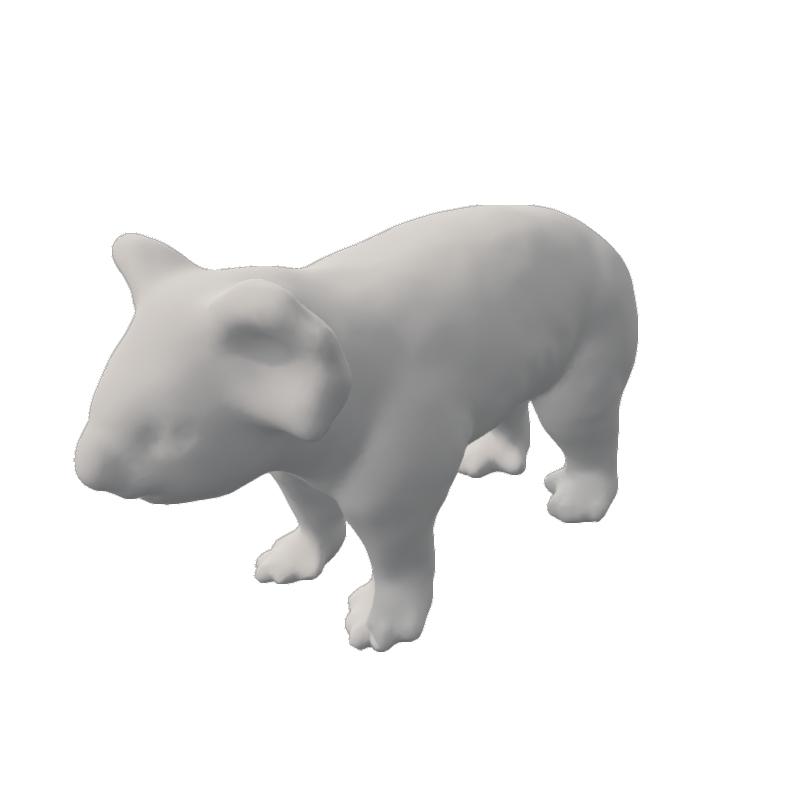} & \cntrnumber{19334} & \cntrnumber{9669} & \cntrdigit{4} & \cntrdigit{8} & \cntrdigit{6.0} &
\cntrtxtmesh{lion statue} & \meshthumb{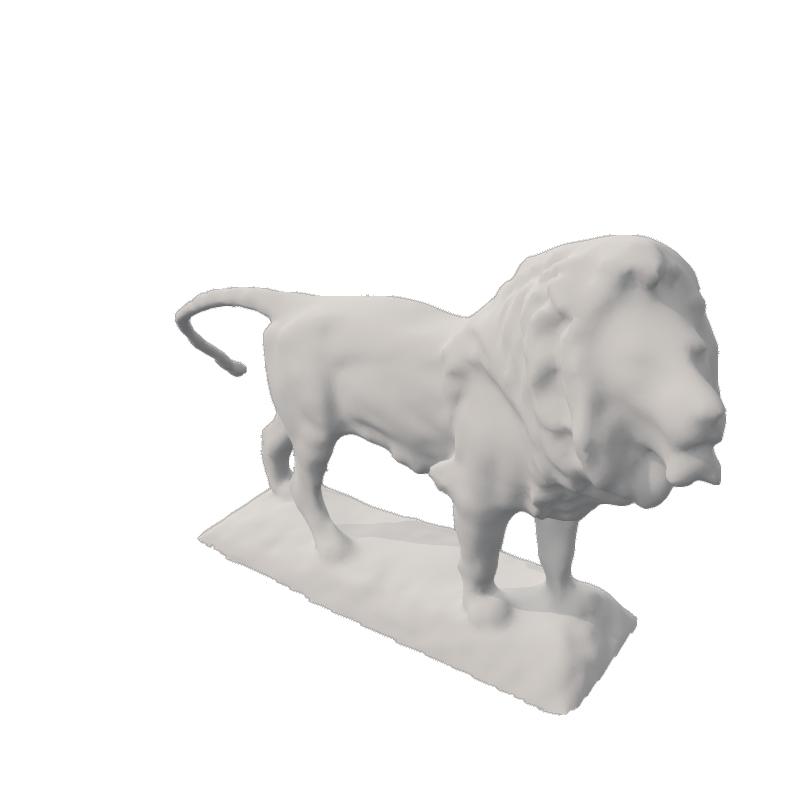} & \cntrnumber{27613} & \cntrnumber{13923} & \cntrdigit{2} & \cntrdigit{9} & \cntrdigit{5.97} \\

\cntrtxtmesh{mountain} & \meshthumb{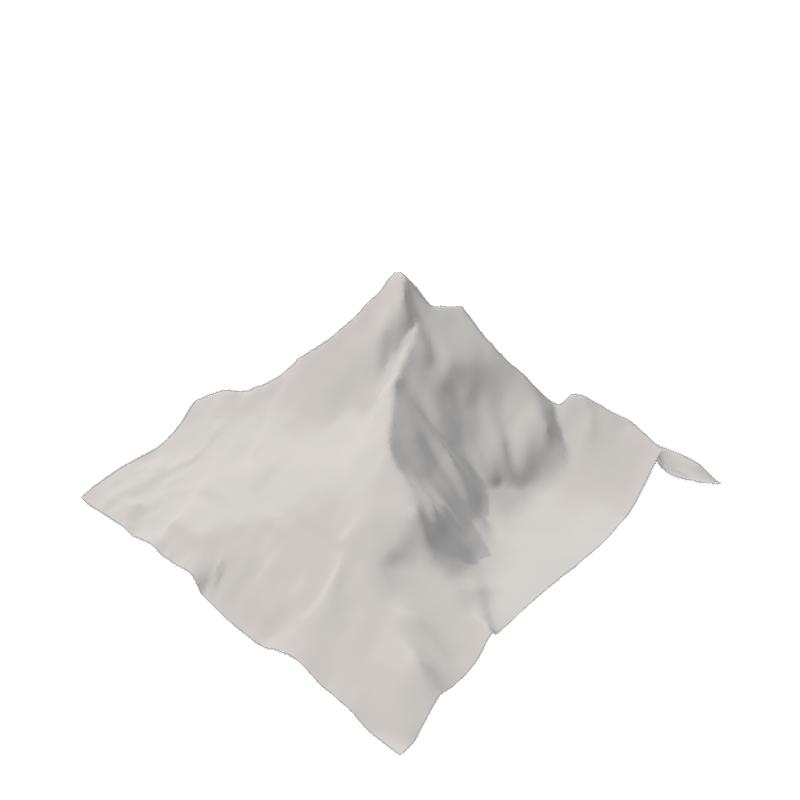} & \cntrnumber{26929} & \cntrnumber{13664} & \cntrdigit{2} & \cntrdigit{8} & \cntrdigit{5.94} &
\cntrtxtmesh{mushroom} & \meshthumb{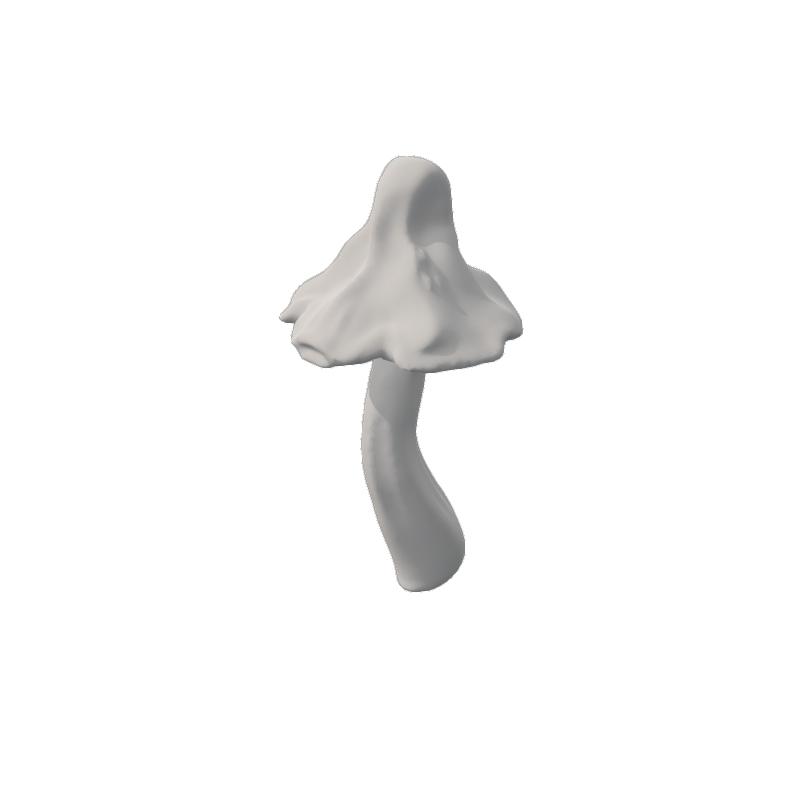} & \cntrnumber{22276} & \cntrnumber{11140} & \cntrdigit{4} & \cntrdigit{8} & \cntrdigit{6.0} \\

\cntrtxtmesh{nefertiti} & \meshthumb{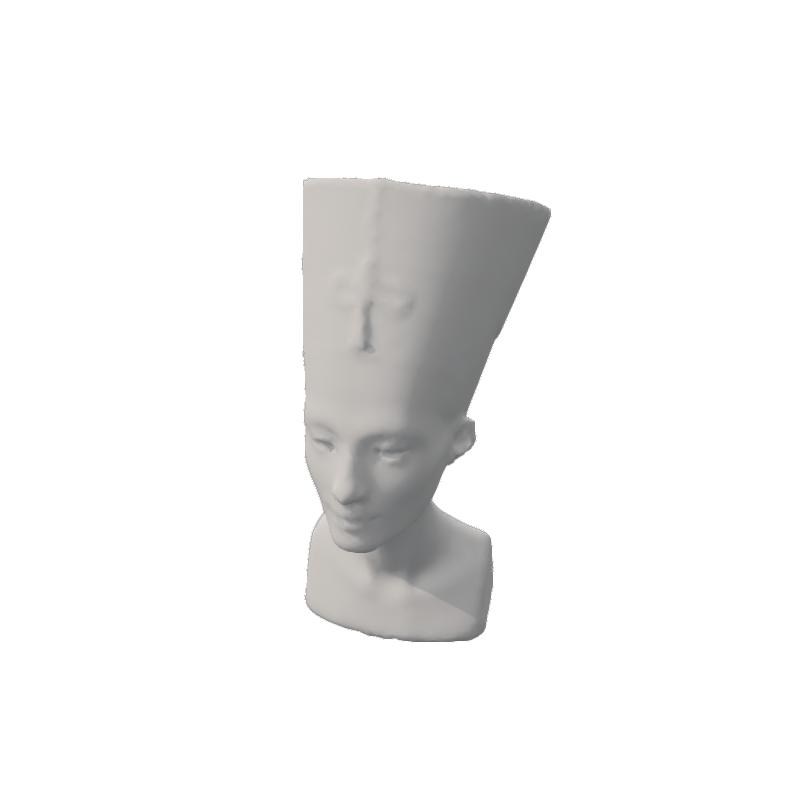} & \cntrnumber{18906} & \cntrnumber{9455} & \cntrdigit{3} & \cntrdigit{9} & \cntrdigit{6.0} &
\cntrtxtmesh{penguin} & \meshthumb{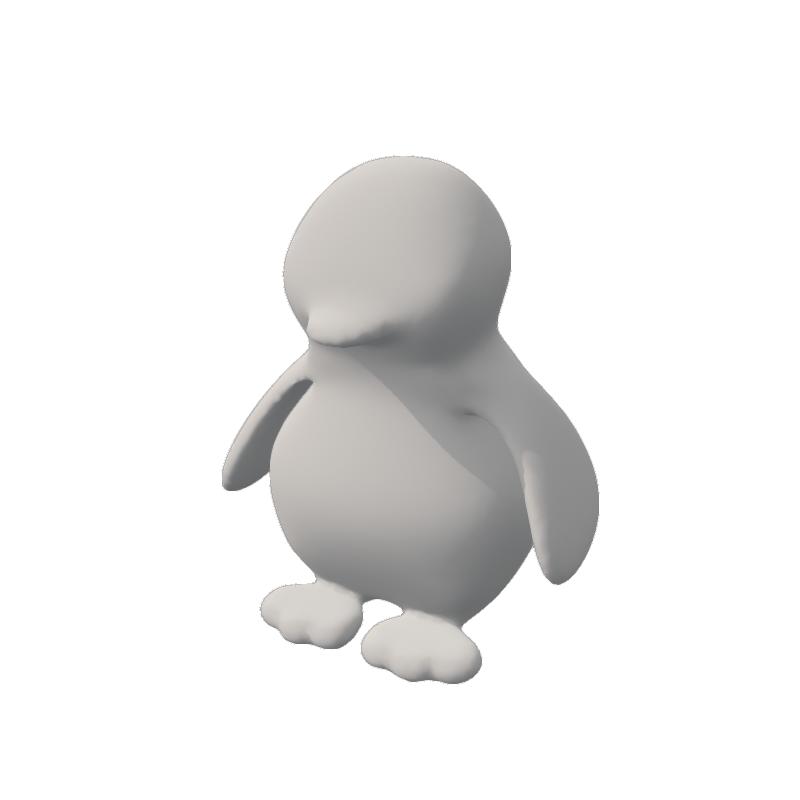} & \cntrnumber{23266} & \cntrnumber{11635} & \cntrdigit{5} & \cntrdigit{8} & \cntrdigit{6.0} \\

\cntrtxtmesh{chair} & \meshthumb{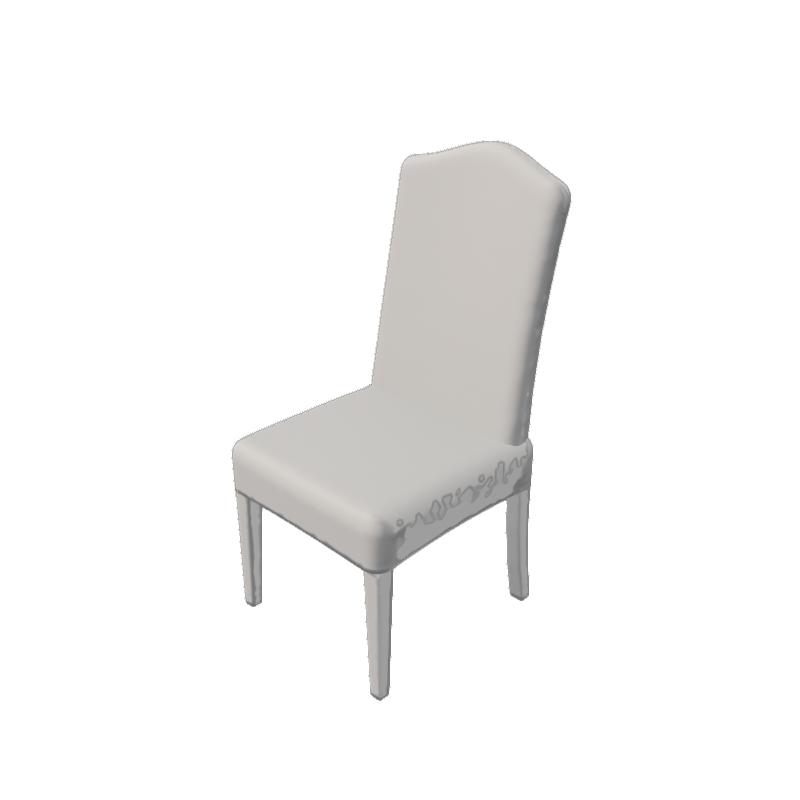} & \cntrnumber{20180} & \cntrnumber{10091} & \cntrdigit{4} & \cntrdigit{12} & \cntrdigit{6.0} &
\cntrtxtmesh{spot} & \meshthumb{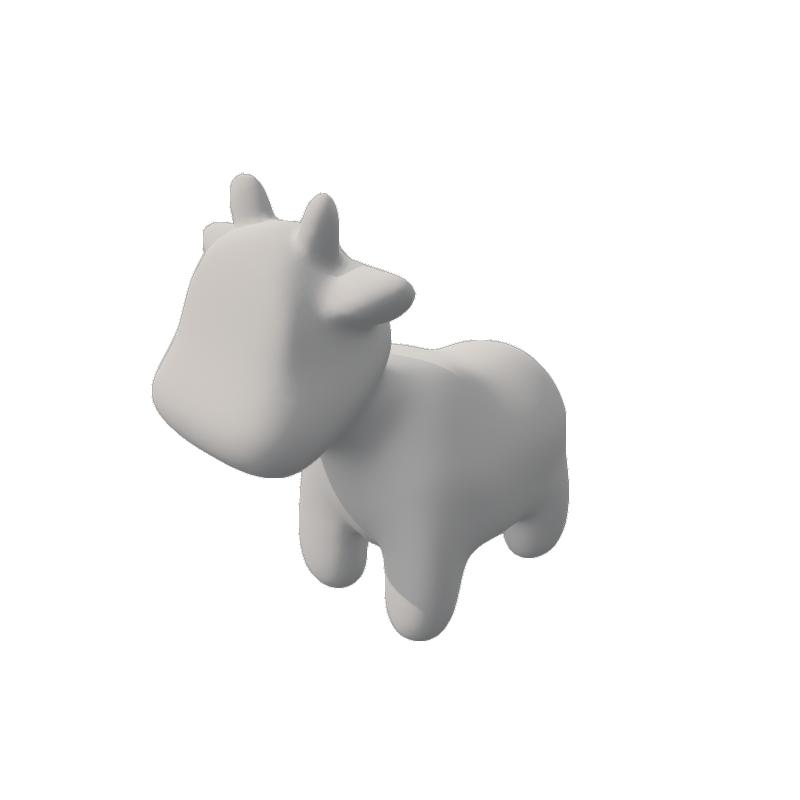} & \cntrnumber{20732} & \cntrnumber{10368} & \cntrdigit{3} & \cntrdigit{8} & \cntrdigit{6.0} \\

\cntrtxtmesh{strawberry} & \meshthumb{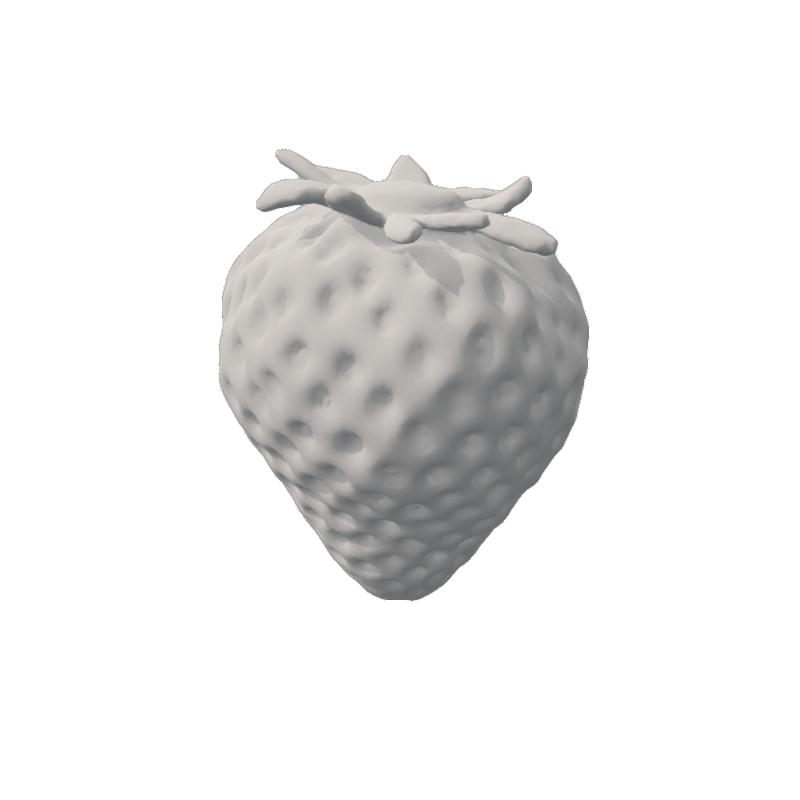} & \cntrnumber{25234} & \cntrnumber{12619} & \cntrdigit{3} & \cntrdigit{11} & \cntrdigit{6.0} &
\cntrtxtmesh{stuffed toy} & \meshthumb{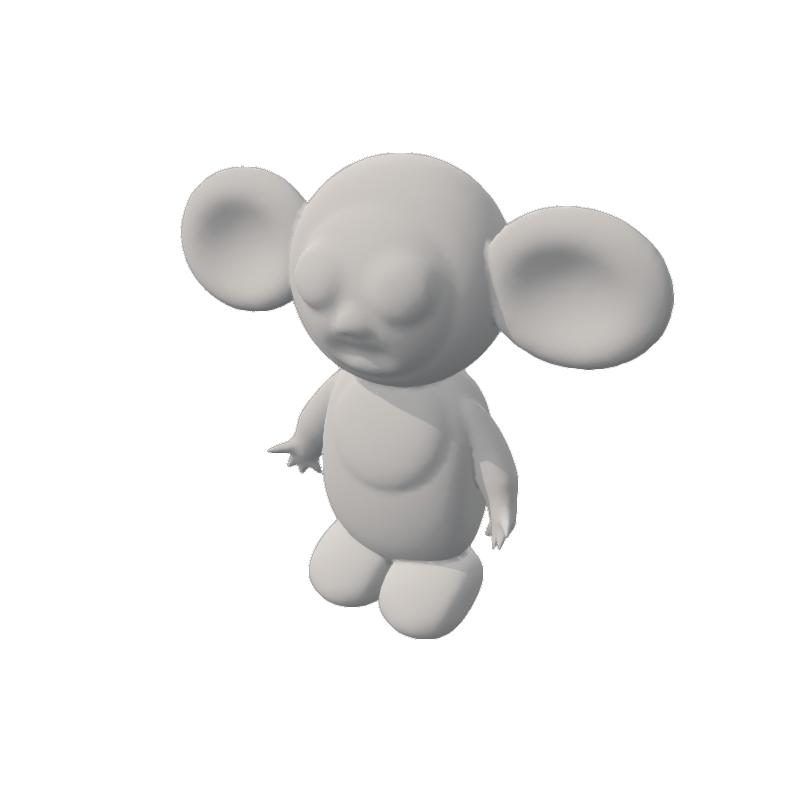} & \cntrnumber{20944} & \cntrnumber{10474} & \cntrdigit{3} & \cntrdigit{9} & \cntrdigit{6.0} \\

\cntrtxtmesh{vase} & \meshthumb{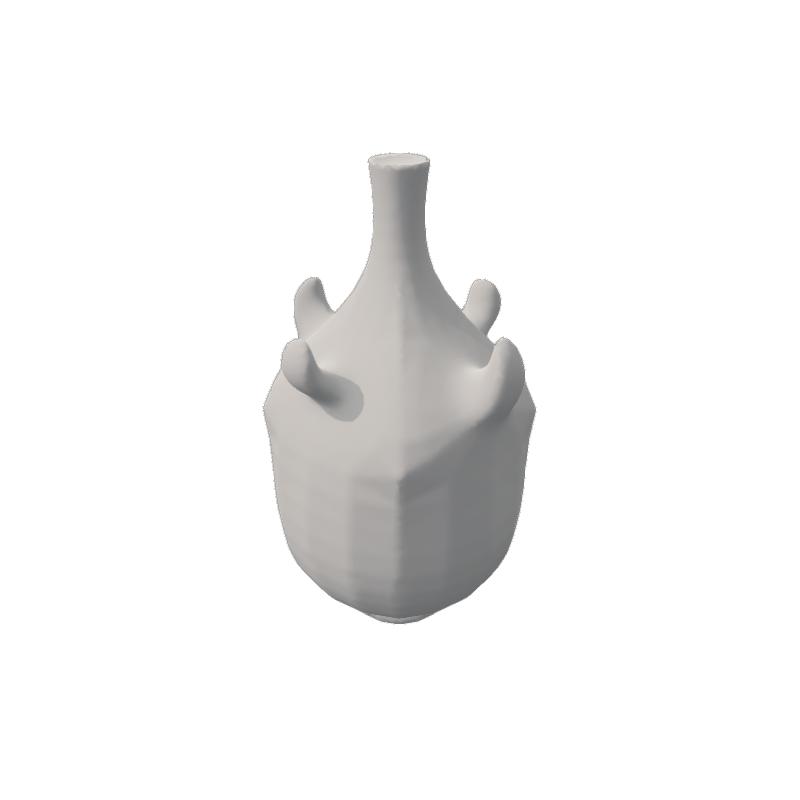} & \cntrnumber{22884} & \cntrnumber{11444} & \cntrdigit{4} & \cntrdigit{9} & \cntrdigit{6.0} &
\cntrtxtmesh{violin} & \meshthumb{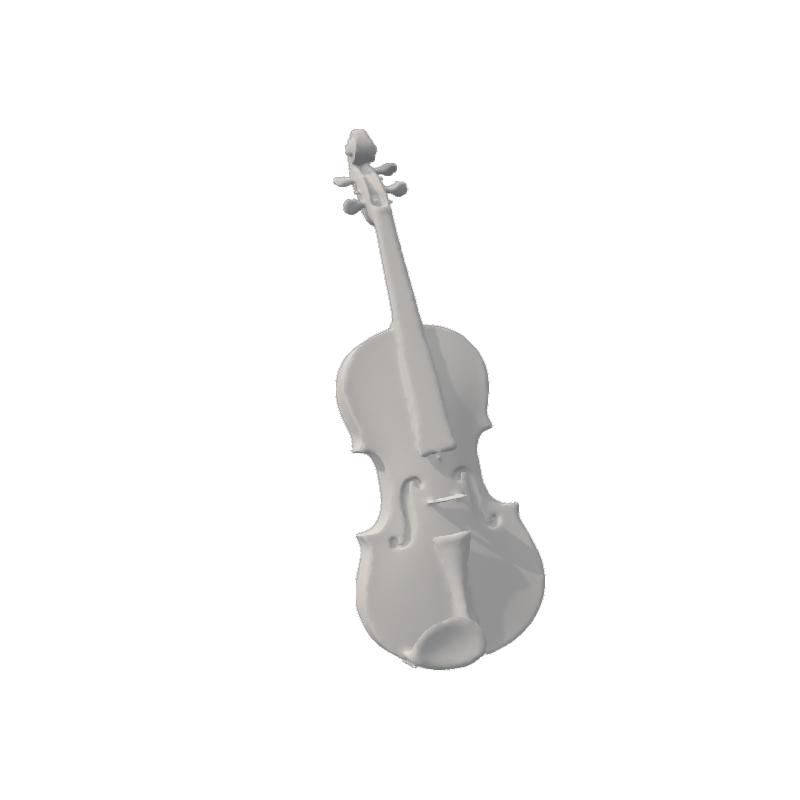} & \cntrnumber{21652} & \cntrnumber{10818} & \cntrdigit{3} & \cntrdigit{10} & \cntrdigit{6.0} \\

\cntrtxtmesh{wingnut} & \meshthumb{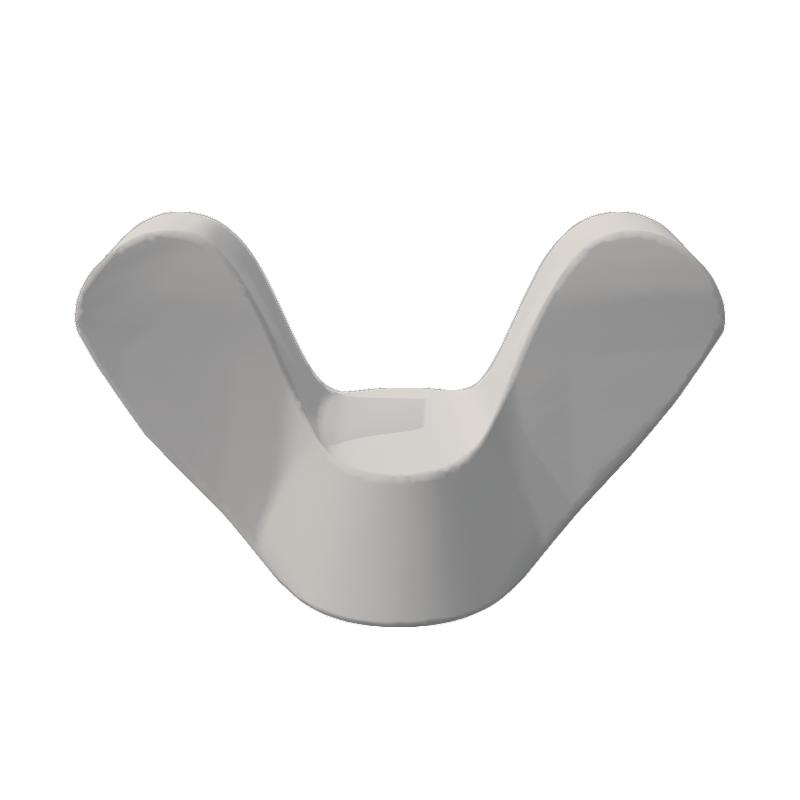} & \cntrnumber{21646} & \cntrnumber{10825} & \cntrdigit{5} & \cntrdigit{8} & \cntrdigit{6.0} &
 &  &  & &  &  & \\
\bottomrule
\end{tabular}
}
\vspace{5pt}
\label{tab:mesh_dataset}
\end{table*}

\section{Ablation Study}
\label{sec:exp-ablation}



\newcommand{\paramablimg}[1]{%
  \ifimgprefix
    \includegraphics[height=60pt]{figures/Experiments/param-abl/#1}%
  \else
    \includegraphics[height=60pt]{figures/Experiments/param-abl/L_#1}%
  \fi
}

\begin{table}[htbp]
\resizebox{\linewidth}{!}{
\begin{tabular}{c||cc}
\multirow{2}{*}{\textbf{Target Albedos}} & \multicolumn{2}{c}{\textbf{Generated Albedos}} \\
 & MeshNCA & MeshNCA-P \\
 \midrule
\paramablimg{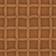}  &  \paramablimg{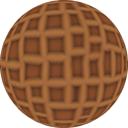}  &  \paramablimg{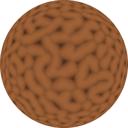} \\
\paramablimg{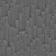}  &  \paramablimg{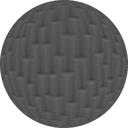}  &  \paramablimg{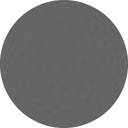} \\
\paramablimg{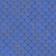}  &  \paramablimg{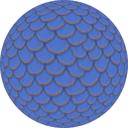}  &  \paramablimg{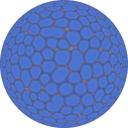}
\end{tabular}
}
\captionof{figure}{Ablation study on parametric \textit{Perception} scheme, denoted as MeshNCA-P. While the scheme integrates directional information during the \textit{Perception} stage, the model did not effectively harness this information, leading to degenerated results.}
\label{fig:param-abl}
\end{table}

\textbf{Parametric and Non-parametric Perception} Spherical Harmonic-based \textit{Perception} stage lies at the core of MeshNCA. To validate the necessity of our proposed method, we design a parametric perception stage and qualitatively compare the results with MeshNCA's. Specifically, the modified \textit{Perception} stage consists of an additional two-layer MLP, taking both features of the center and neighborhood nodes as input. Moreover, to incorporate direction information, the direction vectors between cells are also fed to the MLP. The hidden and output dimensionality is 32 and 64, respectively. The output dimensionality matches the one in our non-parametric \textit{Perception} stage, ensuring fair comparison. We denote the new model as MeshNCA-P. This configuration results in a 25\% increase in trainable parameters compared to MeshNCA. The comparison is given in Figure \ref{fig:param-abl}. MeshNCA-P fails to generate correct patterns while suffering from lower computational efficiency, hindering further applications.

\textbf{Degree of Spherical Harmonics} In our method, spherical harmonics of degrees 0 and 1 are used as the projection bases, considering both efficiency and quality. Here, we present the results in Figure \ref{fig:sh-abl} to demonstrate that the 0-order spherical harmonic is insufficient for generating anisotropic textures while the 2-order one does not show noticeable improvement but increases the size of the model.

\newcommand{\shablimg}[1]{%
  \ifimgprefix
    \includegraphics[height=45pt]{figures/Experiments/sh-abl/#1}%
  \else
    \includegraphics[height=45pt]{figures/Experiments/sh-abl/L_#1}%
  \fi
}

\begin{table}[htbp]
\resizebox{\linewidth}{!}{
\begin{tabular}{c||ccc} \multirow{2}{*}{\parbox[c]{2cm}{\centering\hspace{0pt} \textbf{Target \\ Albedos}}}  & 
 \multicolumn{3}{c}{\textbf{Generated Albedos}} \\
& MeshNCA (SH1) & SH0 & SH2 \\
 \midrule
\shablimg{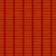}  &  \shablimg{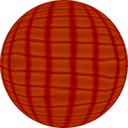}  &  \shablimg{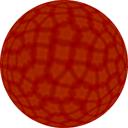}  &  \shablimg{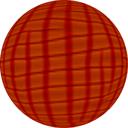} \\
 
     \shablimg{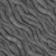}  &  \shablimg{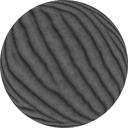}  &  \shablimg{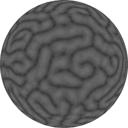}  &  \shablimg{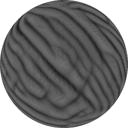} \\

\shablimg{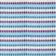}  &  \shablimg{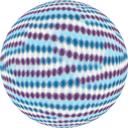}  &  \shablimg{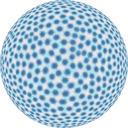}  &  \shablimg{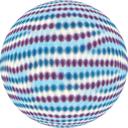}
\end{tabular}
}
\captionof{figure}{Ablation study on the degree of spherical harmonics. Zero-order spherical harmonics do not possess any direction information and can only synthesize isotropic textures, while the second-order ones do not show noticeable improvement compared to MeshNCA. }
\label{fig:sh-abl}
\end{table}

\textbf{Grafting scheme} We graft two MeshNCA models together to produce natural interpolated textures via the same-initialization training scheme. In Figure \ref{fig:graft-abl}, we demonstrate the necessity of our proposed graft training to generate coherent hybrid textures. Either directly training MeshNCA without proper initialization or naive linear interpolation fails to ensure a smooth transition between the two grafted textures.

\newcommand{\graftablimg}[1]{%
  \ifimgprefix
    \includegraphics[height=60pt]{figures/Experiments/graft-abl/#1}%
  \else
    \includegraphics[height=60pt]{figures/Experiments/graft-abl/L_#1}%
  \fi
}

\begin{table*}[htbp]
\centering
\resizebox{\textwidth}{!}{
\begin{tabular}{c|c||ccc} \multirow{2}{*}{\parbox[c]{2cm}{\centering\hspace{0pt} \textbf{Target \\ Albedos}}}  & \multirow{2}{*}{\parbox[c]{2cm}{\centering\hspace{0pt} \textbf{Texture Mask}}}  & 
 \multicolumn{3}{c}{\textbf{Interpolated Albedos}} \\
& & Same-Init & Diff-Init & Direct Interpolation \\
 \midrule
\graftablimg{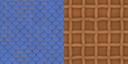}  & \graftablimg{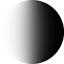}  &  \graftablimg{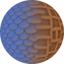}  &  \graftablimg{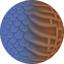}   & \graftablimg{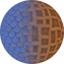} \\
\graftablimg{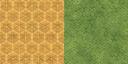}  &  \graftablimg{mask.jpg}  &  \graftablimg{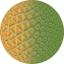}  &  \graftablimg{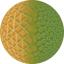}   & \graftablimg{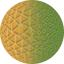} \\
\graftablimg{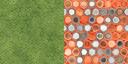}  &  \graftablimg{mask.jpg}  &  \graftablimg{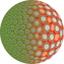}  &  \graftablimg{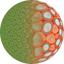}   & \graftablimg{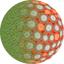} \\
\end{tabular}
}
\captionof{figure}{Ablation study on the graft training scheme. "Same-init" means that two MeshNCA models share the same initialization while "Diff-init" indicates different ones. Moreover, the naive linear interpolation between two textures is denoted as "Direct Interpolation". Our training scheme, "Same-init", produces coherent transition between two textures and results in a new hybrid texture, while the other two schemes generate sub-optimal combination of the two textures.}
\label{fig:graft-abl}
\end{table*}

\textbf{Motion Positional Encoding} Our MeshNCA is accompanied with \textbf{Motion Positional Encoding } (\textbf{MPE}) when performing dynamic texture synthesis. These cell-wise target dynamics guide each cell to move in the desired direction. A lack of such information can lead to sub-optimal dynamic textures. We conduct ablation studies on \textbf{MPE} and present the result in Figure \ref{fig:mpe-abl}.

\newcommand{\mpeablimg}[1]{%
  \ifimgprefix
    \includegraphics[height=42pt]{figures/Experiments/mpe-abl/#1}%
  \else
    \includegraphics[height=42pt]{figures/Experiments/mpe-abl/L_#1}%
  \fi
}

\begin{table}[htbp]
\begin{tabular}{cc||cc} \multicolumn{2}{c||}{\textbf{Target Dynamics}} & \multirow{2}{*}{\textbf{MPE}} & \multirow{2}{*}{\textbf{Plain}} \\
 Vectors & Projections & & \\
 \midrule
\mpeablimg{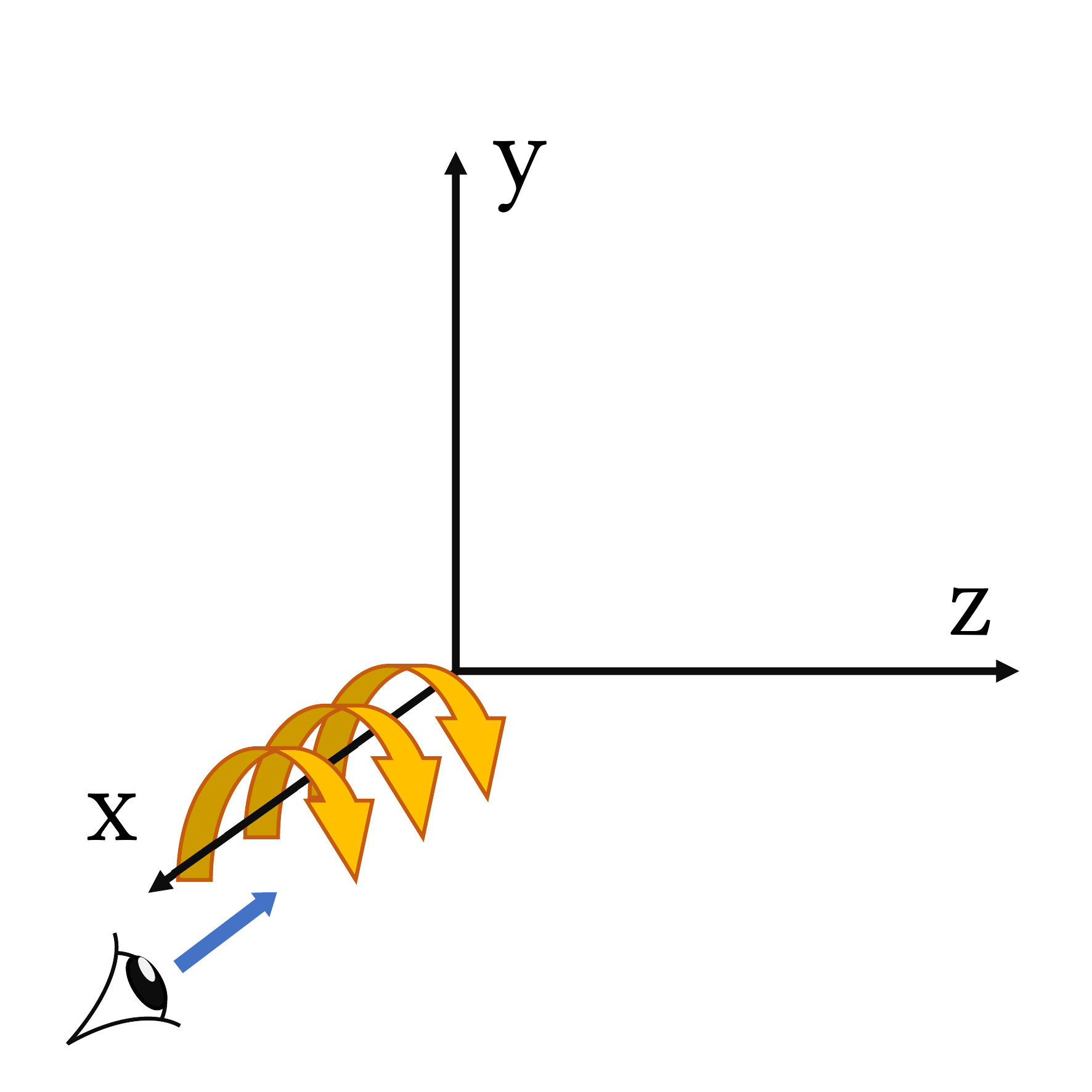}  &  \mpeablimg{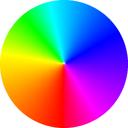}  &  \mpeablimg{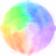}  &  \mpeablimg{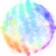} \\
\mpeablimg{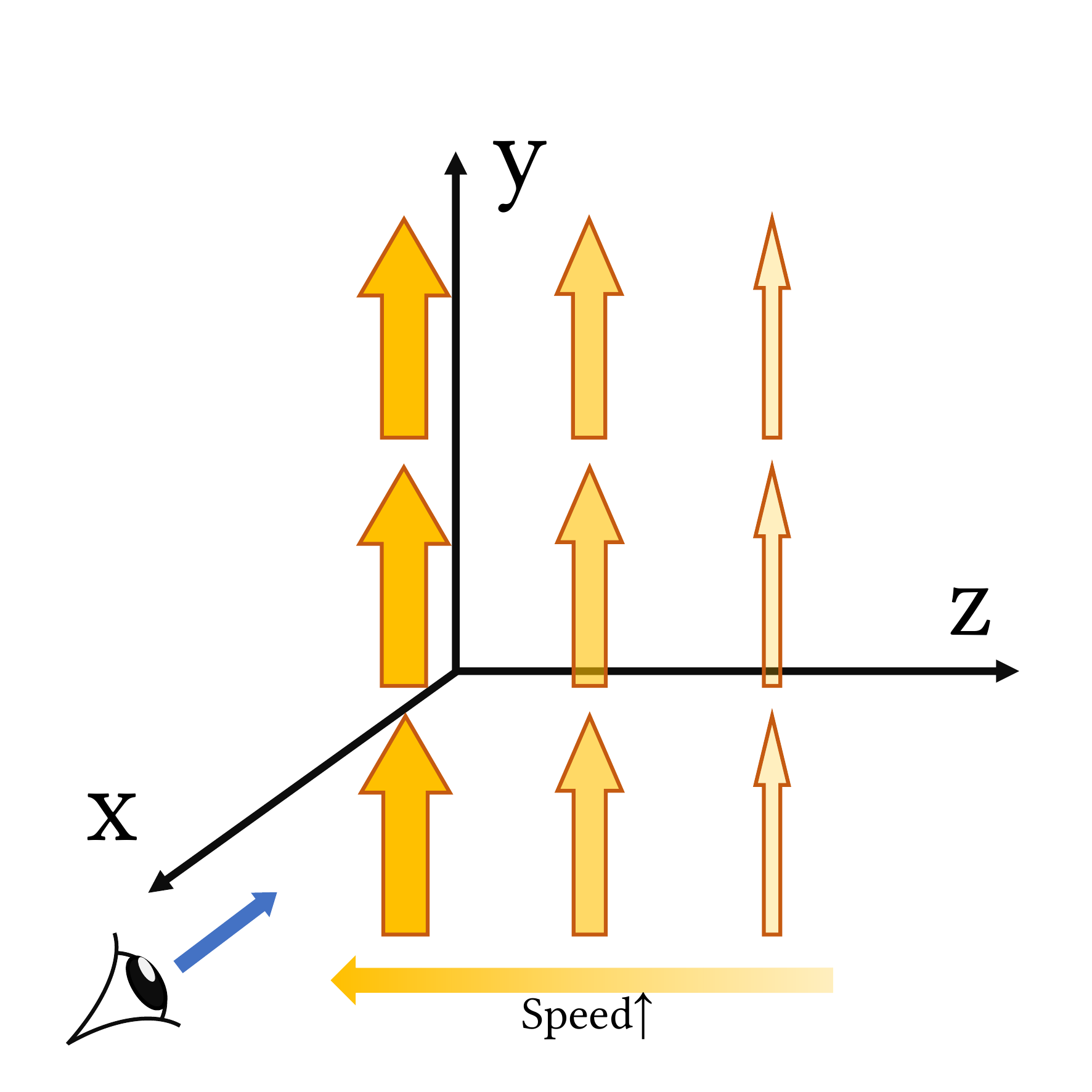}  &  \mpeablimg{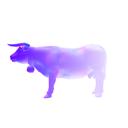}  &  \mpeablimg{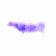}  &  \mpeablimg{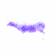} \\
\mpeablimg{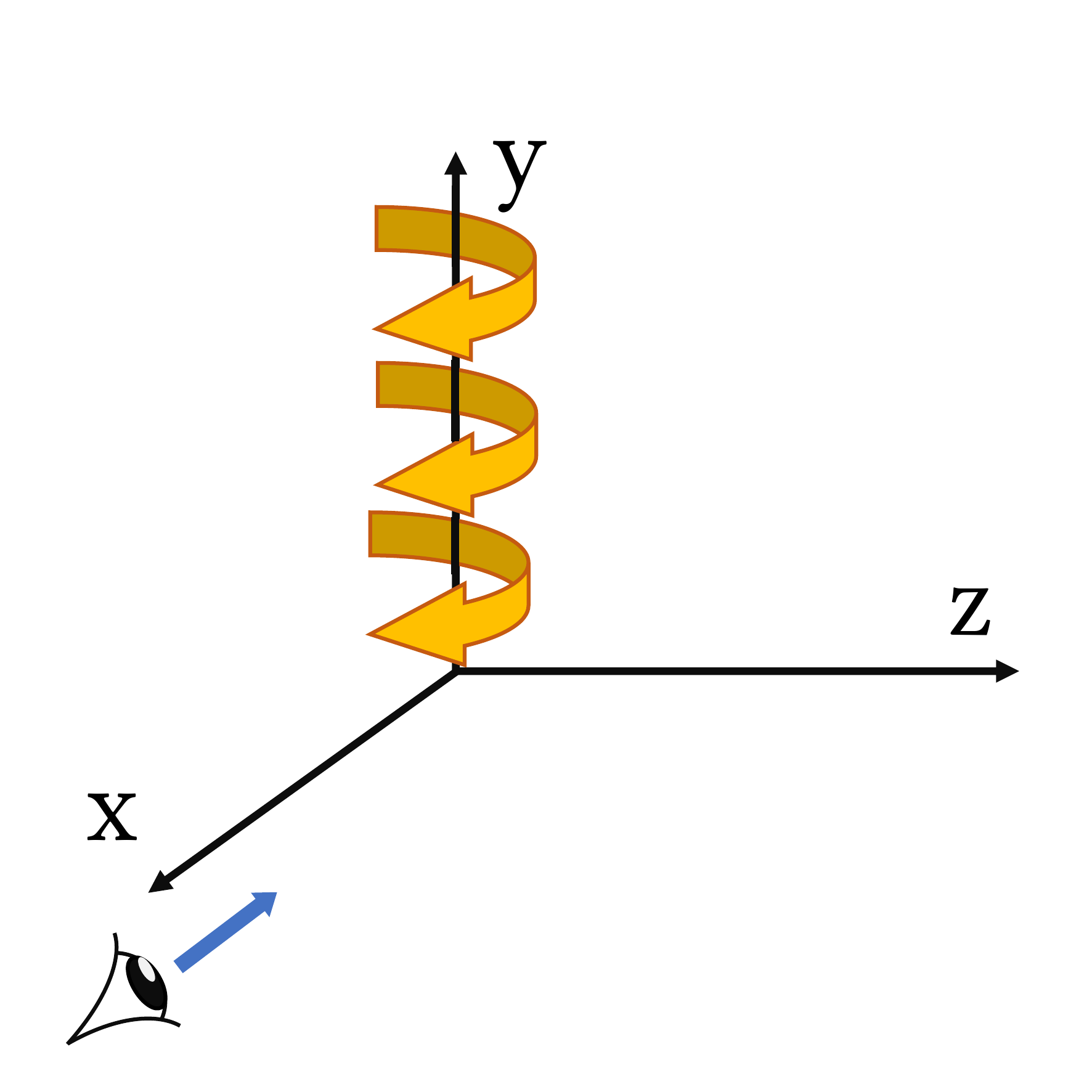}  &  \mpeablimg{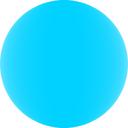}  &  \mpeablimg{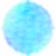}  &  \mpeablimg{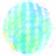} \\
\mpeablimg{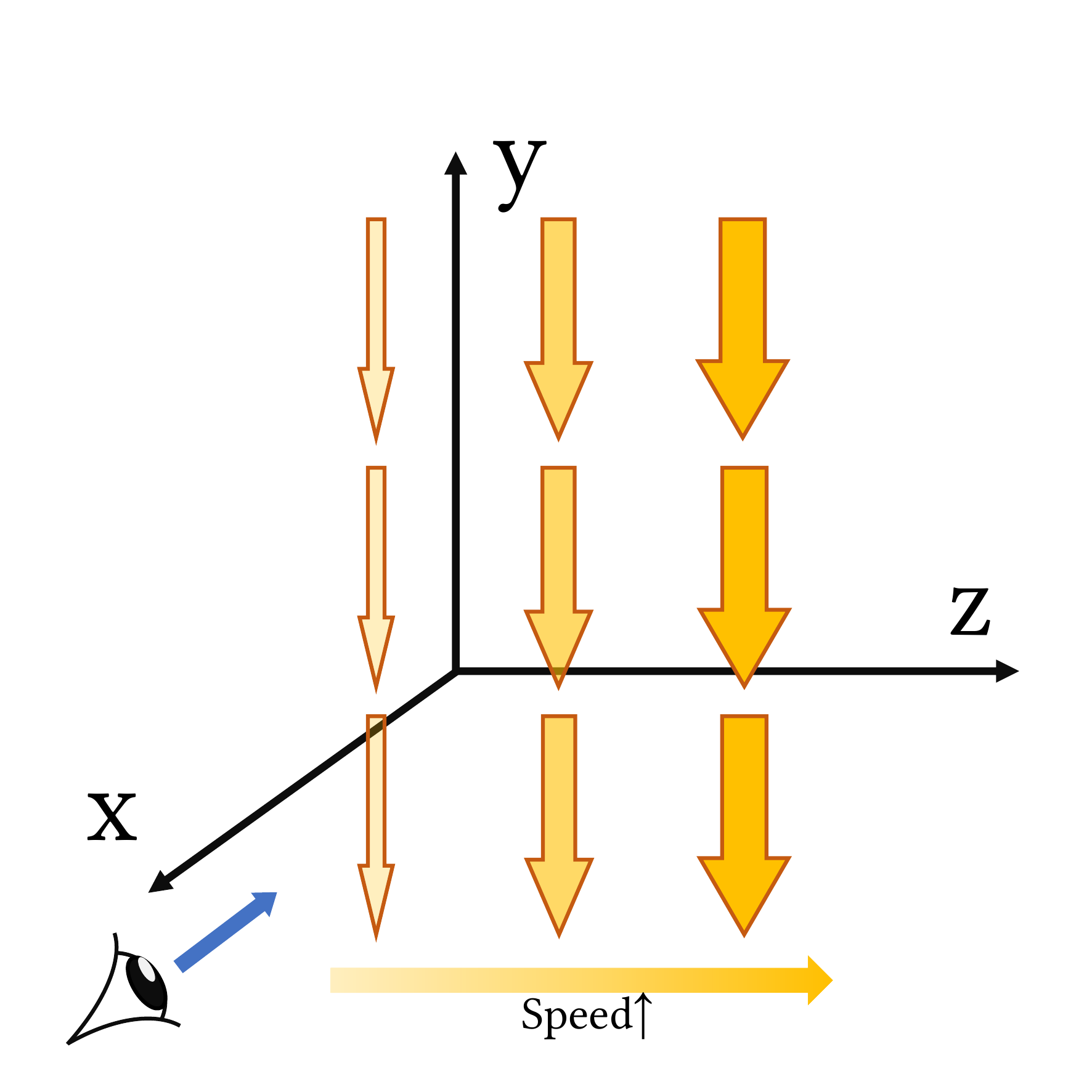}  &  \mpeablimg{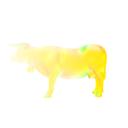}  &  \mpeablimg{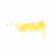}  &  \mpeablimg{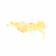}
\end{tabular}
\captionof{figure}{Ablation study on Motion Positional Encoding. The model without the encoding is denoted as \textbf{Plain}. The \textbf{Plain} model struggles to accurately capture intricate motions, including rotation or those exhibiting non-uniform speed changes across vertices. Such a limitation is exhibited both during the training on \icosphere and at test time with new meshes. The underlying textures of the motion demonstration are \textit{Sci-fi\_Wall\_010} and \textit{Sci-Fi\_Wall\_012} in our collected dataset. }
\label{fig:mpe-abl}
\end{table}

\textbf{CLIP Direction Matching} In the \textbf{Text-guided} synthesis, we extend the CLIP direction matching loss \cite{clip-direction-loss} to 3D domain. Here, we compare our scheme with directly optimizing the CLIP score as in previous text-guided mesh manipulation methods \cite{text2mesh, xmesh}. The results are given in Figure \ref{fig:clip-dir-abl}. We can conclude that if directly optimizing the CLIP score, the method can sometimes yield sub-optimal results. Therefore, to ensure more stable training, we choose the direction-matching scheme as introduced in the main paper.

\newcolumntype{H}{>{\centering\arraybackslash} m{25pt} }

\newcommand{\imgclipdirabl}[1]{%
  \ifimgprefix
    \includegraphics[height=50pt]{figures/Experiments/clip-dir-abl/#1}%
  \else
    \includegraphics[height=50pt]{figures/Experiments/clip-dir-abl/#1}%
  \fi
}

\begin{table*}[htbp]
\centering
\resizebox{\textwidth}{!}{
\begin{tabular}{c|cc||c|cc} \textbf{Prompts}  & \textbf{CLIP direction (Ours)} & \textbf{CLIP} & \textbf{Prompts}  & \textbf{CLIP direction (Ours)} & \textbf{CLIP}\\
 \midrule
\rotatebox{90}{\parbox[c]{2cm}{\centering\hspace{0pt} \textbf{Stained Glass}}}  &  \imgtext{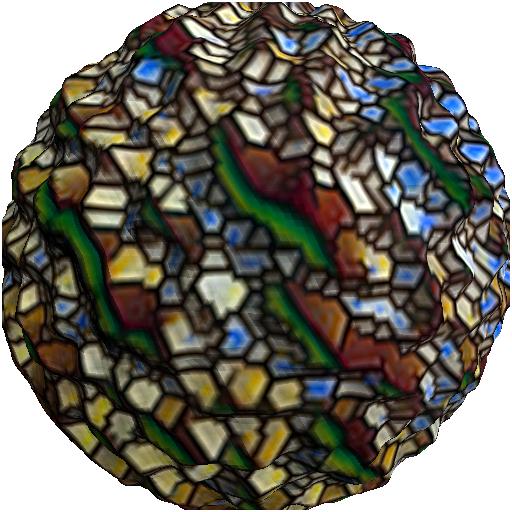}  &  \imgclipdirabl{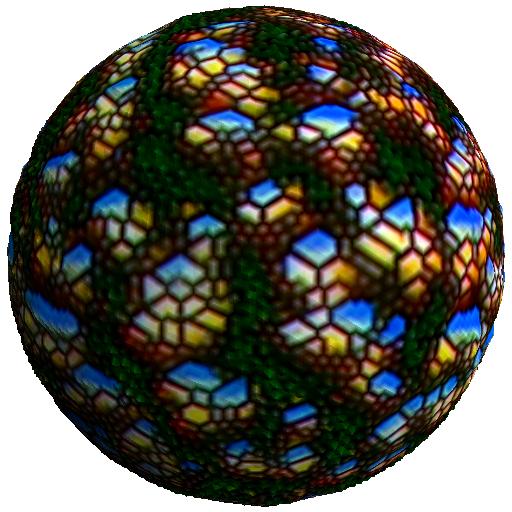} & \rotatebox{90}{\parbox[c]{2cm}{\centering\hspace{0pt} \textbf{Cactus}}}  &  \imgtext{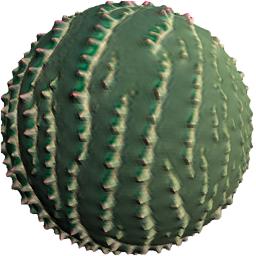}  &  \imgclipdirabl{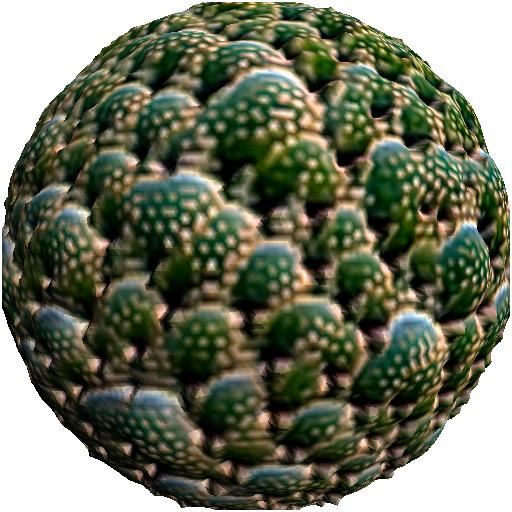} \\
\rotatebox{90}{\parbox[c]{2cm}{\centering\hspace{0pt} \textbf{Marble}}}  &  \imgtext{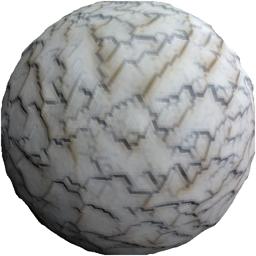}  &  \imgclipdirabl{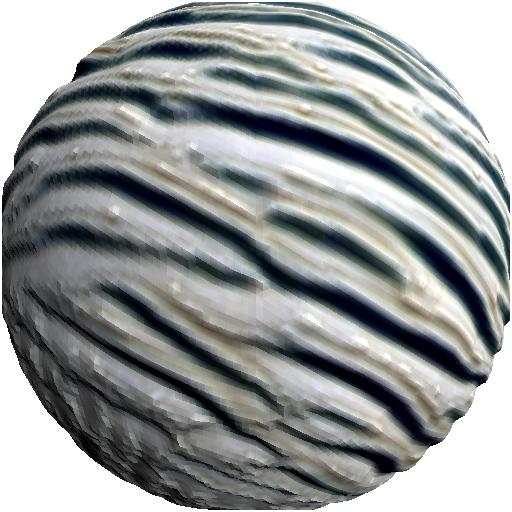} & \rotatebox{90}{\parbox[c]{2cm}{\centering\hspace{0pt} \textbf{Bark}}}  &  \imgtext{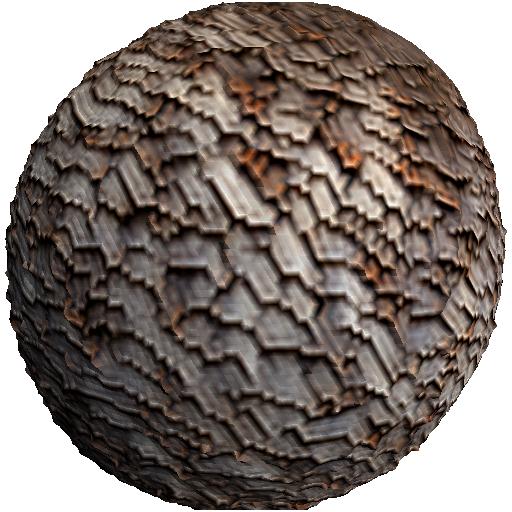}  &  \imgclipdirabl{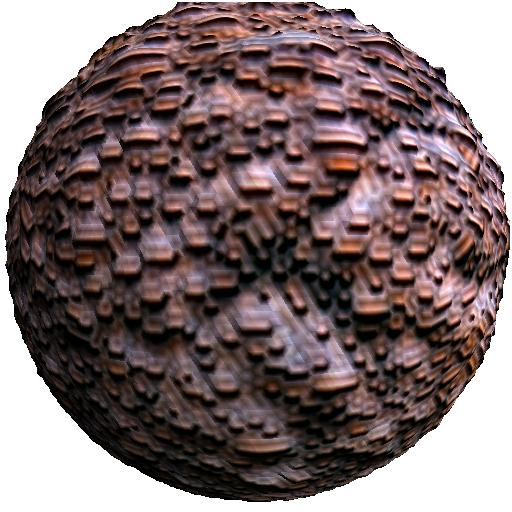}  \\
\rotatebox{90}{\parbox[c]{2cm}{\centering\hspace{0pt} \textbf{Feathers}}}  &  \imgtext{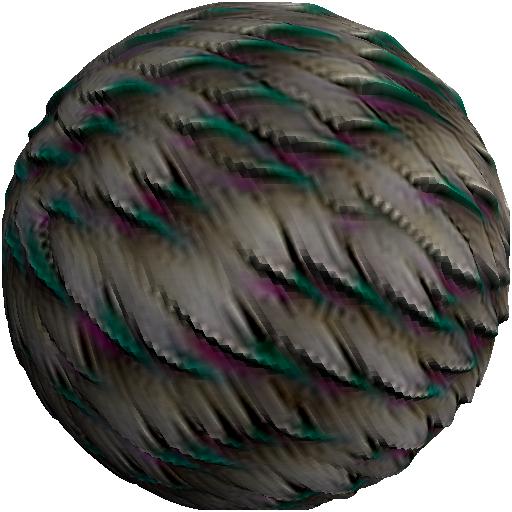}  &  \imgclipdirabl{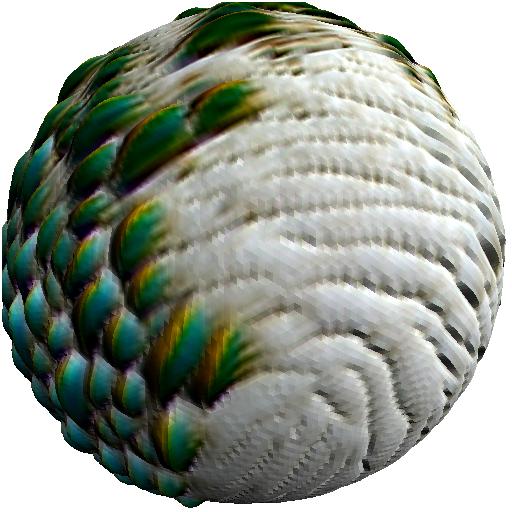} & \rotatebox{90}{\parbox[c]{2cm}{\centering\hspace{0pt} \textbf{Patchwork Leather}}}  &  \imgtext{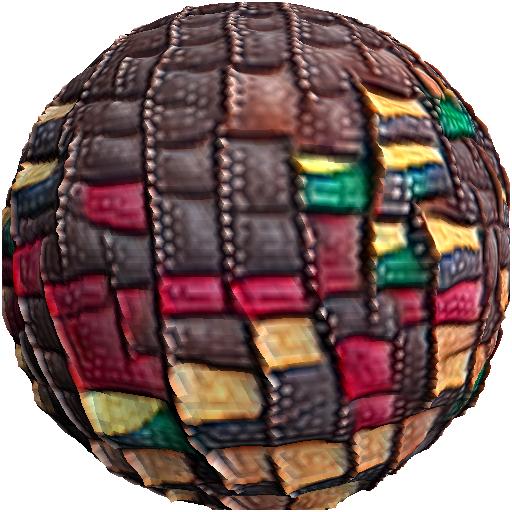}  &  \imgclipdirabl{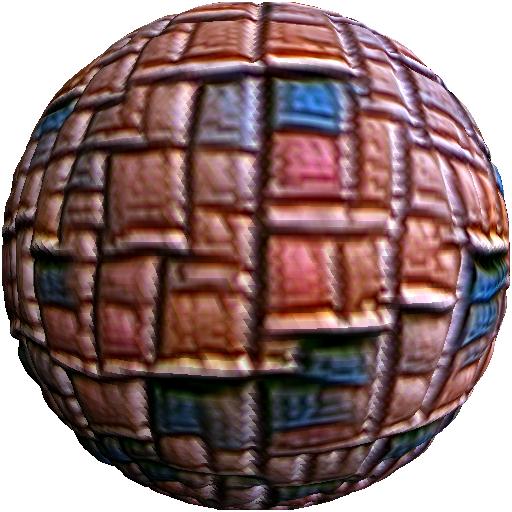} \\
\end{tabular}
}
\captionof{figure}{Ablation study on CLIP direction matching loss in 3D domain. For \textit{stained glass, marble, feathers}, the direct optimization yields sub-optimal results. For \textit{cactus, bark, patchwork leather}, the direct optimization gives satisfactory but another interpretation of the input prompt.}
\label{fig:clip-dir-abl}
\end{table*}
    
\begin{figure*}[ht]
    \centering
    \begin{subfigure}[b]{0.45\textwidth}
        \centering
        \includegraphics[width=\textwidth]{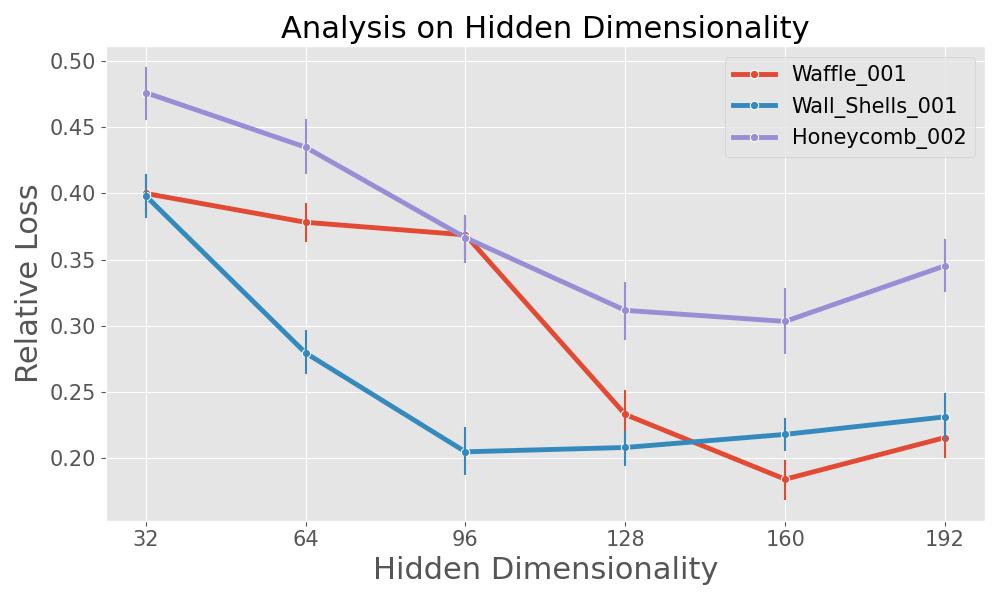}
        \caption{\revv{Ablation study on the hidden dimensionality of the MLP in the MeshNCA \textit{Adaptation} stage. The experiment suggests that a dimensionality of 128 produces a sufficiently small loss.}}
        \label{fig:mlp-abl}
    \end{subfigure}
    \begin{subfigure}[b]{0.45\textwidth}
        \centering
        \includegraphics[width=\textwidth]{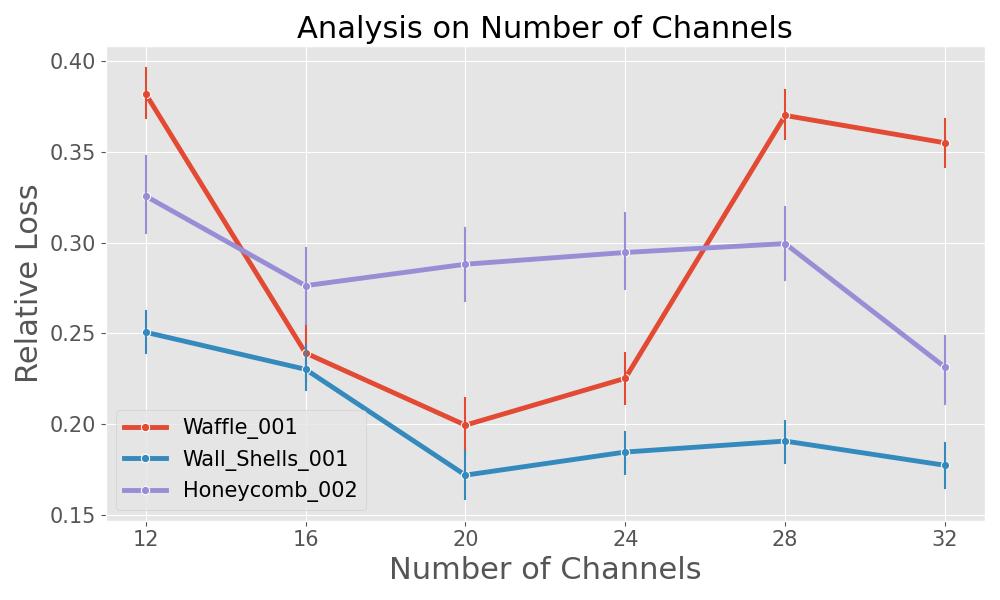}
        \caption{\revv{Ablation study on the number of channels of the cell states. The experiment suggests that choosing 16 as the number of channels leads to a sufficiently small loss.}}
        \label{fig:chn-abl}
    \end{subfigure}
    \caption{\revv{Ablation studies on MLP hyperparameters. Error bar is 95 CI.}}
    \label{fig:mlp-alb-chn}
\end{figure*}

\textbf{Hyperparameters: Number of Hidden Neurons} 
\rev{
MeshNCA parametrizes the \textit{Adaptation} stage using a two-layer MLP. We conduct an ablation study on the hidden dimensionality of this MLP (number of hidden neurons in the output of the first layer). Our goal is to ensure the training sufficiently converges while keeping the model as small as possible. Therefore, we set the hidden dimensionality of the MLP to [32, 64, 96, 128, 160, 192] and train on three textures, \revv{\textit{Waffle\_001, Wall\_Shells\_001, Honeycomb\_002}}, picked from our PBR texture dataset. After training, we record the test loss during video generation for 30 steps, after the model evolves for 50 steps, which is after the model becomes stable. \revv{For each configuration, the loss is deducted by the minimum value to normalize all loss values into similar ranges.} Figure \ref{fig:mlp-abl} shows the results.}

\newcommand{\mlpalbimg}[1]{%
  \ifimgprefix
    \includegraphics[height=90pt]{figures/Experiments/MLP-abl/#1}%
  \else
    \includegraphics[height=90pt]{figures/Experiments/MLP-abl/#1}%
  \fi
}

\rev{Error bar is 95 CI. \revv{The experiment suggests that a dimensionality of 128 produces a sufficiently small loss}. There is no statistically significant difference between the loss recorded on the model with 128, 160, and 196 dimensionalities, as shown by the fact that the pair-wise independent statistical tests all have p > 0.1. Although the choice of 96 appears to be acceptable, we observe more failure cases, such as wrongly aligned texture maps or failed synthesis, than the choice of 128. Therefore, we adopt 128 in all of our experiments.}

\textbf{Hyperparameters: Number of Channels}
\rev{The number of channels of the cell states decides the amount of information that cells can store. We conduct an ablation study on this hyperparameter by setting it to [12, 16, 20, 24, 28, 32] and record the test loss during video generation for 30 steps after the model becomes stable, which corresponds to 50 steps. \revv{For each configuration, the loss is deducted by the minimum value to normalize all loss values into similar ranges.} Results are shown in Figure \ref{fig:chn-abl}. }
\rev{Error bar is 95 CI. Although in the case of \textit{Wall\_Shells\_001}, choosing 20 statistically has a lower test loss, the absolute decrease is not too much compared to the other two textures. Therefore, we choose 16 in the experiments.
}

\begin{table}[htbp]
\resizebox{\linewidth}{!}{
\begin{tabular}{cc||cc} \multicolumn{2}{c||}{\textbf{Target Dynamics}} & \multirow{2}{*}{\textbf{MPE}} & \multirow{2}{*}{\textbf{Plain}} \\
 Vectors & Projections & & \\
 \midrule
\mpeablimg{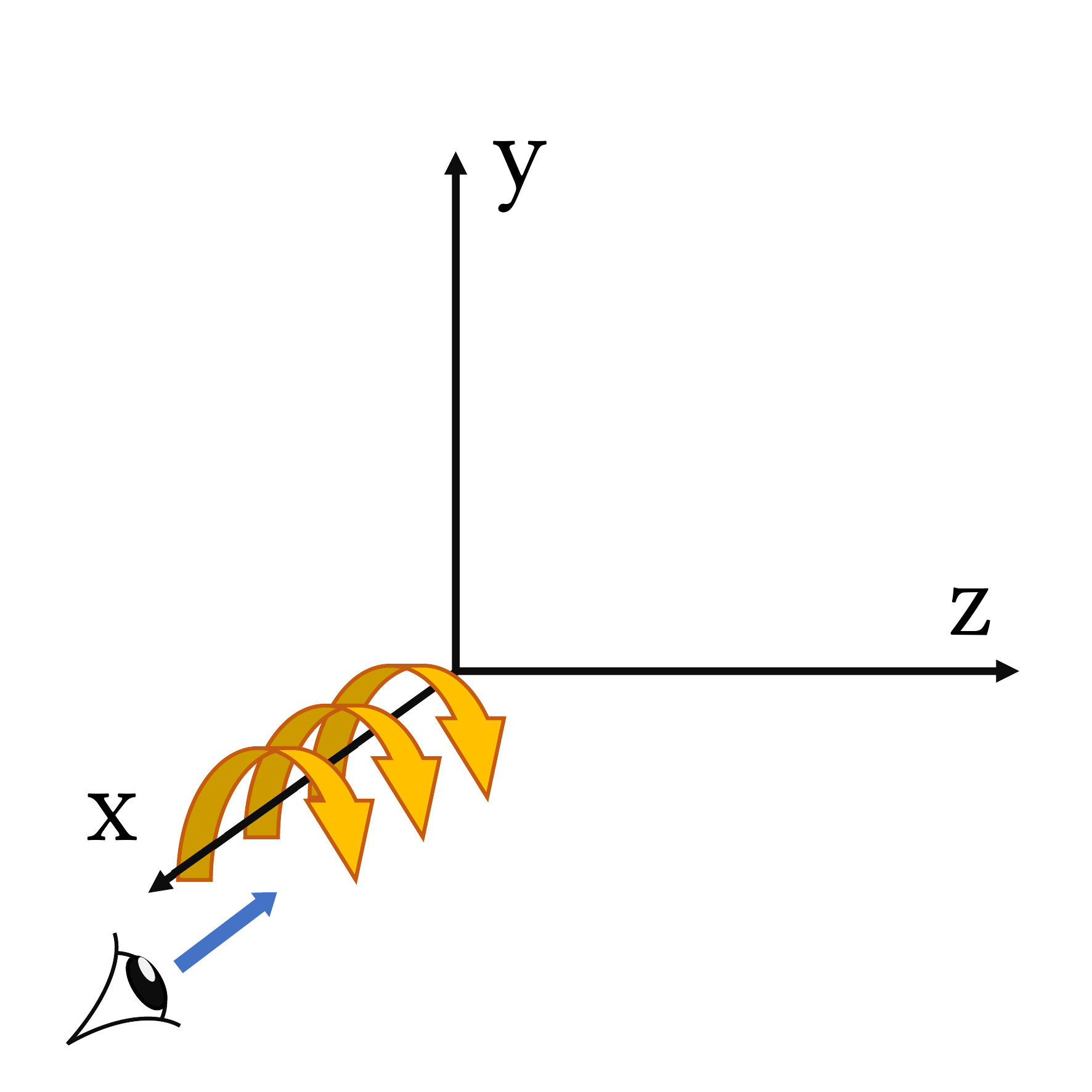}  &  \mpeablimg{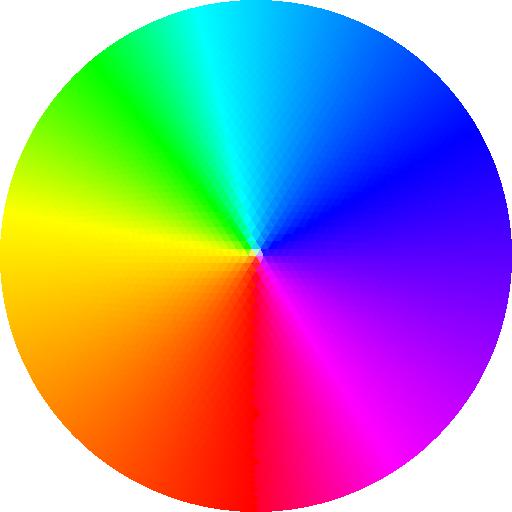}  &  \mpeablimg{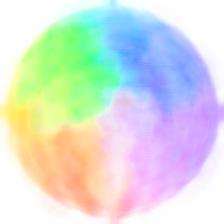}  &  \mpeablimg{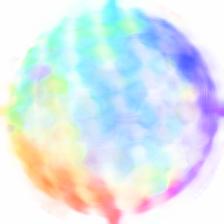} \\
\mpeablimg{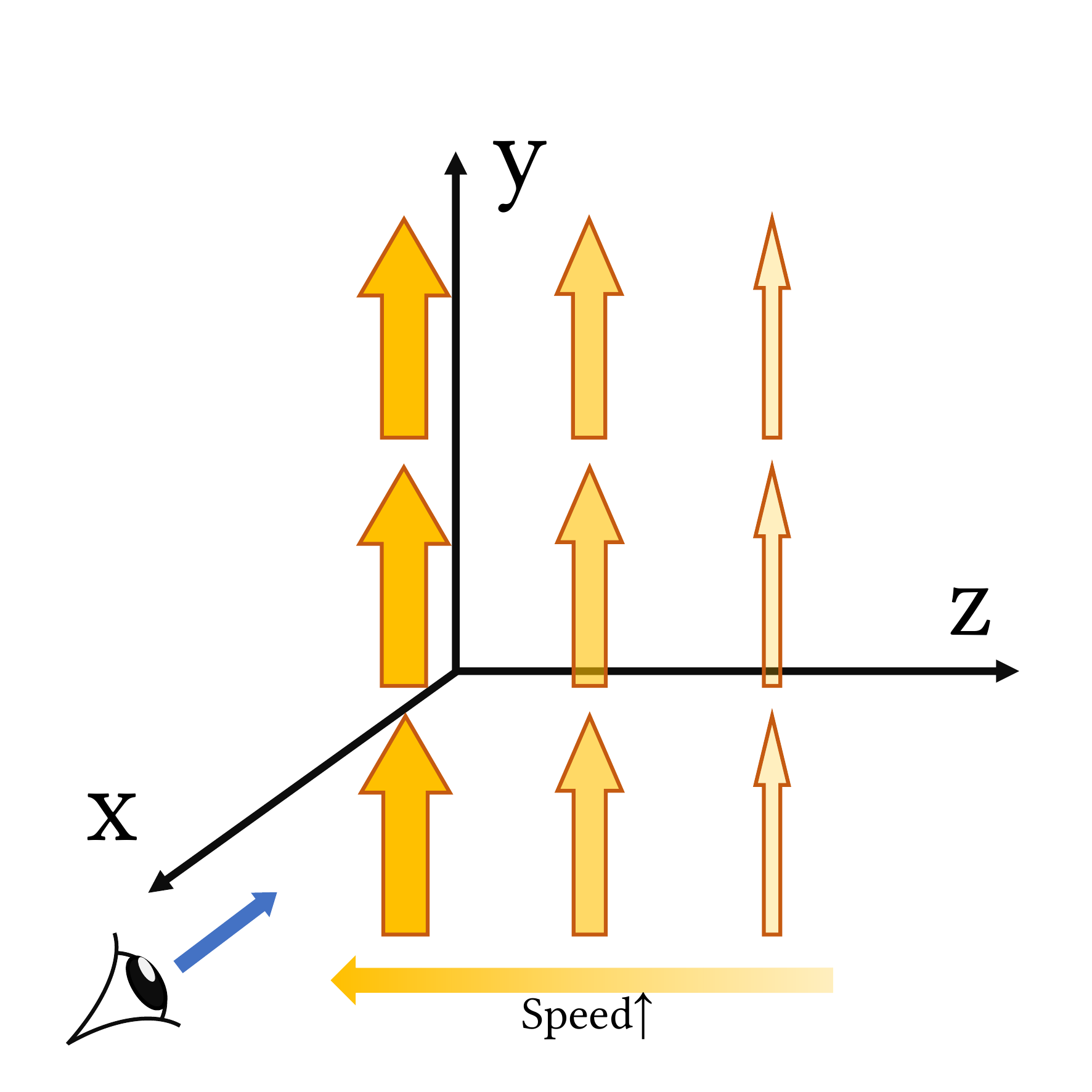}  &  \mpeablimg{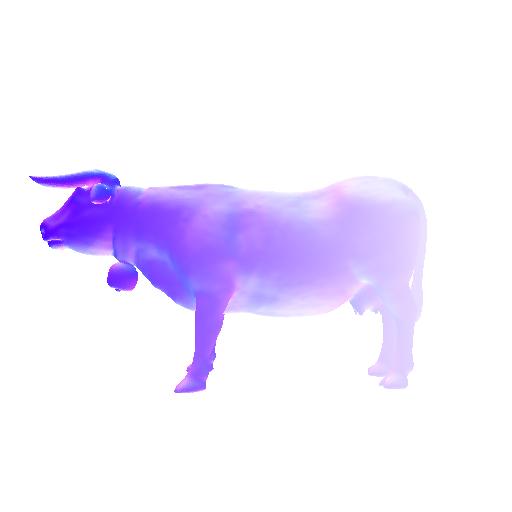}  &  \mpeablimg{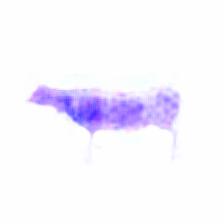}  &  \mpeablimg{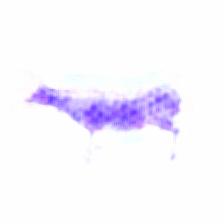} \\
\mpeablimg{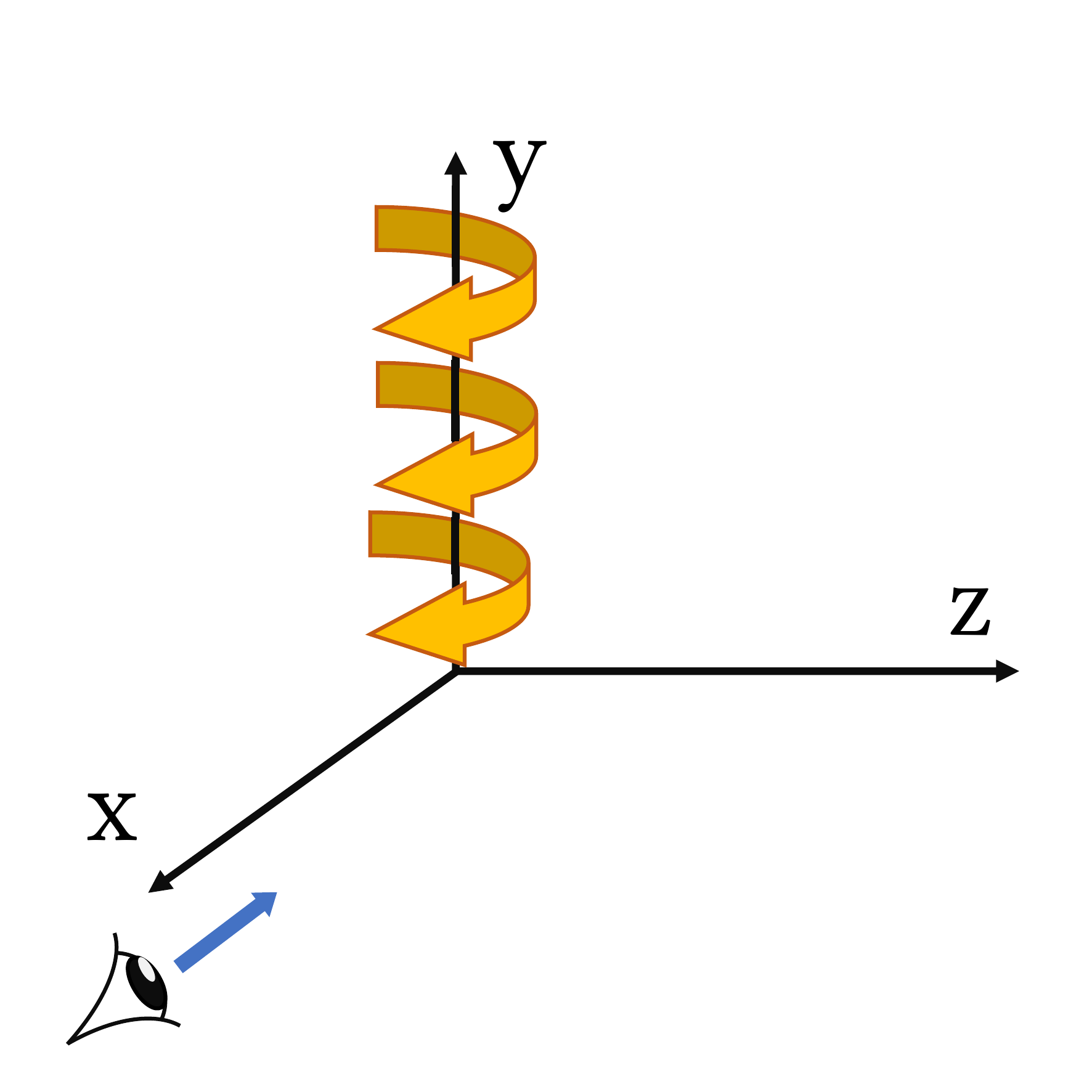}  &  \mpeablimg{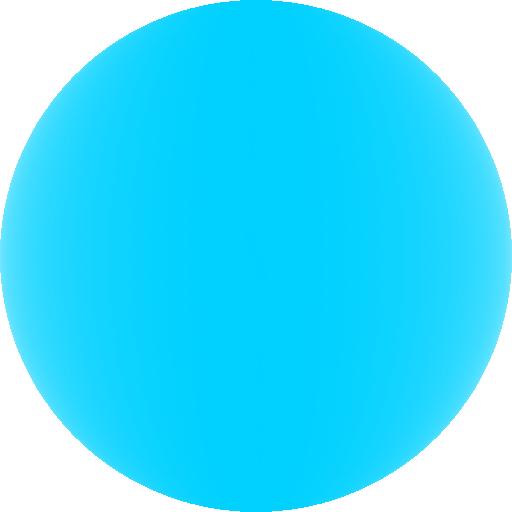}  &  \mpeablimg{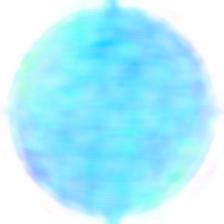}  &  \mpeablimg{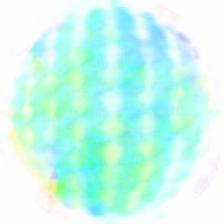} \\
\mpeablimg{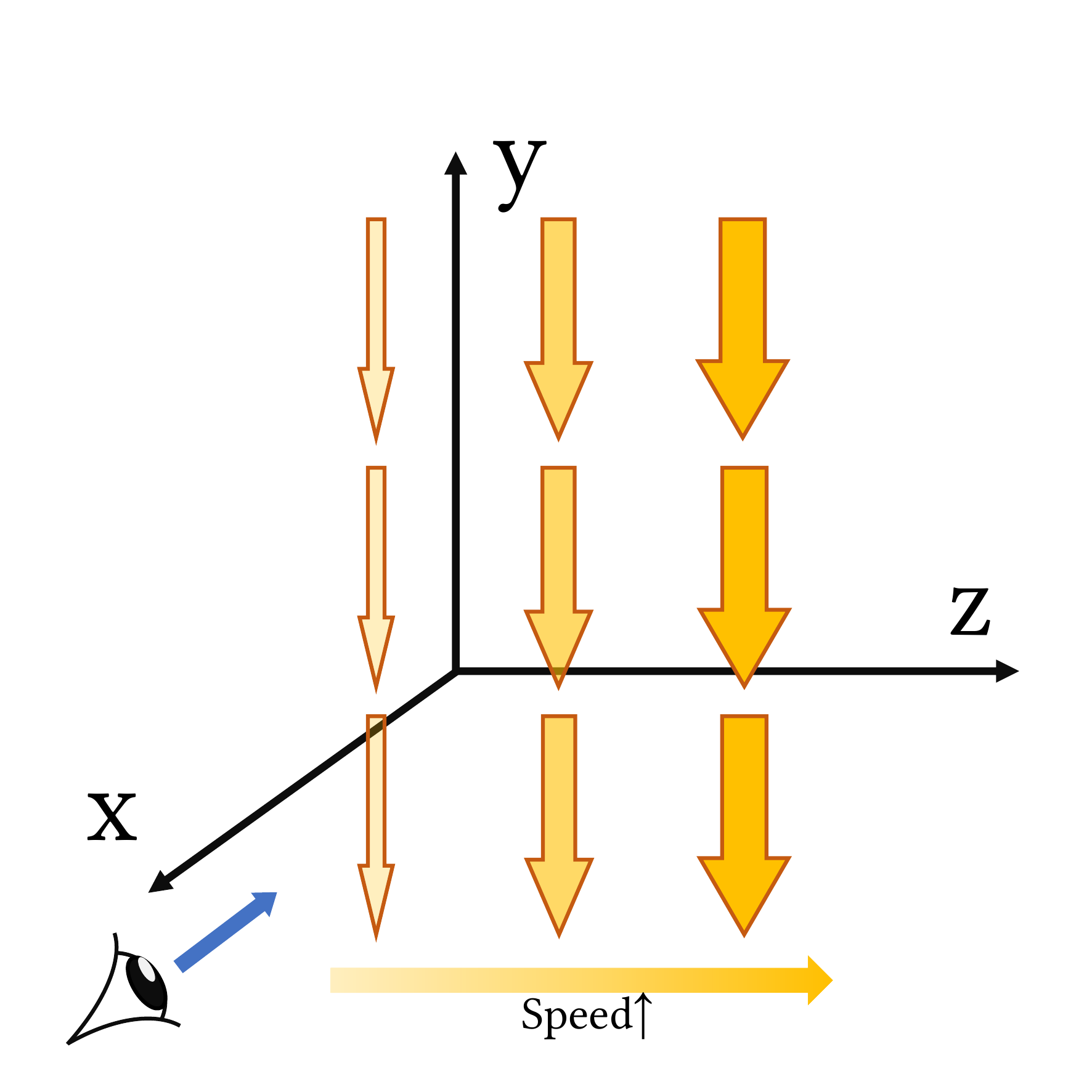}  &  \mpeablimg{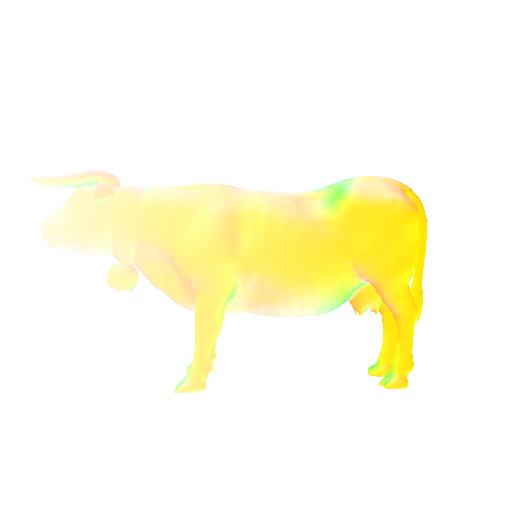}  &  \mpeablimg{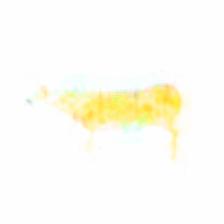}  &  \mpeablimg{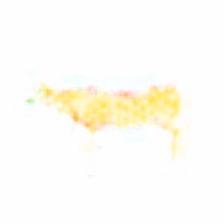}
\end{tabular}
}
\captionof{figure}{Ablation study on Motion Positional Encoding. The model without the encoding is denoted as \textbf{Plain}. The \textbf{Plain} model struggles to accurately capture intricate motions, including rotation or those exhibiting non-uniform speed changes across vertices. Such a limitation is exhibited both during the training on the \icosphere and at test time with new meshes. The underlying textures of the motion demonstration are \textit{Sci-fi\_Wall\_010} and \textit{Sci-Fi\_Wall\_012} in our collected dataset. }
\label{fig:mpe-abl}
\end{table}

\section{Visualization of MeshNCA Limitations}
\subsection{Requirement on Mesh Quality}
\begin{figure}
    \centering
    \includegraphics[width=\linewidth]{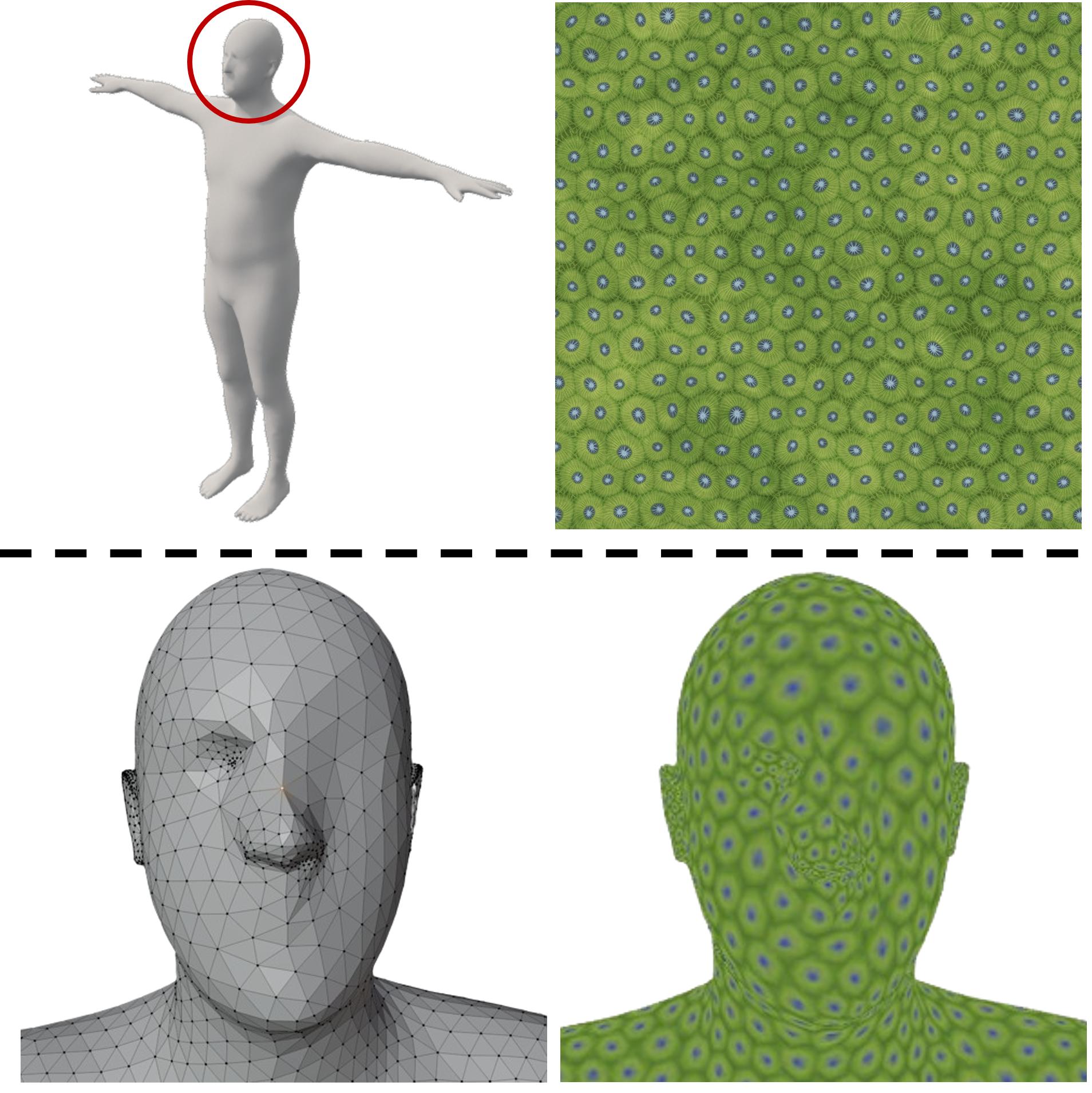}
    \caption{The remeshing algorithm sometimes fails to produce uniform density vertices. Top: Target mesh and the target texture. Bottom: zoomed-in view of the target mesh and the synthesized texture on this mesh. Notice that the synthesized texture does not have uniform density anymore. }
    \label{fig:remesh_failure}
\end{figure}

As mentioned in the Section~\ref{sec:mesh_collection} two important factors for the quality of the synthesized texture are:
\begin{itemize}
    \item Uniform vertex distribution and equal edge lengths
    \item Valence close to 6
\end{itemize}
We remesh our meshes using PyMeshLab \cite{pymeshlab} to ensure these two properties for each mesh, however, in few cases, the remeshing algorithm fails output a mesh with uniform vertex distribution.
This can lead to an undesired appearance on some regions of the mesh. Figure~\ref{fig:remesh_failure} shows one such example. 

\begin{figure}
    \centering
    \includegraphics[width=\linewidth]{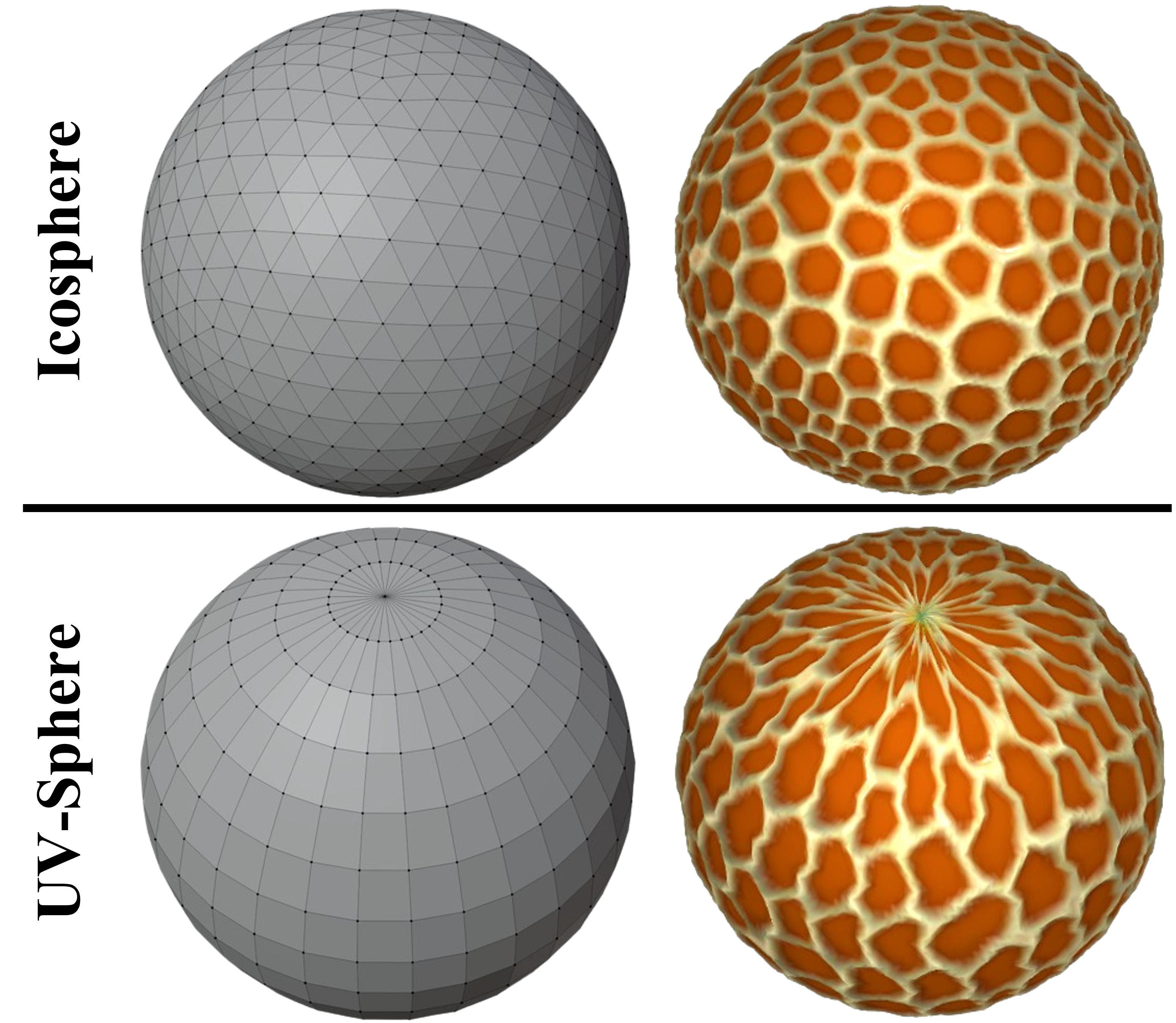}
    \caption{Comparison of the synthesized textures on UV sphere mesh and \icosphere mesh. Notice that the synthesized texture on the UV sphere is distorted near the poles. }
    \label{fig:ico_vs_uv}
\end{figure}

UV sphere is another example that violates both of aforementioned properties. First, the pole vertices of a UV sphere mesh have very high valence. Second, the vertex density is much higher near the poles. 
Figure~\ref{fig:ico_vs_uv} compares the synthesized textures on UV sphere and \icosphere meshes. As UV sphere violates the desired uniform vertex density and valence properties, the synthesized textures appear distorted.

\subsection{Misalignment between texture maps}
\begin{figure}
    \centering
    \includegraphics[width=\linewidth]{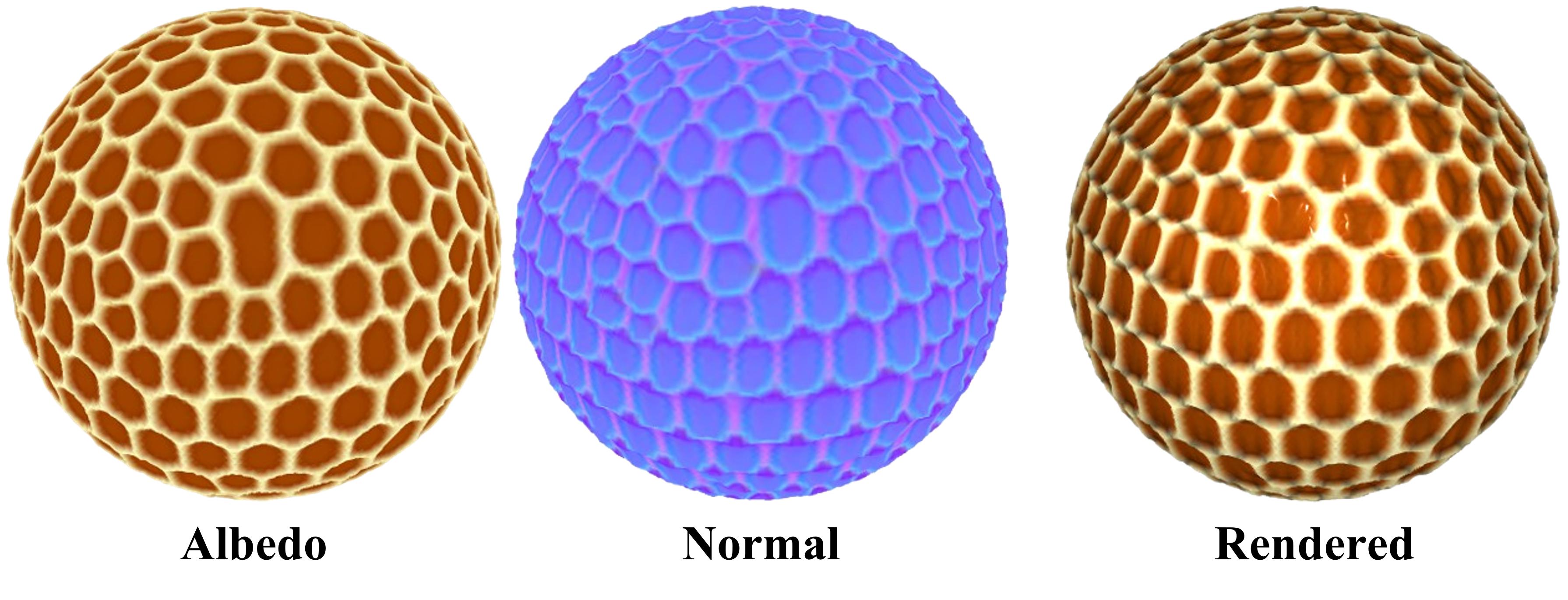}
    \caption{MeshNCA occasionally fails to synthesize correctly aligned texture maps. In this example, the normal map and the albedo are not correctly aligned, leading to an undesired artefact in the synthesized texture.}
    \label{fig:texture_misalignment}
\end{figure}

In our image-guided experiments, we train MeshNCA to simultaneously synthesize 5 texture maps: albedo, normal, height, roughness, and ambient occlusion. 
We apply our appearance loss separately on each of these texture maps and rely on the inductive bias in our MeshNCA model to correctly align all the texture maps in test-time. 
However, rarely, our training fails to converge to a solution where all the texture maps are correctly aligned. Figure~\ref{fig:texture_misalignment} illustrates this problem. Notice that the misalignment between the albedo and normal maps causes the rendered output to look unnatural. We find that this problem is quite rare (happens only on 2 out of the 72 textures, \textit{Honeycomb\_002} and \textit{Abstract\_008}), and re-training MeshNCA with a different random seed can be one solution to achieve a MeshNCA model with correctly aligned texture maps. 

\subsection{Agnostic to Mesh Semantics}
MeshNCA is unaware of the mesh semantics as it only relies on local communication. Therefore, it synthesizes more homogeneous textures than the methods that directly train on the target mesh as shown in Figure \ref{fig:limitation-vis}. X-mesh can synthesize the alien-like eye structure on the mesh when trained using prompt \textit{An image of an alien made of marble}, while MeshNCA cannot synthesize textures related to the concept of alien.
\begin{table}[]
\resizebox{\linewidth}{!}{%
    \centering
    \begin{tabular}{cc}
        Ours & \rev{\small{\citet{xmesh}}}  \\
        \midrule
        \includegraphics[height=60pt]{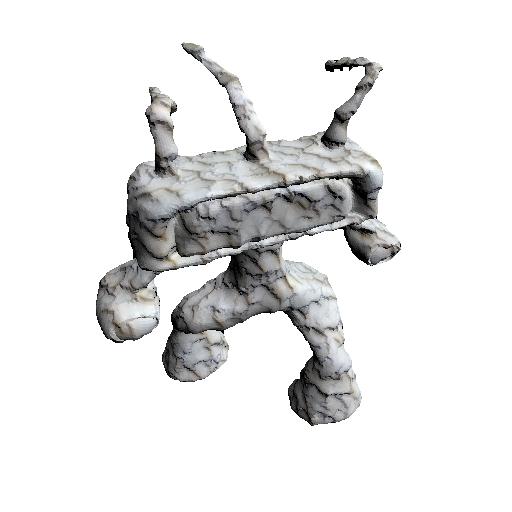} & 
        \includegraphics[height=60pt]{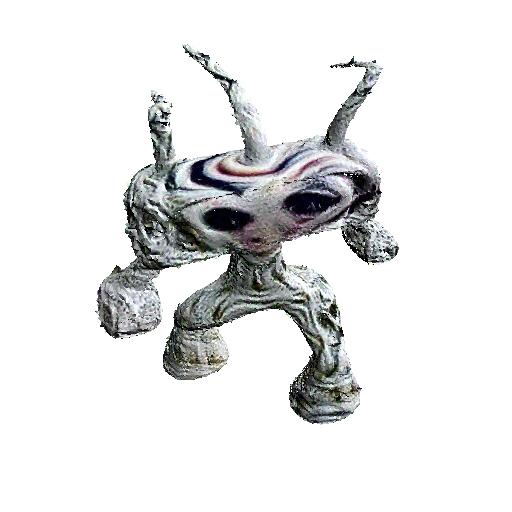}  \\
    \end{tabular}
    }
    \captionof{figure}{Visualization of the non-semantic texture synthesized by MeshNCA. X-mesh \cite{xmesh} can synthesize alien-like eyes one the mesh while MeshNCA textures the mesh more homogeneously.}
    \label{fig:limitation-vis}
\end{table}

\section{Implementation Details}
For all the training experiments, we use a multi-step learning rate scheduler, which decays the learning rate at epoch 1000 and 2000 with a decay rate of 0.3. We implement the overflow loss and the pooling strategy used in previous Neural Cellular Automata literature \cite{mordvintsev2020growing, dynca, niklasson2021self-sothtml} to stabilize the training. Given a cell state of MeshNCA $\mathbf{S} \in \mathbb{R}^{N \times C}$, the overflow loss  $\mathcal{L}_{over}$ of $\mathbf{S}$ is defined as:

$$\mathcal{L}_{over} = \frac{1}{NC} \sum^{N}_{i} \sum^{C}_{c} \left |\mathbf{S}^{t}_{i,c} - clip_{[-1,1]} (\mathbf{S}^{t}_{i,c}) \right |$$ 
\begin{equation}
    \quad \quad \quad clip_{[a,b]} (x) = max(a, min(x,b)), \; \; \; \; \; \; \; \; \; \; \;
\label{eq:overflow-loss}
\end{equation}
where $S_{i,c}$ is one of the entries of the MeshNCA cell state tensor $\mathbf{S}$. The weight of the overflow loss is 10000. 

For \textbf{Image-guided} synthesis, the pool size is 256. The seed inject frequency is every 16 epochs. When there is no motion target, we set the MeshNCA step range, namely the range of the number of steps of MeshNCA in one epoch to $[15, 25]$. With the motion target, the range becomes $[32, 64]$. 

For \textbf{Text-guided} synthesis, the pool size is 64 and the seed inject frequency is every 32 epochs to ensure more stable training. In both \textbf{Text} and \textbf{Text+Motion} experiments, the MeshNCA step range is $[15, 25]$.

\subsection{Additional Loss used in Text-guided Training}
When performing the mesh geometry changing during \textbf{Text-guided} synthesis, we add a constraint term $\mathcal{L}_{hf}$ to the final loss to reduce the high-frequency noise. The loss is defined in Equation \ref{eq:laplacian-clip-loss}.

\begin{equation}
    \mathcal{L}_{hf} = max \left (0.0, \frac{1}{N}\sum_{i=1}^N \Delta C_{i}^{geo} - \xi \right ).
    \label{eq:laplacian-clip-loss}
\end{equation}

Recall that $N$ is the total number of cells. $\xi$ is a hyperparameter controlling the strength of the constraint. We set it to 0.01. The $\Delta C_{i}^{geo}$ stands for the uniform Laplacian of the geometric channel of cell $i$ generated by MeshNCA, which is given in Equation \ref{eq:laplacian-compute}.

\begin{equation}
    \Delta C_{i}^{geo}=\sum_{j \in \mathcal{N}(i)} (C_{j}^{geo} - C_{i}^{geo}).
    \label{eq:laplacian-compute}
\end{equation}

\subsection{Vector Fields In Dynamic Texture Synthesis}
Here we formalize the mathematical expression of all the global vector fields used in our 3D dynamic texture synthesis experiments. 


\newcommand{\centeredtxtx}[1]{
\begin{tabular}{l}
\parbox{6.0cm}{\vspace{-50pt} \centering #1}
\end{tabular}
}
Let $\bar{V}^t(x,y,z)$ be the un-normalized motion vector at point $(x,y,z)$, and $\bar{V}^t_{i}$ be the vector field evaluated at the position of the $i$-th mesh vertex. 
To obtain the target vector field $V^t$, we L2-normalize $\bar{V}^t$ using the following equation:
\begin{equation}
 V^t=\frac{\bar{V}^t}{\frac{1}{N}\sum_{i}^N||\bar{V}^t_{i}||_{2}}   
\end{equation}
$N$ is the number of cells/vertices.
We define the vector fields in Table \ref{tab:vec-field1}.

\begin{table*}[]
\caption{Vector field definition. The vector field with gradually changing strength is properly shifted to ensure consistent motion direction.}
\label{tab:vec-field1}
\resizebox{\textwidth}{!}{
\begin{tabular}{cc||cc}
\textbf{Vector Field} & \textbf{Formula} & \textbf{Vector Field} & \textbf{Formula}  \\
 \midrule 
 \motionimg{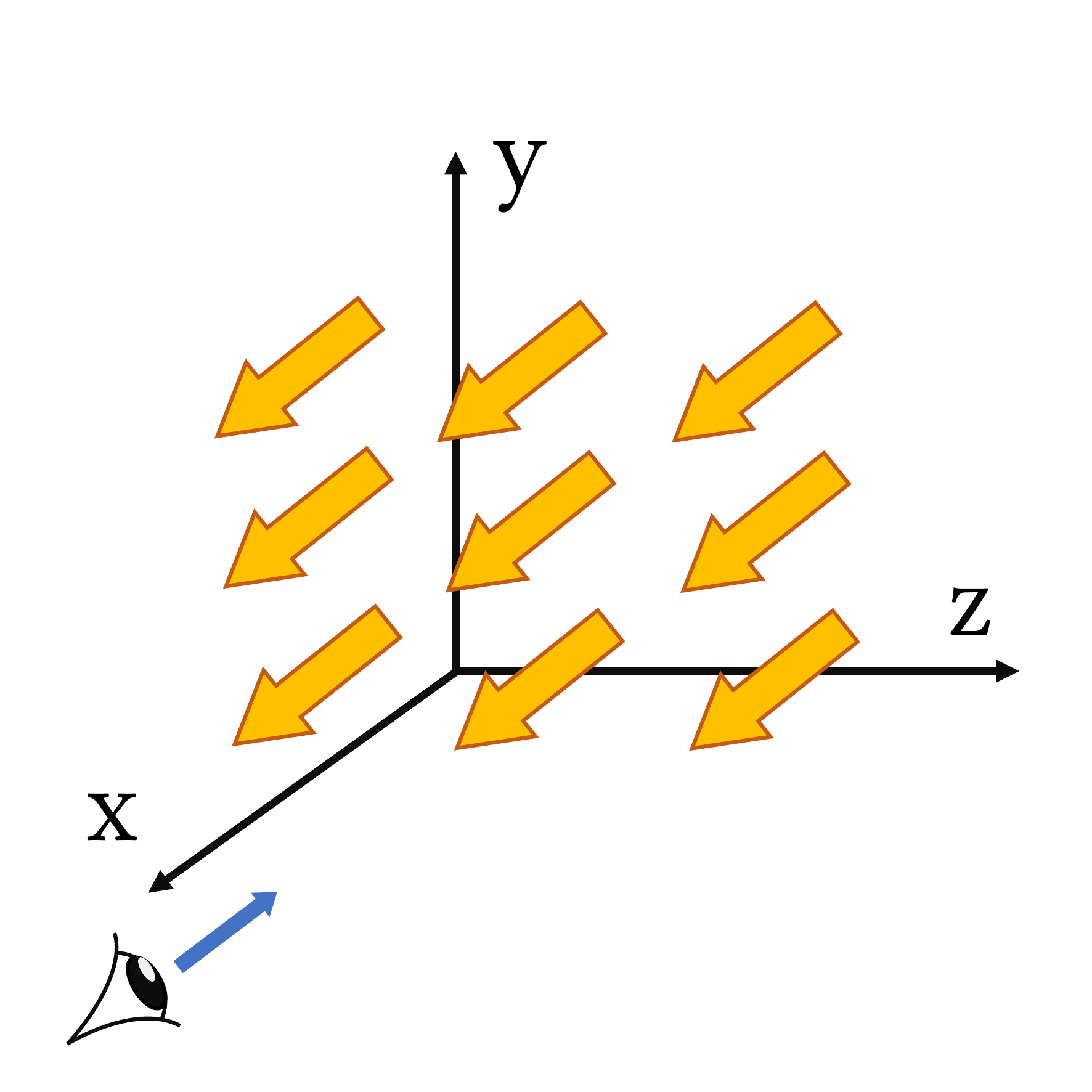}  &
 \centeredtxtx{$\bar{V}^t(x,y,z)=(1,0,0)$} 
 & 
 \motionimg{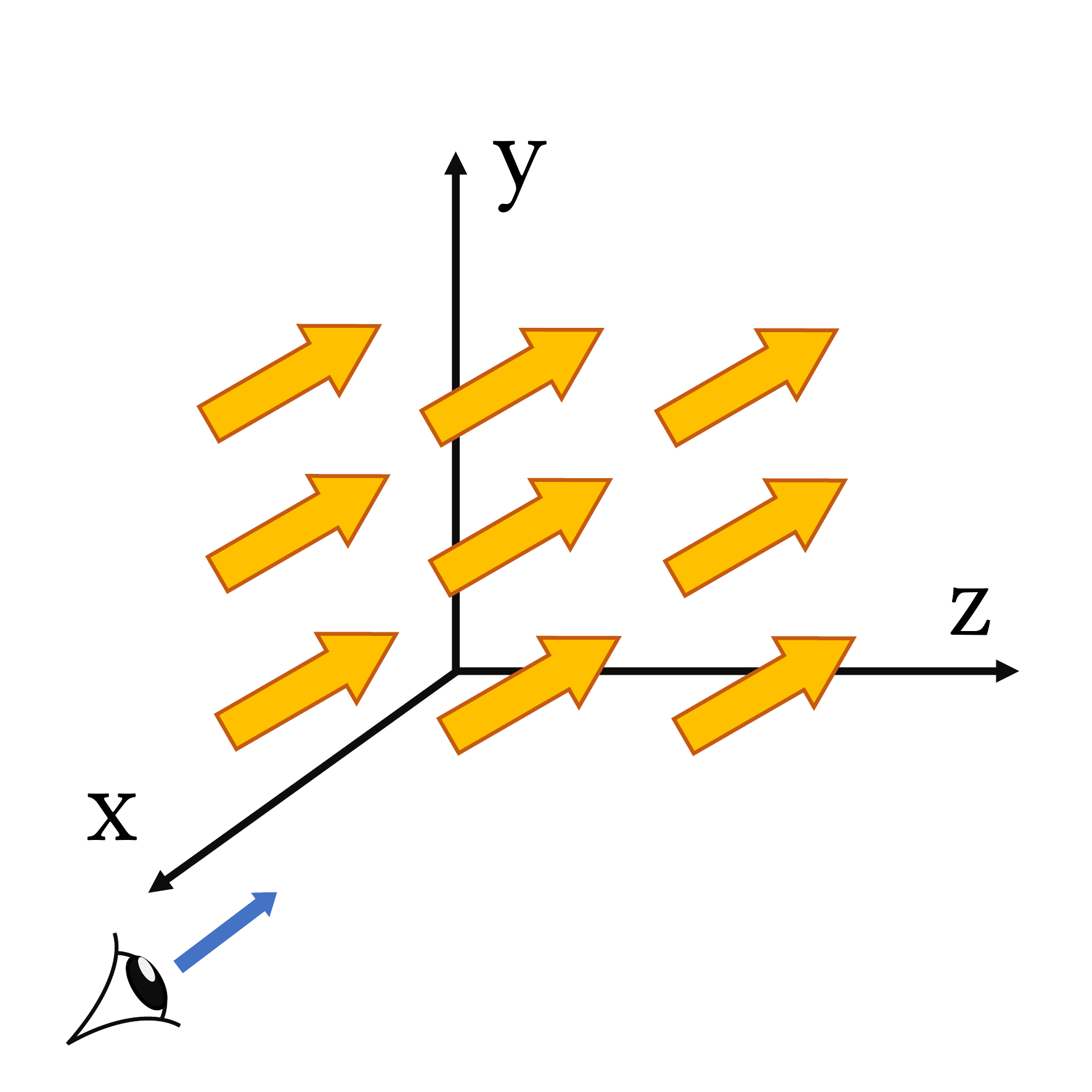}  &
 \centeredtxtx{$\bar{V}^t(x,y,z) = (-1, 0, 0)$} 
 \\

 \motionimg{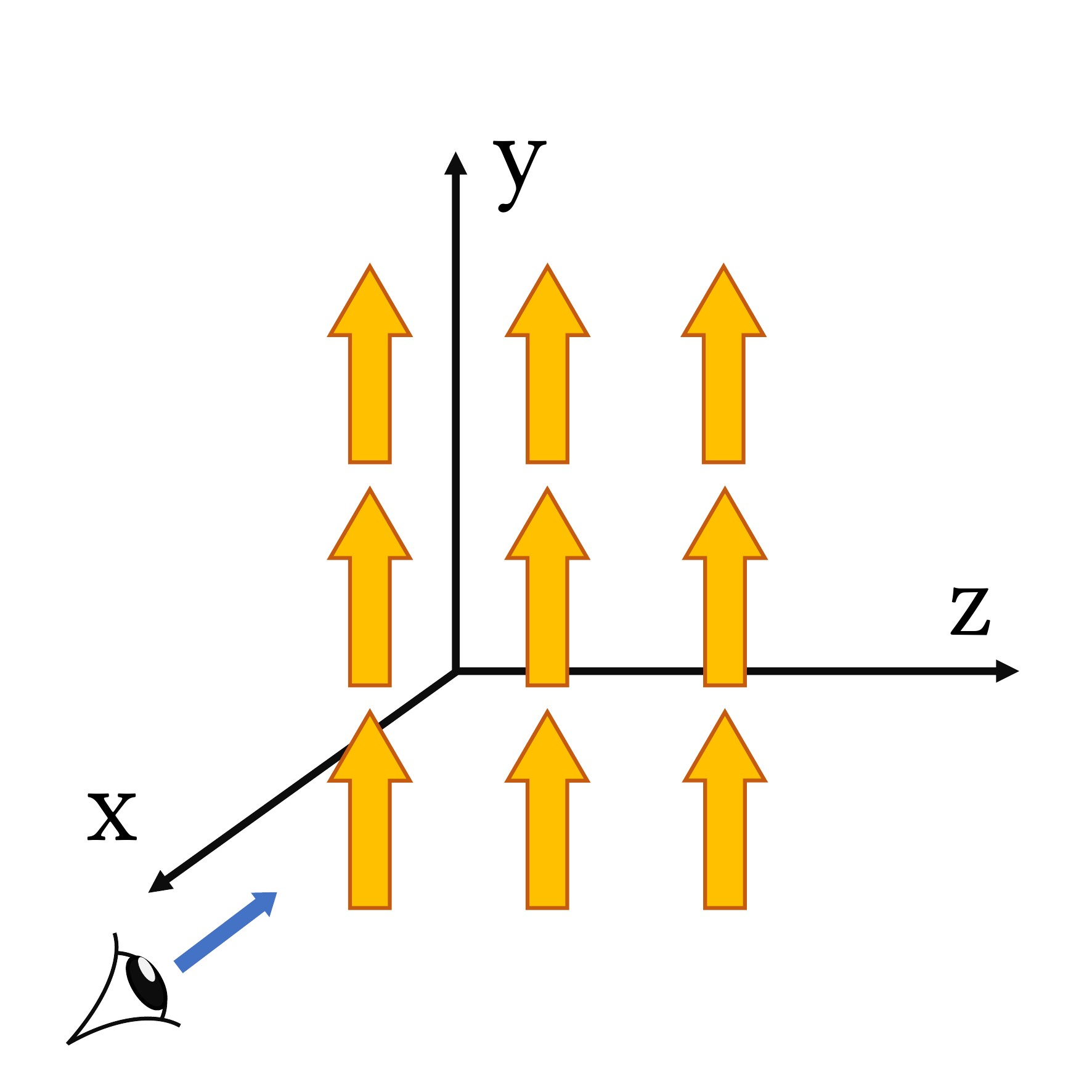}  &  
 \centeredtxtx{$\bar{V}^t_{i}=(0, 1, 0)$} 
 &  \motionimg{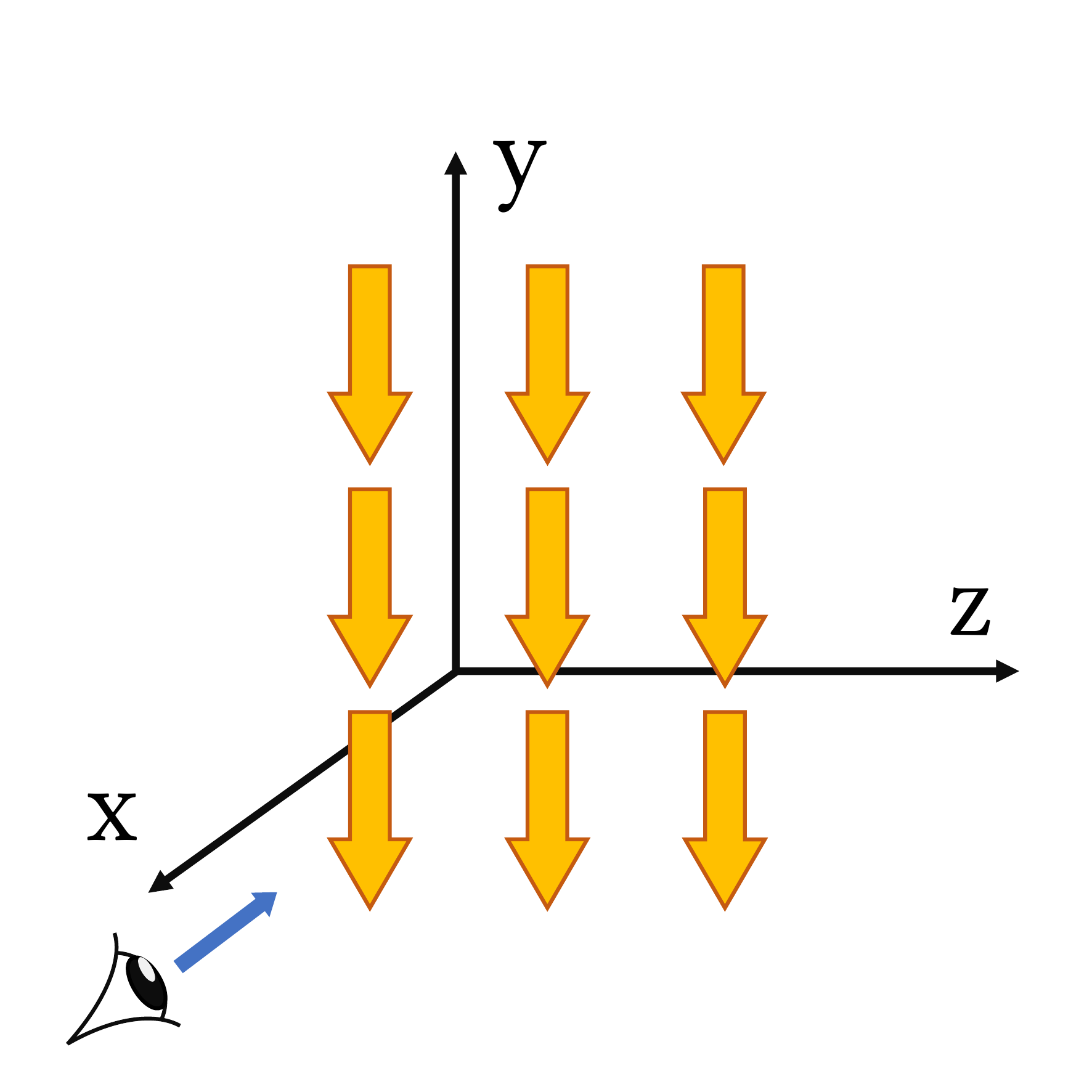}  &  
 \centeredtxtx{$\bar{V}^t(x,y,z)=(0, -1, 0)$} \\
 
 \motionimg{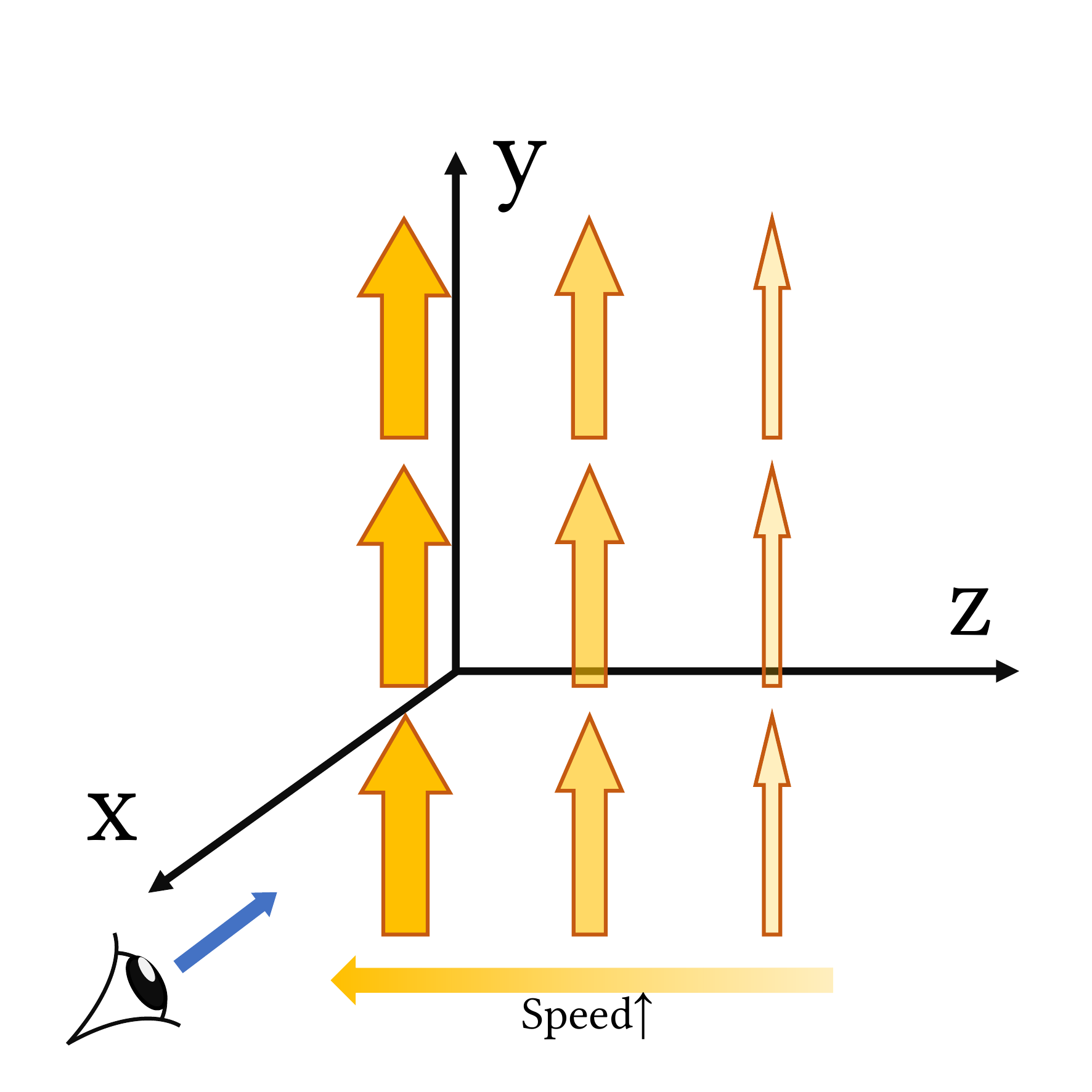}  &  
 \centeredtxtx{$\bar{V}^t(x, y, z)=(0, z + 1.0, 0)$} 
 &  \motionimg{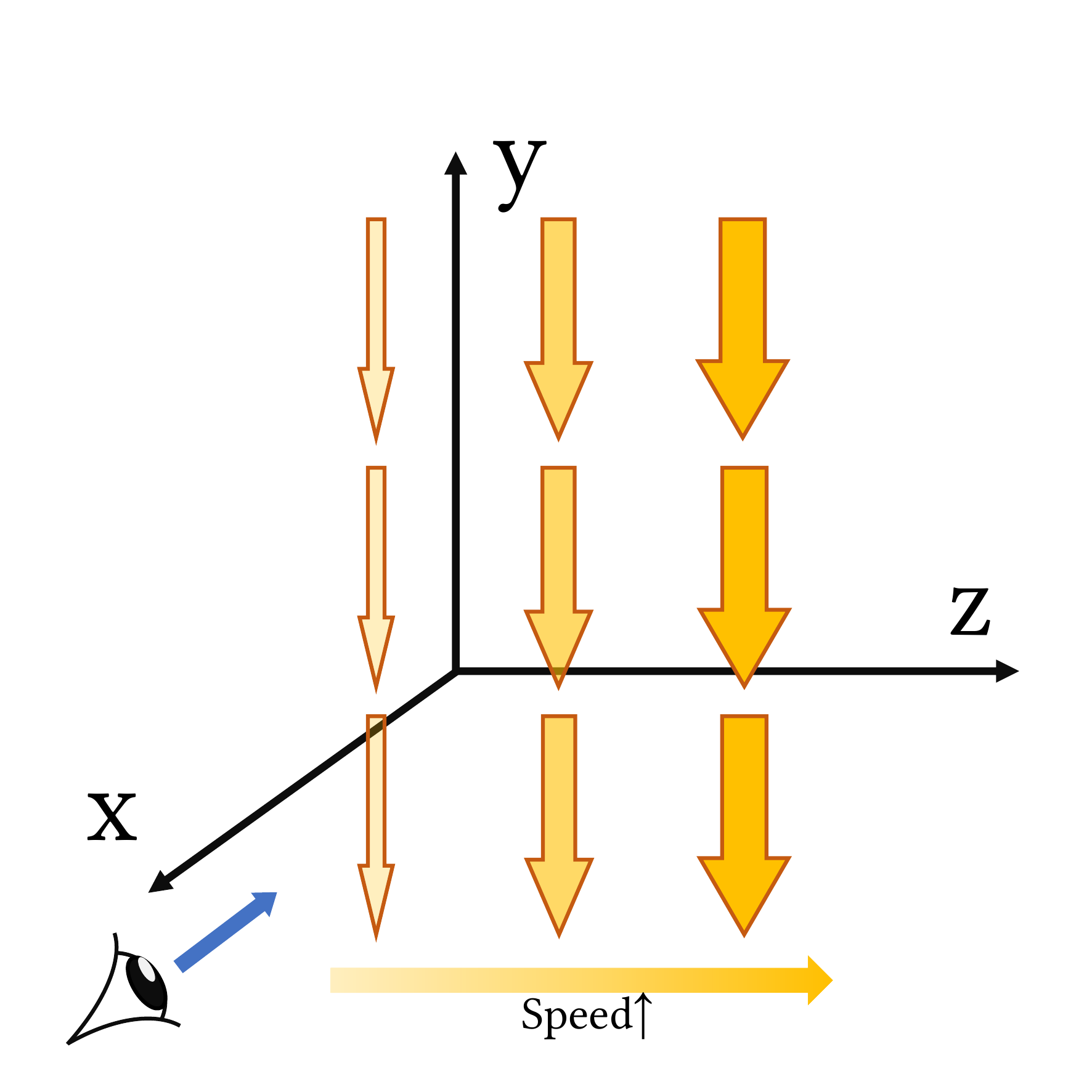}  &
 \centeredtxtx{$\bar{V}^t(x, y, z)=(0, -z - 1.0, 0)$} 
 \\

  \motionimg{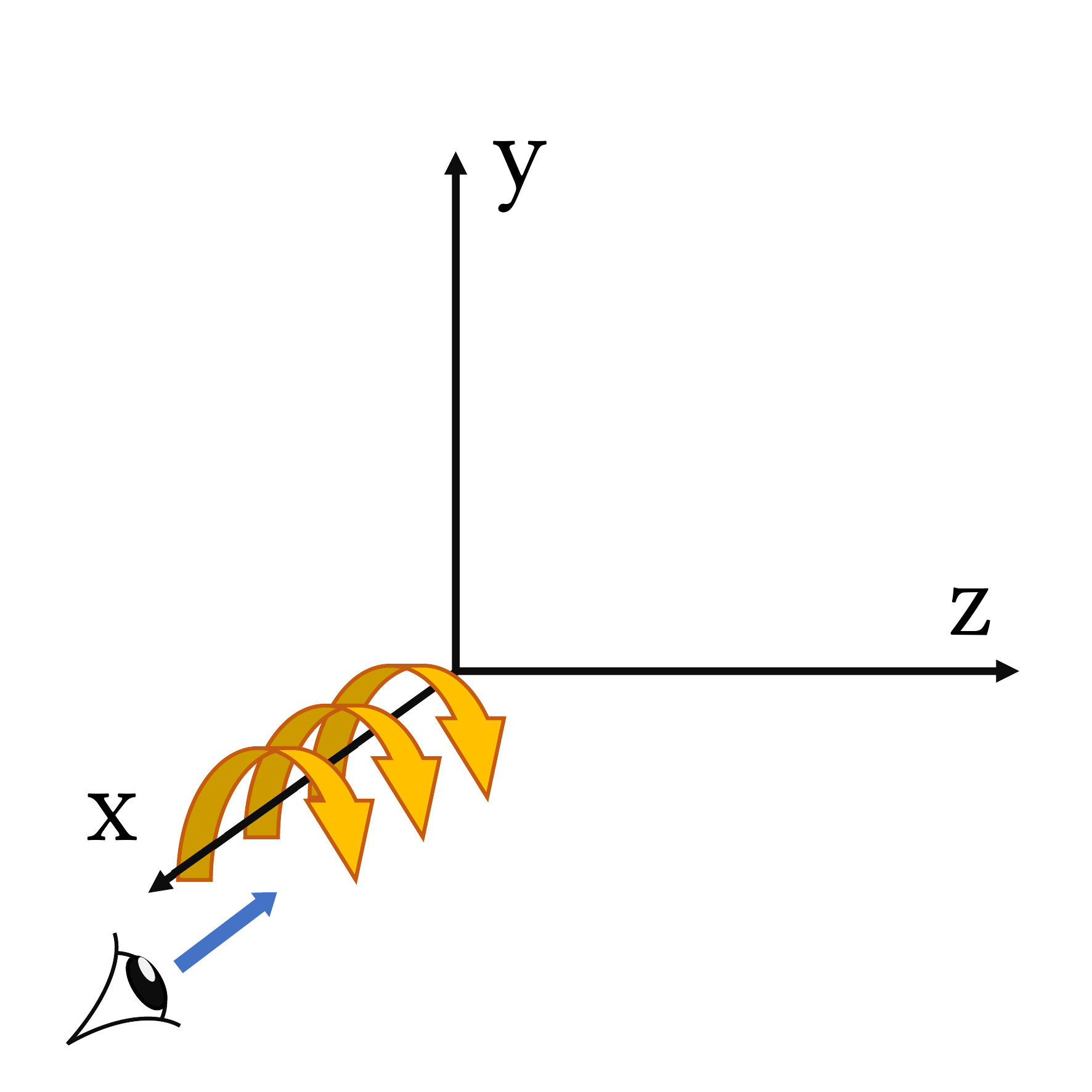}  &  
  \centeredtxtx{$\bar{V}^t(x,y,z)=(0,\frac{-z}{\sqrt{x^2+y^2+z^2}},\frac{y}{\sqrt{x^2+y^2+z^2}})$} 
  &  \motionimg{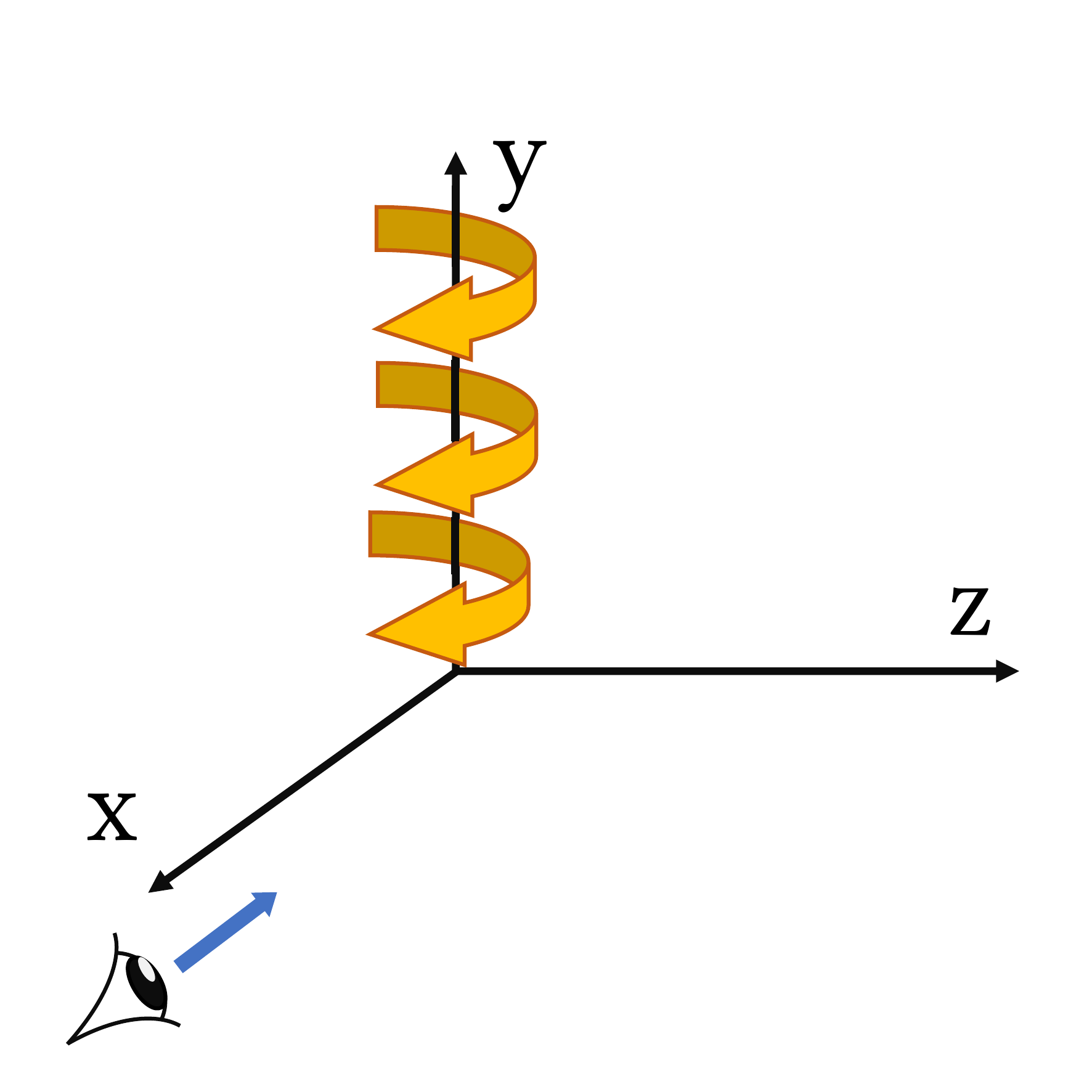}  
  &  \centeredtxtx{$\bar{V}^t(x, y, z) =  (\frac{z}{\sqrt{x^2+y^2+z^2}},0,\frac{-x}{\sqrt{x^2+y^2+z^2}})$} 
  \\
\end{tabular}
}
\end{table*}


\section{Details of User Study}

\begin{table}[]
\resizebox{\linewidth}{!}{
\begin{tabular}{l}
\toprule
\includegraphics[width=\linewidth]{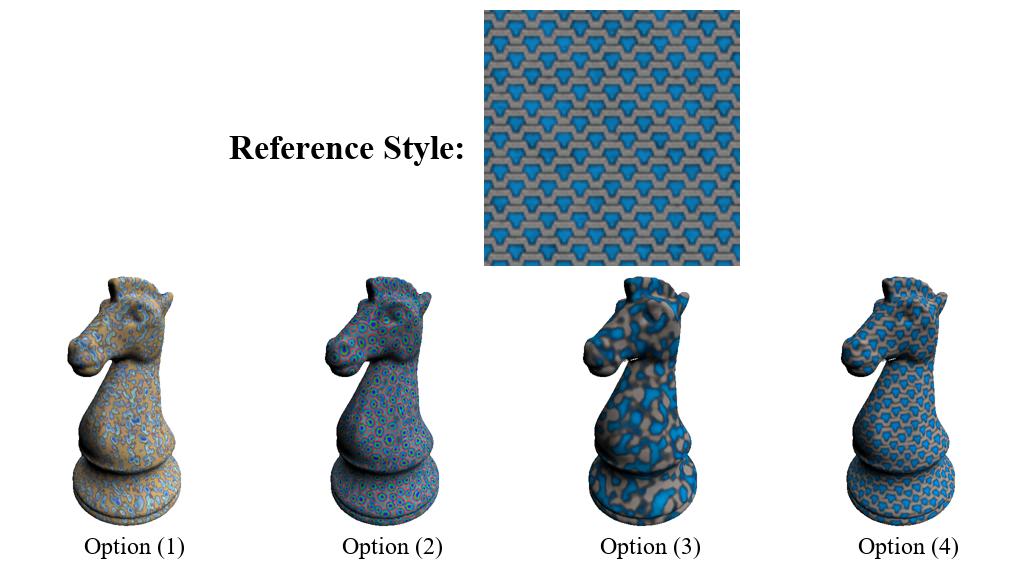}
 \\
 \midrule
\includegraphics[width=\linewidth]{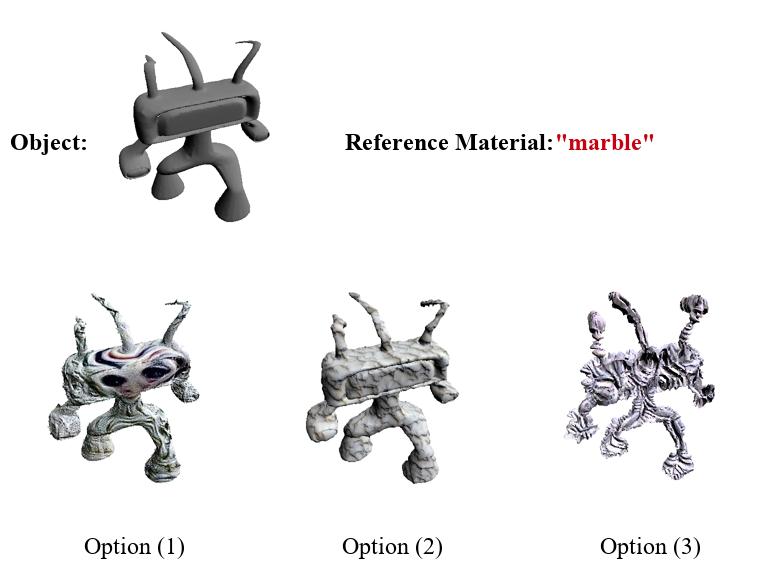} \\
\bottomrule
\end{tabular}
}
\captionof{figure}{\rev{Sample questions from our user studies. (Top) For the image-guided user study, we show the target texture on top as the reference style and ask the users to choose the option that better matches the reference style in terms of both color and structure. (Bottom) For the second user study, we show the original mesh rendered in gray and the target prompt as the reference material and ask the users to choose the option that better matches the following description: "An image of an \textit{Object} made of \textit{Reference Material}".}}
\label{fig:user-study-questions}
\end{table}

\subsection{Image-guided Texture Synthesis}
\rev{In our image-guided user study, we show images like the rows of Figure~\ref{fig:image-comparison-train} and ask the users to choose the option that better matches the reference texture both in terms of color and structure. The goal of this user study is to evaluate the texture quality of different methods alongside their generalizability to unseen meshes. We use a subset of our texture dataset with 24 samples and for each method synthesize their corresponding textures on the 6 meshes shown in Figure~\ref{fig:image-comparison-test-supp1},\ref{fig:image-comparison-test-supp2},\ref{fig:image-comparison-test-supp3},\ref{fig:image-comparison-test-supp33},\ref{fig:image-comparison-test-supp4},\ref{fig:image-comparison-test-supp5},\ref{fig:image-comparison-test-supp6}, \ref{fig:image-comparison-test-supp7} to create 144 questions in total. 
For a given target texture and mesh pairs, we show the stylized mesh by different methods and ask the following question from the participants: \textit{Which option better matches the style of the reference image?}. 
Notice that we shuffle the order of the methods to eliminate any bias. Each questionnaire consisted of 40 randomly selected questions and we received responses from 61 unique participants in total.}

\subsection{Text-guided Texture Synthesis}
\rev{As the baseline methods \cite{text2mesh, xmesh} for text-guided synthesis do not claim to have any generalization and perform very poorly on unseen meshes, we directly train them on the target meshes for creating the questions for our user study. Notice that, however, MeshNCA is still only trained on the \icosphere mesh. In other words, this user study is in favor of the baseline methods as they are directly trained on the target mesh while MeshNCA results are a practice of generalization to unseen meshes. Figure~\ref{fig:user-study-vis} shows some samples from the mentioned results. We select 10 different prompts and 5 different meshes and create a single questionnaire with 50 questions in total. For a given prompt and a mesh, we show the stylized meshes by different methods in random order alongside the input mesh to the participants and ask the following question: 
\textit{Choose the option that better matches the following description: An Image of an "Object" made of "Reference Material"}.
The results of our user study further demonstrate a drawback of the previously proposed metric in \cite{xmesh} to quantify the quality of text-guided 3D texture synthesis. We refer the readers to Section~\ref{sec:quant-text-metric} for details of the metric.}

\newcommand{\imguserstudy}[1]{%
  \ifimgprefix
    \includegraphics[height=60pt]{figures/Experiments/user-study/#1}%
  \else
    \includegraphics[height=60pt]{figures/Experiments/user-study/#1}%
  \fi
}

\begin{table}[]
\resizebox{\linewidth}{!}{%
    \centering
    \begin{tabular}{cc|ccc}
        Prompt & Object & Ours & \rev{\small{\citet{text2mesh}}} & \rev{\small{\citet{xmesh}}}  \\
        \midrule
        \rotatebox{90}{\parbox[c]{2cm}{\centering\hspace{0pt} \textbf{Jelly Beans}}} & \rotatebox{90}{\parbox[c]{2cm}{\centering\hspace{0pt} \textbf{Vase}}} & \imguserstudy{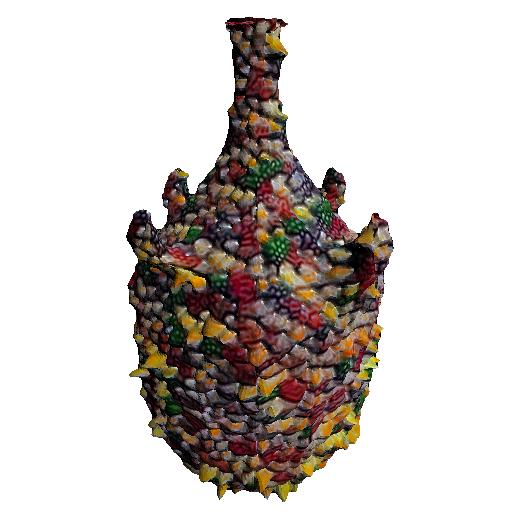} & 
        \imguserstudy{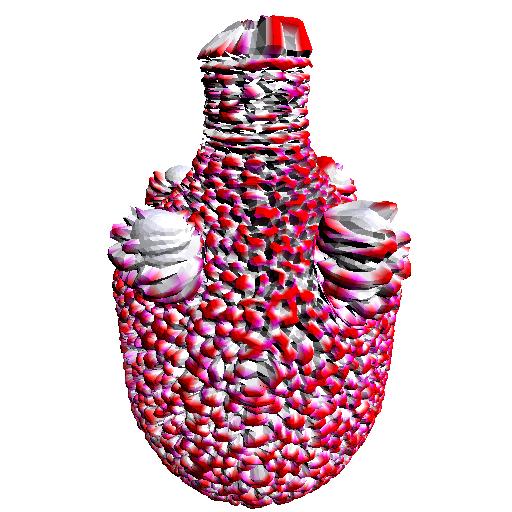} & 
        \imguserstudy{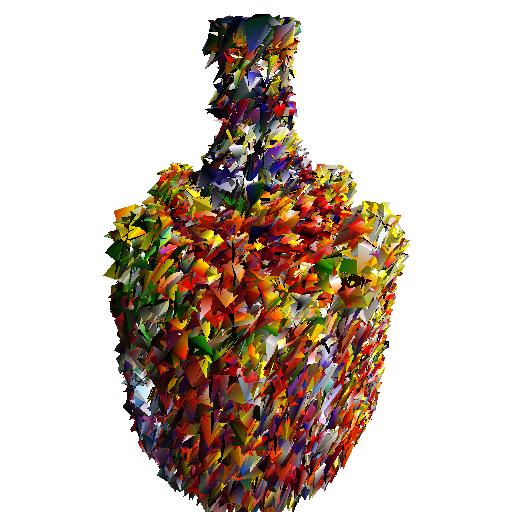} \\
        \rotatebox{90}{\parbox[c]{2cm}{\centering\hspace{0pt} \textbf{Patchwork Leather}}} & \rotatebox{90}{\parbox[c]{2cm}{\centering\hspace{0pt} \textbf{Mug}}} & \imguserstudy{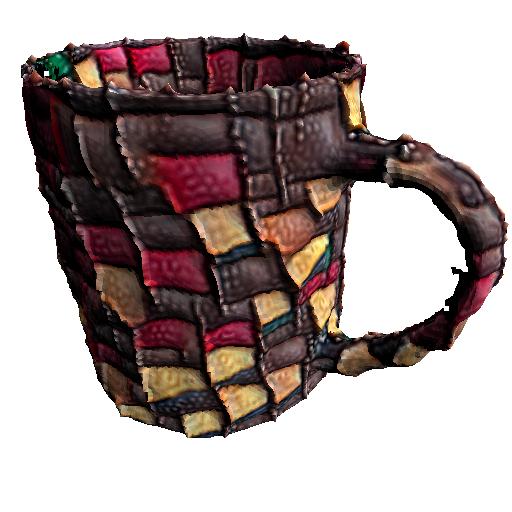} & 
        \imguserstudy{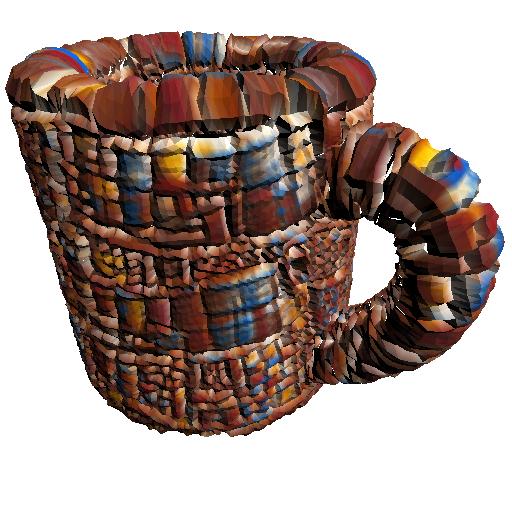} & 
        \imguserstudy{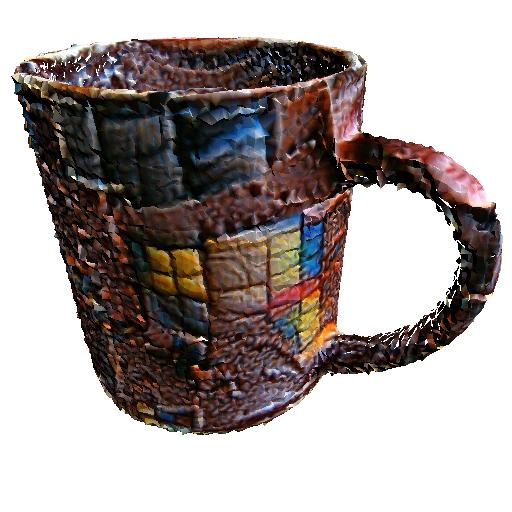} \\
        \rotatebox{90}{\parbox[c]{2cm}{\centering\hspace{0pt} \textbf{Colorful Crochet}}} & \rotatebox{90}{\parbox[c]{2cm}{\centering\hspace{0pt} \textbf{Chair}}} & \imguserstudy{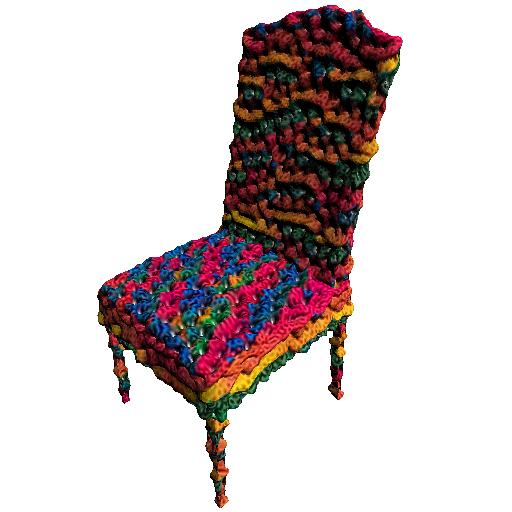} & 
        \imguserstudy{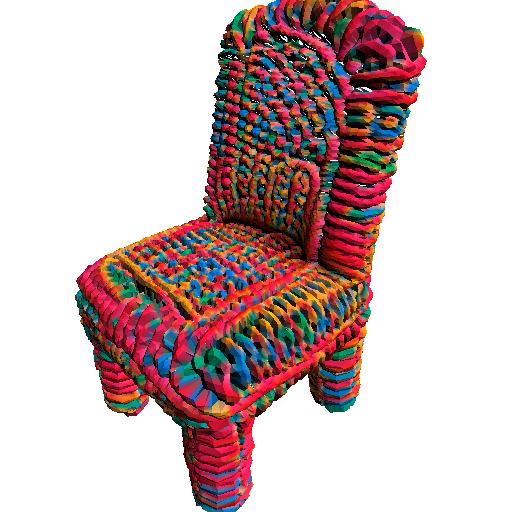} & 
        \imguserstudy{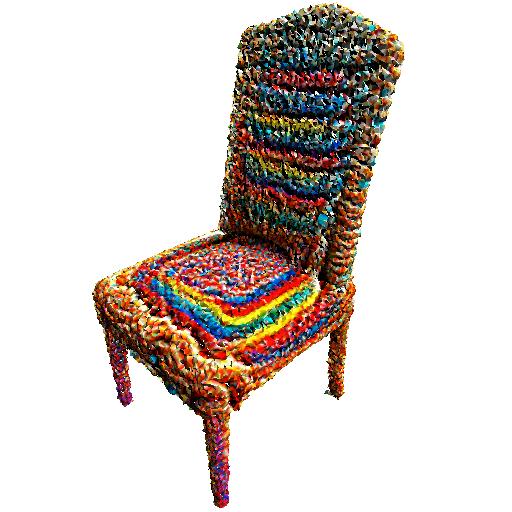} \\
    \end{tabular}
    }
    \captionof{figure}{\rev{Materials of user study in \textbf{Text-guided} synthesis. We train the methods of \cite{text2mesh} and  \cite{xmesh} on all combinations of prompt and object (mesh) pairs using the prompt template: "\textit{An image of a/an <Object> made of <Prompt>}". Note that our results do not involve any retraining, but only direct generalization from the sphere mesh.}}
    \label{fig:user-study-vis}
\end{table}

\section{Additional Comparisons}

\subsection{Quantitative Comparison on Text-guided Synthesis}
\label{sec:quant-text-metric}
The two text-guided mesh texturing methods we choose for comparison, Text2Mesh \cite{text2mesh} and X-Mesh \cite{xmesh}, are excellent text-to-3D manipulation methods despite not possessing any generalization capability. Here, we conduct an automatic quantitative comparison between MeshNCA and these two methods. As their method cannot generalize to unseen meshes, we have to train those two models on each specific text-mesh combination. However, for MeshNCA, we still use the test-time generalization results of the model trained on the sphere mesh.

In \cite{xmesh}, the authors propose two quantitative metrics, MES and ITS for automatically evaluating the quality of a text-guided texture synthesis method. MES compares the multi-view test-time CLIP score of the synthesized results, and ITS computes the number of training epochs required for a certain model to obtain a target MES. Since ITS is more bound with the text2mesh-like framework, namely an MLP trained under the CLIP guidance with some mesh preprocessing, we focus on the quality comparison using MES. However, we find this metric prefers high-frequency and sometimes unpleasant results, potentially due to the complexity of the loss landscape of the CLIP model. More importantly, using different conceptualizations of the CLIP score, the conclusion drawn from this method varies a lot. Specifically, we choose 4 different expert models that use ViT-B-32, ViT-L-14, ViT-H-14, ViT-g-14 as the backbone from OpenCLIP \cite{openclip}. ViT-B-32, ViT-L-14 are trained on the LAION-400M \cite{laion400m} dataset and ViT-H-14, ViT-g-14 are trained on the LAION-2B \cite{laion2b} dataset. We choose 10 textures, \textit{"stained glass", "cactus", "feathers", "bark", "sandstone", "marble", "patchwork leather", "colorful crochet", "moss", "jelly beans"}, and 5 meshes, \textit{"chair", "bunny", "mug", "vase", "alien"}, resulting in 50 mesh-texture combinations. The training and testing prompts use the same template: \textit{An image of a/an <Object> made of <Prompt>}. We summarize the results averaged over all texture prompts in Table \ref{tab:quant-text-sphere} for the sphere mesh training and the average score over all mesh-texture combinations in Table \ref{tab:quant-text-allmesh} for other meshes. 

\newcommand{\imgtextquant}[1]{%
  \ifimgprefix
    \includegraphics[height=60pt]{figures/Experiments/comparison/text-quant/#1}%
  \else
    \includegraphics[height=60pt]{figures/Experiments/comparison/text-quant/#1}%
  \fi
}





\begin{table}[]
    \centering
    \caption{MES of different models on \icosphere mesh training. }
    \label{tab:quant-text-sphere}
    \resizebox{\linewidth}{!}{
    \begin{tabular}{c|cccc}
       Methods  & ViT-B-32 & ViT-L-14 & ViT-H-14 & ViT-g-14 \\
       \midrule
        \rev{\small{\citet{text2mesh}}} & 32.47 & 36.74 & 35.52 & 32.02 \\
        \midrule
        \rev{\small{\citet{xmesh}}} & 34.60 & 39.05 & \textbf{38.16} & 34.43 \\
        \midrule
        Ours & \textbf{35.05} & \textbf{40.47} & 37.89 & \textbf{35.58} \\
    \end{tabular}
    }

\end{table}





\begin{table}[]
    \centering
    \caption{MES of different models on all mesh-prompt combinations. Our results are based on direct generalization from the \icosphere mesh while the other methods are trained separately on each mesh and prompt combination.}
    \label{tab:quant-text-allmesh}
    \resizebox{\linewidth}{!}{
    \begin{tabular}{c|cccc}
       Methods  & ViT-B-32 & ViT-L-14 & ViT-H-14 & ViT-g-14 \\
       \midrule
        \rev{\small{\citet{text2mesh}}} & 29.09 & 35.37 & 32.15 & 29.14 \\
        \midrule
        \rev{\small{\citet{xmesh}}} & \textbf{31.09} & \textbf{39.17} & \textbf{34.74} & \textbf{31.14} \\
        \midrule
        Ours & 28.80 & 36.34 & 33.87 & 30.24 \\
    \end{tabular}
    }

\end{table}

\begin{table}[]
\resizebox{\linewidth}{!}{%
\begin{tabular}{c||cc}
\textbf{Prompts} & \textbf{Ours} & \rev{\small{\citet{xmesh}}} \\
 \midrule
\rotatebox{90}{\parbox[c]{2cm}{\centering\hspace{0pt} \textbf{Jelly Beans}}}  & \imgtextquant{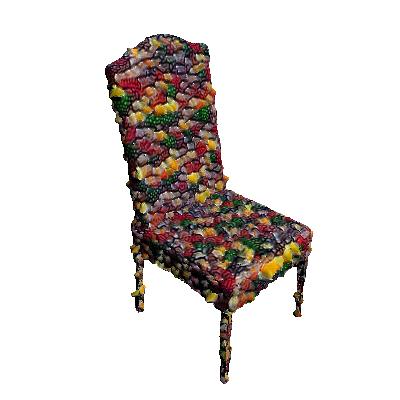} & \imgtextquant{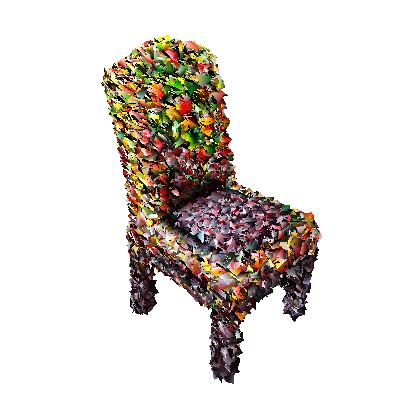} \\
 & 32.47 & 37.53 \\

 \rotatebox{90}{\parbox[c]{2cm}{\centering\hspace{0pt} \textbf{Moss}}}  & \imgtextquant{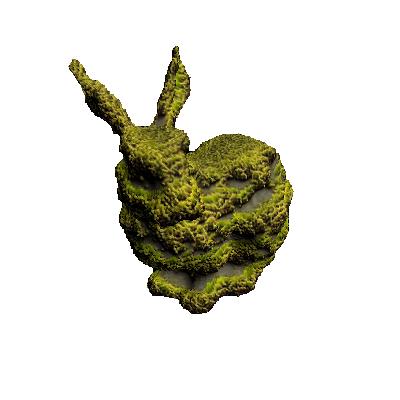} & \imgtextquant{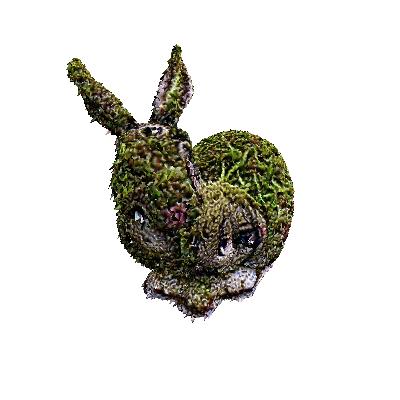} \\
 & 34.17 & 38.02 \\

 \rotatebox{90}{\parbox[c]{2cm}{\centering\hspace{0pt} \textbf{Stained Glass}}}  & \imgtextquant{meshnca-vase-glass.jpg} & \imgtextquant{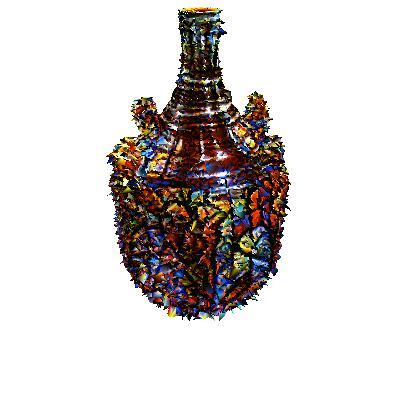} \\
 & 25.64 & 30.39 \\
\end{tabular}
}
\captionof{figure}{Visual results and corresponding MES on \textit{jelly beans-chair}, \textit{moss-bunny}, \textit{Stained Glass-vase}. X-mesh \cite{xmesh} much higher MES than our method, while the visual results of X-mesh contain high-frequency noisy patterns. }
\label{fig:text-quant-vis}
\end{table}

\begin{table}[]
    \caption{MES-dir of different models on sphere mesh training. }
    \resizebox{\linewidth}{!}{
    \begin{tabular}{c|cccc}
       Methods  & ViT-B-32 & ViT-L-14 & ViT-H-14 & ViT-g-14 \\
       \midrule
        \rev{\small{\citet{text2mesh}}} & 26.61 & 23.98 & 24.05 & 26.22 \\
        \midrule
        \rev{\small{\citet{xmesh}}} & 27.06 & 24.91 & 24.68 & 27.18 \\
        \midrule
        Ours & \textbf{30.32} & \textbf{30.62} & \textbf{29.67} & \textbf{32.42} \\
    \end{tabular}
    }
    
    \label{tab:quant-text-sphere-dir}
\end{table}





\begin{table}[]
    \centering
    \caption{MES-dir of different models on all mesh-texture combinations. Our results are based on direct generalization while the others are after retraining.}
    \resizebox{\linewidth}{!}{
    \begin{tabular}{c|cccc}
       Methods  & ViT-B-32 & ViT-L-14 & ViT-H-14 & ViT-g-14 \\
       \midrule
        \rev{\small{\citet{text2mesh}}} & 19.83 & 17.82 & 17.03 & 17.59 \\
        \midrule
        \rev{\small{\citet{xmesh}}} & 21.41 & 19.70 & 18.45 & 18.47 \\
        \midrule
        Ours & \textbf{21.67} & \textbf{20.76} & \textbf{19.90} & \textbf{19.88} \\
    \end{tabular}
    }
    
    \label{tab:quant-text-allmesh-dir}
\end{table}

We can clearly observe from Table~\ref{tab:quant-text-sphere} that MeshNCA achieves the best MES on the sphere mesh, although it is not trained towards maximizing CLIP score like the other two methods. In Table~\ref{tab:quant-text-allmesh}, X-mesh dominates the result, indicating that it converges to a better local minimum on the CLIP loss landscape. However, when we see some visualization of the results in Figure~\ref{fig:text-quant-vis}, we can see that the results from X-mesh contain unpleasant noise and choppy textures while achieving much higher MES than us. We assume the reason to be X-mesh overfitting to the CLIP model and a better local minimum on the CLIP loss landscape does not necessarily mean a higher-quality result. As shown in the user study in our main paper, our generalization results are recognized as having the highest quality compared to the other two methods. 

Furthermore, we use a different conceptualization of the CLIP score used in MES. We test the CLIP direction matching score using the same framework of MES, denoted as MES-dir. The CLIP direction matching score is computed the same as our CLIP direction matching loss. We set the negative prompts as the mesh name, and the negative image to be the mesh without any texturing. The results of MES-dir are reported in Table~\ref{tab:quant-text-sphere-dir} for sphere training and \ref{tab:quant-text-allmesh-dir} for other mesh-texture combinations. We can see that this time MeshNCA dominates two tables, indicating that MeshNCA captures the directional concept of the texture in CLIP embedding space. From this analysis and the visual results, we can conclude that the metric proposed in \cite{xmesh} is a weak necessary condition for proving the method has high-quality results and does not consider the overfitting problem in the CLIP model. Moreover, the conclusion drawn from the MES-dir metric aligns better with our user study. We can infer that in this case, the MES-dir approximates the golden standards, namely human judgment, better than the previous MES metric, providing a more reliable result.

\subsection{Direct Synthesis}
\label{sec:supp-compare-direct}

We present more results of MeshNCA and \cite{text2mesh}, \cite{diff-program-rdsystem}, and \cite{on-demand-solid-texture} on direct synthesis on the sphere mesh in Figures \ref{fig:image-comparison-train-supp1}, \ref{fig:image-comparison-train-supp2}, and \ref{fig:image-comparison-train-supp3}.

\subsection{Generalization}
\label{sec:supp-compare-gen}

We present more results of MeshNCA and \cite{text2mesh}, \cite{diff-program-rdsystem}, and \cite{on-demand-solid-texture} on generalization on new meshes after being trained on the sphere mesh in Figures \ref{fig:image-comparison-test-supp1}, \ref{fig:image-comparison-test-supp2}, \ref{fig:image-comparison-test-supp33}, \ref{fig:image-comparison-test-supp3}, \ref{fig:image-comparison-test-supp4}, \ref{fig:image-comparison-test-supp5}, \ref{fig:image-comparison-test-supp6}, and \ref{fig:image-comparison-test-supp7}.

\section{Additional Results}

\subsection{Image-guided Synthesis}
\rev{We show a few more results of \textbf{Image-guided} synthesis in Figure. Full results are available at our \textcolor{red}{\href{https://meshnca.github.io/}{demo}}.}

\subsection{Text-guided Synthesis}
We show more results of \textbf{Text-guided} synthesis in Figures \ref{fig:text-synthesis-result-supp1},\ref{fig:text-synthesis-result-supp2}, and \ref{fig:text-synthesis-result-supp3}.

\subsection{Grafting}
We invite readers to play with the grafting brush tool in our online \textcolor{pinkl}{\href{https://meshnca.github.io/}{demo}} to see the grafting results of MeshNCA trained under image targets.

We present more results of MeshNCA grafting when it is trained under the guidance of text prompts, in Figures \ref{fig:text-graft-result-supp1}, and \ref{fig:text-graft-result-supp2}.

\subsection{Direct Training on Target Meshes}

Our method, though not necessary, can also be trained on the specific mesh-texture combination, as in \cite{text2mesh} and \cite{xmesh}. We show some results in Figures \ref{fig:text-train-direct-mesh-result-supp1}, and \ref{fig:text-train-direct-mesh-result-supp2}. The models are trained with initialization from the pre-trained model of each prompt on the \icosphere mesh.

\subsection{Dynamic Texture Synthesis}
\label{sec:supp-motion-image-text}
In order to better show the results of MeshNCA on 3D dynamic texture synthesis, we provide the link to the motion videos in Figure \ref{fig:image-motion-result-supp1} with image target, and Figure
\ref{fig:text-motion-result-supp1} for text-guided dynamic texture synthesis.

In the video, the user will see the dynamic texture results on multiple meshes while the model is only trained on the sphere mesh. We provide an anchor view on the mountain mesh, the last column of each video, where the camera is on the red point shown in Figure \ref{fig:anchor-view}. The view corresponds to an azimuth of 240 degrees (counter-clockwise) and an elevation of 15 degrees, treating positive x direction as 0 azimuth and elevation. Therefore, in that view, the motion \textbf{PositiveX} should go to the upper right of the canvas. For all other meshes, we first fix the camera at positive x direction for around 2 seconds and then rotate the camera counter-clockwise, until it reaches the negative x direction.

\begin{table*}[]
    \resizebox{\textwidth}{!}{%
    \begin{tabular}{cc}
       \includegraphics[width=100pt]{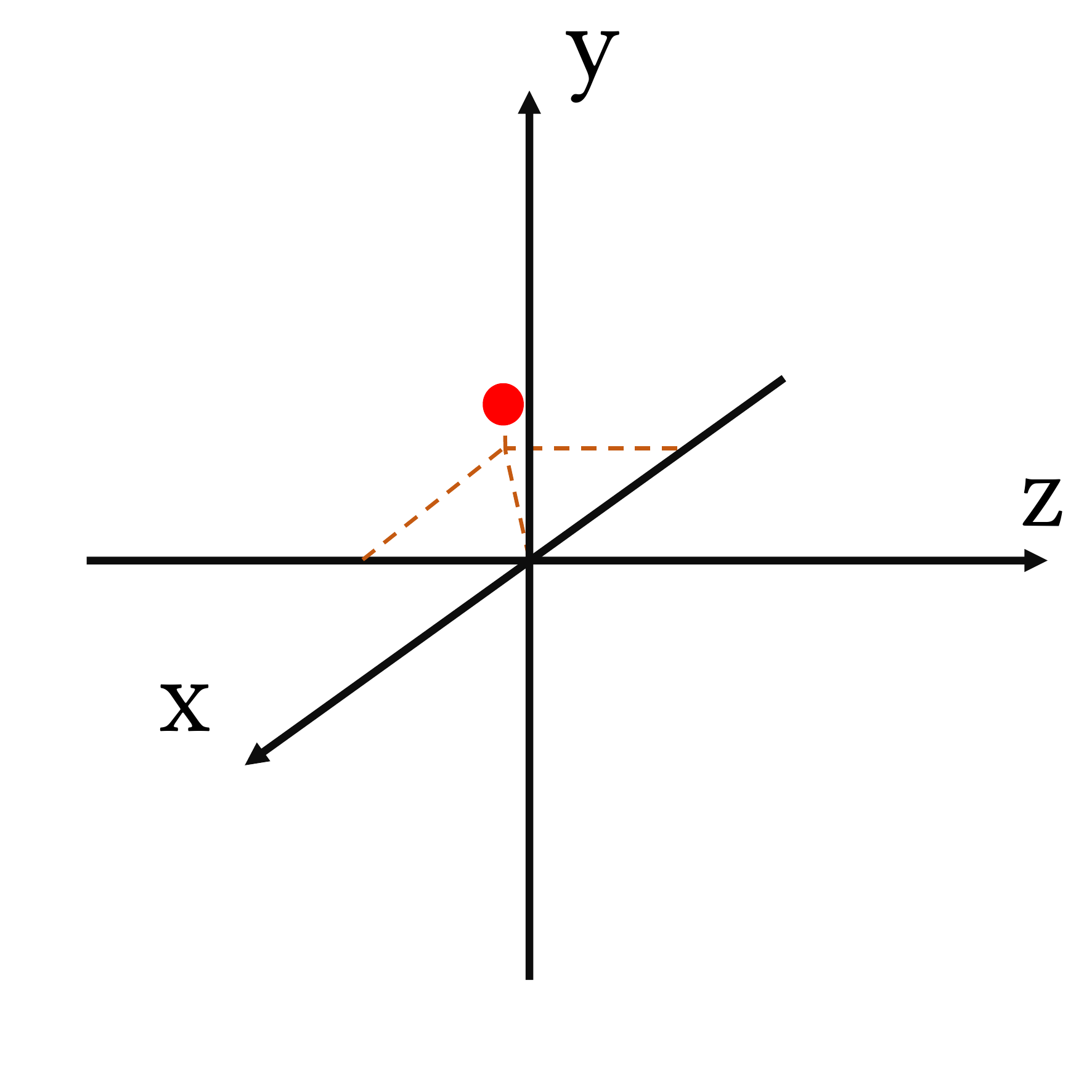}  &  \includegraphics[width=100pt] {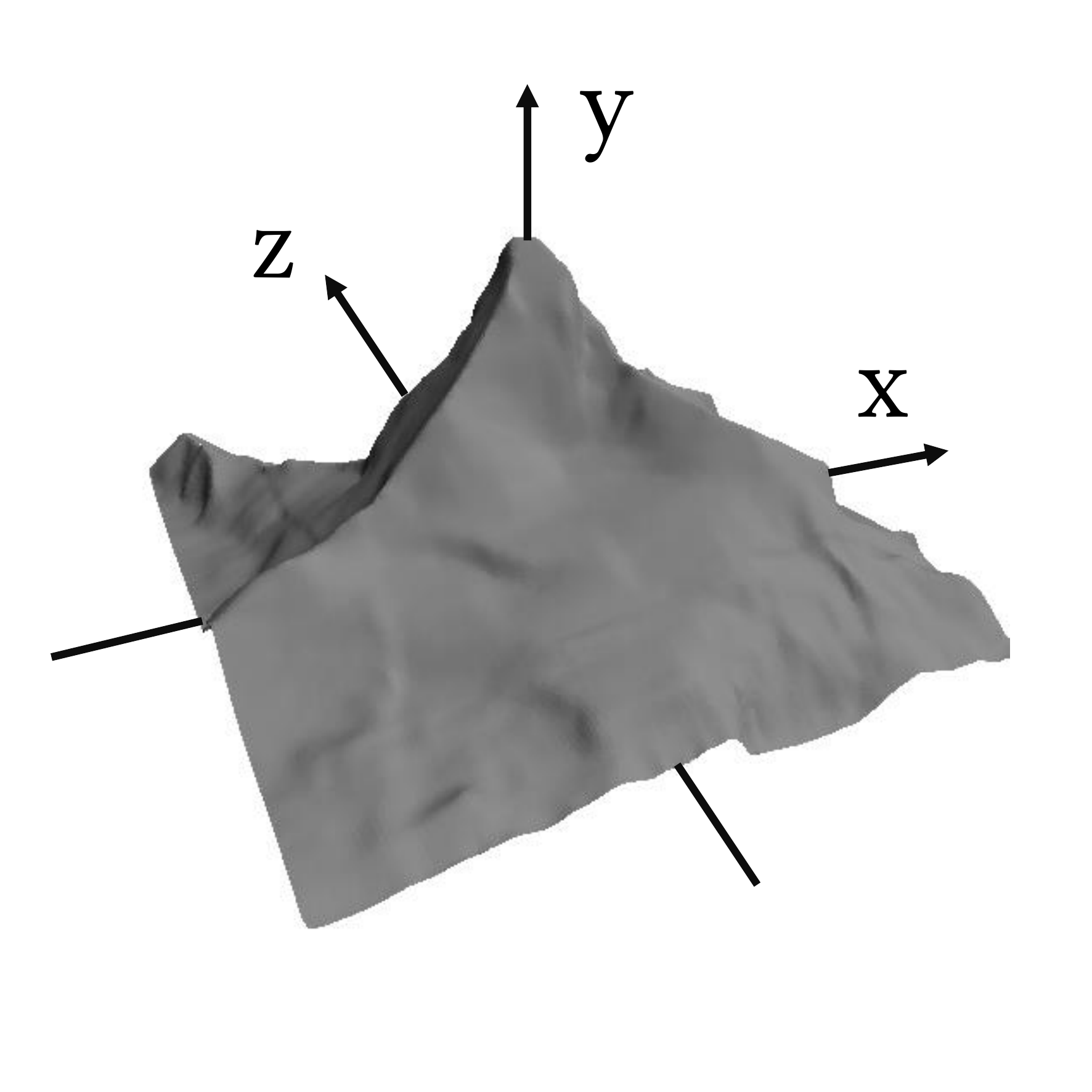} \\
       (a) & (b)
    \end{tabular}
    }
    \captionof{figure}{Figure (a) shows the anchor view on mountain mesh, represented by the red point. Figure (b) is the mountain mesh rendered using that view. }
    \label{fig:anchor-view}
\end{table*}

\begin{table*}[]
\resizebox{\textwidth}{!}{%
\begin{tabular}{cc||cc||ccc}
 \multicolumn{2}{c||}{\textbf{Target Texture Maps}}  &  \multicolumn{2}{c||}{\textbf{Icosphere Train}}  &  \multicolumn{3}{c}{\textbf{Test-time Generalization}}  \\
 Albedo & Attributes &  Albedo & Attributes &  Albedo & Attributes & Rendered \\
 \midrule
\imgimg{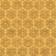}  &  \imgimg{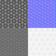}  &  \imgimg{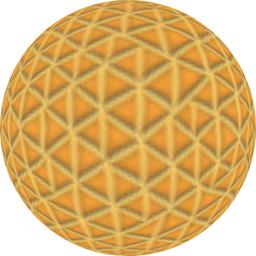}  &  \imgimg{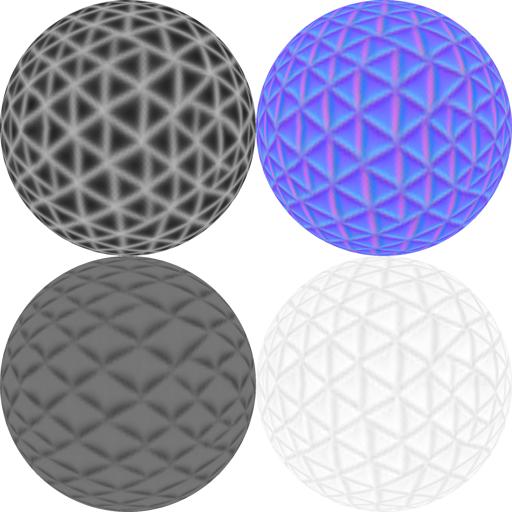}  &  \imgimg{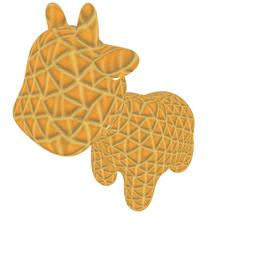}  &  \imgimg{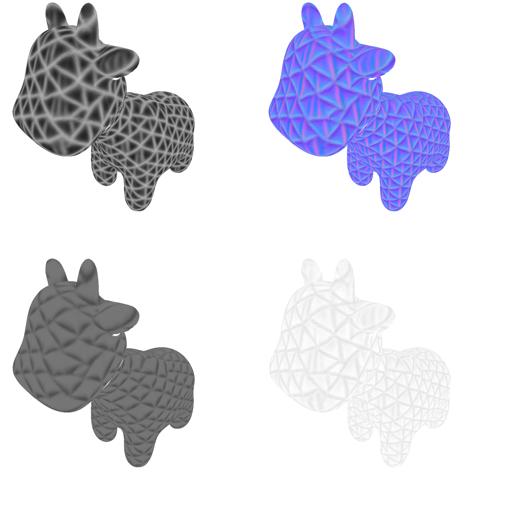}  &  \imgimg{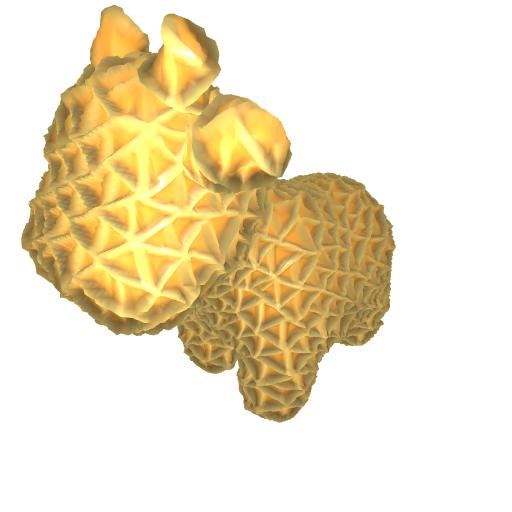} \\
\imgimg{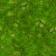}  &  \imgimg{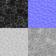}  &  \imgimg{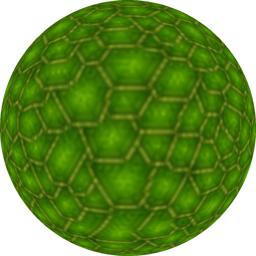}  &  \imgimg{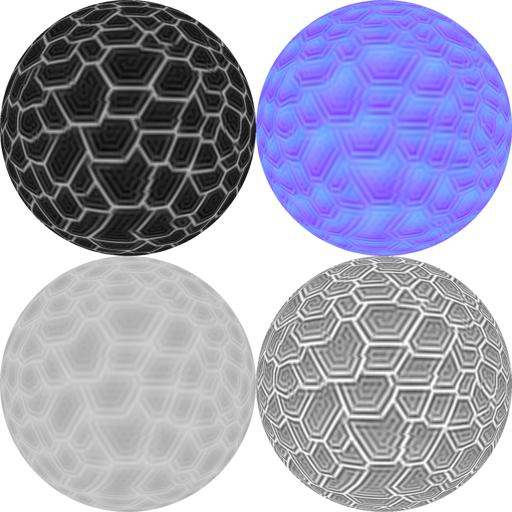}  &  \imgimg{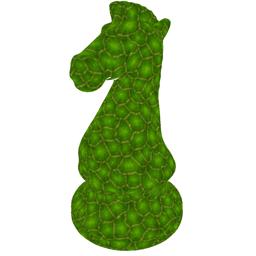}  &  \imgimg{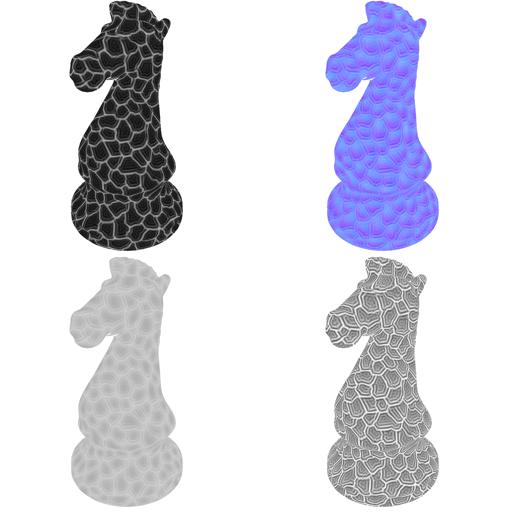}  &  \imgimg{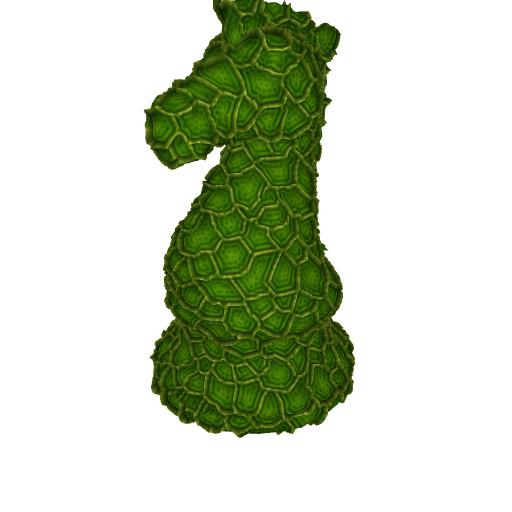} \\
\imgimg{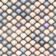}  &  \imgimg{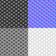}  &  \imgimg{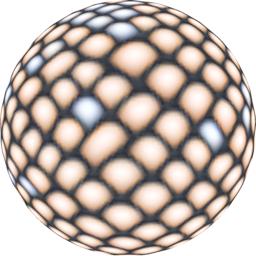}  &  \imgimg{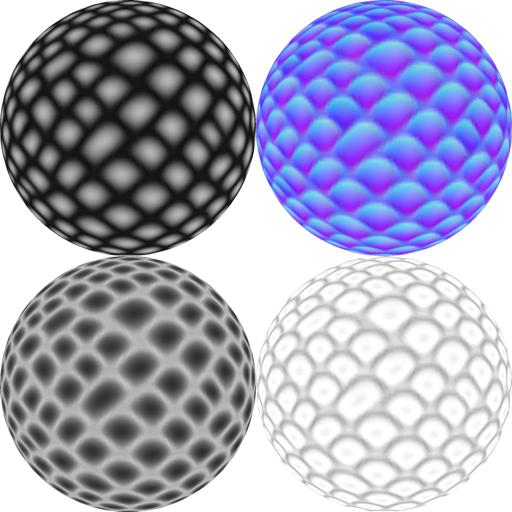}  &  \imgimg{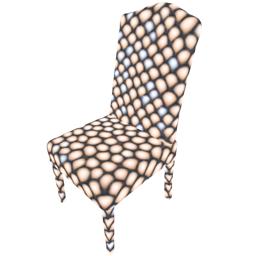}  &  \imgimg{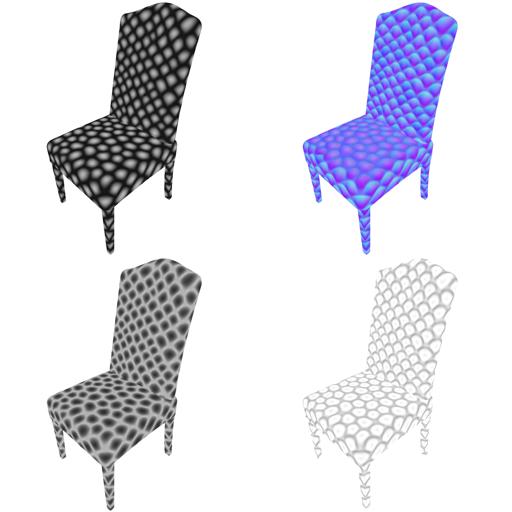}  &  \imgimg{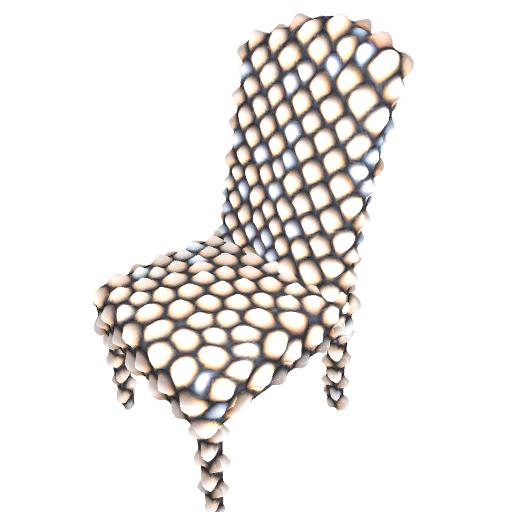} \\
\imgimg{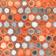}  &  \imgimg{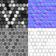}  &  \imgimg{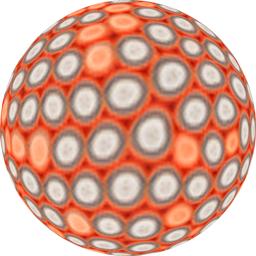}  &  \imgimg{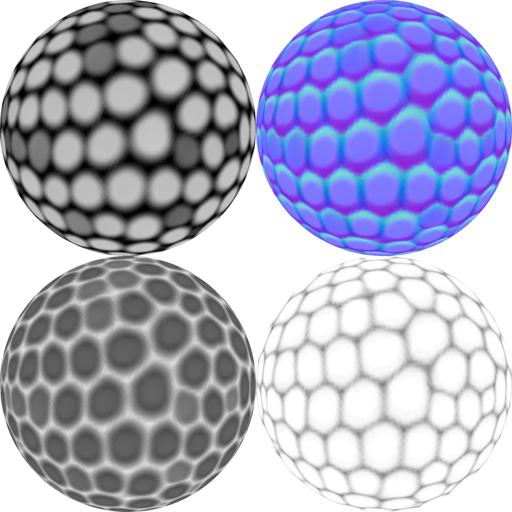}  &  \imgimg{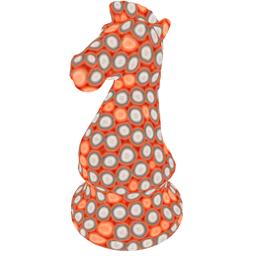}  &  \imgimg{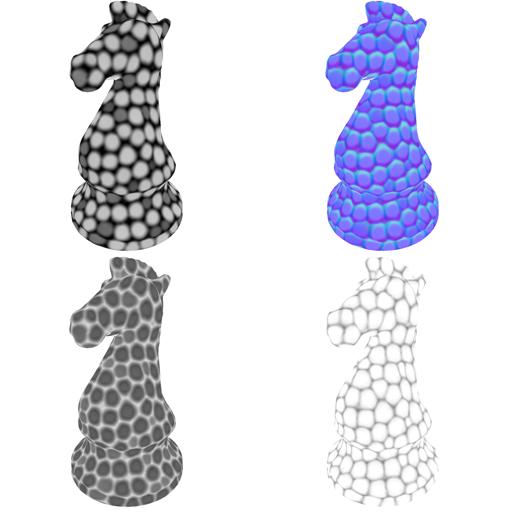}  &  \imgimg{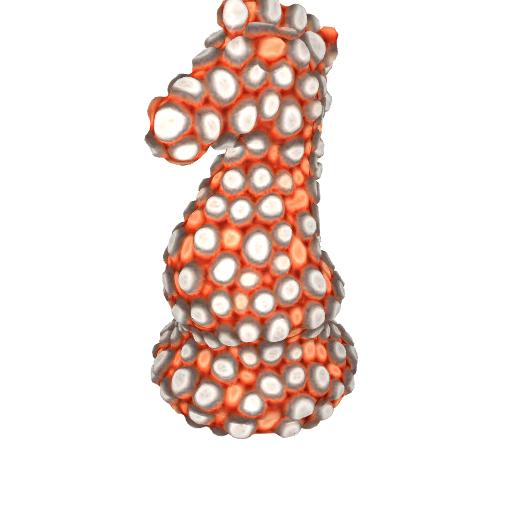}
\end{tabular}
}
\captionof{figure}{\rev{Results of \textbf{Image-guided} 3D texture synthesis. The attributes are height, normal, roughness, and ambient occlusion maps, from top to bottom, left to right. MeshNCA is trained only on the icosphere mesh. At test time, it generalizes to unseen meshes with aligned texture maps. After rendering, the synthesized texture maps show correct shading effects.}}
\label{fig:image-synthesis-result-supp}

\end{table*}

\newcolumntype{H}{>{\centering\arraybackslash} m{25pt} }

\newcommand{\imgtextsupp}[1]{%
  \ifimgprefix
    \includegraphics[height=55pt]{figures/Experiments/supp/text-synthesis/#1}%
  \else
    \includegraphics[height=55pt]{figures/Experiments/supp/text-synthesis/#1}%
  \fi
}

\begin{table*}[]
\begin{tabular}{c||c||cccccc}
\textbf{Prompts} & \textbf{Train} & \multicolumn{6}{c}{\textbf{Test}}  \\
 \midrule
\rotatebox{90}{\parbox[c]{2cm}{\centering\hspace{0pt} \textbf{Animal Fur}}}  &  \imgtextsupp{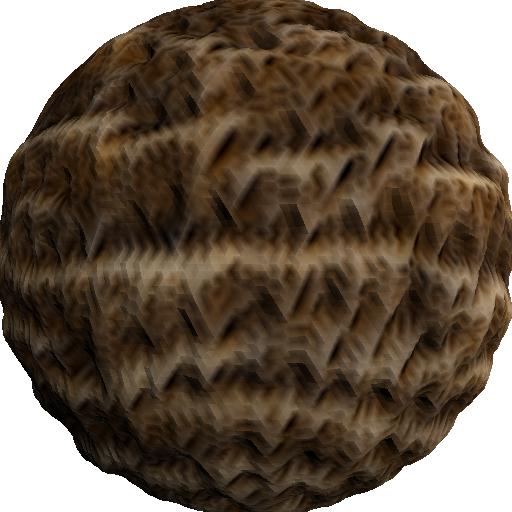}  & \imgtextsupp{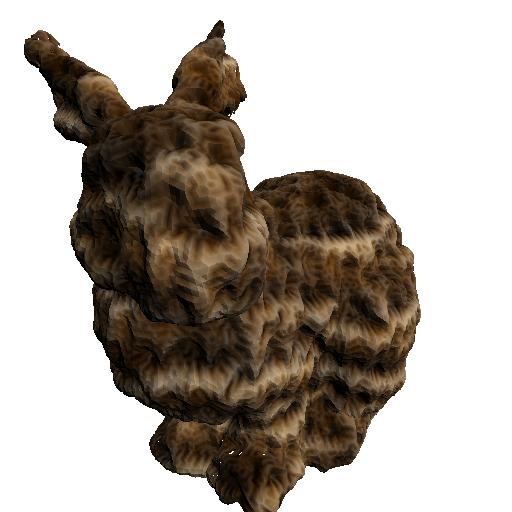} & \imgtextsupp{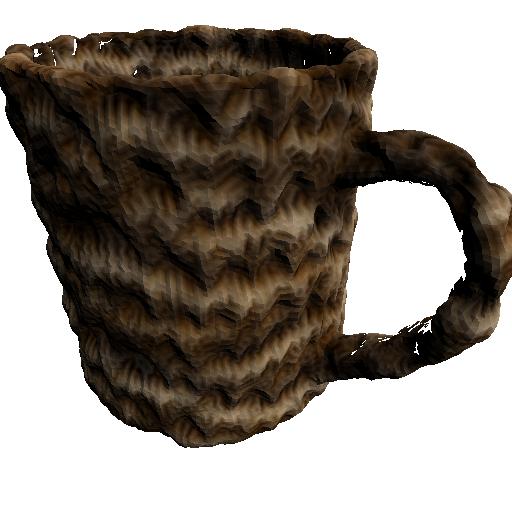} & \imgtextsupp{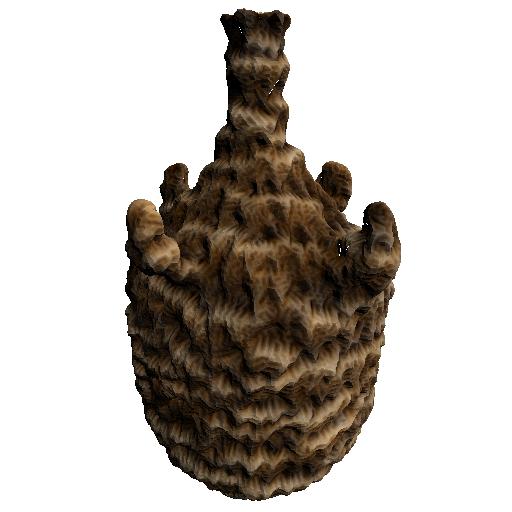} & \imgtextsupp{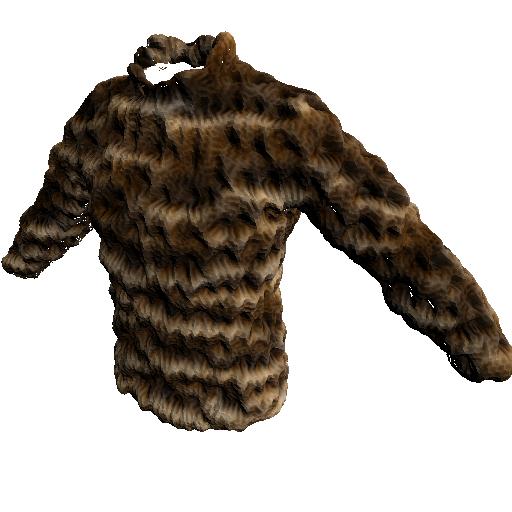} & \imgtextsupp{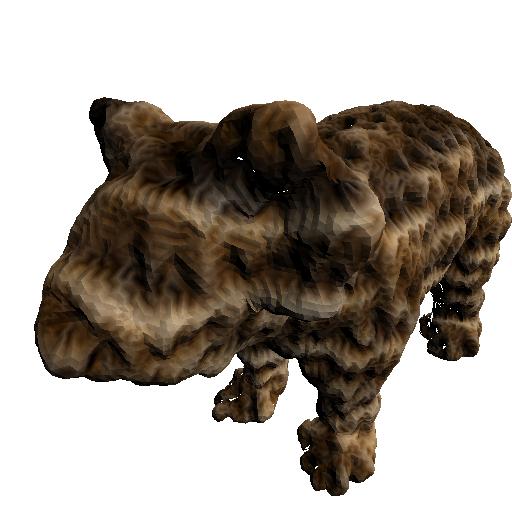} & \imgtextsupp{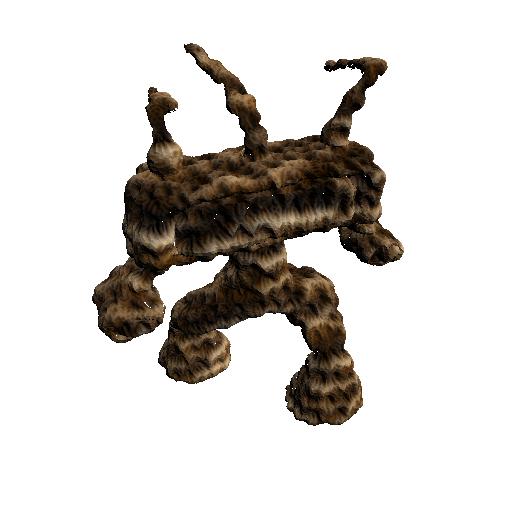}
\\
\rotatebox{90}{\parbox[c]{2cm}{\centering\hspace{0pt} \textbf{Cactus}}}  &  \imgtextsupp{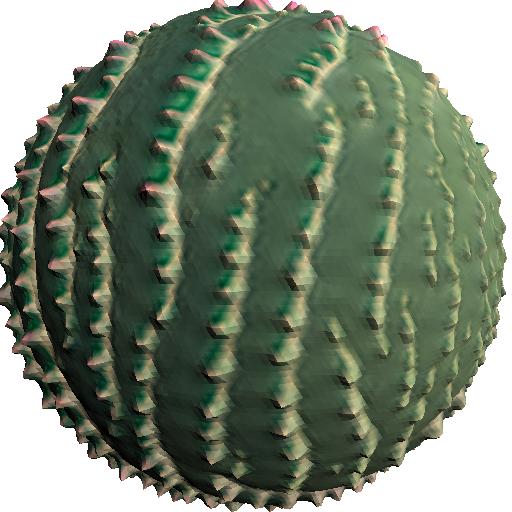}  & \imgtextsupp{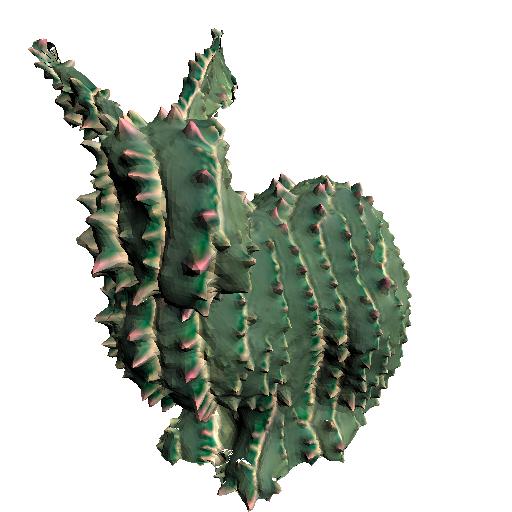} & \imgtextsupp{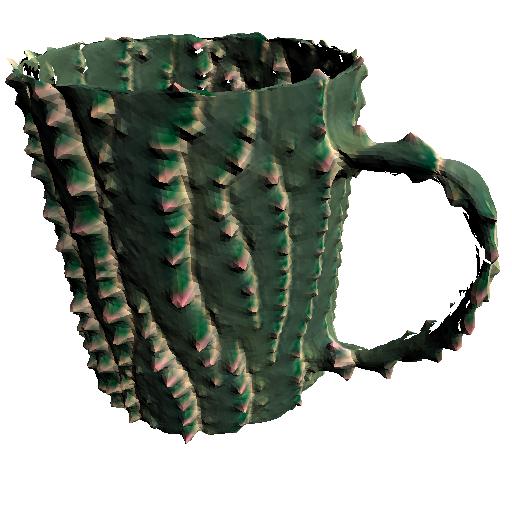} & \imgtextsupp{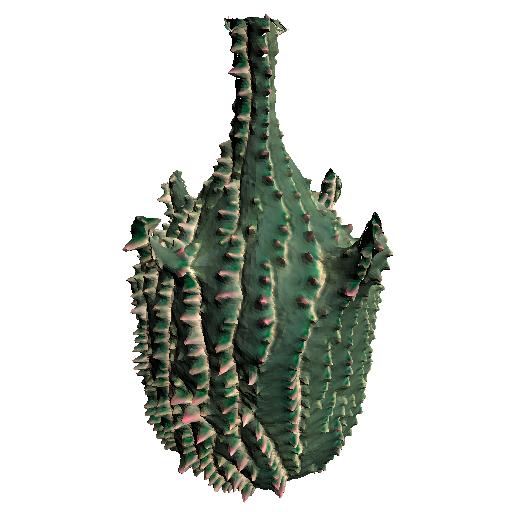} & \imgtextsupp{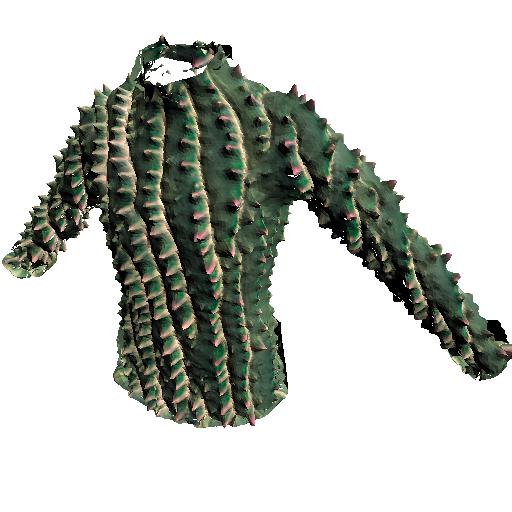} & \imgtextsupp{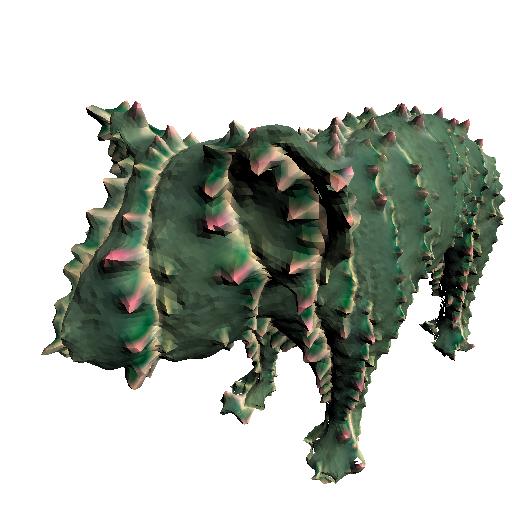} & \imgtextsupp{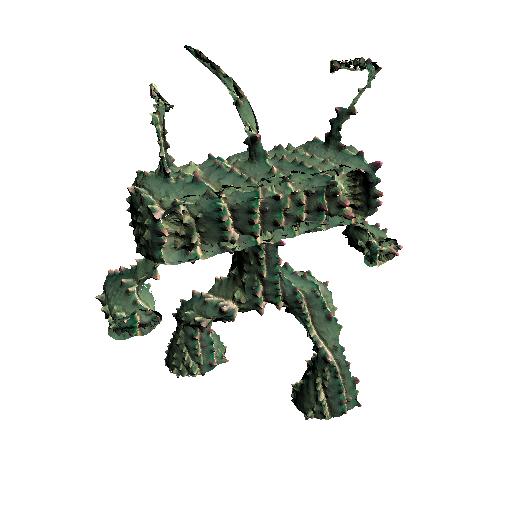}
\\
\rotatebox{90}{\parbox[c]{2cm}{\centering\hspace{0pt} \textbf{Feathers}}}  &  \imgtextsupp{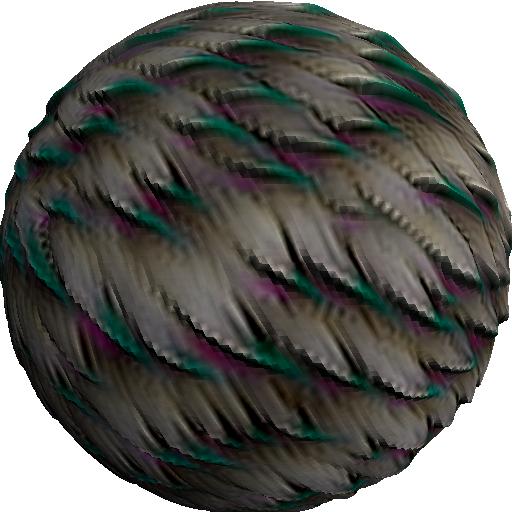}  & \imgtextsupp{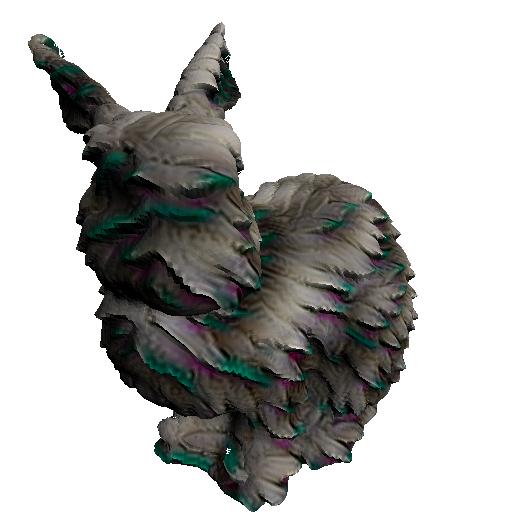} & \imgtextsupp{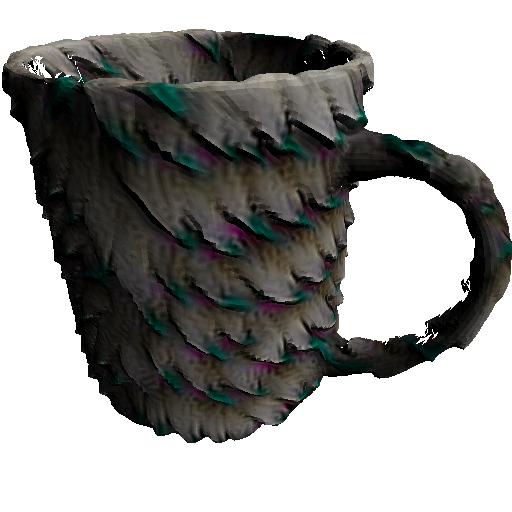} & \imgtextsupp{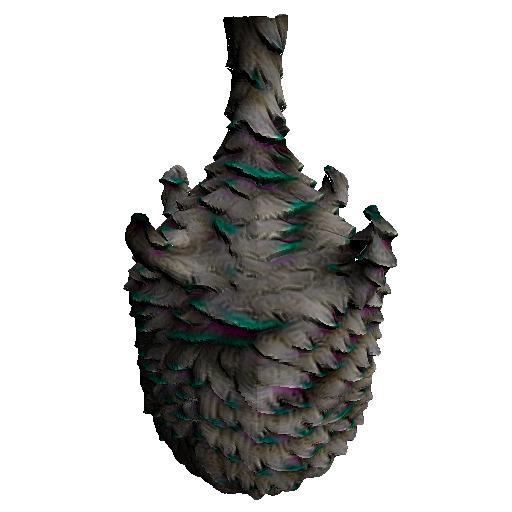} & \imgtextsupp{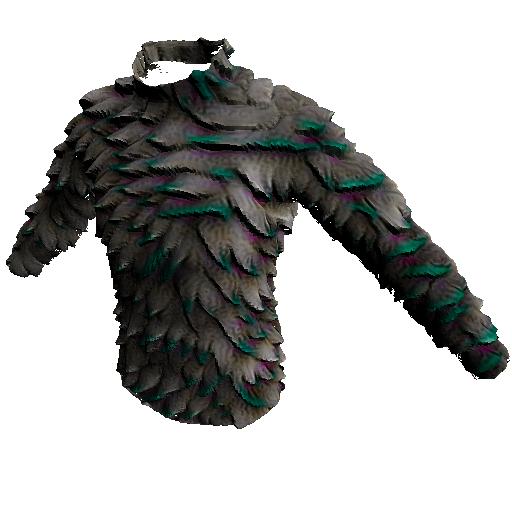} & \imgtextsupp{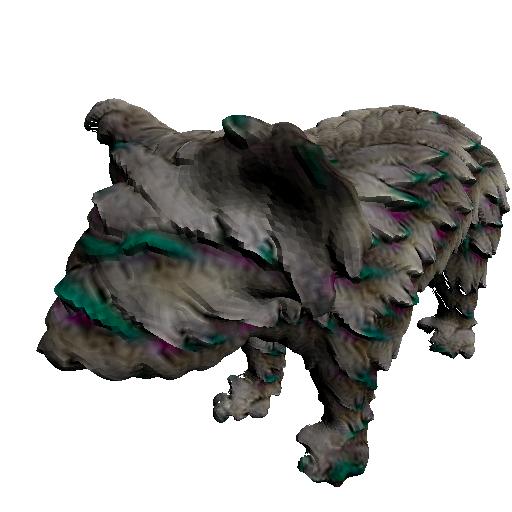} & \imgtextsupp{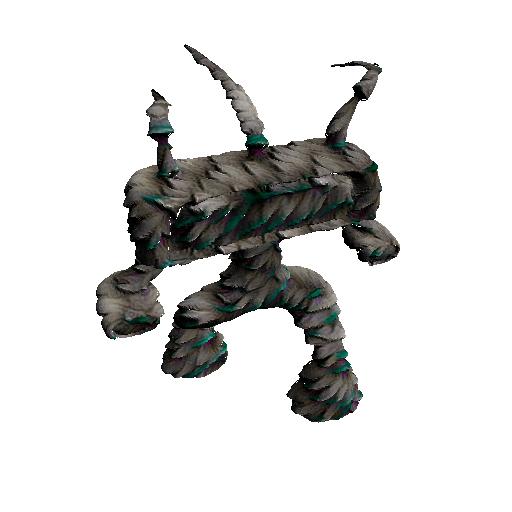}
\\
\rotatebox{90}{\parbox[c]{2cm}{\centering\hspace{0pt} \textbf{Patchwork Leather}}}  &  \imgtextsupp{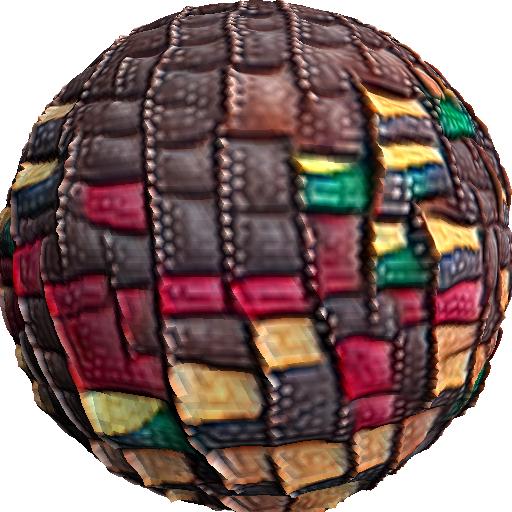}  & \imgtextsupp{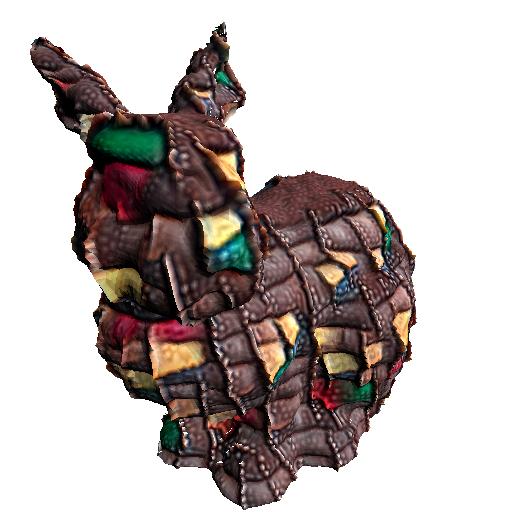} & \imgtextsupp{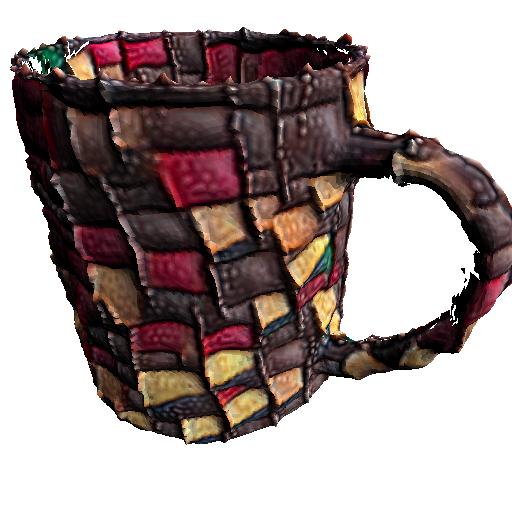} & \imgtextsupp{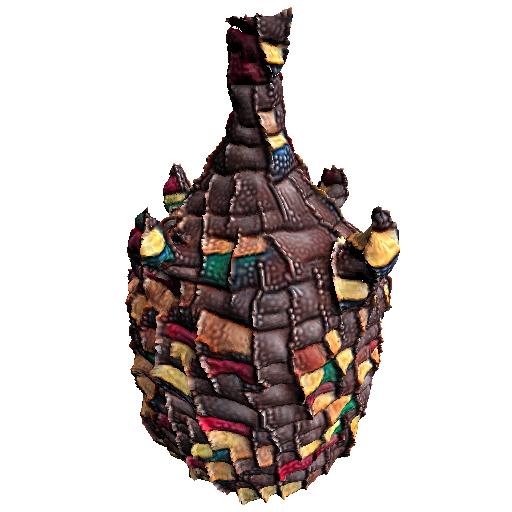} & \imgtextsupp{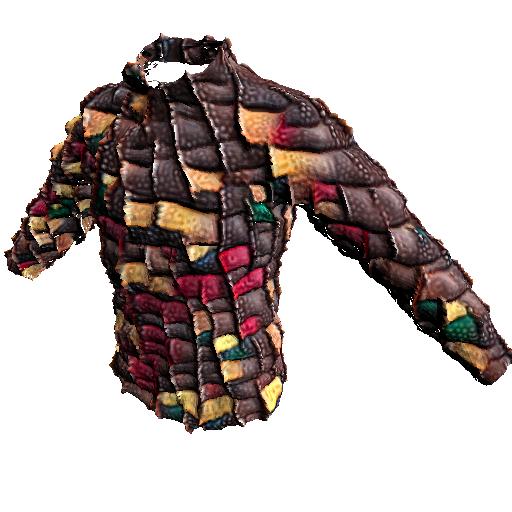} & \imgtextsupp{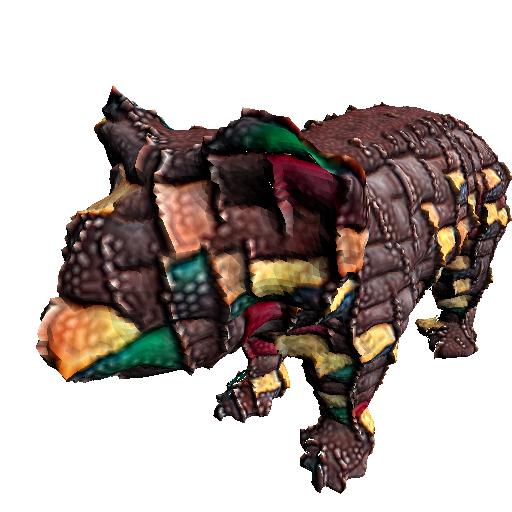} & \imgtextsupp{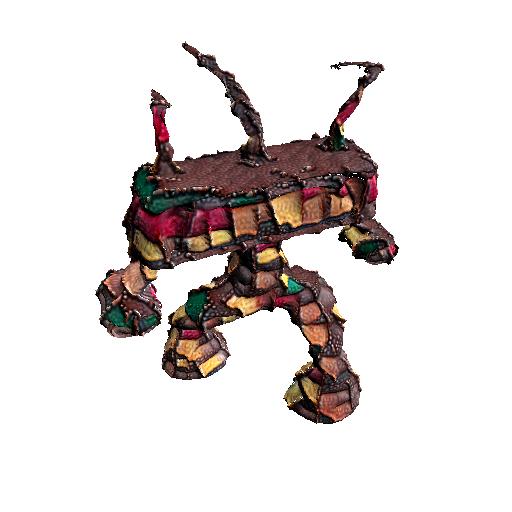}
\\
\rotatebox{90}{\parbox[c]{2cm}{\centering\hspace{0pt} \textbf{Stained Glass}}}  &  \imgtextsupp{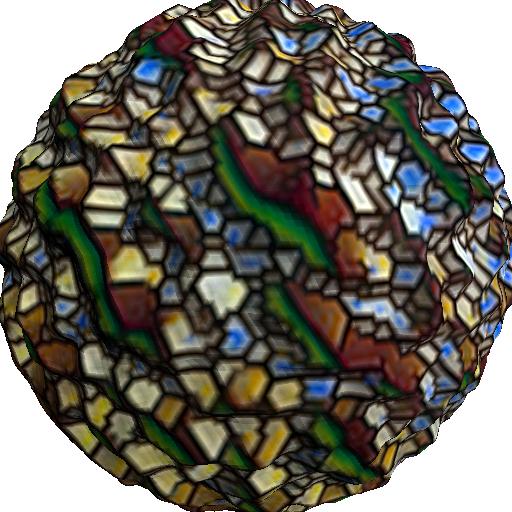}  & \imgtextsupp{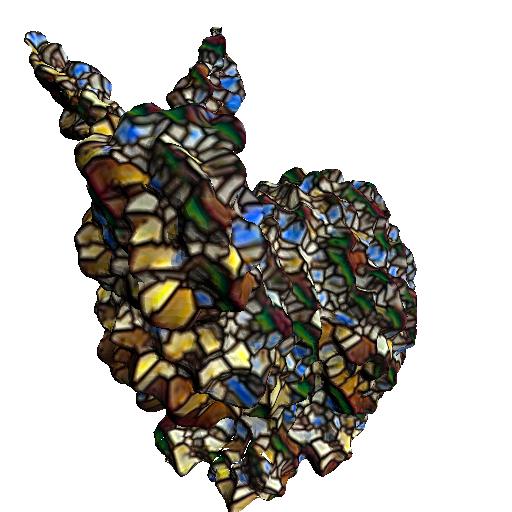} & \imgtextsupp{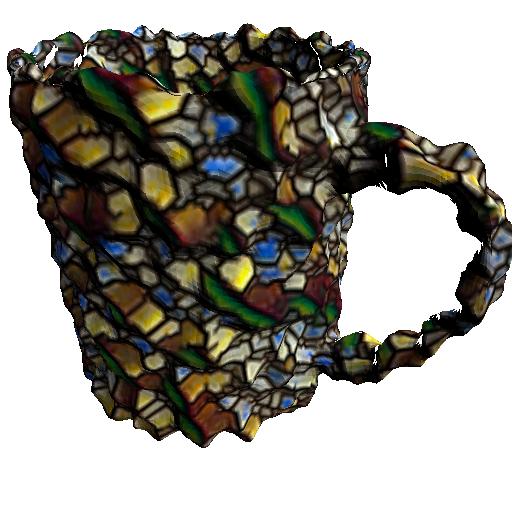} & \imgtextsupp{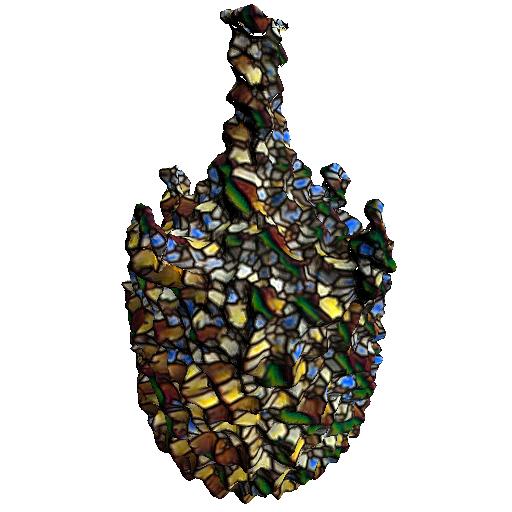} & \imgtextsupp{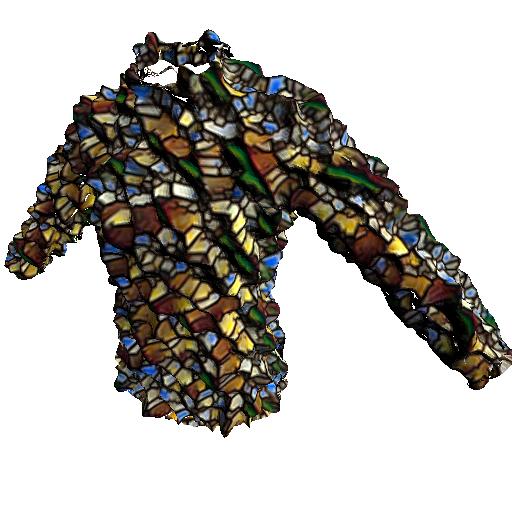} & \imgtextsupp{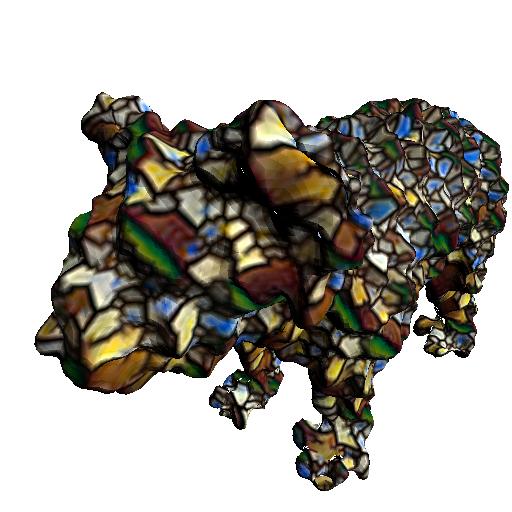} & \imgtextsupp{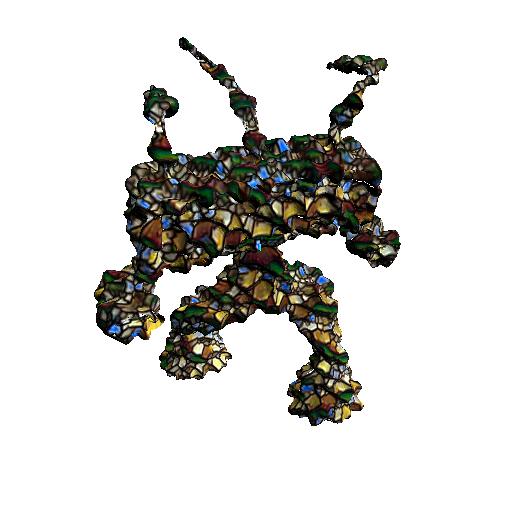}
\\
\rotatebox{90}{\parbox[c]{2cm}{\centering\hspace{0pt} \textbf{Marble}}}  &  \imgtextsupp{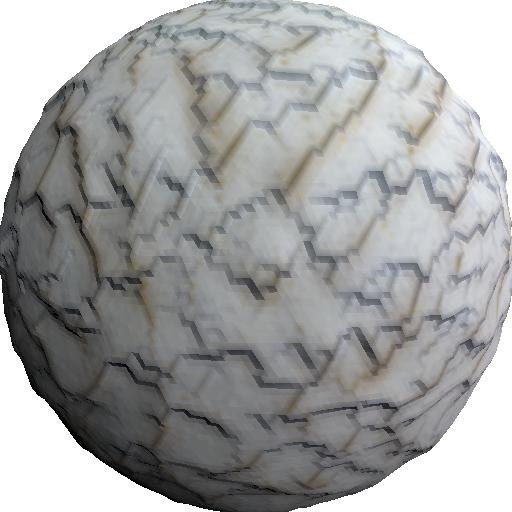}  & \imgtextsupp{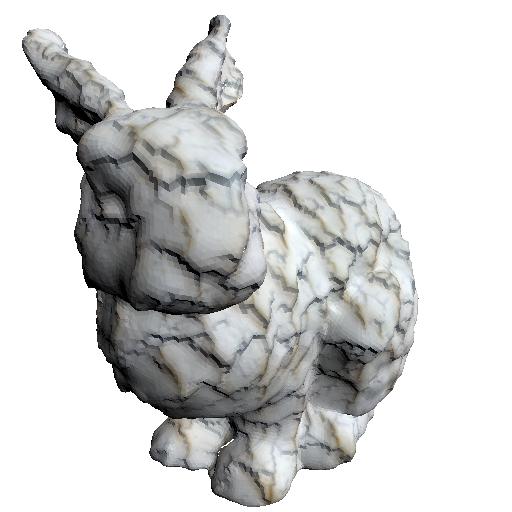} & \imgtextsupp{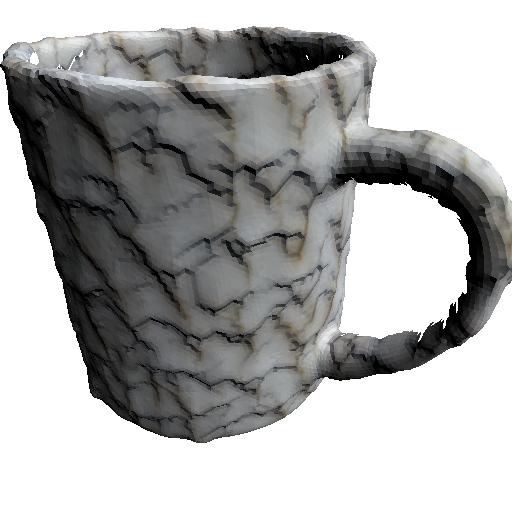} & \imgtextsupp{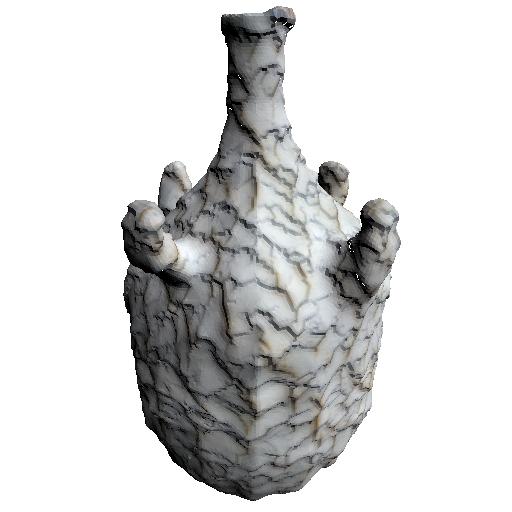} & \imgtextsupp{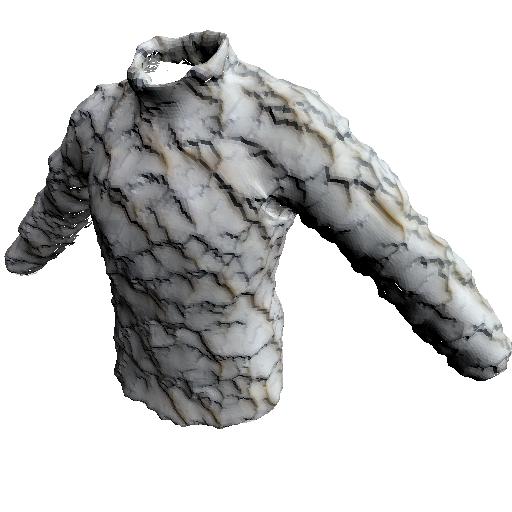} & \imgtextsupp{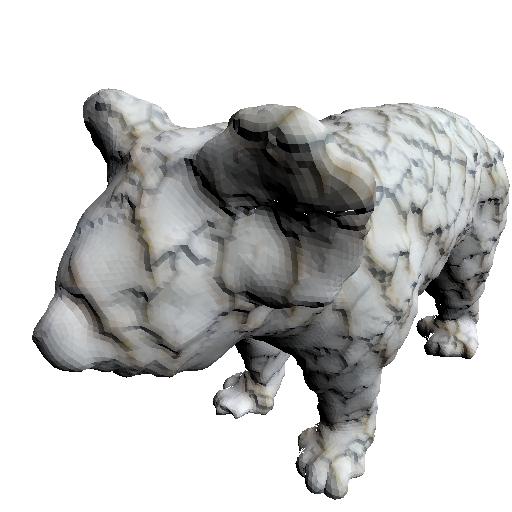} & \imgtextsupp{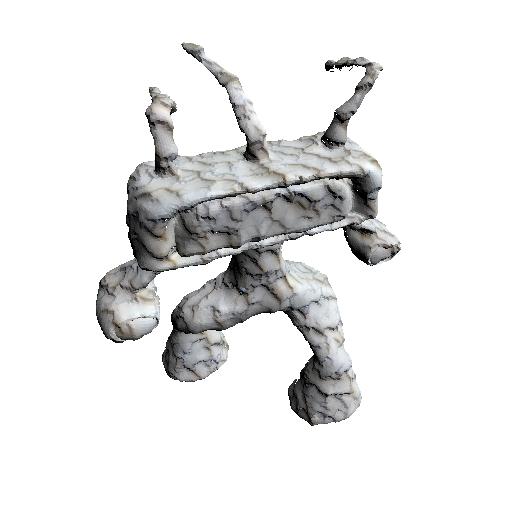}
\\
\rotatebox{90}{\parbox[c]{2cm}{\centering\hspace{0pt} \textbf{Bark}}}  &  \imgtextsupp{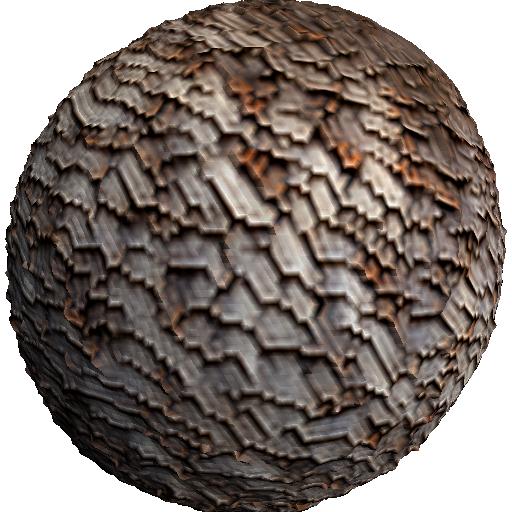}  & \imgtextsupp{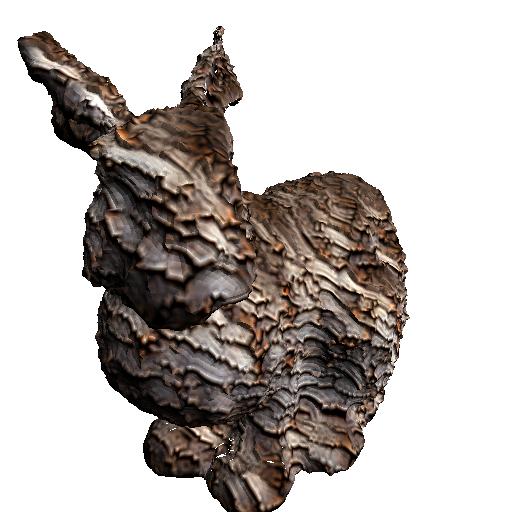} & \imgtextsupp{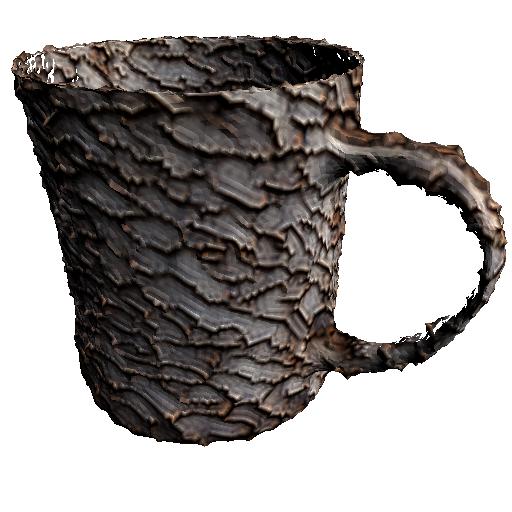} & \imgtextsupp{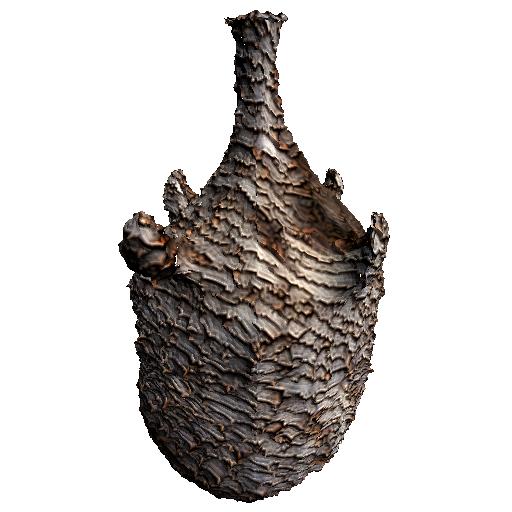} & \imgtextsupp{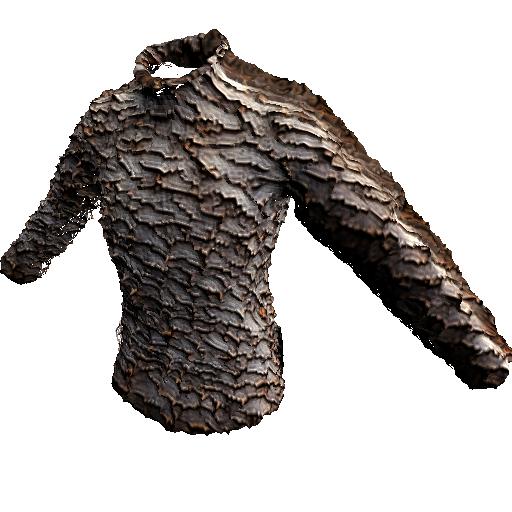} & \imgtextsupp{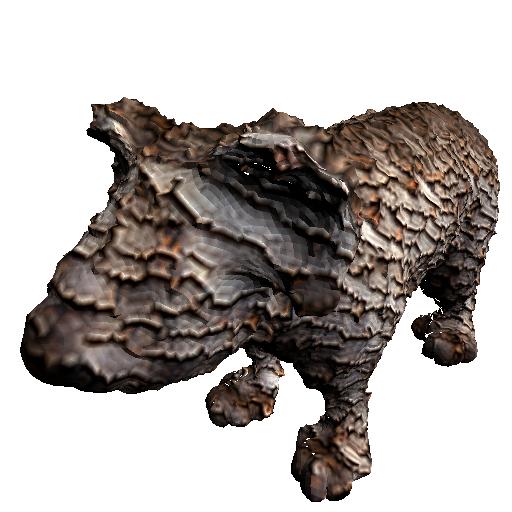} & \imgtextsupp{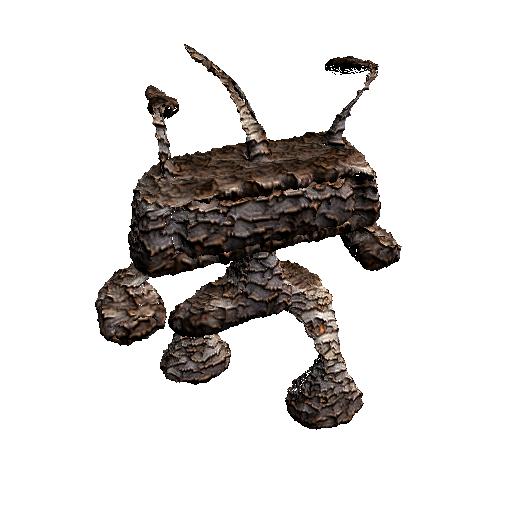}
\end{tabular}
\captionof{figure}{Results of \textbf{Text-guided} 3D texture synthesis.}
\label{fig:text-synthesis-result-supp1}
\end{table*}

\begin{table*}[]
\begin{tabular}{c||c||cccccc}
\textbf{Prompts} & \textbf{Train} & \multicolumn{6}{c}{\textbf{Test}}  \\
 \midrule
\rotatebox{90}{\parbox[c]{2cm}{\centering\hspace{0pt} \textbf{Wicker}}}  &  \imgtextsupp{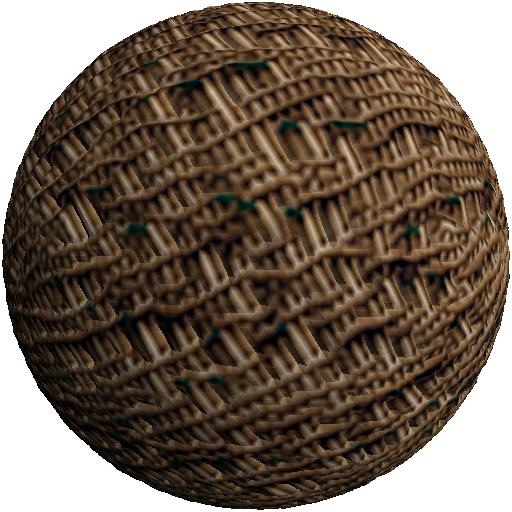}  & \imgtextsupp{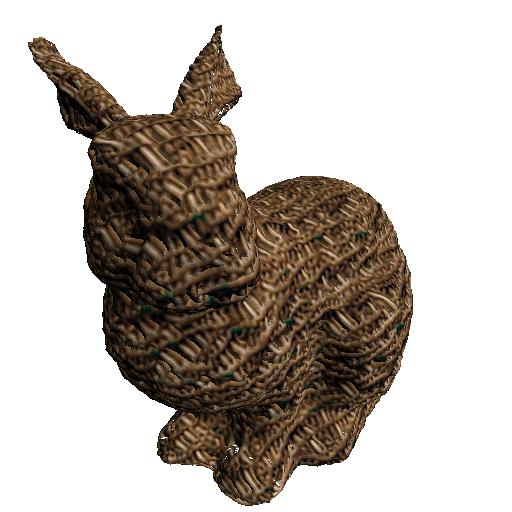} & \imgtextsupp{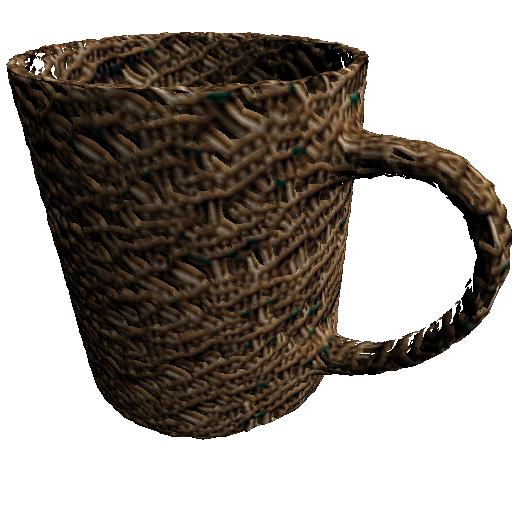} & \imgtextsupp{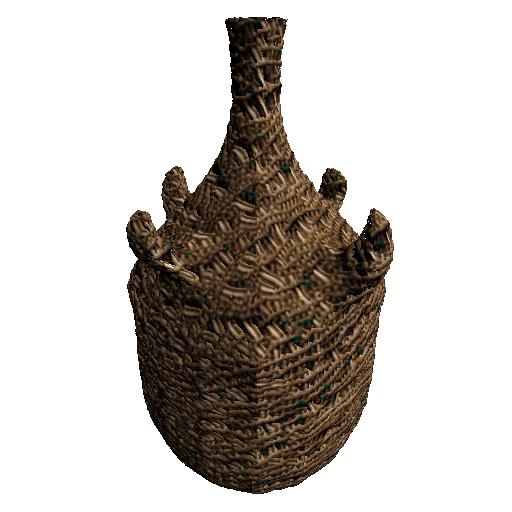} & \imgtextsupp{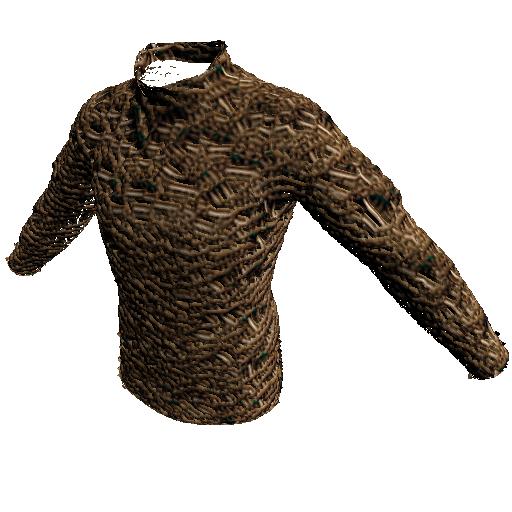} & \imgtextsupp{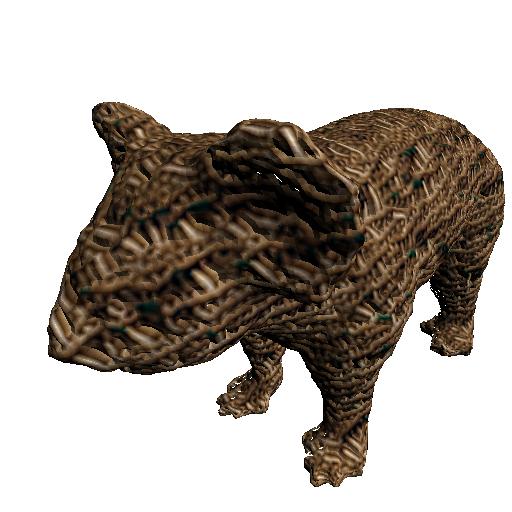} & \imgtextsupp{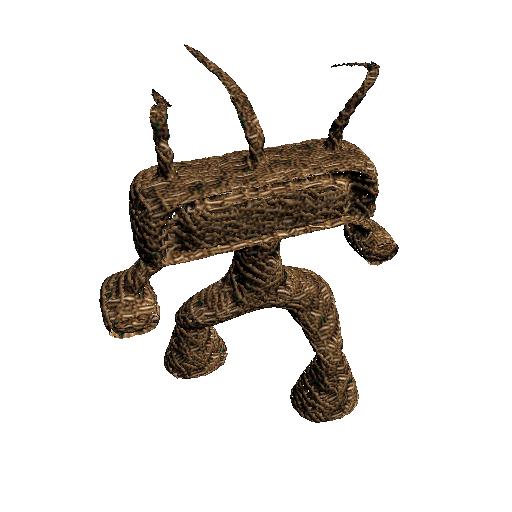}
\\
\rotatebox{90}{\parbox[c]{2cm}{\centering\hspace{0pt} \textbf{Brick}}}  &  \imgtextsupp{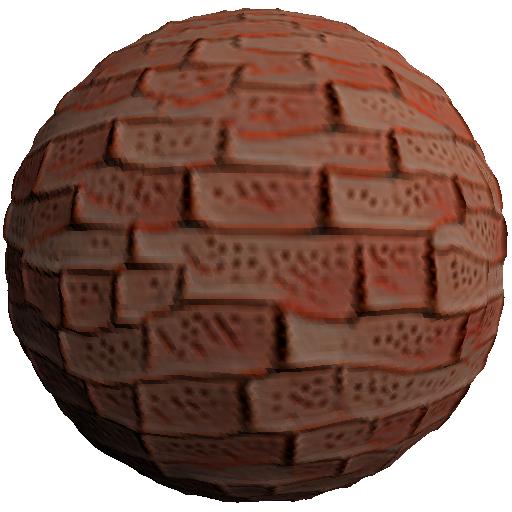}  & \imgtextsupp{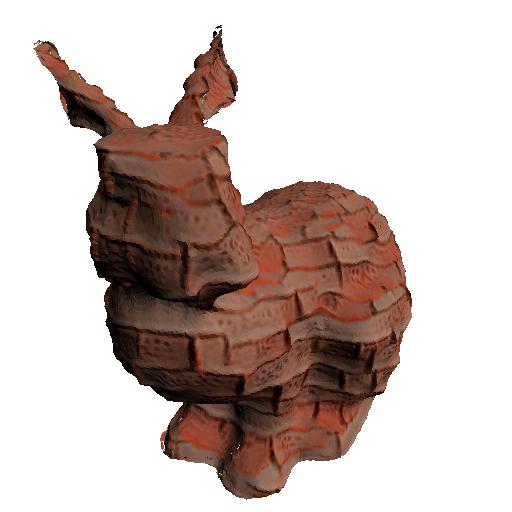} & \imgtextsupp{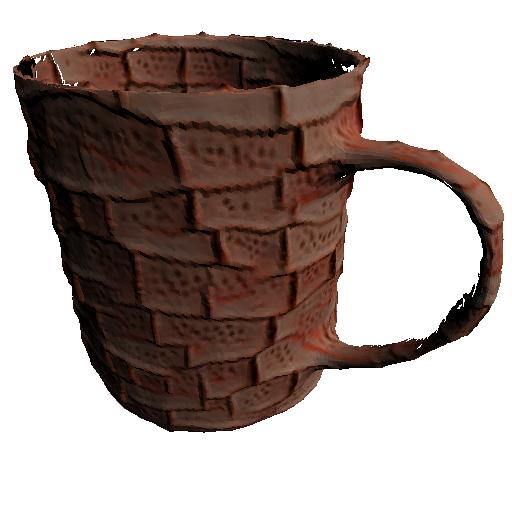} & \imgtextsupp{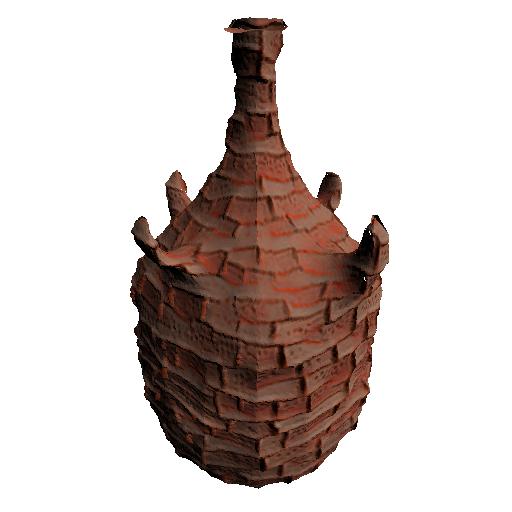} & \imgtextsupp{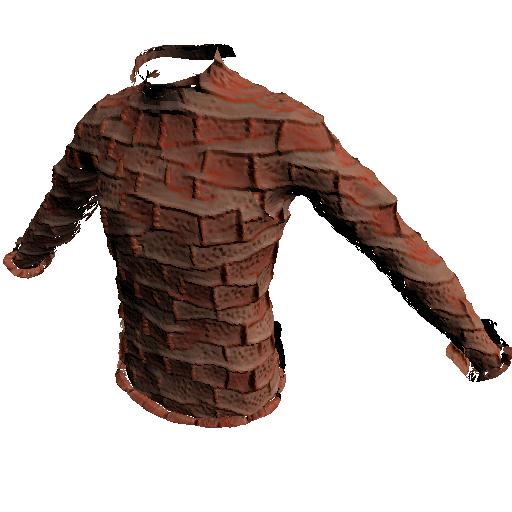} & \imgtextsupp{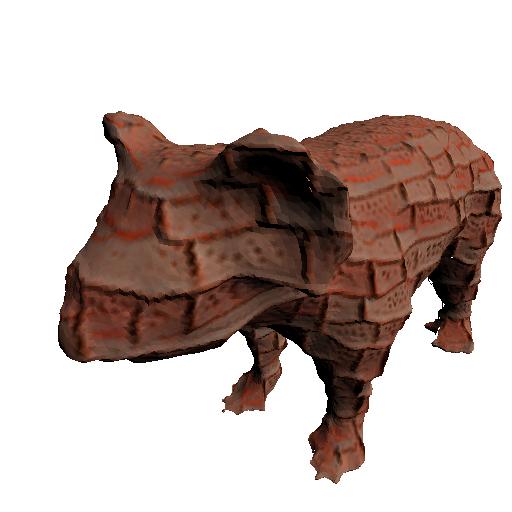} & \imgtextsupp{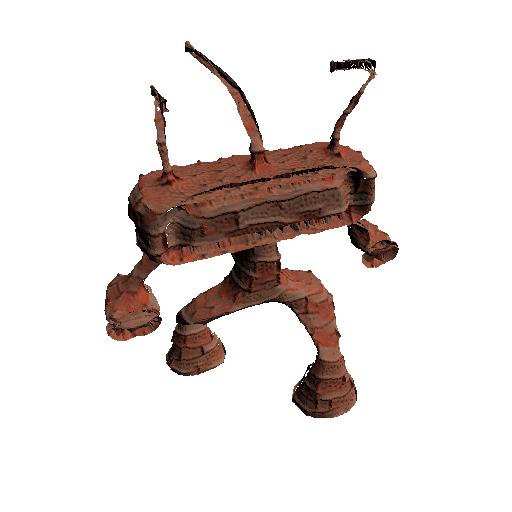}
\\
\rotatebox{90}{\parbox[c]{2cm}{\centering\hspace{0pt} \textbf{Ceramic Tiles}}}  &  \imgtextsupp{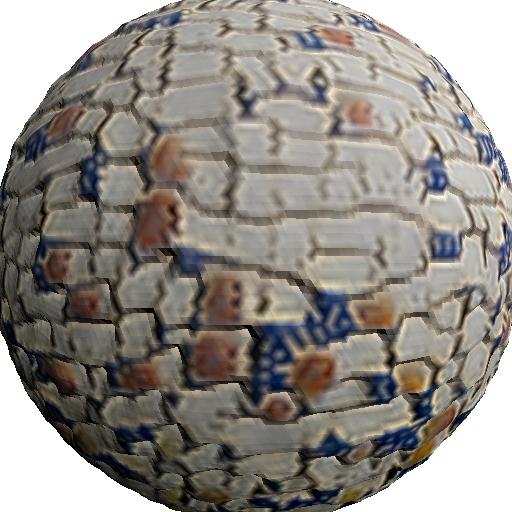}  & \imgtextsupp{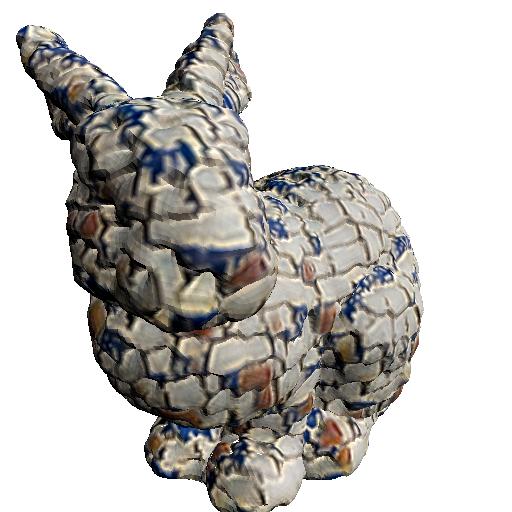} & \imgtextsupp{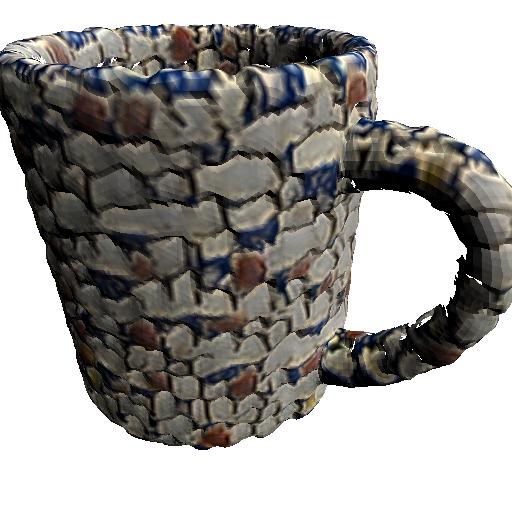} & \imgtextsupp{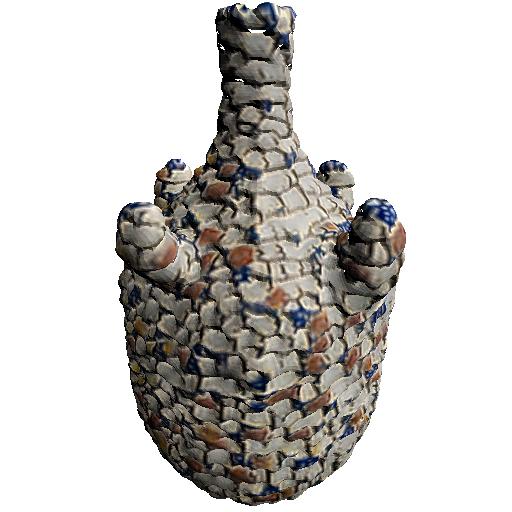} & \imgtextsupp{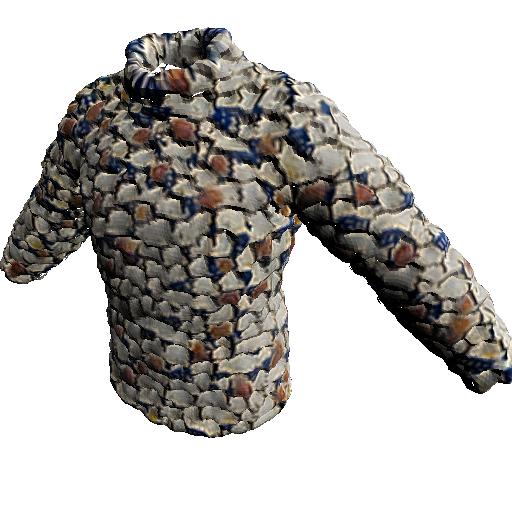} & \imgtextsupp{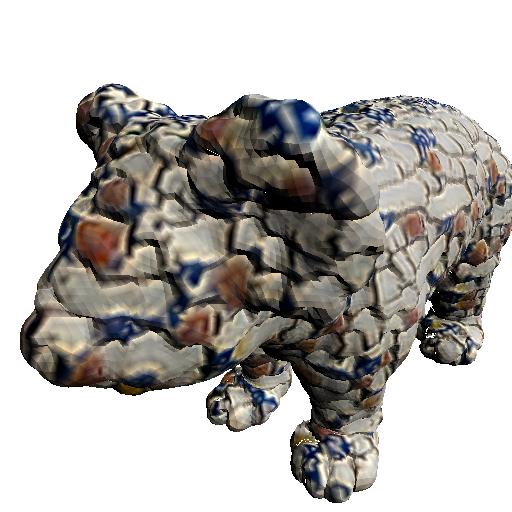} & \imgtextsupp{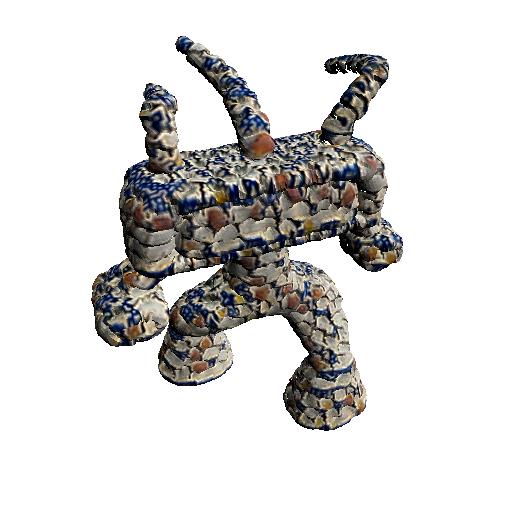}
\\
\rotatebox{90}{\parbox[c]{2cm}{\centering\hspace{0pt} \textbf{Chainmail}}}  &  \imgtextsupp{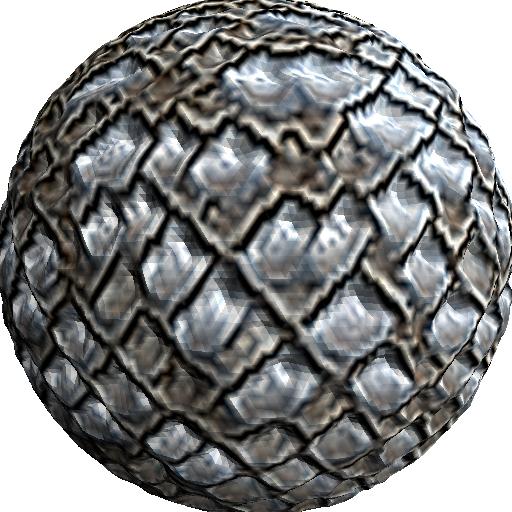}  & \imgtextsupp{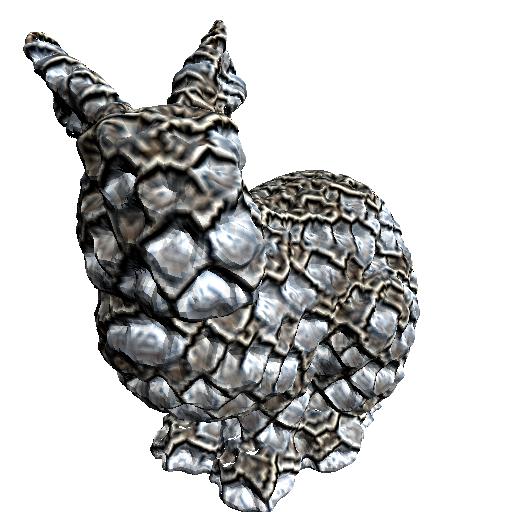} & \imgtextsupp{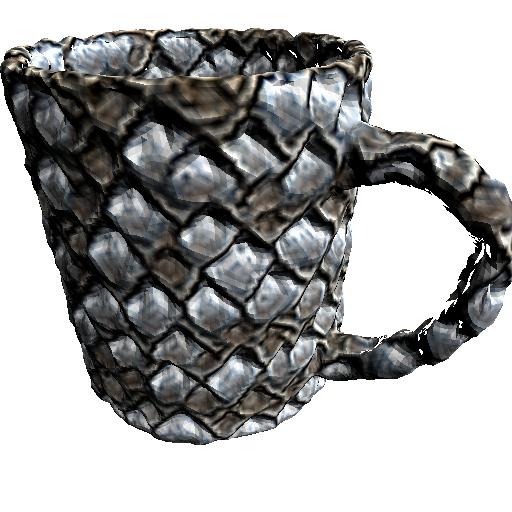} & \imgtextsupp{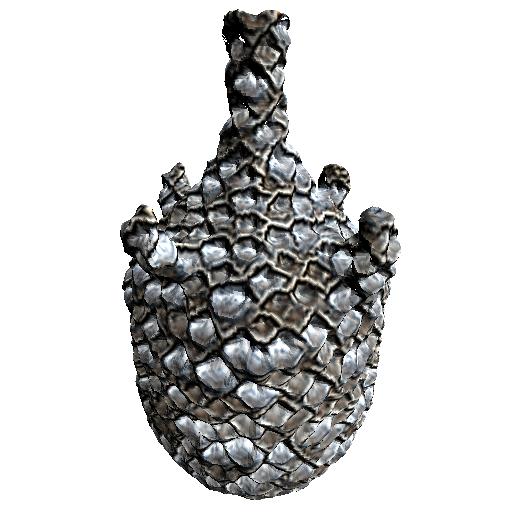} & \imgtextsupp{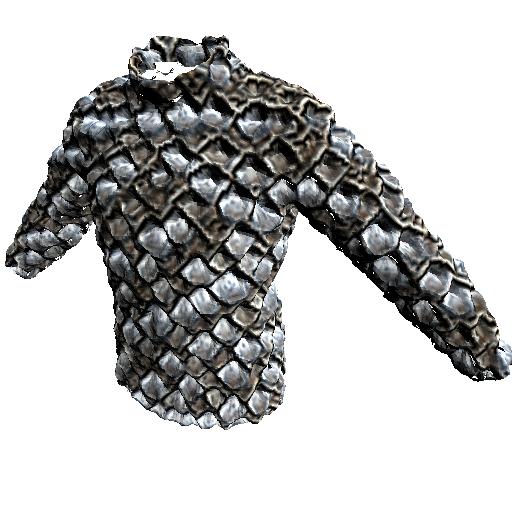} & \imgtextsupp{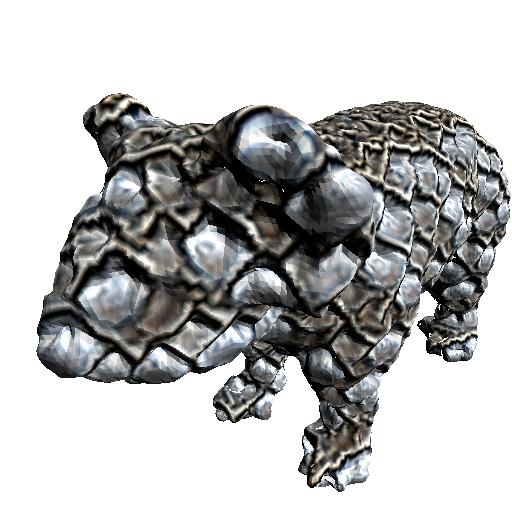} & \imgtextsupp{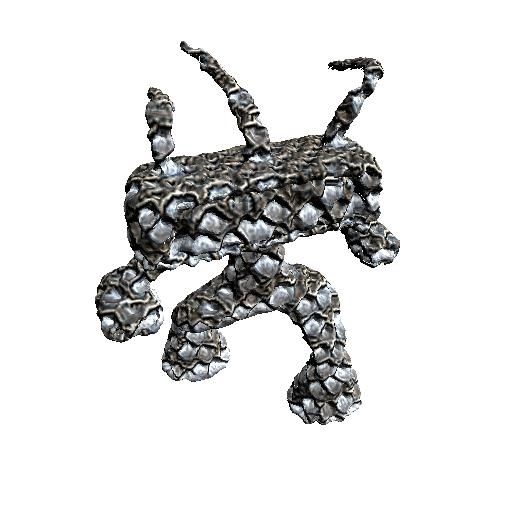}
\\
\rotatebox{90}{\parbox[c]{2cm}{\centering\hspace{0pt} \textbf{Colorful Crochet}}}  &  \imgtextsupp{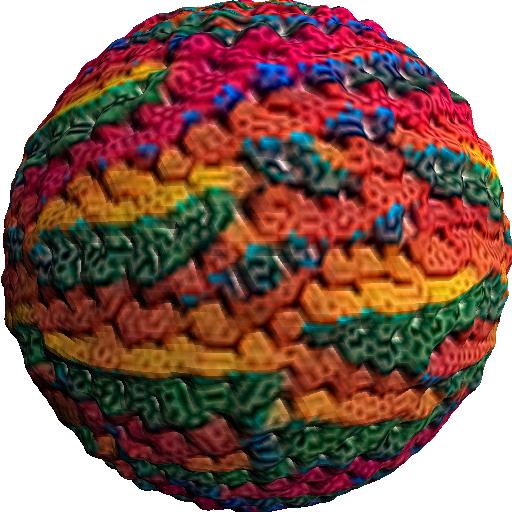}  & \imgtextsupp{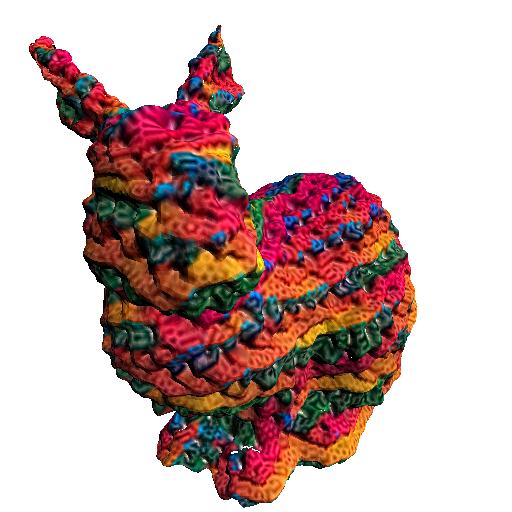} & \imgtextsupp{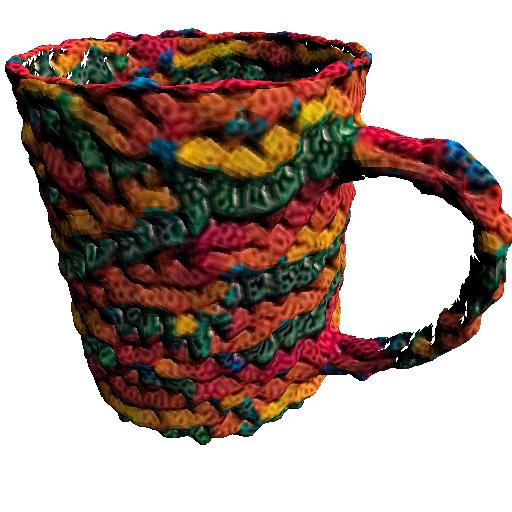} & \imgtextsupp{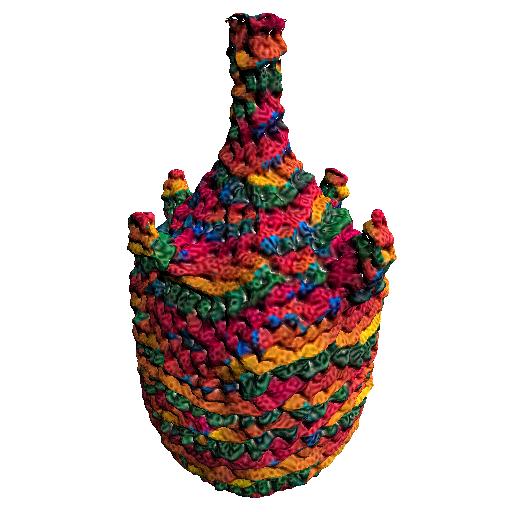} & \imgtextsupp{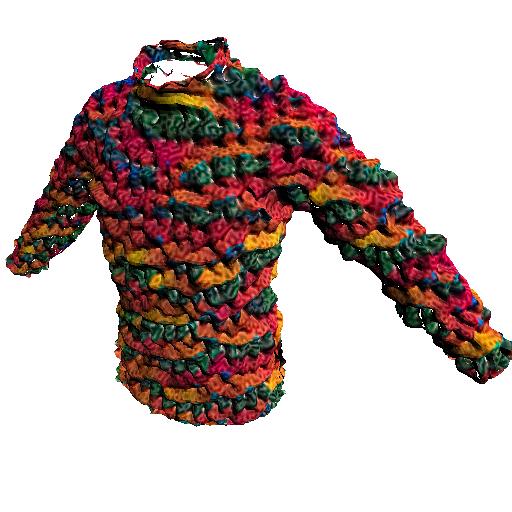} & \imgtextsupp{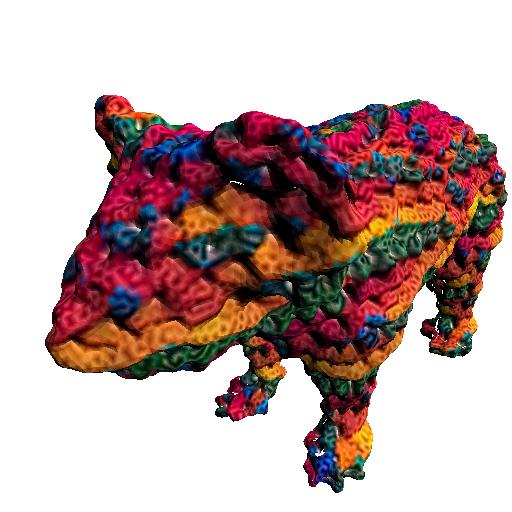} & \imgtextsupp{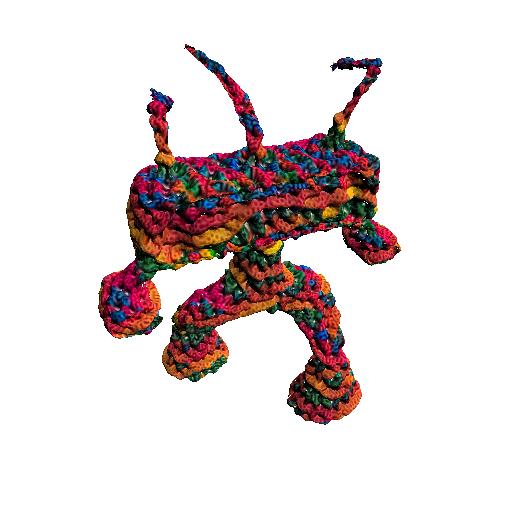}
\\
\rotatebox{90}{\parbox[c]{2cm}{\centering\hspace{0pt} \textbf{Colorful Shells}}}  &  \imgtextsupp{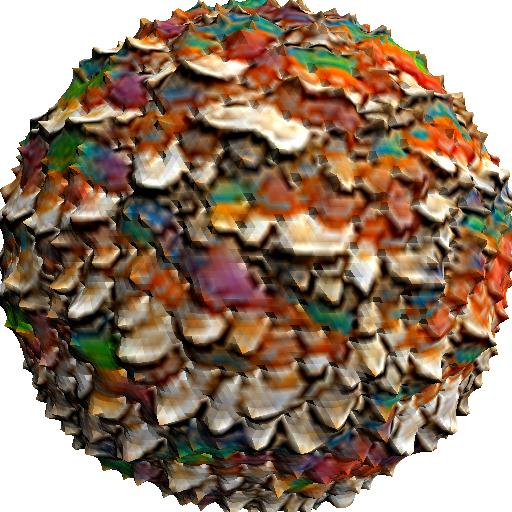}  & \imgtextsupp{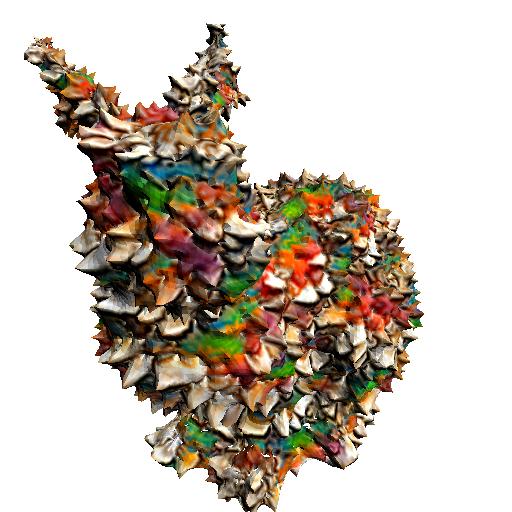} & \imgtextsupp{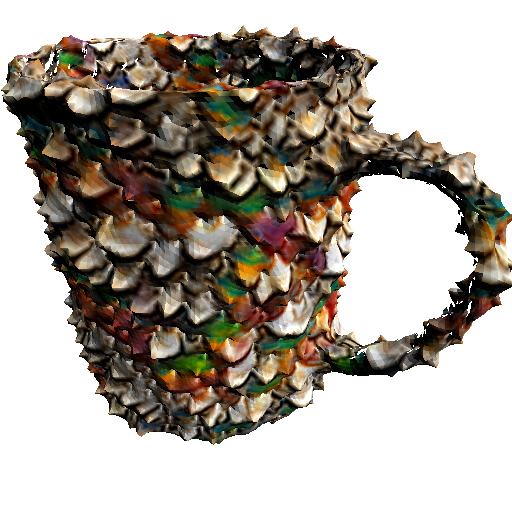} & \imgtextsupp{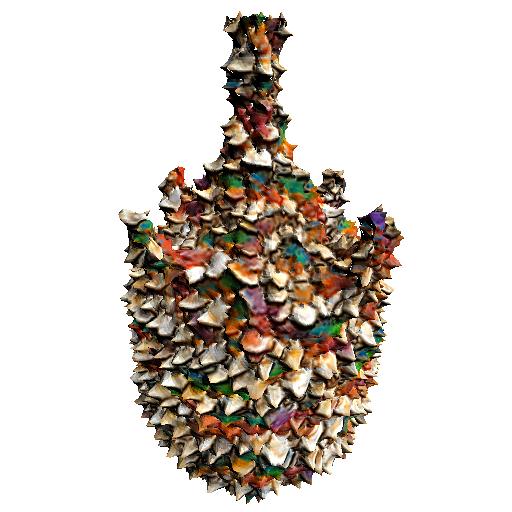} & \imgtextsupp{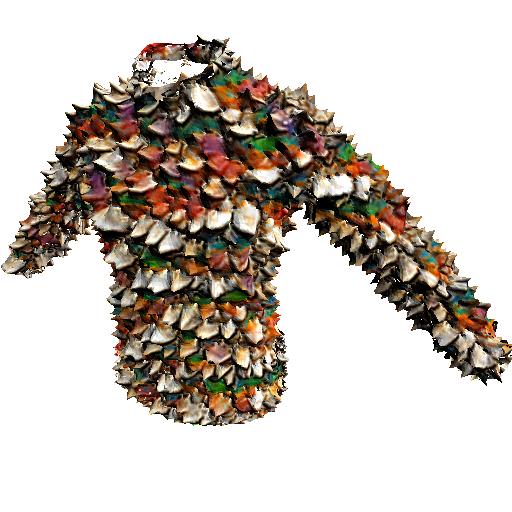} & \imgtextsupp{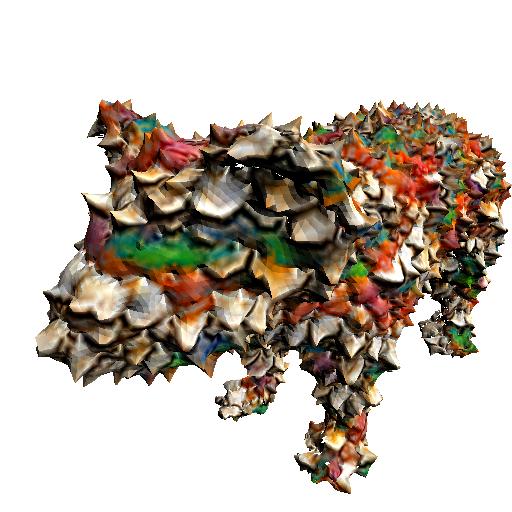} & \imgtextsupp{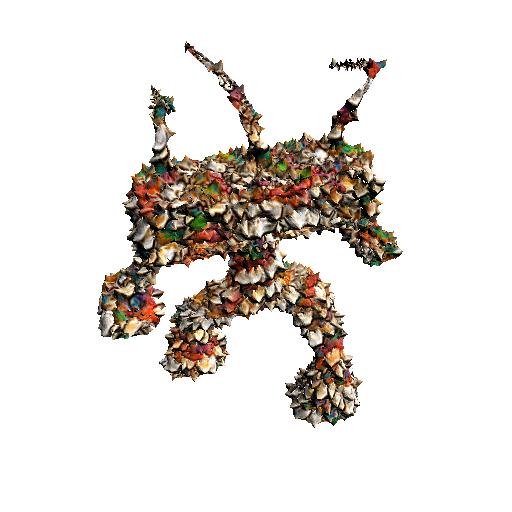}
\\
\rotatebox{90}{\parbox[c]{2cm}{\centering\hspace{0pt} \textbf{Crumpled Paper}}}  &  \imgtextsupp{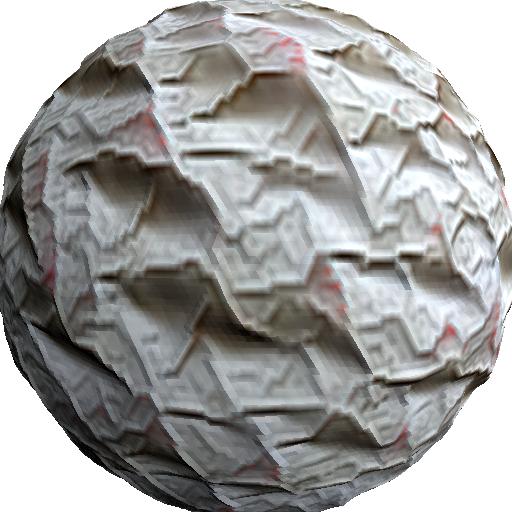}  & \imgtextsupp{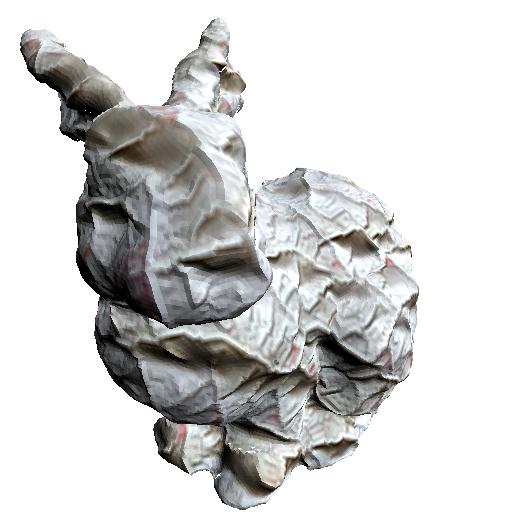} & \imgtextsupp{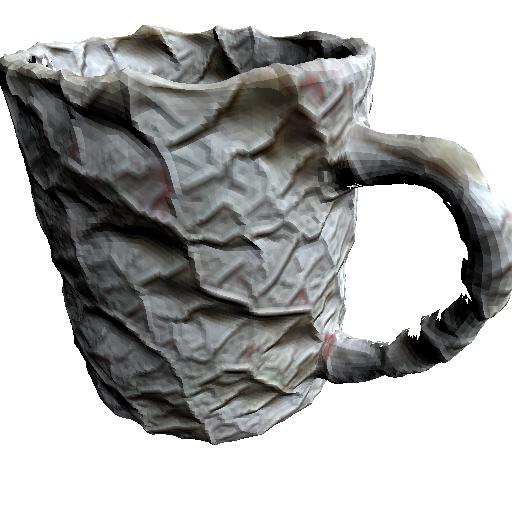} & \imgtextsupp{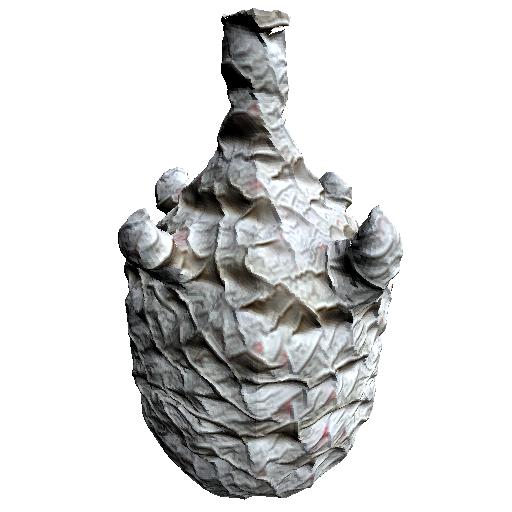} & \imgtextsupp{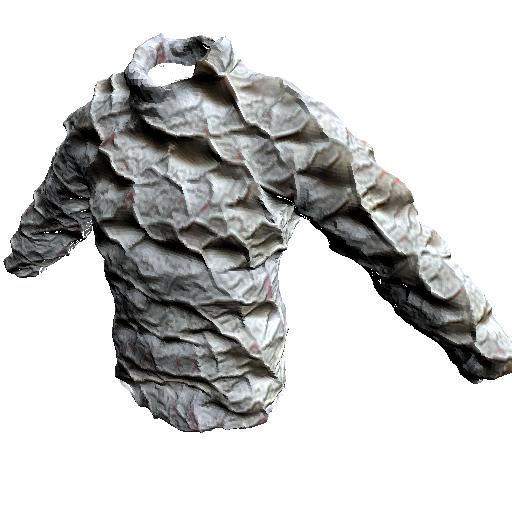} & \imgtextsupp{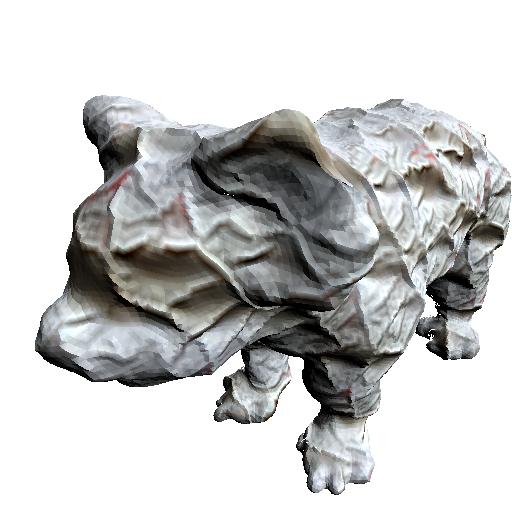} & \imgtextsupp{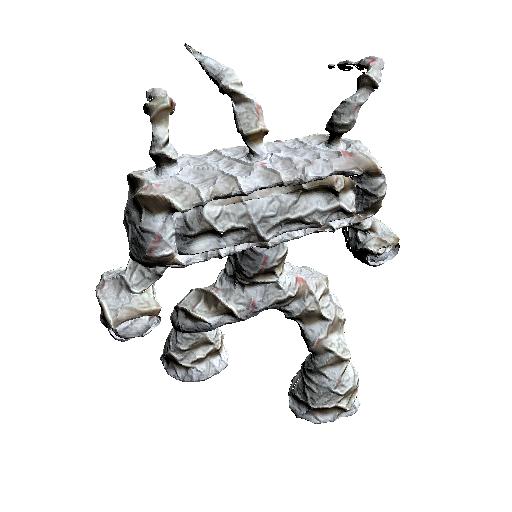}
\end{tabular}
\captionof{figure}{Results of \textbf{Text-guided} 3D texture synthesis.}
\label{fig:text-synthesis-result-supp2}
\end{table*}

\begin{table*}[]
\begin{tabular}{c||c||cccccc}
\textbf{Prompts} & \textbf{Train} & \multicolumn{6}{c}{\textbf{Test}}  \\
 \midrule
\rotatebox{90}{\parbox[c]{2cm}{\centering\hspace{0pt} \textbf{Moss}}}  &  \imgtextsupp{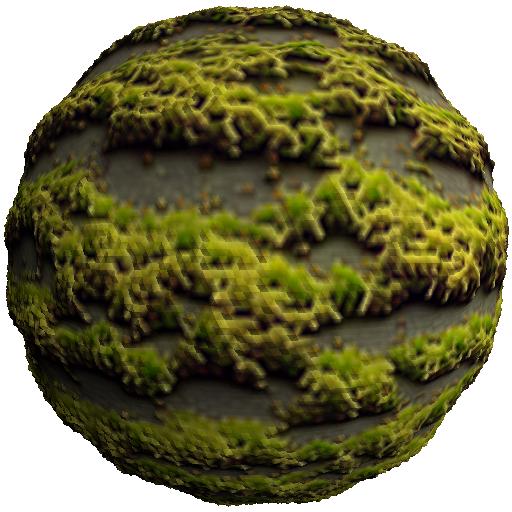}  & \imgtextsupp{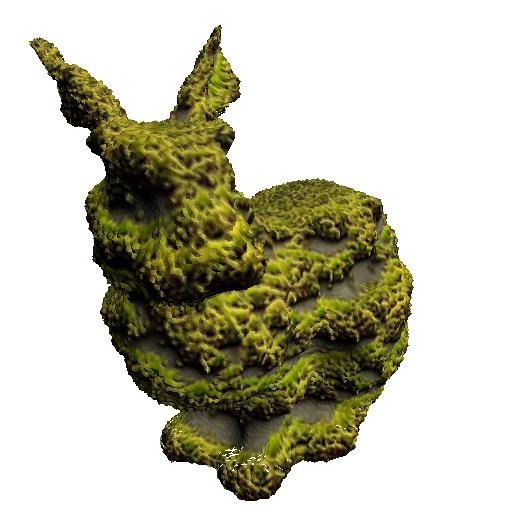} & \imgtextsupp{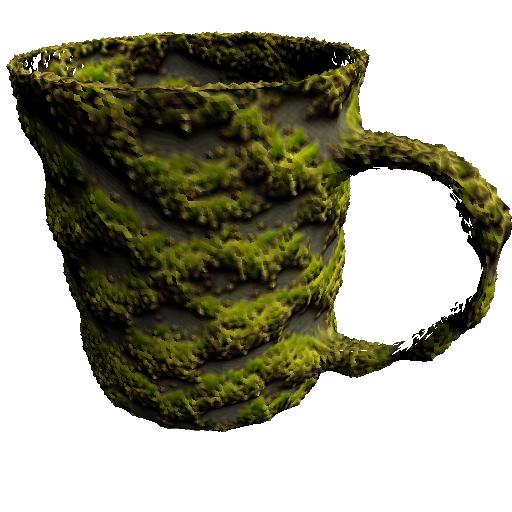} & \imgtextsupp{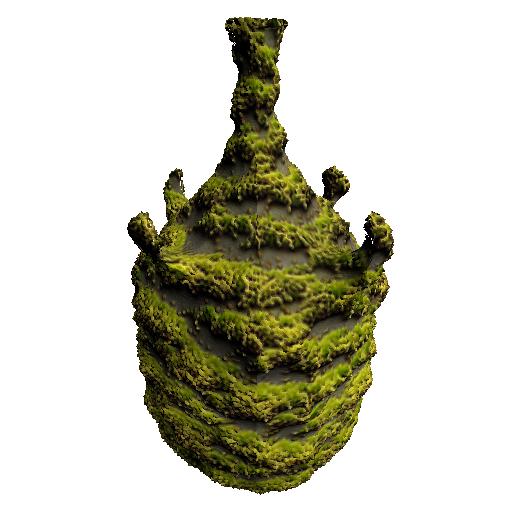} & \imgtextsupp{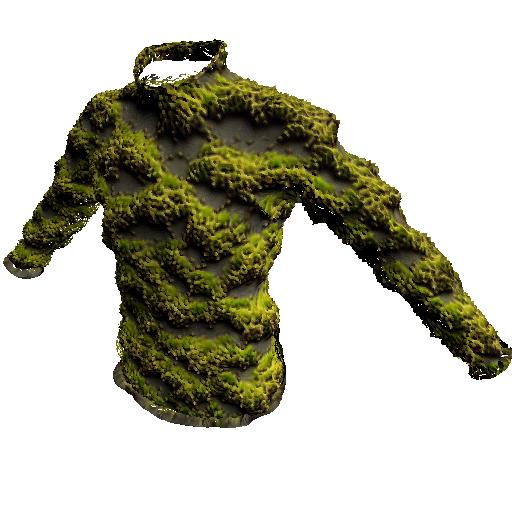} & \imgtextsupp{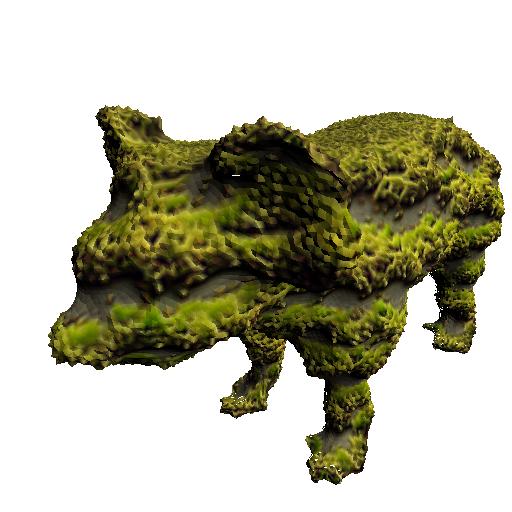} & \imgtextsupp{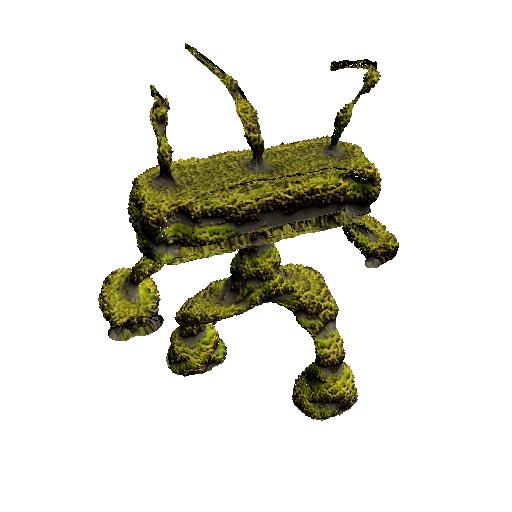}
\\
\rotatebox{90}{\parbox[c]{2cm}{\centering\hspace{0pt} \textbf{Jelly Beans}}}  &  \imgtextsupp{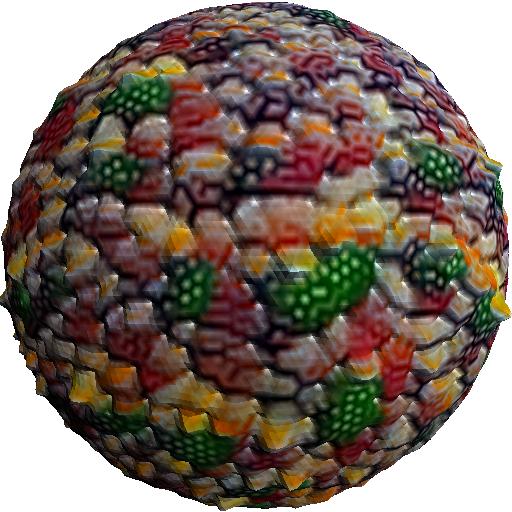}  & \imgtextsupp{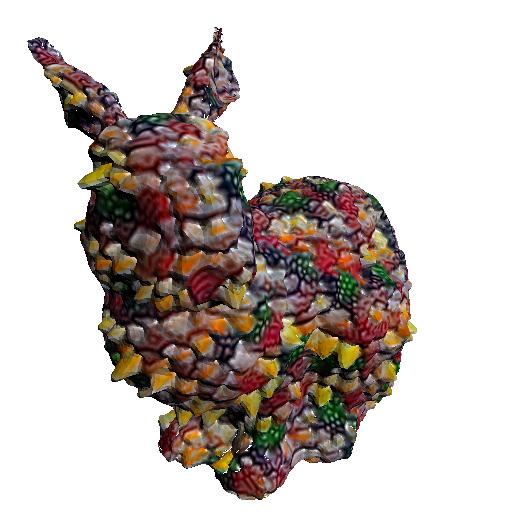} & \imgtextsupp{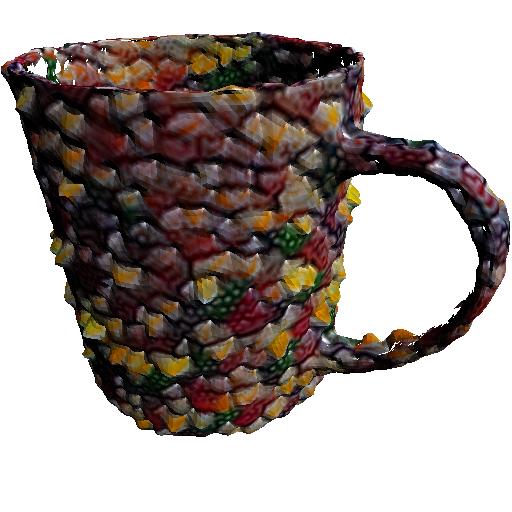} & \imgtextsupp{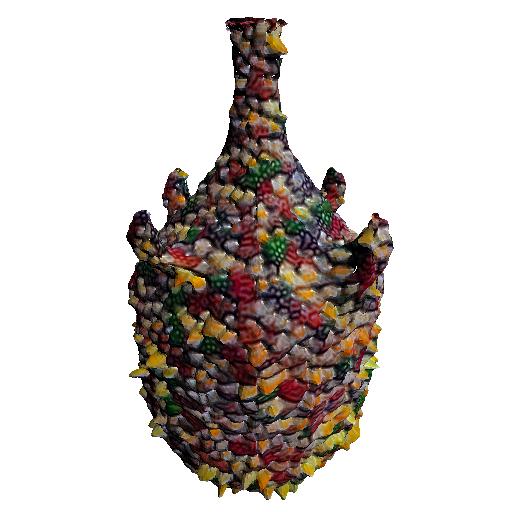} & \imgtextsupp{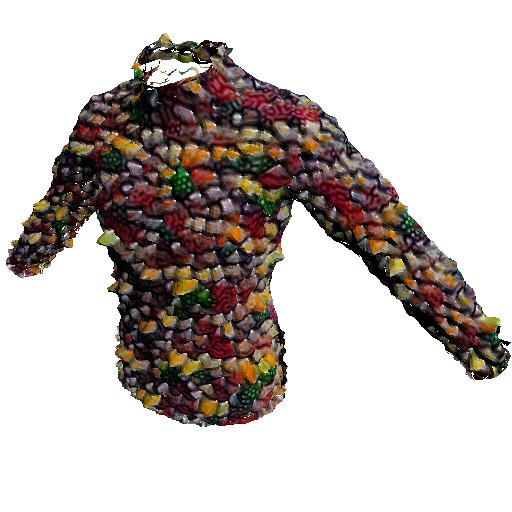} & \imgtextsupp{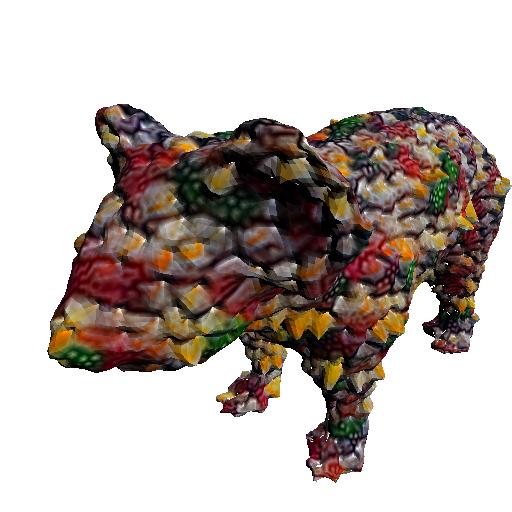} & \imgtextsupp{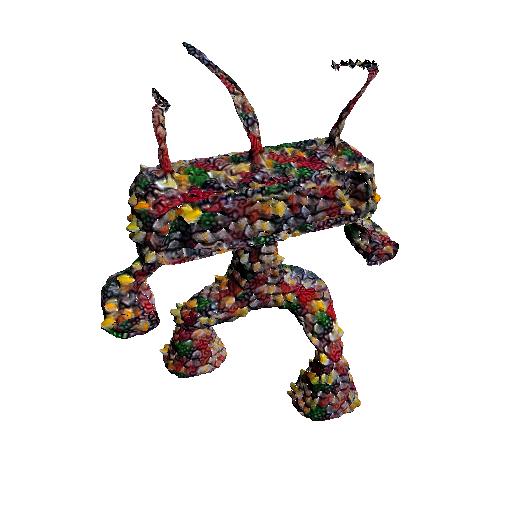}
\\
\rotatebox{90}{\parbox[c]{2cm}{\centering\hspace{0pt} \textbf{Quilted Fabric}}}  &  \imgtextsupp{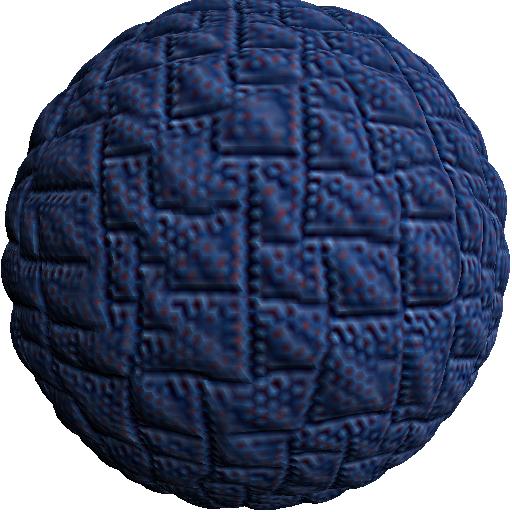}  & \imgtextsupp{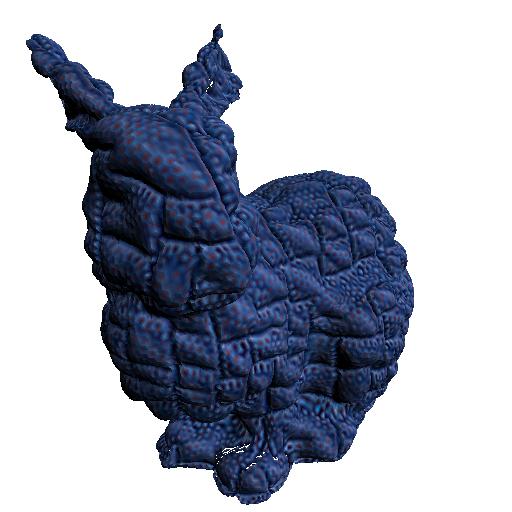} & \imgtextsupp{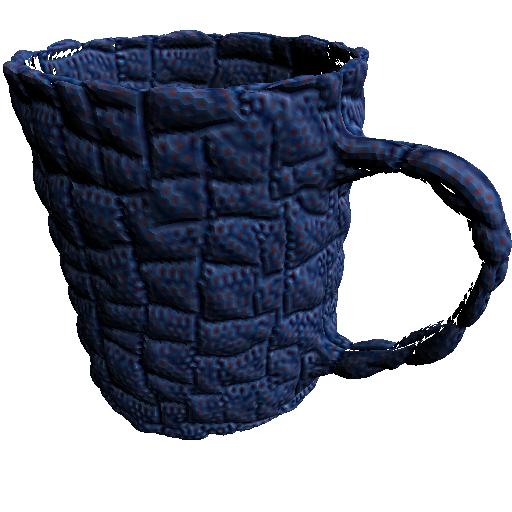} & \imgtextsupp{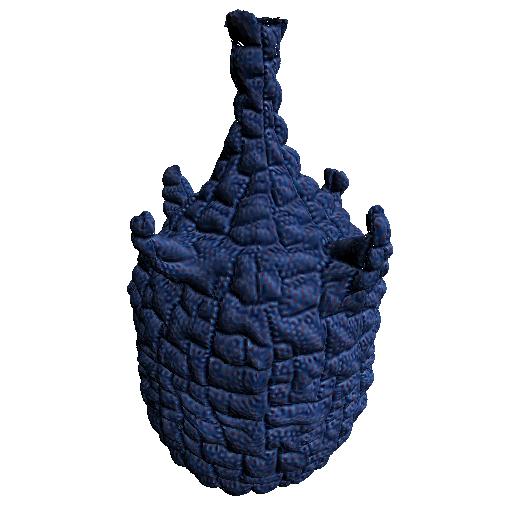} & \imgtextsupp{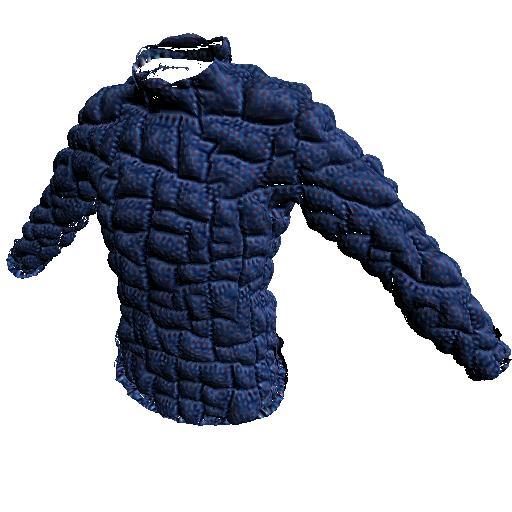} & \imgtextsupp{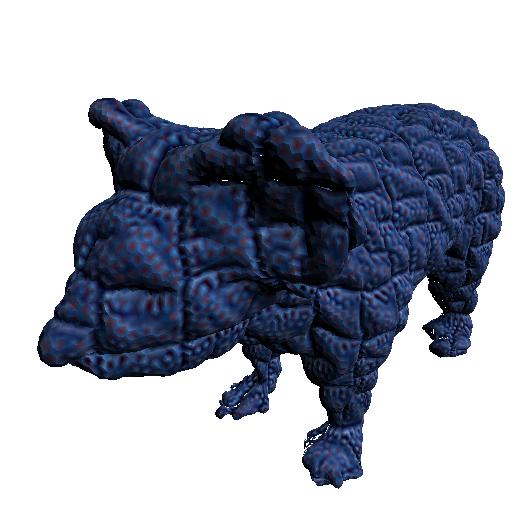} & \imgtextsupp{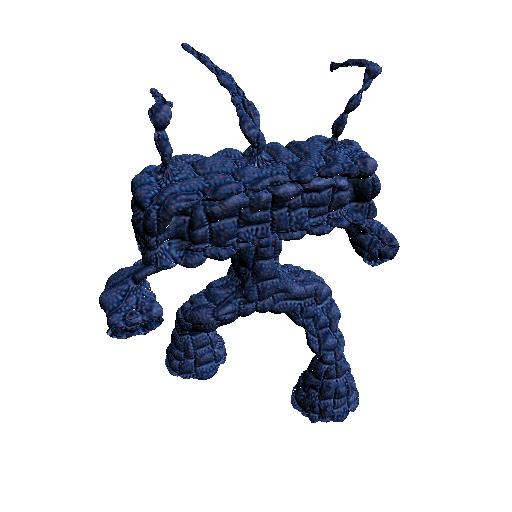}
\\
\rotatebox{90}{\parbox[c]{2cm}{\centering\hspace{0pt} \textbf{Satin}}}  &  \imgtextsupp{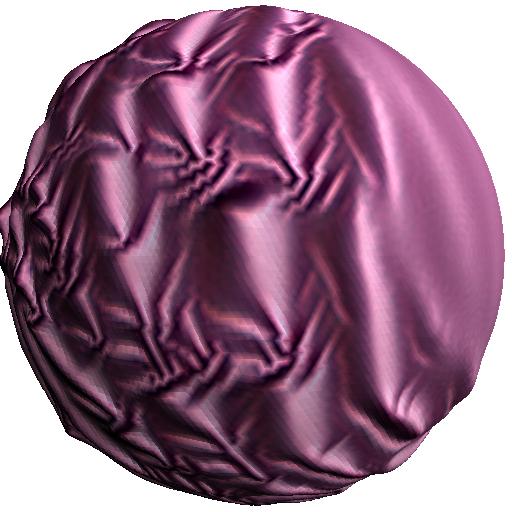}  & \imgtextsupp{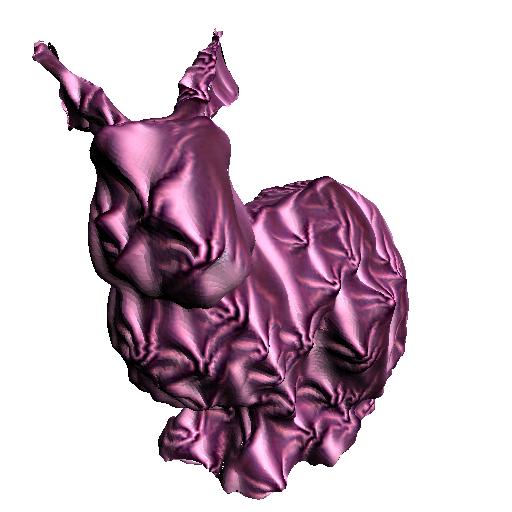} & \imgtextsupp{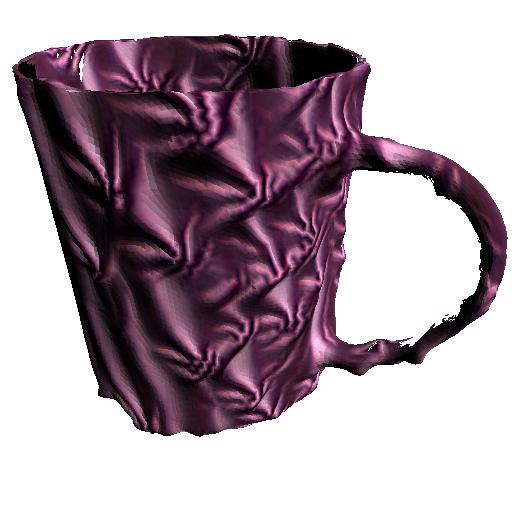} & \imgtextsupp{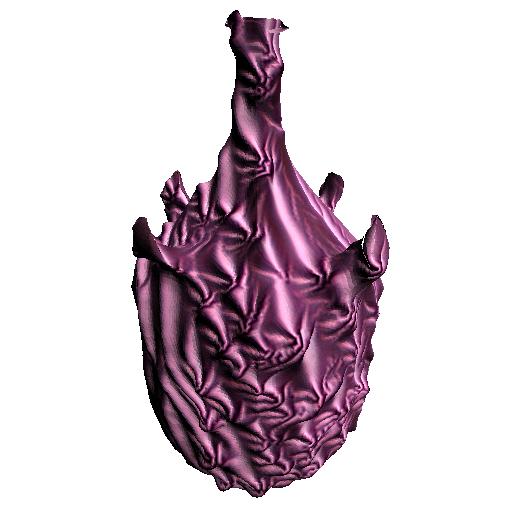} & \imgtextsupp{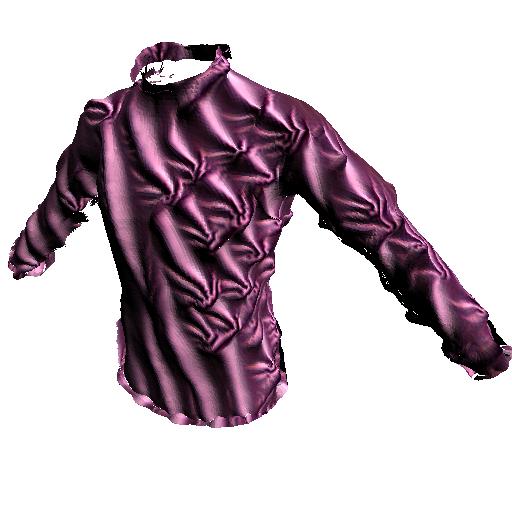} & \imgtextsupp{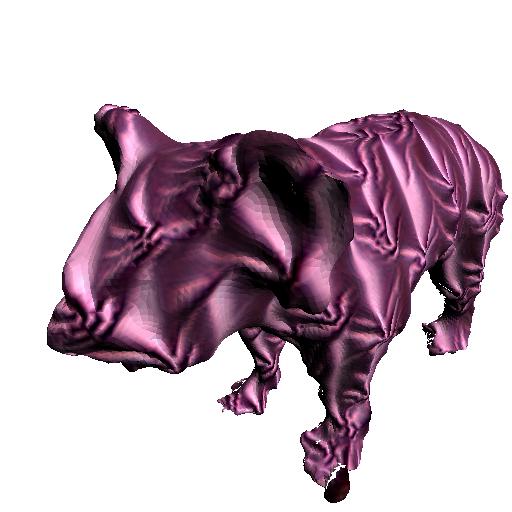} & \imgtextsupp{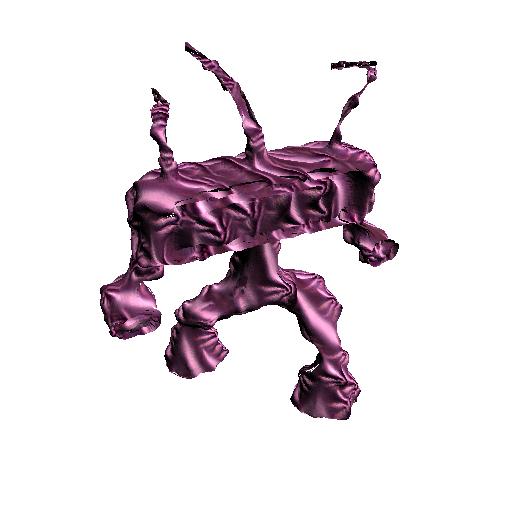}
\\
\rotatebox{90}{\parbox[c]{2cm}{\centering\hspace{0pt} \textbf{Woven Straw}}}  &  \imgtextsupp{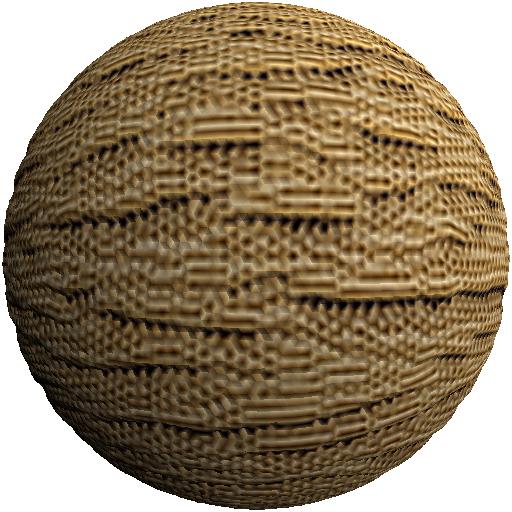}  & \imgtextsupp{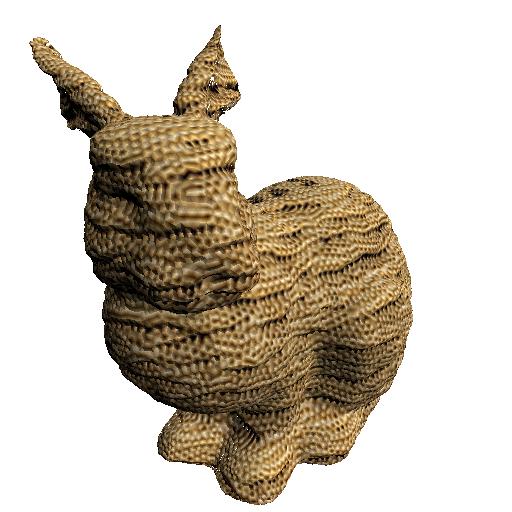} & \imgtextsupp{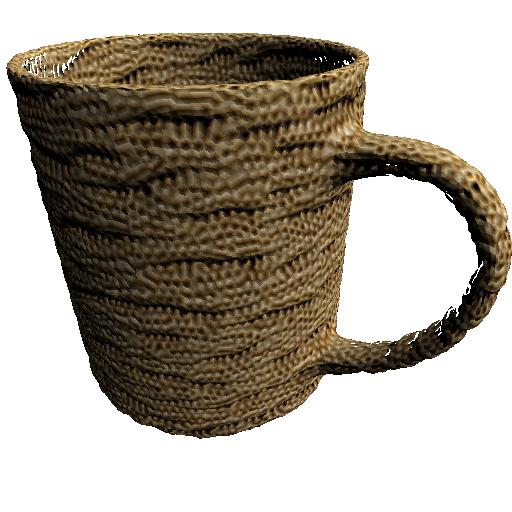} & \imgtextsupp{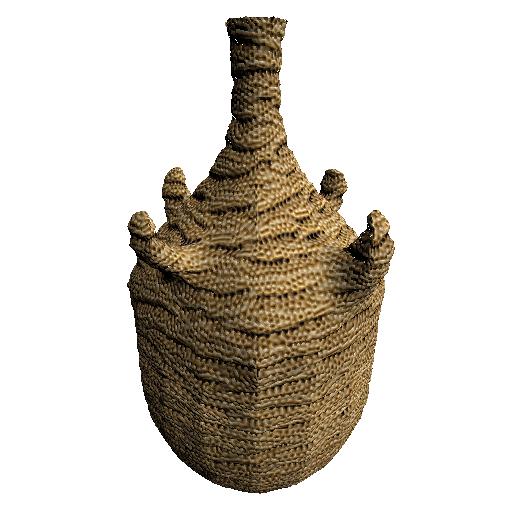} & \imgtextsupp{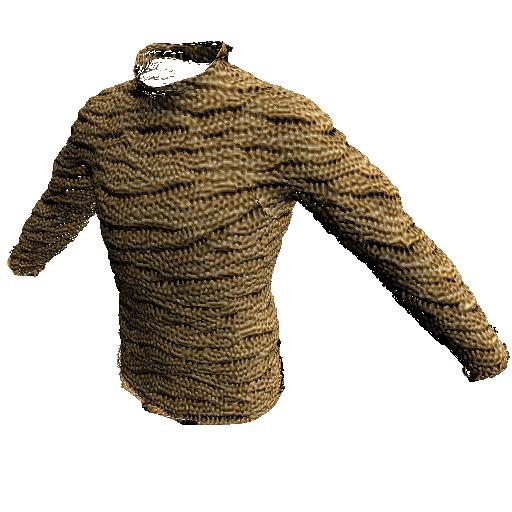} & \imgtextsupp{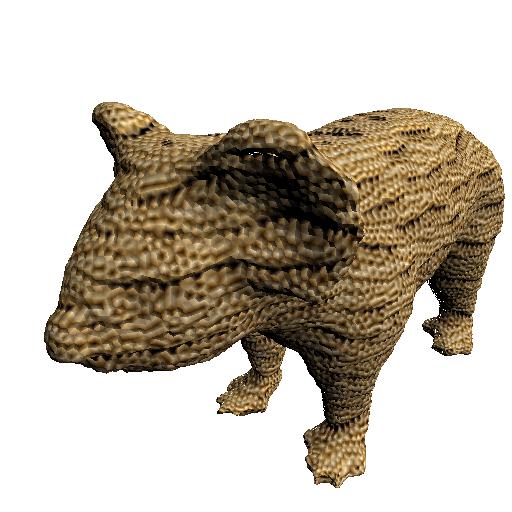} & \imgtextsupp{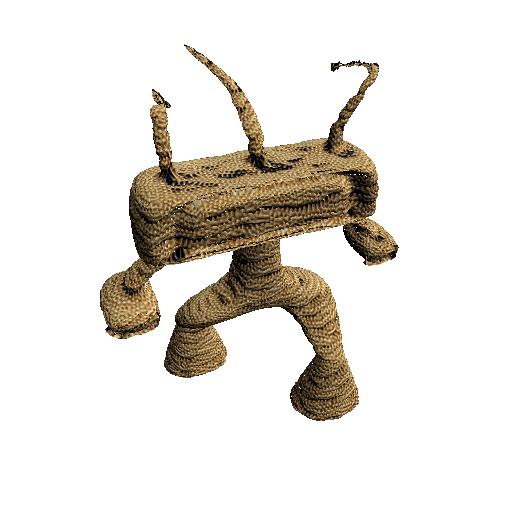}
\\
\rotatebox{90}{\parbox[c]{2cm}{\centering\hspace{0pt} \textbf{Gem}}}  &  \imgtextsupp{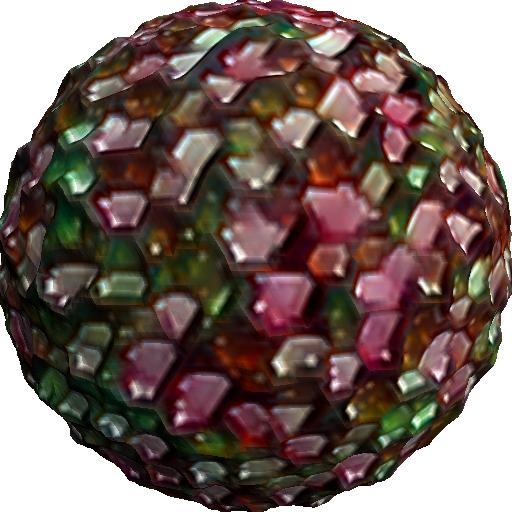}  & \imgtextsupp{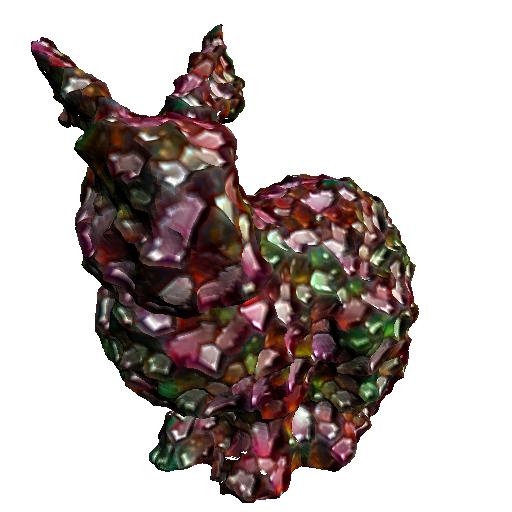} & \imgtextsupp{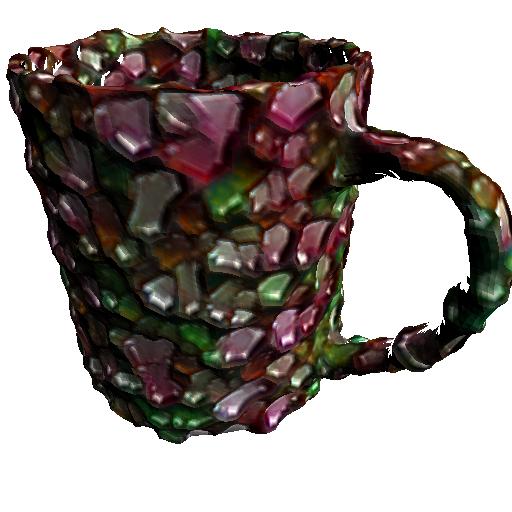} & \imgtextsupp{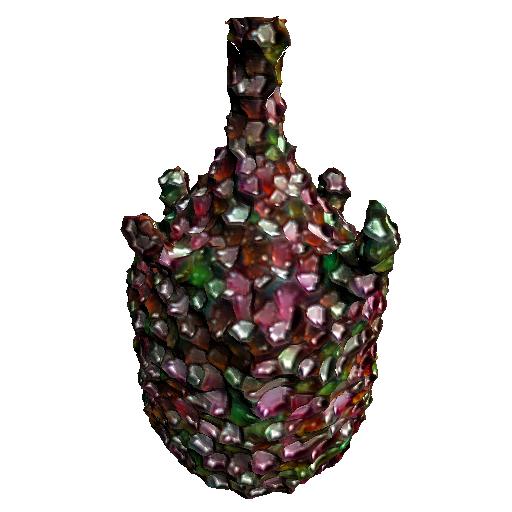} & \imgtextsupp{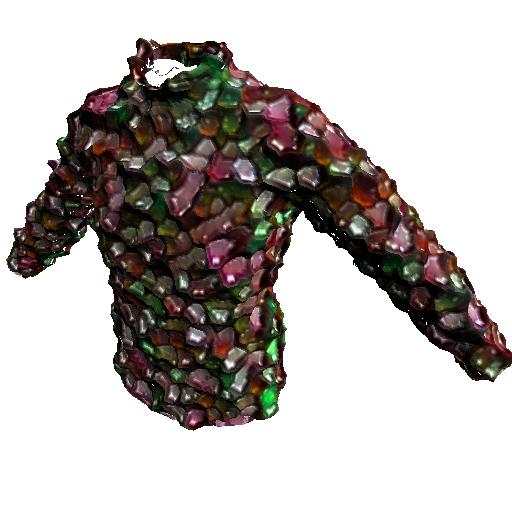} & \imgtextsupp{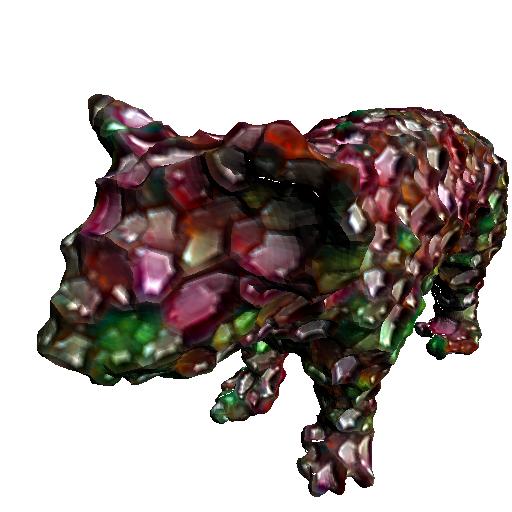} & \imgtextsupp{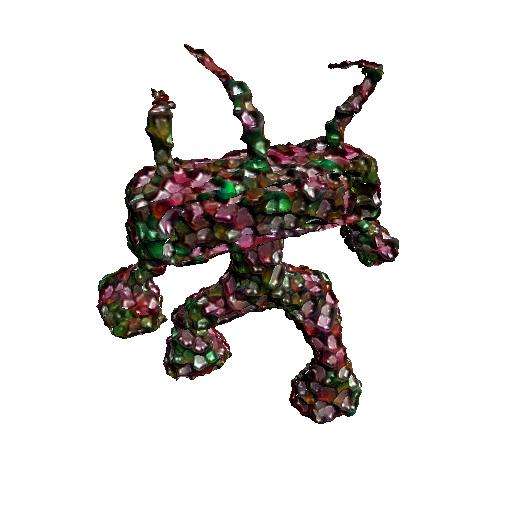}
\\
\rotatebox{90}{\parbox[c]{2cm}{\centering\hspace{0pt} \textbf{Sandstone}}}  &  \imgtextsupp{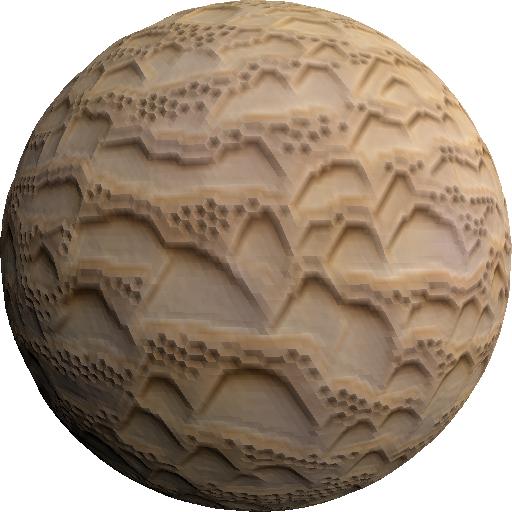}  & \imgtextsupp{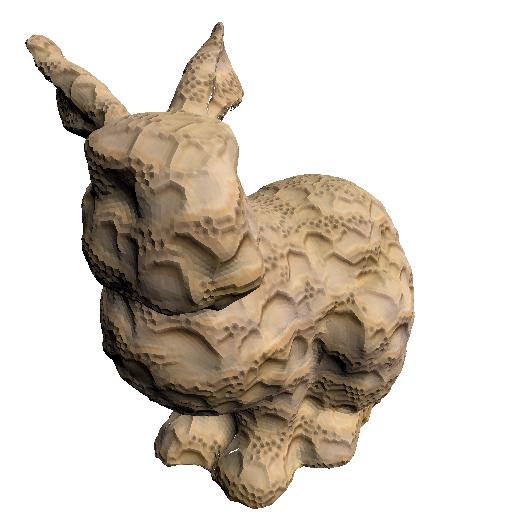} & \imgtextsupp{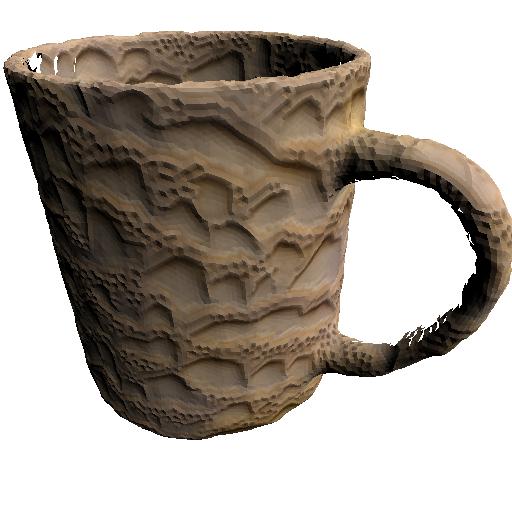} & \imgtextsupp{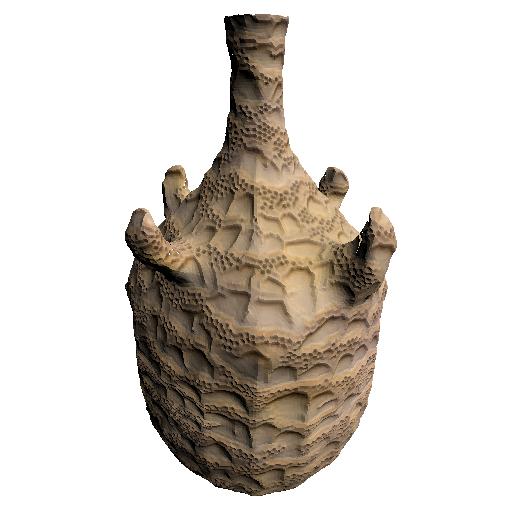} & \imgtextsupp{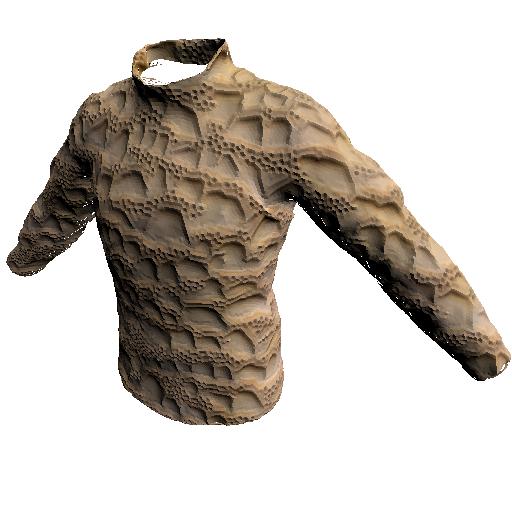} & \imgtextsupp{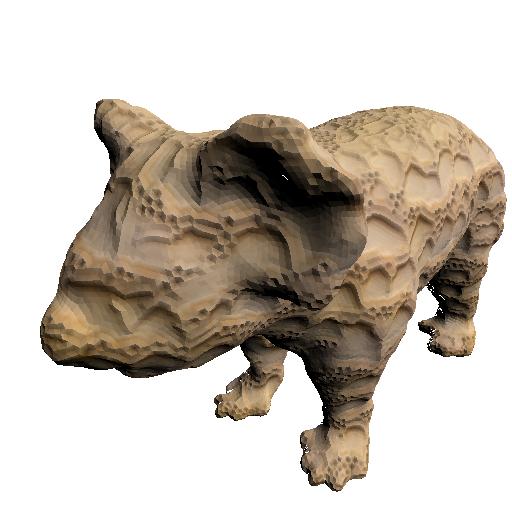} & \imgtextsupp{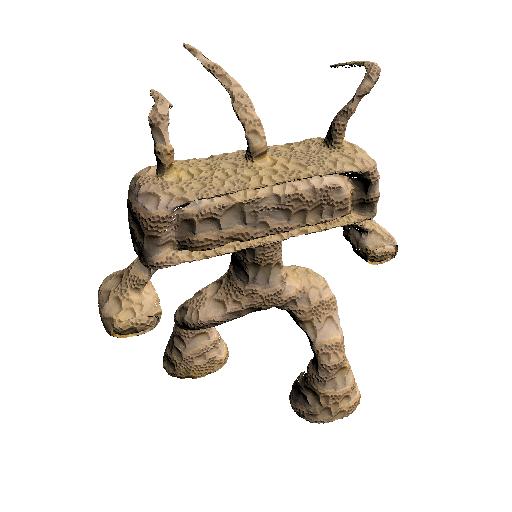}
\end{tabular}
\captionof{figure}{Results of \textbf{Text-guided} 3D texture synthesis.}
\label{fig:text-synthesis-result-supp3}
\end{table*}

\newcommand{\centeredtxt}[1]{
\begin{tabular}{l}
\parbox{2.2cm}{\vspace{-50pt} \centering #1}
\end{tabular}
}

\begin{table*}[]
\resizebox{\textwidth}{!}{
\begin{tabular}{ccc||ccc}
\textbf{Target Appearance} & \textbf{Target Motion} & \textbf{Filename} &
\textbf{Target Appearance} & \textbf{Target Motion} & \textbf{Filename} \\
 \midrule 
 
 \imgimg{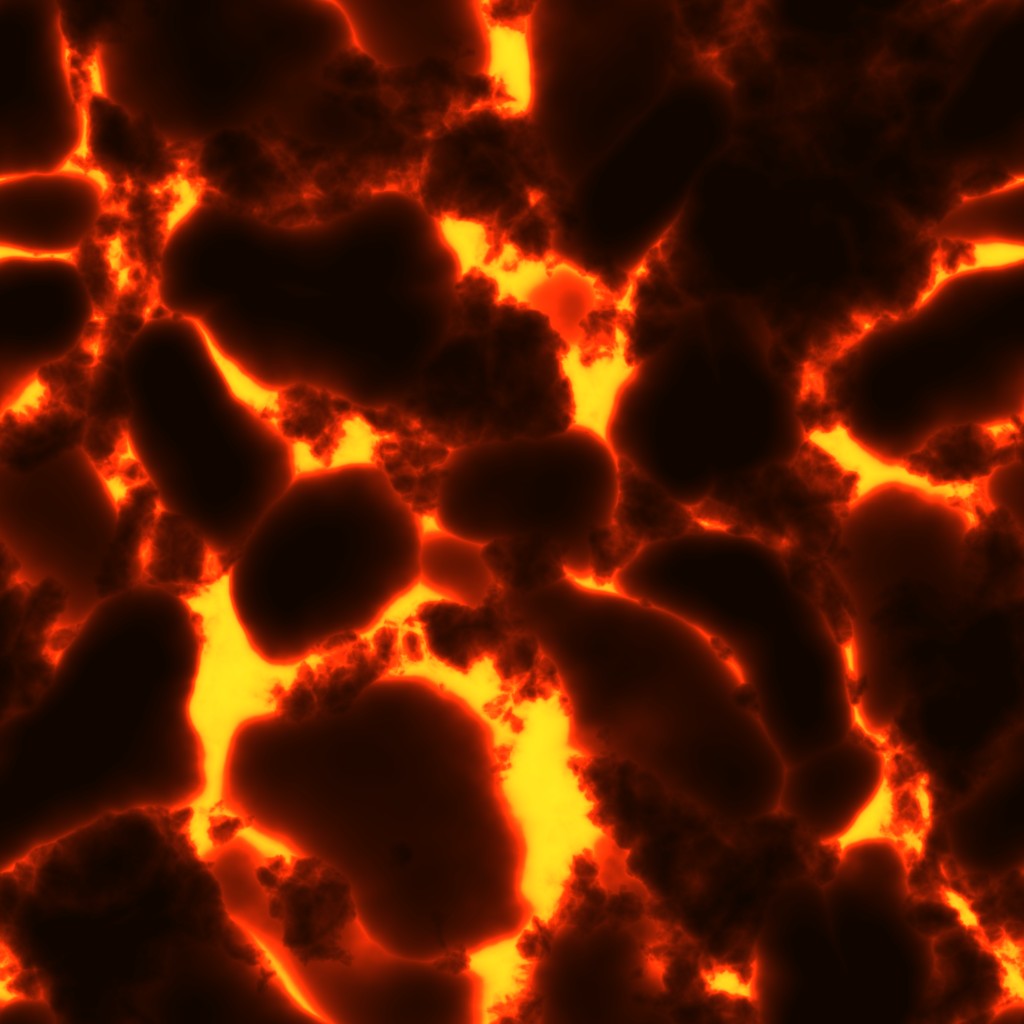}  &  \motionimg{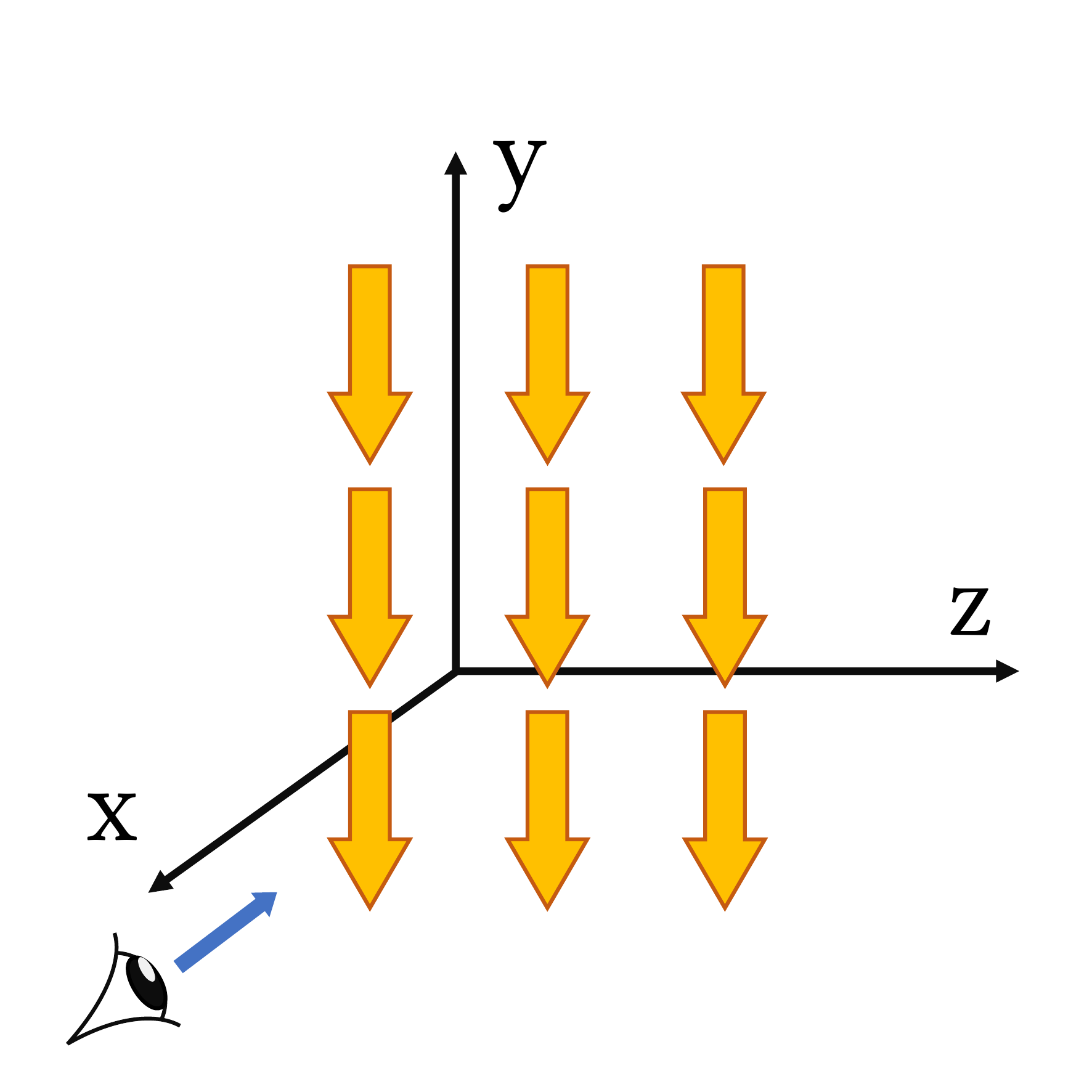} & \centeredtxt{\href{https://meshnca.github.io/supplementary/Image+Motion/Lava_005_Downward.mp4}{Lava\_005 \\ Downward}} &
 \imgimg{albedo_Sci-Fi\_Wall_012.jpg}  & \motionimg{grad_0_270_-1_-1.pdf}  &  \centeredtxt{\href{https://meshnca.github.io/supplementary/Image+Motion/Sci-Fi_Wall_012_Downward.mp4}{Sci-fi\_Wall \\ 012\_Downward}} \\

\imgimg{albedo_Lava\_005.jpg}  &  \motionimg{grad_0_0_-1_-1.pdf}  &  \centeredtxt{\href{https://meshnca.github.io/supplementary/Image+Motion/Lava_005_PositiveX.mp4}{Lava\_005 \\ PositiveX}} &
\imgimg{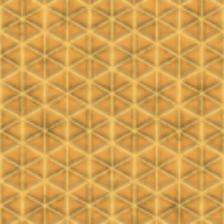}  &  \motionimg{grad_0_0_-1_-1.pdf}  &  \centeredtxt{\href{https://meshnca.github.io/supplementary/Image+Motion/Sci-Fi_Wall_012_PositiveX.mp4}{Sci-Fi\_Wall \\ 012\_PositiveX}} \\

\imgimg{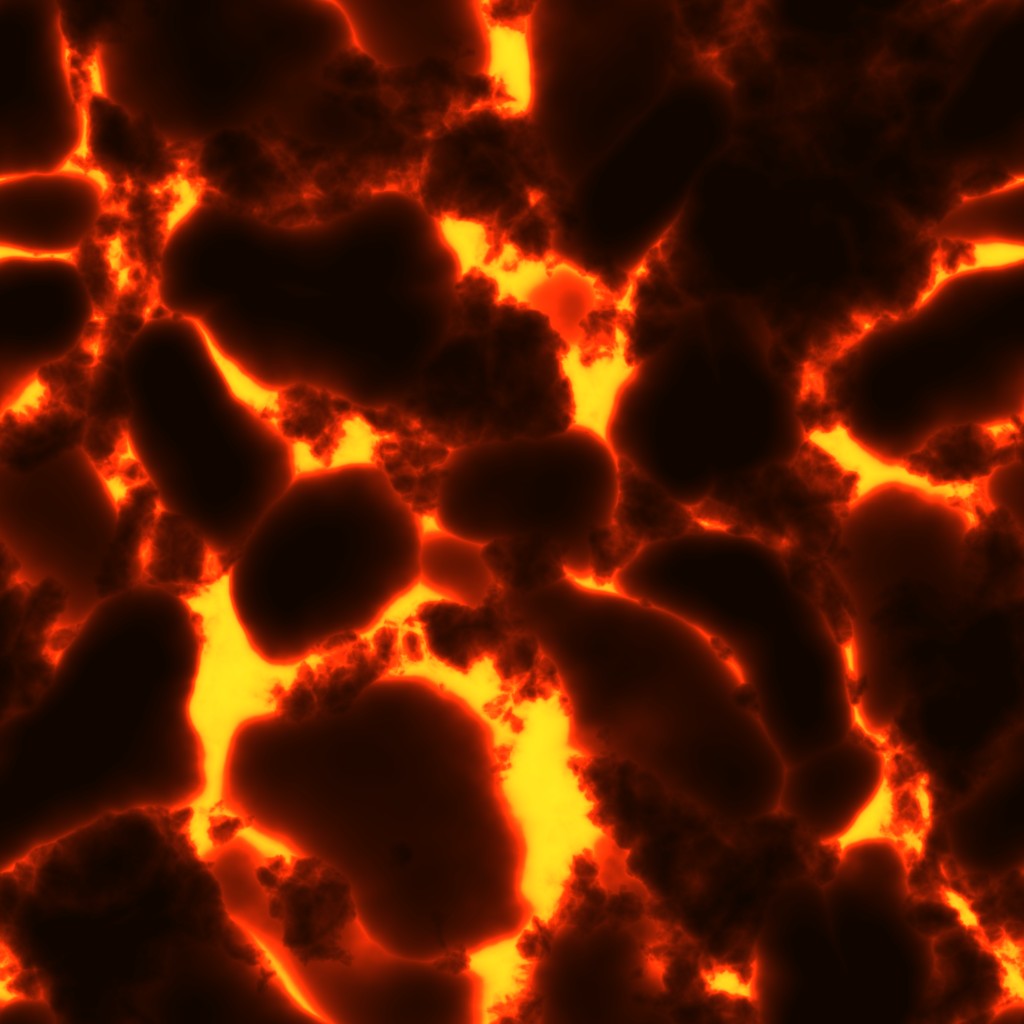}  &  \motionimg{circular_y.pdf}  &  \centeredtxt{\href{https://meshnca.github.io/supplementary/Image+Motion/Lava_005_CircularY.mp4}{Lava\_005 \\ CircularY}} &
\imgimg{albedo_Sci-Fi_Wall_012.jpg}  &  \motionimg{circular_y.pdf}  &  \centeredtxt{\href{https://meshnca.github.io/supplementary/Image+Motion/Sci-Fi_Wall_012_CircularY.mp4}{Sci-Fi\_Wall \\ 012\_CircularY}} \\

\imgimg{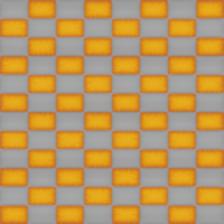}  &  \motionimg{grad_0_270_-1_-1.pdf}  &  \centeredtxt{\href{https://meshnca.github.io/supplementary/Image+Motion/Sci-fi_Wall_010_Downward.mp4}{Sci-fi\_Wall \\ 010\_Donward}} & 
\imgimg{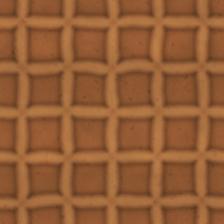}  &  \motionimg{grad_0_270_-1_-1.pdf}  &  \centeredtxt{\href{https://meshnca.github.io/supplementary/Image+Motion/Waffle_001_Downward.mp4}{Waffle\_001 \\ Downward}} \\

\imgimg{albedo_Sci-fi_Wall_010.jpg}  &  \motionimg{grad_0_0_-1_-1.pdf}  &  \centeredtxt{\href{https://meshnca.github.io/supplementary/Image+Motion/Sci-fi_Wall_010_PositiveX.mp4}{Sci-fi\_Wall \\ 010\_PositiveX}} & 
\imgimg{albedo_Waffle_001.jpg}  &  \motionimg{grad_0_0_-1_-1.pdf}  &  \centeredtxt{\href{https://meshnca.github.io/supplementary/Image+Motion/Waffle_001_PositiveX.mp4}{Waffle\_001 \\ PositiveX}} \\

\imgimg{albedo_Sci-fi\_Wall_010.jpg}  &  \motionimg{circular_y.pdf}  &  \centeredtxt{\href{https://meshnca.github.io/supplementary/Image+Motion/Sci-fi_Wall_010_CircularY.mp4}{Sci-fi\_Wall \\ 010\_CircularY}} &  
\imgimg{albedo_Waffle_001.jpg}  &  \motionimg{circular_y.pdf}  &  \centeredtxt{\href{https://meshnca.github.io/supplementary/Image+Motion/Waffle_001_CircularY.mp4}{Waffle\_001 \\ CircularY}} \\

\imgimg{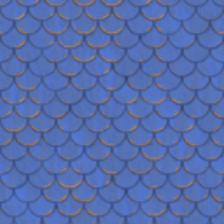}  &  \motionimg{grad_0_270_-1_-1.pdf}  &  \centeredtxt{\href{https://meshnca.github.io/supplementary/Image+Motion/Stylized_Wood_Tiles_001_Downward.mp4}{Stylized\_Wood \\ Tiles\_001 \\ Downward}} & \imgimg{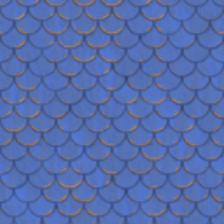}  &  \motionimg{circular_y.pdf}  &  \centeredtxt{\href{https://meshnca.github.io/supplementary/Image+Motion/Stylized_Wood_Tiles_001_CircularY.mp4}{Stylized\_Wood \\ Tiles\_001 \\ CircularY}} \\
\end{tabular}
}
\captionof{figure}{Results of dynamic texture synthesis with image guidance. All videos are available at \href{https://meshnca.github.io/supplementary/Image+Motion/}{https://meshnca.github.io/supplementary/Image+motion/}}
\label{fig:image-motion-result-supp1}
\end{table*}

\begin{table*}[]
\resizebox{\textwidth}{!}{
\begin{tabular}{ccc||ccc}
\textbf{Target Appearance} & \textbf{Target Motion} & \textbf{Filename} &
\textbf{Target Appearance} & \textbf{Target Motion} & \textbf{Filename} \\
 \midrule
\rotatebox{90}{\parbox[c]{2cm}{\centering\hspace{0pt} \textbf{Moss}}}  &  \motionimg{grad_0_270_-1_-1.pdf}  &  \centeredtxt{\href{https://meshnca.github.io/supplementary/Text+Motion/moss_Downward.mp4}{moss \\ Downward}} &
\rotatebox{90}{\parbox[c]{2cm}{\centering\hspace{0pt} \textbf{Gem}}}  &  \motionimg{grad_0_270_-1_-1.pdf}  &  \centeredtxt{\href{https://meshnca.github.io/supplementary/Text+Motion/gem_Downward.mp4}{gem \\ Downward}} \\

\rotatebox{90}{\parbox[c]{2cm}{\centering\hspace{0pt} \textbf{Moss}}}  &  \motionimg{grad_0_0_-1_-1.pdf}  &  \centeredtxt{\href{https://meshnca.github.io/supplementary/Text+Motion/moss_PositiveX.mp4}{moss \\ PositiveX}} &
\rotatebox{90}{\parbox[c]{2cm}{\centering\hspace{0pt} \textbf{Gem}}}  &  \motionimg{grad_0_0_-1_-1.pdf}  &  \centeredtxt{\href{https://meshnca.github.io/supplementary/Text+Motion/gem_PositiveX.mp4}{gem \\ PositiveX}} \\

\rotatebox{90}{\parbox[c]{2cm}{\centering\hspace{0pt} \textbf{Moss}}}  &  \motionimg{circular_y.pdf}  &  \centeredtxt{\href{https://meshnca.github.io/supplementary/Text+Motion/moss_CircularY.mp4}{moss \\ CircularY}} & 
\rotatebox{90}{\parbox[c]{2cm}{\centering\hspace{0pt} \textbf{Gem}}}  &  \motionimg{circular_y.pdf}  &  \centeredtxt{\href{https://meshnca.github.io/supplementary/Text+Motion/gem_CircularY.mp4}{gem \\ CircularY}} \\

\rotatebox{90}{\parbox[c]{2cm}{\centering\hspace{0pt} \textbf{Jelly Beans}}}  &  \motionimg{grad_0_270_-1_-1.pdf}  &  \centeredtxt{\href{https://meshnca.github.io/supplementary/Text+Motion/jelly\%20beans_Downward.mp4}{jelly beans \\ Downward}} &
\rotatebox{90}{\parbox[c]{2cm}{\centering\hspace{0pt} \textbf{Colorful Crochet}}}  &  \motionimg{grad_0_270_-1_-1.pdf}  &  \centeredtxt{\href{https://meshnca.github.io/supplementary/Text+Motion/colorful\%20crochet_Downward.mp4}{colorful crochet \\ Downward}} \\

\rotatebox{90}{\parbox[c]{2cm}{\centering\hspace{0pt} \textbf{Jelly Beans}}}  &  \motionimg{grad_0_0_-1_-1.pdf}  &  \centeredtxt{\href{https://meshnca.github.io/supplementary/Text+Motion/jelly\%20beans_PositiveX.mp4}{jelly beans \\ PositiveX}} &
\rotatebox{90}{\parbox[c]{2cm}{\centering\hspace{0pt} \textbf{Colorful Crochet}}}  &  \motionimg{grad_0_0_-1_-1.pdf}  &  \centeredtxt{\href{https://meshnca.github.io/supplementary/Text+Motion/colorful\%20crochet_PositiveX.mp4}{colorful crochet \\ PositiveX}} \\

\rotatebox{90}{\parbox[c]{2cm}{\centering\hspace{0pt} \textbf{Jelly Beans}}}  &  \motionimg{circular_y.pdf}  &  \centeredtxt{\href{https://meshnca.github.io/supplementary/Text+Motion/jelly\%20beans_CircularY.mp4}{jelly beans \\ CircularY}} &
\rotatebox{90}{\parbox[c]{2cm}{\centering\hspace{0pt} \textbf{Colorful Crochet}}}  &  \motionimg{circular_y.pdf}  &  \centeredtxt{\href{https://meshnca.github.io/supplementary/Text+Motion/colorful\%20crochet_CircularY.mp4}{colorful crochet \\ CircularY}} \\

\end{tabular}
}
\captionof{figure}{Results of dynamic texture synthesis with text guidance. All videos are available at \href{https://meshnca.github.io/supplementary/CLIP+motion/}{https://meshnca.github.io/supplementary/Text+motion/} }
\label{fig:text-motion-result-supp1}
\end{table*}

\newcommand{\imgtextgraft}[1]{%
  \ifimgprefix
    \includegraphics[height=58pt]{figures/Experiments/supp/text-graft/#1}%
  \else
    \includegraphics[height=58pt]{figures/Experiments/supp/text-graft/#1}%
  \fi
}

\begin{table*}[]
\begin{tabular}{ccc||ccc||ccc}
$\mathbf{Prompt_{left}}$ & $\mathbf{Prompt_{right}}$  & \textbf{Graft Result} & $\mathbf{Prompt_{left}}$ & $\mathbf{Prompt_{right}}$  & \textbf{Graft Result} & $\mathbf{Prompt_{left}}$ & $\mathbf{Prompt_{right}}$  & \textbf{Graft Result}  \\
\midrule
\rotatebox{90}{\parbox[c]{2cm}{\centering\hspace{0pt} \textbf{Stained Glass}}}  &  \rotatebox{90}{\parbox[c]{2cm}{\centering\hspace{0pt} \textbf{Cactus}}}  &  \imgtextgraft{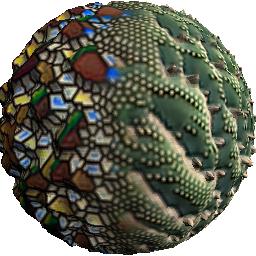} &\rotatebox{90}{\parbox[c]{2cm}{\centering\hspace{0pt} \textbf{Colorful Crochet}}}  &  \rotatebox{90}{\parbox[c]{2cm}{\centering\hspace{0pt} \textbf{Cactus}}}  &  \imgtextgraft{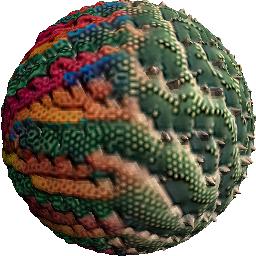} &\rotatebox{90}{\parbox[c]{2cm}{\centering\hspace{0pt} \textbf{Feathers}}}  &  \rotatebox{90}{\parbox[c]{2cm}{\centering\hspace{0pt} \textbf{Cactus}}}  &  \imgtextgraft{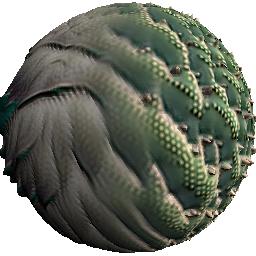} \\
\rotatebox{90}{\parbox[c]{2cm}{\centering\hspace{0pt} \textbf{Bark}}}  &  \rotatebox{90}{\parbox[c]{2cm}{\centering\hspace{0pt} \textbf{Cactus}}}  &  \imgtextgraft{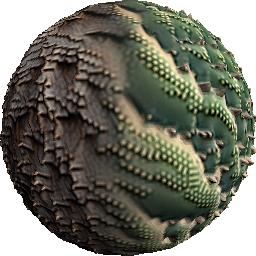} &\rotatebox{90}{\parbox[c]{2cm}{\centering\hspace{0pt} \textbf{Sandstone}}}  &  \rotatebox{90}{\parbox[c]{2cm}{\centering\hspace{0pt} \textbf{Cactus}}}  &  \imgtextgraft{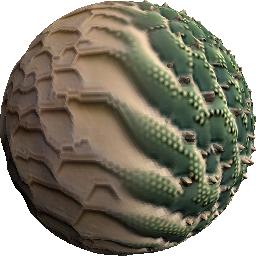} &\rotatebox{90}{\parbox[c]{2cm}{\centering\hspace{0pt} \textbf{Marble}}}  &  \rotatebox{90}{\parbox[c]{2cm}{\centering\hspace{0pt} \textbf{Cactus}}}  &  \imgtextgraft{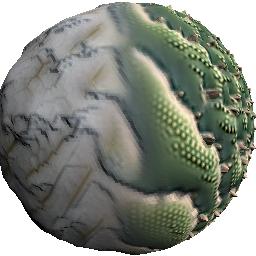} \\
\rotatebox{90}{\parbox[c]{2cm}{\centering\hspace{0pt} \textbf{Patchwork Leather}}}  &  \rotatebox{90}{\parbox[c]{2cm}{\centering\hspace{0pt} \textbf{Cactus}}}  &  \imgtextgraft{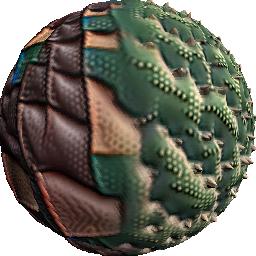} &\rotatebox{90}{\parbox[c]{2cm}{\centering\hspace{0pt} \textbf{Moss}}}  &  \rotatebox{90}{\parbox[c]{2cm}{\centering\hspace{0pt} \textbf{Cactus}}}  &  \imgtextgraft{cactus_moss.jpg} &\rotatebox{90}{\parbox[c]{2cm}{\centering\hspace{0pt} \textbf{Jelly Beans}}}  &  \rotatebox{90}{\parbox[c]{2cm}{\centering\hspace{0pt} \textbf{Cactus}}}  &  \imgtextgraft{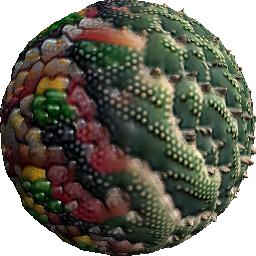} \\
\rotatebox{90}{\parbox[c]{2cm}{\centering\hspace{0pt} \textbf{Colorful Crochet}}}  &  \rotatebox{90}{\parbox[c]{2cm}{\centering\hspace{0pt} \textbf{Stained Glass}}}  &  \imgtextgraft{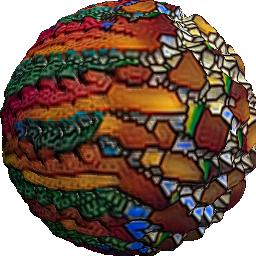} &\rotatebox{90}{\parbox[c]{2cm}{\centering\hspace{0pt} \textbf{Feathers}}}  &  \rotatebox{90}{\parbox[c]{2cm}{\centering\hspace{0pt} \textbf{Stained Glass}}}  &  \imgtextgraft{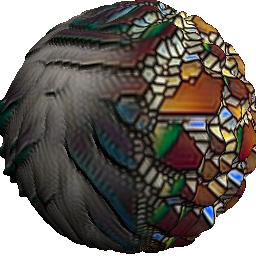} &\rotatebox{90}{\parbox[c]{2cm}{\centering\hspace{0pt} \textbf{Bark}}}  &  \rotatebox{90}{\parbox[c]{2cm}{\centering\hspace{0pt} \textbf{Stained Glass}}}  &  \imgtextgraft{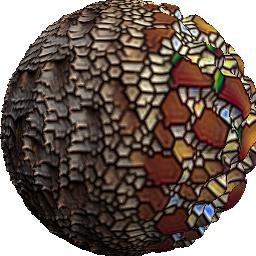} \\
\rotatebox{90}{\parbox[c]{2cm}{\centering\hspace{0pt} \textbf{Sandstone}}}  &  \rotatebox{90}{\parbox[c]{2cm}{\centering\hspace{0pt} \textbf{Stained Glass}}}  &  \imgtextgraft{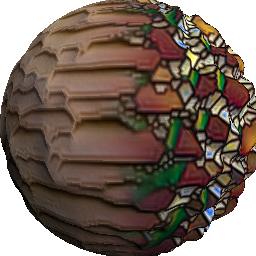} &\rotatebox{90}{\parbox[c]{2cm}{\centering\hspace{0pt} \textbf{Marble}}}  &  \rotatebox{90}{\parbox[c]{2cm}{\centering\hspace{0pt} \textbf{Stained Glass}}}  &  \imgtextgraft{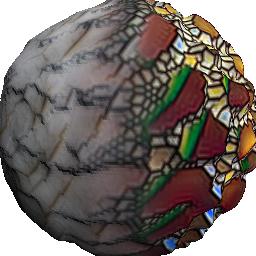} &\rotatebox{90}{\parbox[c]{2cm}{\centering\hspace{0pt} \textbf{Patchwork Leather}}}  &  \rotatebox{90}{\parbox[c]{2cm}{\centering\hspace{0pt} \textbf{Stained Glass}}}  &  \imgtextgraft{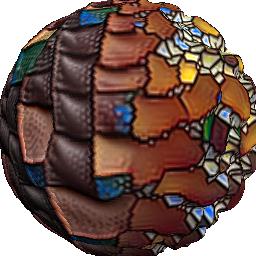} \\
\rotatebox{90}{\parbox[c]{2cm}{\centering\hspace{0pt} \textbf{Moss}}}  &  \rotatebox{90}{\parbox[c]{2cm}{\centering\hspace{0pt} \textbf{Stained Glass}}}  &  \imgtextgraft{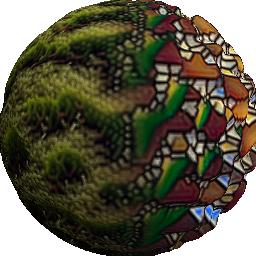} &\rotatebox{90}{\parbox[c]{2cm}{\centering\hspace{0pt} \textbf{Jelly Beans}}}  &  \rotatebox{90}{\parbox[c]{2cm}{\centering\hspace{0pt} \textbf{Stained Glass}}}  &  \imgtextgraft{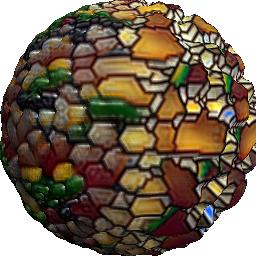} &\rotatebox{90}{\parbox[c]{2cm}{\centering\hspace{0pt} \textbf{Feathers}}}  &  \rotatebox{90}{\parbox[c]{2cm}{\centering\hspace{0pt} \textbf{Colorful Crochet}}}  &  \imgtextgraft{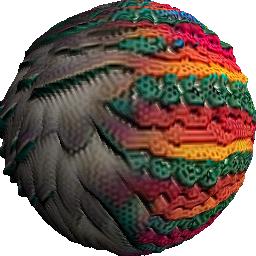} \\
\rotatebox{90}{\parbox[c]{2cm}{\centering\hspace{0pt} \textbf{Bark}}}  &  \rotatebox{90}{\parbox[c]{2cm}{\centering\hspace{0pt} \textbf{Colorful Crochet}}}  &  \imgtextgraft{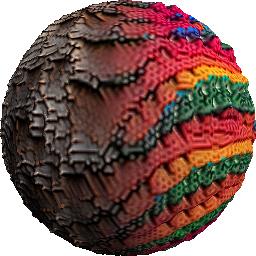} &\rotatebox{90}{\parbox[c]{2cm}{\centering\hspace{0pt} \textbf{Sandstone}}}  &  \rotatebox{90}{\parbox[c]{2cm}{\centering\hspace{0pt} \textbf{Colorful Crochet}}}  &  \imgtextgraft{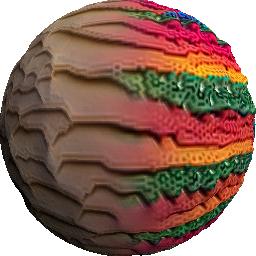} &\rotatebox{90}{\parbox[c]{2cm}{\centering\hspace{0pt} \textbf{Marble}}}  &  \rotatebox{90}{\parbox[c]{2cm}{\centering\hspace{0pt} \textbf{Colorful Crochet}}}  &  \imgtextgraft{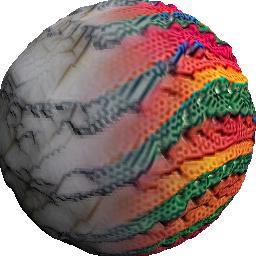} \\
 \midrule
\end{tabular}
\captionof{figure}{Results of grafting in \textbf{Text-guided} 3D texture synthesis.}
\label{fig:text-graft-result-supp1}
\end{table*}

\begin{table*}[]
\begin{tabular}{ccc||ccc||ccc}
$\mathbf{Prompt_{left}}$ & $\mathbf{Prompt_{right}}$  & \textbf{Graft Result} & $\mathbf{Prompt_{left}}$ & $\mathbf{Prompt_{right}}$  & \textbf{Graft Result} & $\mathbf{Prompt_{left}}$ & $\mathbf{Prompt_{right}}$  & \textbf{Graft Result}  \\
\midrule
\rotatebox{90}{\parbox[c]{2cm}{\centering\hspace{0pt} \textbf{Patchwork Leather}}}  &  \rotatebox{90}{\parbox[c]{2cm}{\centering\hspace{0pt} \textbf{Colorful Crochet}}}  &  \imgtextgraft{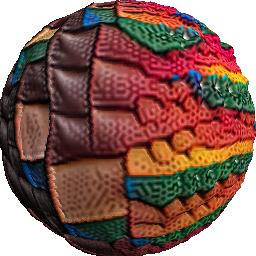} &\rotatebox{90}{\parbox[c]{2cm}{\centering\hspace{0pt} \textbf{Moss}}}  &  \rotatebox{90}{\parbox[c]{2cm}{\centering\hspace{0pt} \textbf{Colorful Crochet}}}  &  \imgtextgraft{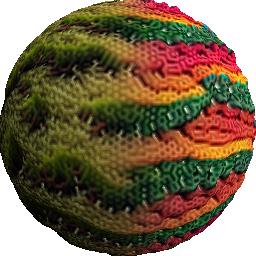} &\rotatebox{90}{\parbox[c]{2cm}{\centering\hspace{0pt} \textbf{Jelly Beans}}}  &  \rotatebox{90}{\parbox[c]{2cm}{\centering\hspace{0pt} \textbf{Colorful Crochet}}}  &  \imgtextgraft{colorful_crochet_jelly_beans.jpg} \\
\rotatebox{90}{\parbox[c]{2cm}{\centering\hspace{0pt} \textbf{Bark}}}  &  \rotatebox{90}{\parbox[c]{2cm}{\centering\hspace{0pt} \textbf{Feathers}}}  &  \imgtextgraft{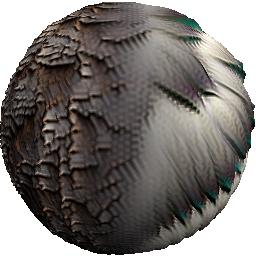} &\rotatebox{90}{\parbox[c]{2cm}{\centering\hspace{0pt} \textbf{Sandstone}}}  &  \rotatebox{90}{\parbox[c]{2cm}{\centering\hspace{0pt} \textbf{Feathers}}}  &  \imgtextgraft{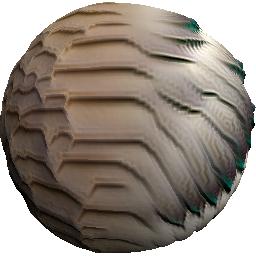} &\rotatebox{90}{\parbox[c]{2cm}{\centering\hspace{0pt} \textbf{Marble}}}  &  \rotatebox{90}{\parbox[c]{2cm}{\centering\hspace{0pt} \textbf{Feathers}}}  &  \imgtextgraft{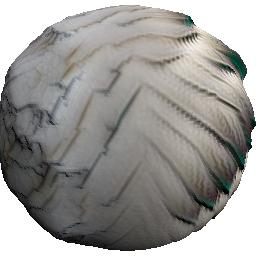} \\
\rotatebox{90}{\parbox[c]{2cm}{\centering\hspace{0pt} \textbf{Patchwork Leather}}}  &  \rotatebox{90}{\parbox[c]{2cm}{\centering\hspace{0pt} \textbf{Feathers}}}  &  \imgtextgraft{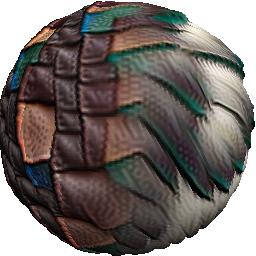} &\rotatebox{90}{\parbox[c]{2cm}{\centering\hspace{0pt} \textbf{Moss}}}  &  \rotatebox{90}{\parbox[c]{2cm}{\centering\hspace{0pt} \textbf{Feathers}}}  &  \imgtextgraft{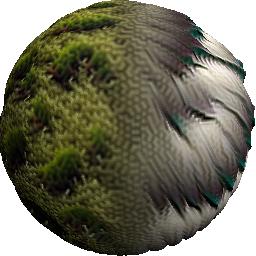} &\rotatebox{90}{\parbox[c]{2cm}{\centering\hspace{0pt} \textbf{Jelly Beans}}}  &  \rotatebox{90}{\parbox[c]{2cm}{\centering\hspace{0pt} \textbf{Feathers}}}  &  \imgtextgraft{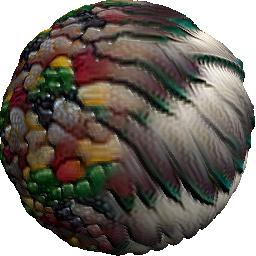} \\
\rotatebox{90}{\parbox[c]{2cm}{\centering\hspace{0pt} \textbf{Sandstone}}}  &  \rotatebox{90}{\parbox[c]{2cm}{\centering\hspace{0pt} \textbf{Bark}}}  &  \imgtextgraft{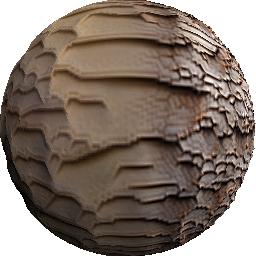} &\rotatebox{90}{\parbox[c]{2cm}{\centering\hspace{0pt} \textbf{Marble}}}  &  \rotatebox{90}{\parbox[c]{2cm}{\centering\hspace{0pt} \textbf{Bark}}}  &  \imgtextgraft{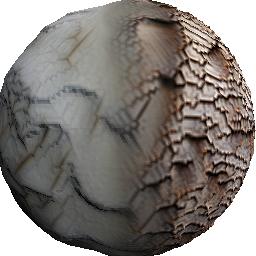} &\rotatebox{90}{\parbox[c]{2cm}{\centering\hspace{0pt} \textbf{Patchwork Leather}}}  &  \rotatebox{90}{\parbox[c]{2cm}{\centering\hspace{0pt} \textbf{Bark}}}  &  \imgtextgraft{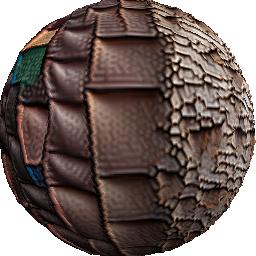} \\
\rotatebox{90}{\parbox[c]{2cm}{\centering\hspace{0pt} \textbf{Moss}}}  &  \rotatebox{90}{\parbox[c]{2cm}{\centering\hspace{0pt} \textbf{Bark}}}  &  \imgtextgraft{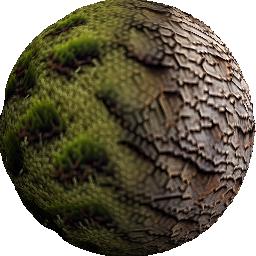} &\rotatebox{90}{\parbox[c]{2cm}{\centering\hspace{0pt} \textbf{Jelly Beans}}}  &  \rotatebox{90}{\parbox[c]{2cm}{\centering\hspace{0pt} \textbf{Bark}}}  &  \imgtextgraft{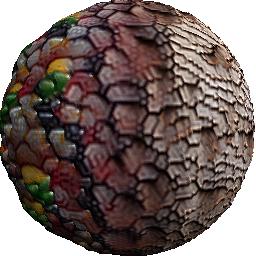} &\rotatebox{90}{\parbox[c]{2cm}{\centering\hspace{0pt} \textbf{Marble}}}  &  \rotatebox{90}{\parbox[c]{2cm}{\centering\hspace{0pt} \textbf{Sandstone}}}  &  \imgtextgraft{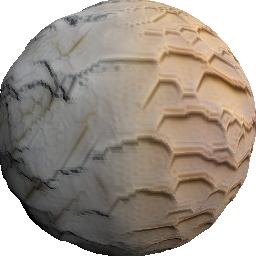} \\
\rotatebox{90}{\parbox[c]{2cm}{\centering\hspace{0pt} \textbf{Patchwork Leather}}}  &  \rotatebox{90}{\parbox[c]{2cm}{\centering\hspace{0pt} \textbf{Sandstone}}}  &  \imgtextgraft{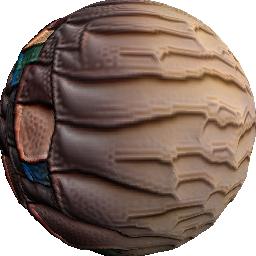} &\rotatebox{90}{\parbox[c]{2cm}{\centering\hspace{0pt} \textbf{Moss}}}  &  \rotatebox{90}{\parbox[c]{2cm}{\centering\hspace{0pt} \textbf{Sandstone}}}  &  \imgtextgraft{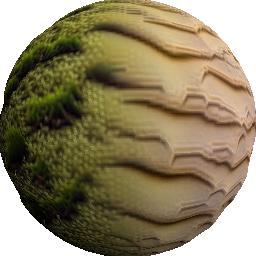} &\rotatebox{90}{\parbox[c]{2cm}{\centering\hspace{0pt} \textbf{Jelly Beans}}}  &  \rotatebox{90}{\parbox[c]{2cm}{\centering\hspace{0pt} \textbf{Sandstone}}}  &  \imgtextgraft{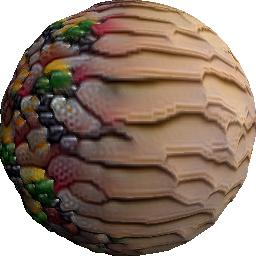} \\
\rotatebox{90}{\parbox[c]{2cm}{\centering\hspace{0pt} \textbf{Patchwork Leather}}}  &  \rotatebox{90}{\parbox[c]{2cm}{\centering\hspace{0pt} \textbf{Marble}}}  &  \imgtextgraft{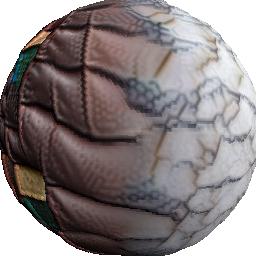} &\rotatebox{90}{\parbox[c]{2cm}{\centering\hspace{0pt} \textbf{Moss}}}  &  \rotatebox{90}{\parbox[c]{2cm}{\centering\hspace{0pt} \textbf{Marble}}}  &  \imgtextgraft{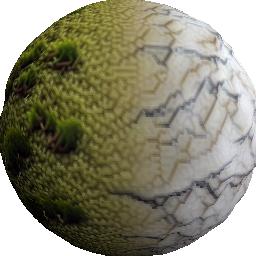} &\rotatebox{90}{\parbox[c]{2cm}{\centering\hspace{0pt} \textbf{Jelly Beans}}}  &  \rotatebox{90}{\parbox[c]{2cm}{\centering\hspace{0pt} \textbf{Marble}}}  &  \imgtextgraft{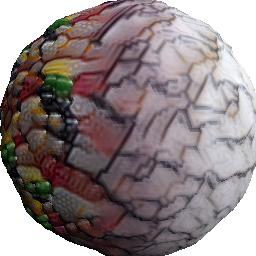} \\
\rotatebox{90}{\parbox[c]{2cm}{\centering\hspace{0pt} \textbf{Moss}}}  &  \rotatebox{90}{\parbox[c]{2cm}{\centering\hspace{0pt} \textbf{Patchwork Leather}}}  &  \imgtextgraft{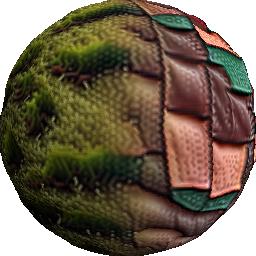} &\rotatebox{90}{\parbox[c]{2cm}{\centering\hspace{0pt} \textbf{Jelly Beans}}}  &  \rotatebox{90}{\parbox[c]{2cm}{\centering\hspace{0pt} \textbf{Patchwork Leather}}}  &  \imgtextgraft{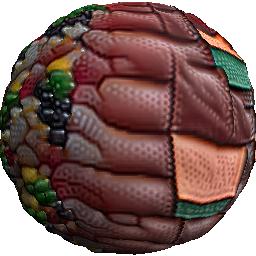} &\rotatebox{90}{\parbox[c]{2cm}{\centering\hspace{0pt} \textbf{Jelly Beans}}}  &  \rotatebox{90}{\parbox[c]{2cm}{\centering\hspace{0pt} \textbf{Moss}}}  &  \imgtextgraft{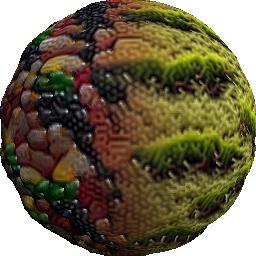} \\
 \midrule
\end{tabular}
\captionof{figure}{Results of grafting in \textbf{Text-guided} 3D texture synthesis.}
\label{fig:text-graft-result-supp2}
\end{table*}

\newcommand{\imgtexttrainmesh}[1]{%
  \ifimgprefix
    \includegraphics[height=55pt]{figures/Experiments/supp/text-train-direct-mesh/#1}%
  \else
    \includegraphics[height=55pt]{figures/Experiments/supp/text-train-direct-mesh/#1}%
  \fi
}

\begin{table*}[]
\begin{tabular}{cccc||cccc}
\textbf{Prompt} & Train & Generalization & Direct Train & \textbf{Prompt} & Train & Generalization & Direct Train  \\
\midrule
\rotatebox{90}{\parbox[c]{2cm}{\centering\hspace{0pt} \textbf{Stained Glass}}}  &  \imgtextsupp{color_img_train_stained_glass.jpg}  &  \imgtextsupp{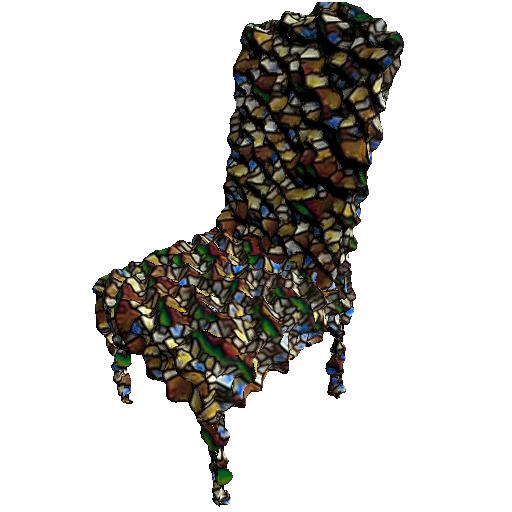}  &  \imgtexttrainmesh{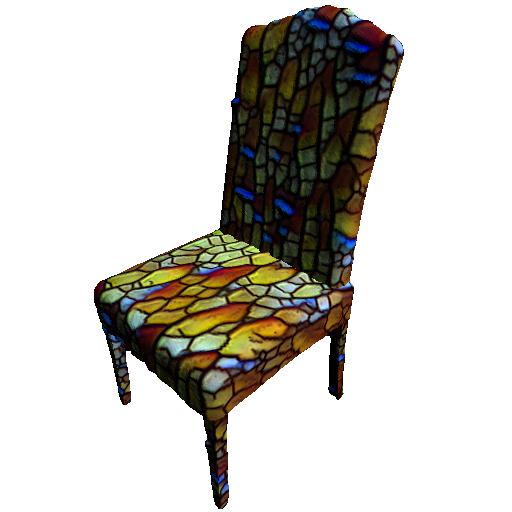} &\rotatebox{90}{\parbox[c]{2cm}{\centering\hspace{0pt} \textbf{Bark}}}  &  \imgtextsupp{color_img_train_bark.jpg}  &  \imgtextsupp{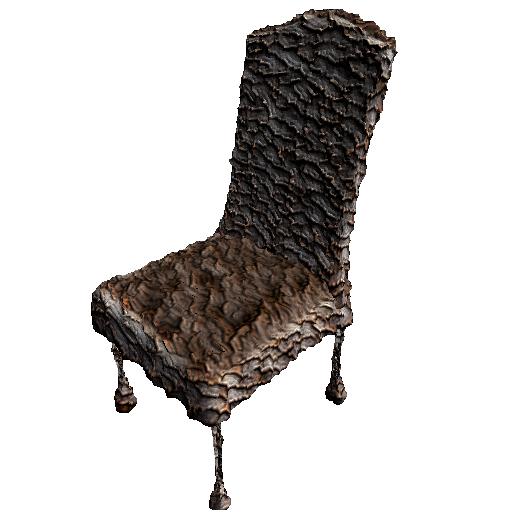}  &  \imgtexttrainmesh{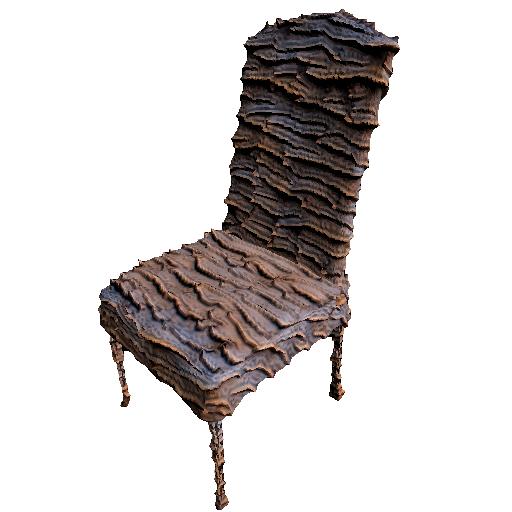} \\
\rotatebox{90}{\parbox[c]{2cm}{\centering\hspace{0pt} \textbf{Marble}}}  &  \imgtextsupp{color_img_train_marble.jpg}  &  \imgtextsupp{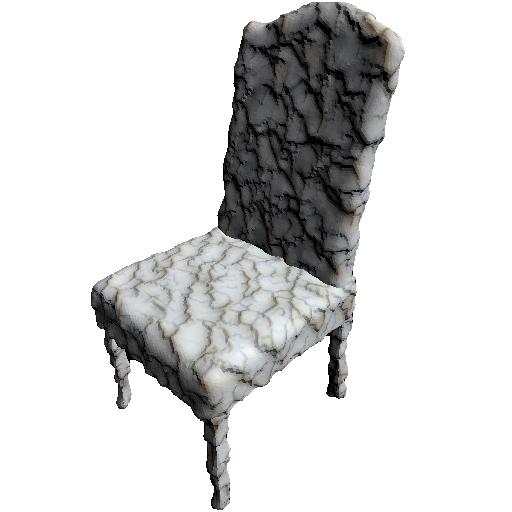}  &  \imgtexttrainmesh{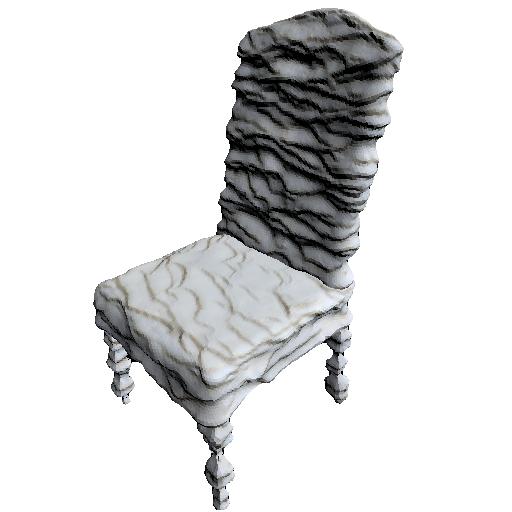} &\rotatebox{90}{\parbox[c]{2cm}{\centering\hspace{0pt} \textbf{Cactus}}}  &  \imgtextsupp{color_img_train_cactus.jpg}  &  \imgtextsupp{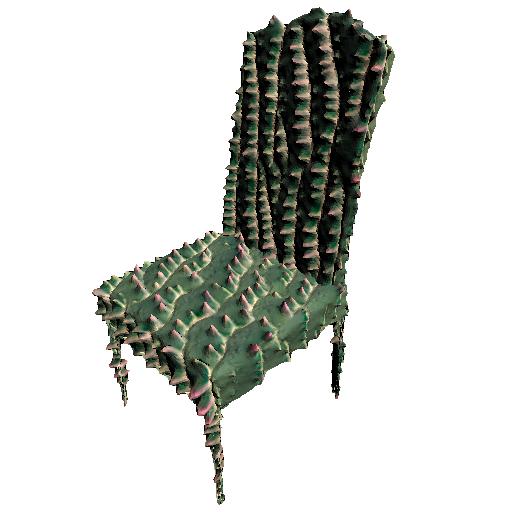}  &  \imgtexttrainmesh{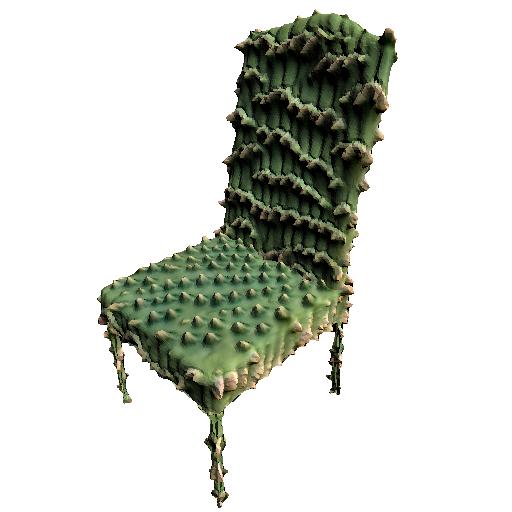} \\
\rotatebox{90}{\parbox[c]{2cm}{\centering\hspace{0pt} \textbf{Feathers}}}  &  \imgtextsupp{color_img_train_feathers.jpg}  &  \imgtextsupp{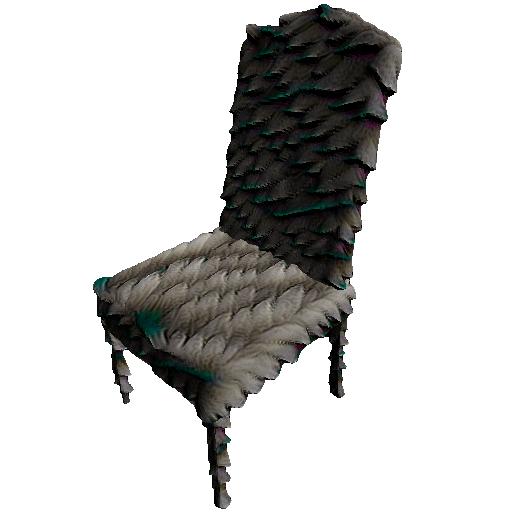}  &  \imgtexttrainmesh{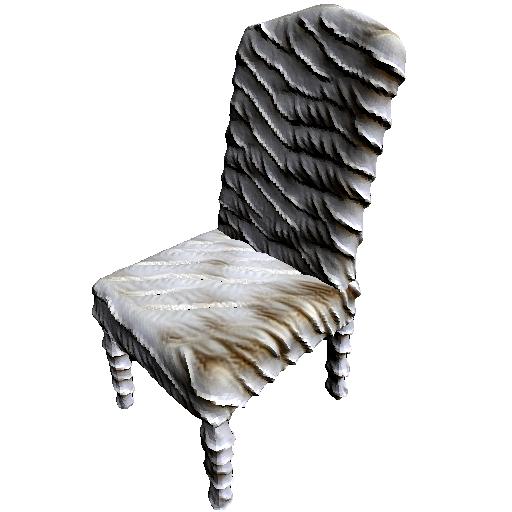} &\rotatebox{90}{\parbox[c]{2cm}{\centering\hspace{0pt} \textbf{Sandstone}}}  &  \imgtextsupp{color_img_train_sandstone.jpg}  &  \imgtextsupp{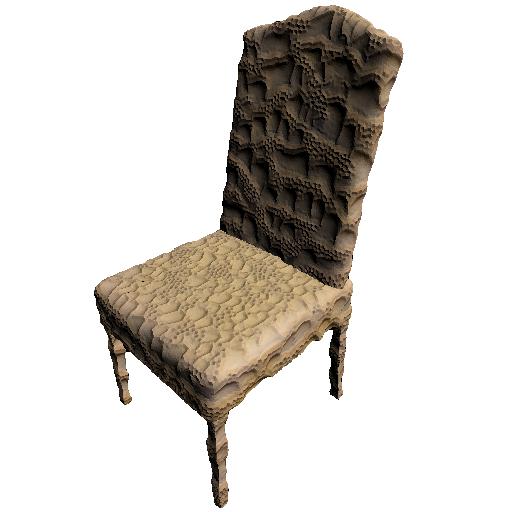}  &  \imgtexttrainmesh{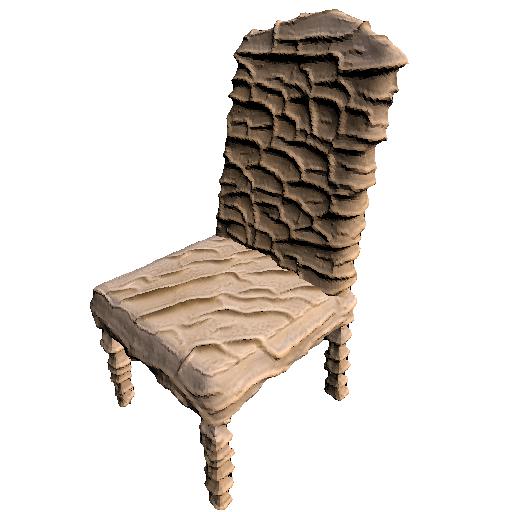} \\
\rotatebox{90}{\parbox[c]{2cm}{\centering\hspace{0pt} \textbf{Patchwork Leather}}}  &  \imgtextsupp{color_img_train_patchwork_leather.jpg}  &  \imgtextsupp{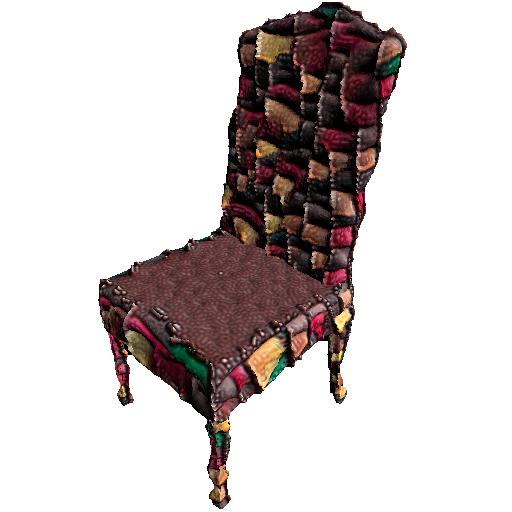}  &  \imgtexttrainmesh{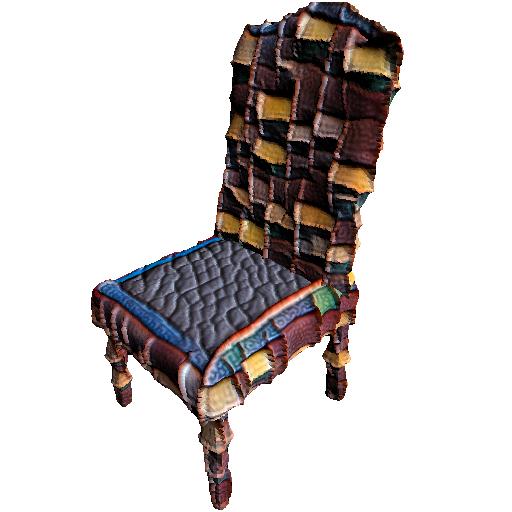} &\rotatebox{90}{\parbox[c]{2cm}{\centering\hspace{0pt} \textbf{Colorful Crochet}}}  &  \imgtextsupp{color_img_train_colorful_crochet.jpg}  &  \imgtextsupp{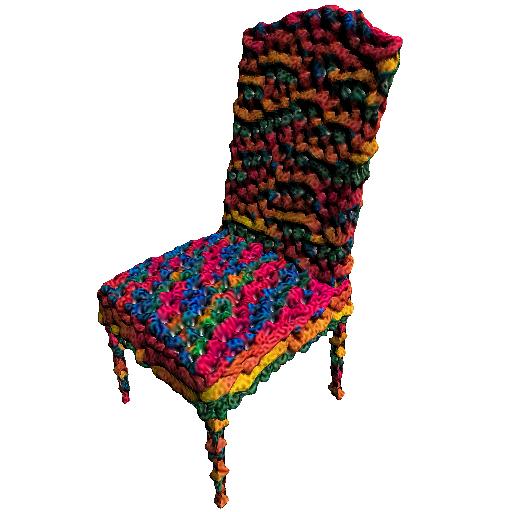}  &  \imgtexttrainmesh{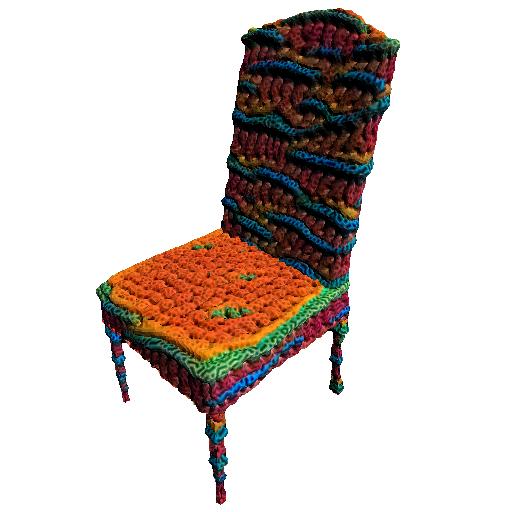} \\
\rotatebox{90}{\parbox[c]{2cm}{\centering\hspace{0pt} \textbf{Moss}}}  &  \imgtextsupp{color_img_train_moss.jpg}  &  \imgtextsupp{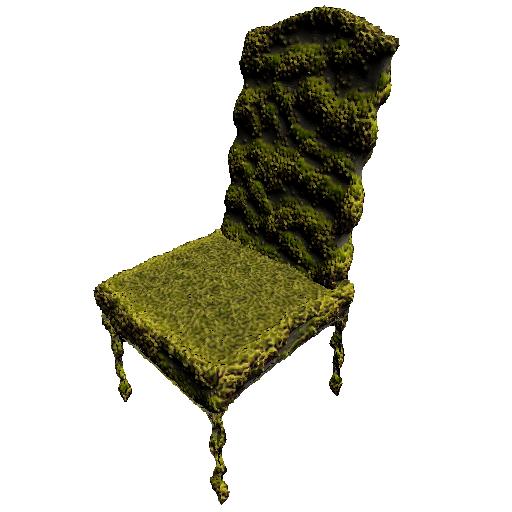}  &  \imgtexttrainmesh{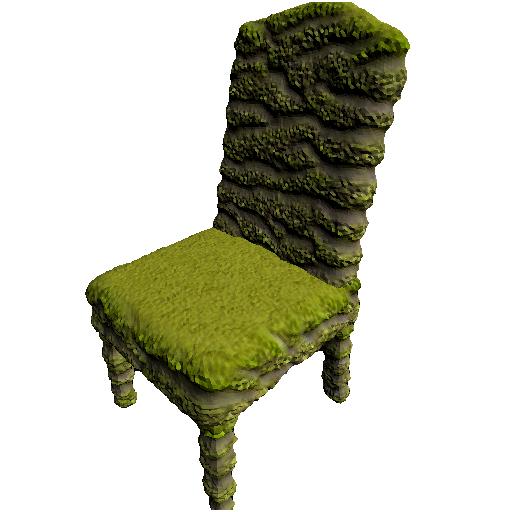} &\rotatebox{90}{\parbox[c]{2cm}{\centering\hspace{0pt} \textbf{Jelly Beans}}}  &  \imgtextsupp{color_img_train_jelly_beans.jpg}  &  \imgtextsupp{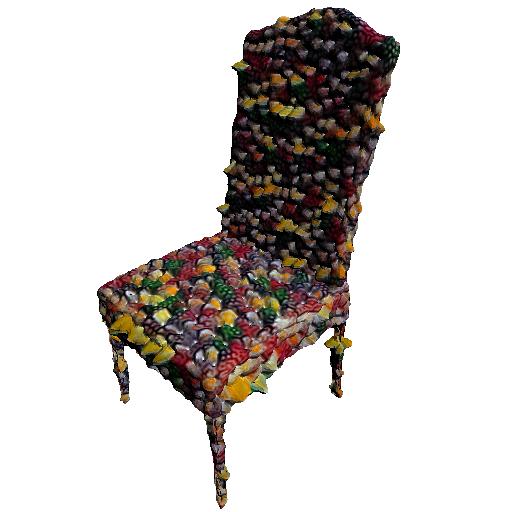}  &  \imgtexttrainmesh{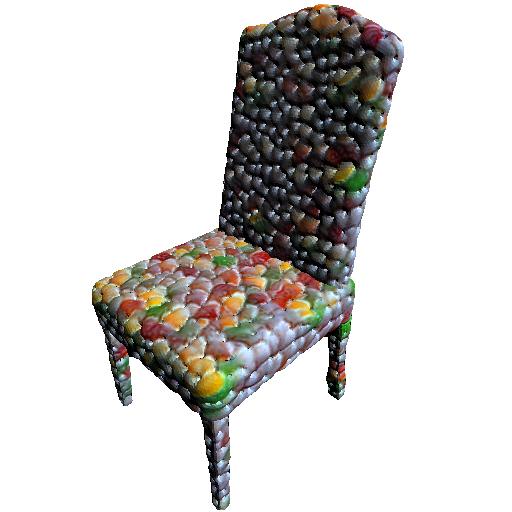} \\
\rotatebox{90}{\parbox[c]{2cm}{\centering\hspace{0pt} \textbf{Stained Glass}}}  &  \imgtextsupp{color_img_train_stained_glass.jpg}  &  \imgtextsupp{color_img_ood_stained_glass_mug.jpg}  &  \imgtexttrainmesh{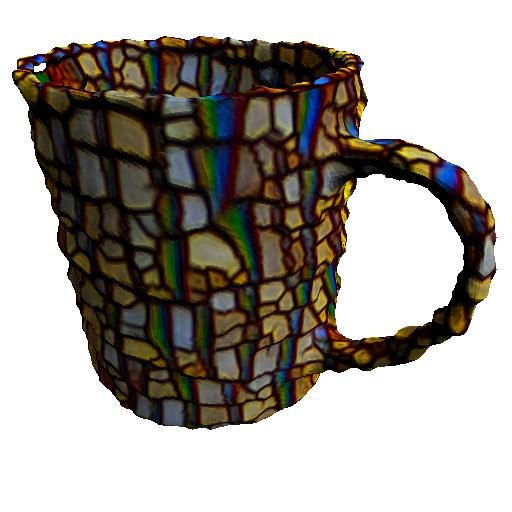} &\rotatebox{90}{\parbox[c]{2cm}{\centering\hspace{0pt} \textbf{Bark}}}  &  \imgtextsupp{color_img_train_bark.jpg}  &  \imgtextsupp{color_img_ood_bark_mug.jpg}  &  \imgtexttrainmesh{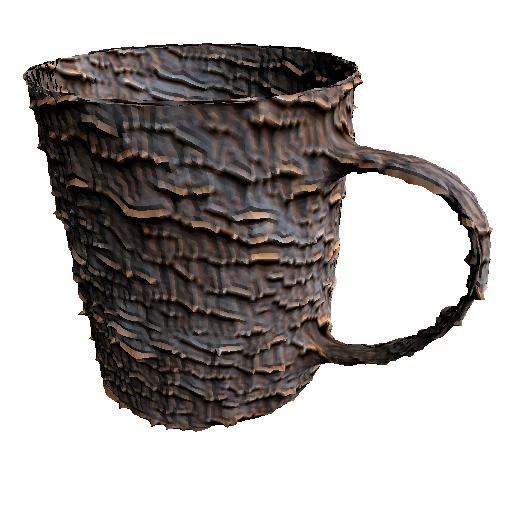} \\
\rotatebox{90}{\parbox[c]{2cm}{\centering\hspace{0pt} \textbf{Marble}}}  &  \imgtextsupp{color_img_train_marble.jpg}  &  \imgtextsupp{color_img_ood_marble_mug.jpg}  &  \imgtexttrainmesh{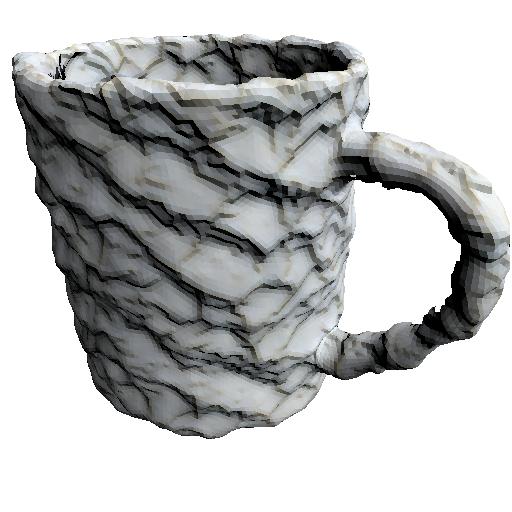} &\rotatebox{90}{\parbox[c]{2cm}{\centering\hspace{0pt} \textbf{Cactus}}}  &  \imgtextsupp{color_img_train_cactus.jpg}  &  \imgtextsupp{color_img_ood_cactus_mug.jpg}  &  \imgtexttrainmesh{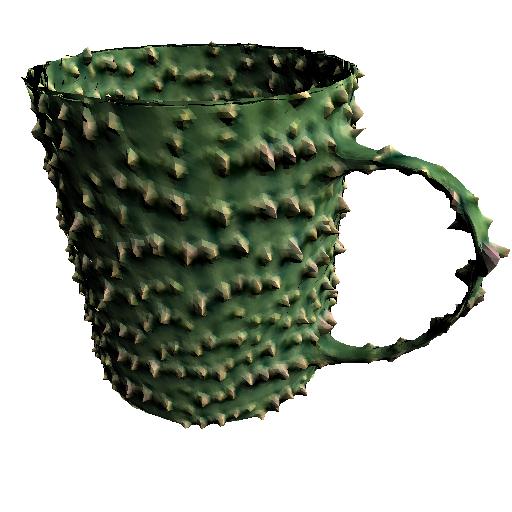} \\
\rotatebox{90}{\parbox[c]{2cm}{\centering\hspace{0pt} \textbf{Feathers}}}  &  \imgtextsupp{color_img_train_feathers.jpg}  &  \imgtextsupp{color_img_ood_feathers_mug.jpg}  &  \imgtexttrainmesh{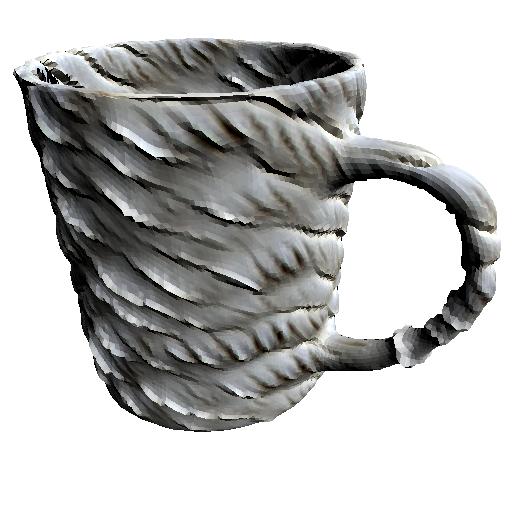} &\rotatebox{90}{\parbox[c]{2cm}{\centering\hspace{0pt} \textbf{Sandstone}}}  &  \imgtextsupp{color_img_train_sandstone.jpg}  &  \imgtextsupp{color_img_ood_sandstone_mug.jpg}  &  \imgtexttrainmesh{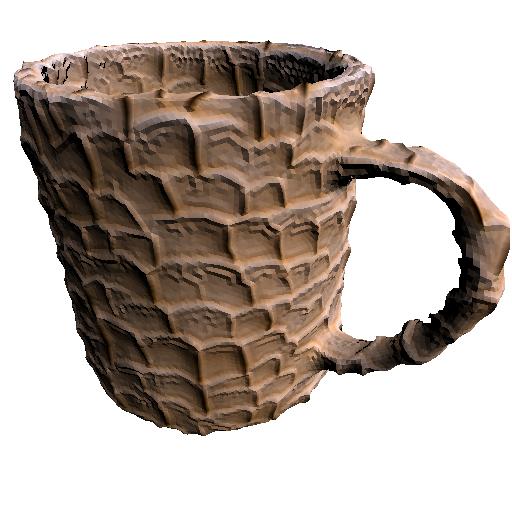} \\
\end{tabular}
\captionof{figure}{Results of direct training on target meshes in \textbf{Text-guided} 3D texture synthesis.}
\label{fig:text-train-direct-mesh-result-supp1}
\end{table*}

\begin{table*}[]
\begin{tabular}{cccc||cccc}
\textbf{Prompt} & Train & Generalization & Direct Train & \textbf{Prompt} & Train & Generalization & Direct Train  \\
\midrule

\rotatebox{90}{\parbox[c]{2cm}{\centering\hspace{0pt} \textbf{Patchwork Leather}}}  &  \imgtextsupp{color_img_train_patchwork_leather.jpg}  &  \imgtextsupp{color_img_ood_patchwork_leather_mug.jpg}  &  \imgtexttrainmesh{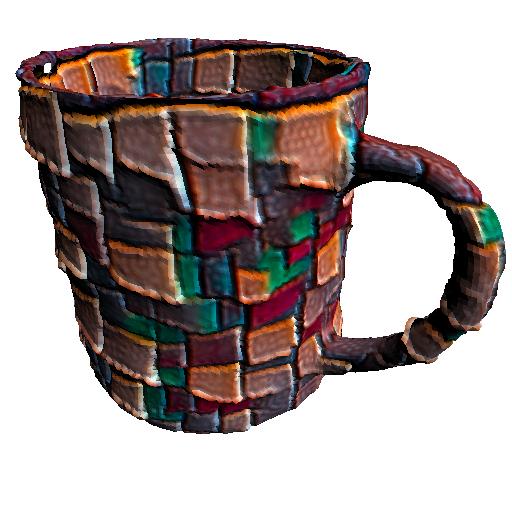} &\rotatebox{90}{\parbox[c]{2cm}{\centering\hspace{0pt} \textbf{Colorful Crochet}}}  &  \imgtextsupp{color_img_train_colorful_crochet.jpg}  &  \imgtextsupp{color_img_ood_colorful_crochet_mug.jpg}  &  \imgtexttrainmesh{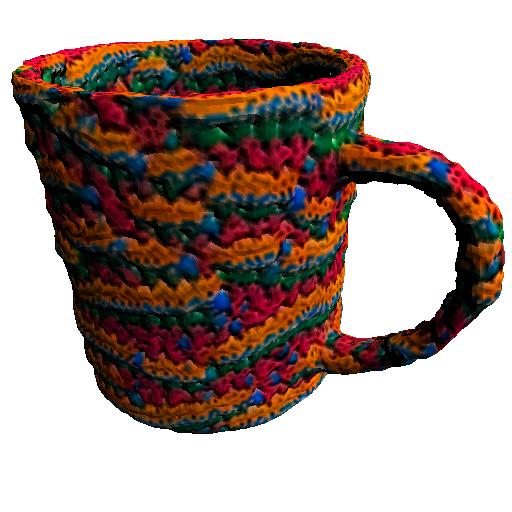} \\
\rotatebox{90}{\parbox[c]{2cm}{\centering\hspace{0pt} \textbf{Moss}}}  &  \imgtextsupp{color_img_train_moss.jpg}  &  \imgtextsupp{color_img_ood_moss_mug.jpg}  &  \imgtexttrainmesh{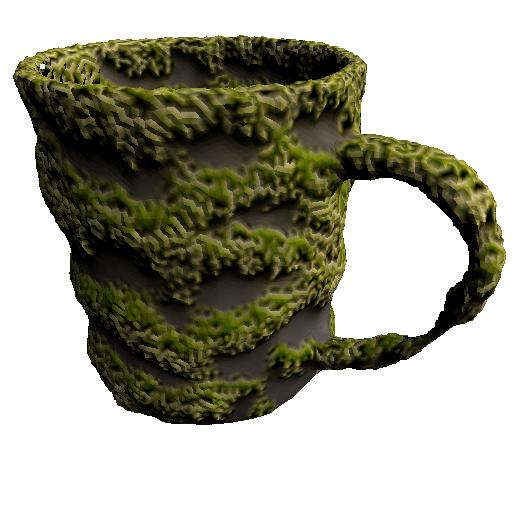} &\rotatebox{90}{\parbox[c]{2cm}{\centering\hspace{0pt} \textbf{Jelly Beans}}}  &  \imgtextsupp{color_img_train_jelly_beans.jpg}  &  \imgtextsupp{color_img_ood_jelly_beans_mug.jpg}  &  \imgtexttrainmesh{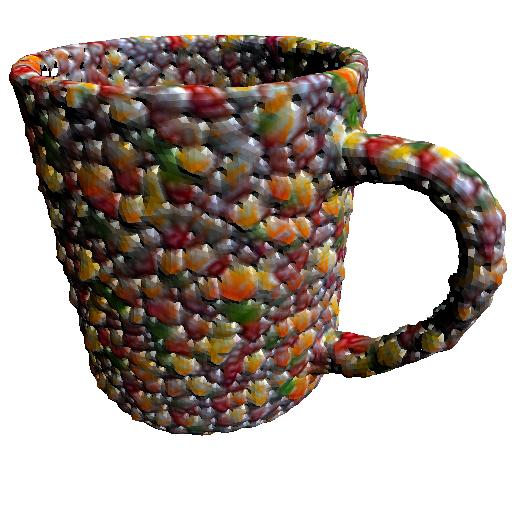} \\
\rotatebox{90}{\parbox[c]{2cm}{\centering\hspace{0pt} \textbf{Stained Glass}}}  &  \imgtextsupp{color_img_train_stained_glass.jpg}  &  \imgtextsupp{color_img_ood_stained_glass_vase.jpg}  &  \imgtexttrainmesh{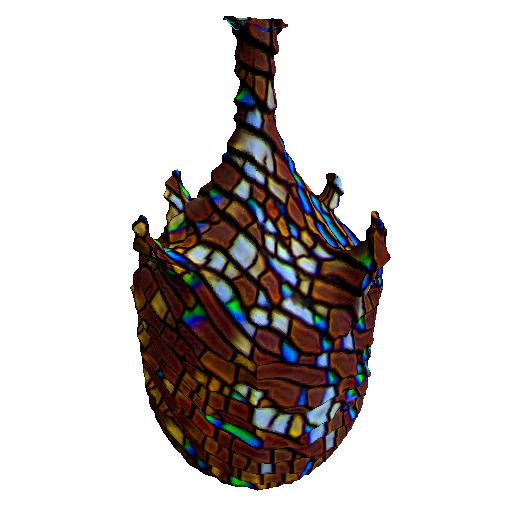} &\rotatebox{90}{\parbox[c]{2cm}{\centering\hspace{0pt} \textbf{Bark}}}  &  \imgtextsupp{color_img_train_bark.jpg}  &  \imgtextsupp{color_img_ood_bark_vase.jpg}  &  \imgtexttrainmesh{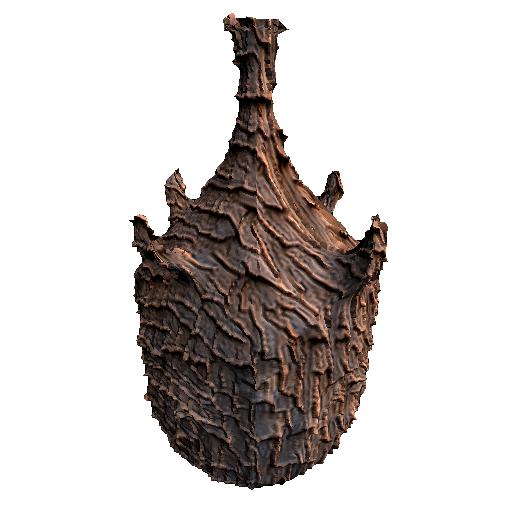} \\
\rotatebox{90}{\parbox[c]{2cm}{\centering\hspace{0pt} \textbf{Marble}}}  &  \imgtextsupp{color_img_train_marble.jpg}  &  \imgtextsupp{color_img_ood_marble_vase.jpg}  &  \imgtexttrainmesh{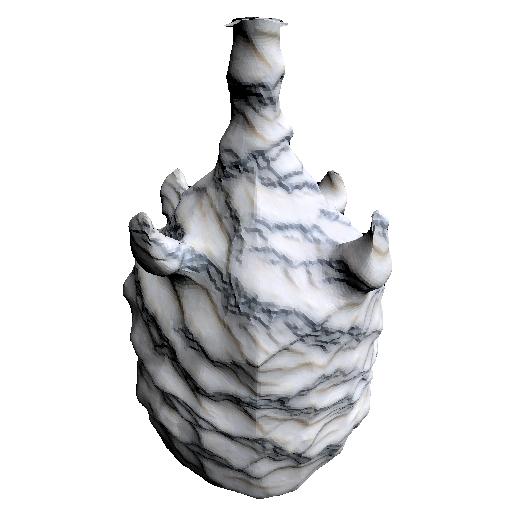} &\rotatebox{90}{\parbox[c]{2cm}{\centering\hspace{0pt} \textbf{Cactus}}}  &  \imgtextsupp{color_img_train_cactus.jpg}  &  \imgtextsupp{color_img_ood_cactus_vase.jpg}  &  \imgtexttrainmesh{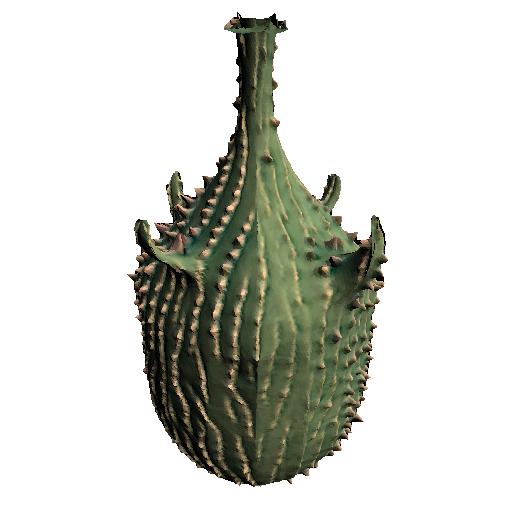} \\
\rotatebox{90}{\parbox[c]{2cm}{\centering\hspace{0pt} \textbf{Feathers}}}  &  \imgtextsupp{color_img_train_feathers.jpg}  &  \imgtextsupp{color_img_ood_feathers_vase.jpg}  &  \imgtexttrainmesh{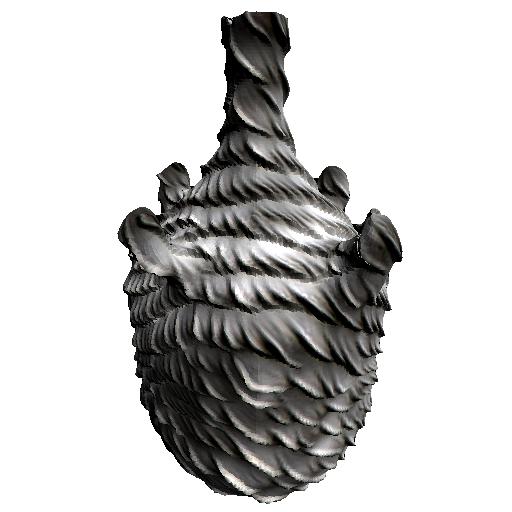} &\rotatebox{90}{\parbox[c]{2cm}{\centering\hspace{0pt} \textbf{Sandstone}}}  &  \imgtextsupp{color_img_train_sandstone.jpg}  &  \imgtextsupp{color_img_ood_sandstone_vase.jpg}  &  \imgtexttrainmesh{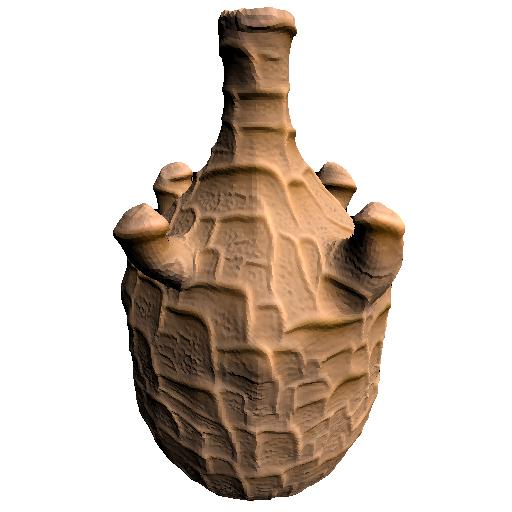} \\
\rotatebox{90}{\parbox[c]{2cm}{\centering\hspace{0pt} \textbf{Patchwork Leather}}}  &  \imgtextsupp{color_img_train_patchwork_leather.jpg}  &  \imgtextsupp{color_img_ood_patchwork_leather_vase.jpg}  &  \imgtexttrainmesh{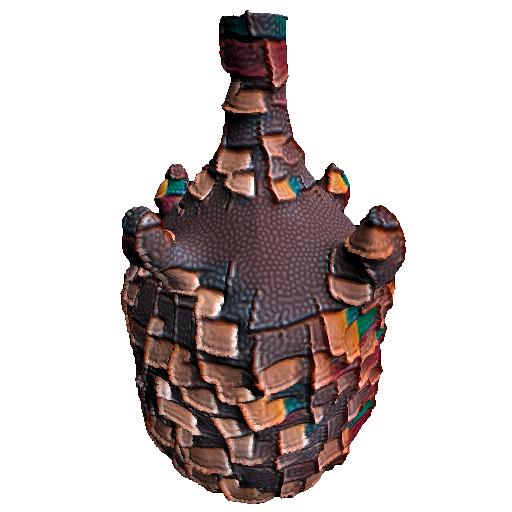} &\rotatebox{90}{\parbox[c]{2cm}{\centering\hspace{0pt} \textbf{Colorful Crochet}}}  &  \imgtextsupp{color_img_train_colorful_crochet.jpg}  &  \imgtextsupp{color_img_ood_colorful_crochet_vase.jpg}  &  \imgtexttrainmesh{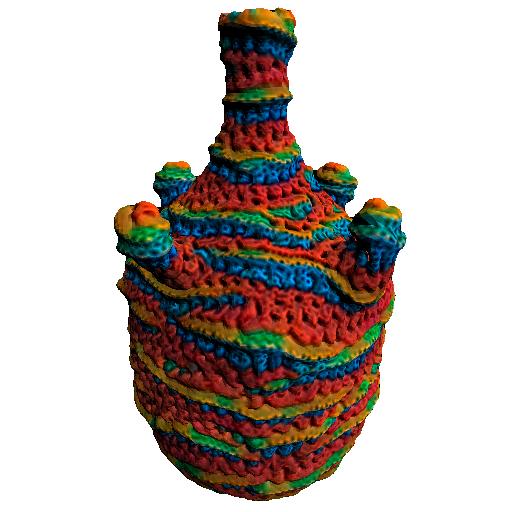} \\
\rotatebox{90}{\parbox[c]{2cm}{\centering\hspace{0pt} \textbf{Moss}}}  &  \imgtextsupp{color_img_train_moss.jpg}  &  \imgtextsupp{color_img_ood_moss_vase.jpg}  &  \imgtexttrainmesh{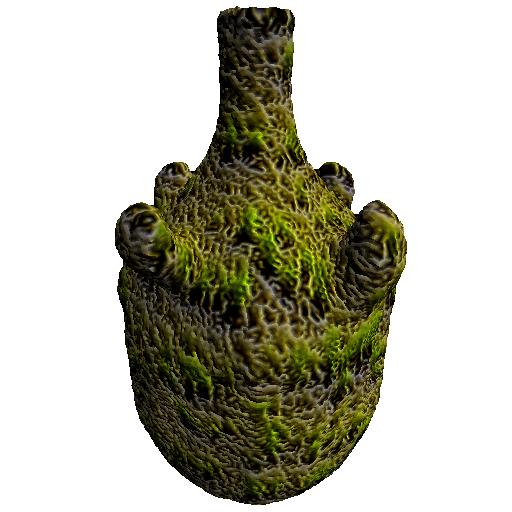} &\rotatebox{90}{\parbox[c]{2cm}{\centering\hspace{0pt} \textbf{Jelly Beans}}}  &  \imgtextsupp{color_img_train_jelly_beans.jpg}  &  \imgtextsupp{color_img_ood_jelly_beans_vase.jpg}  &  \imgtexttrainmesh{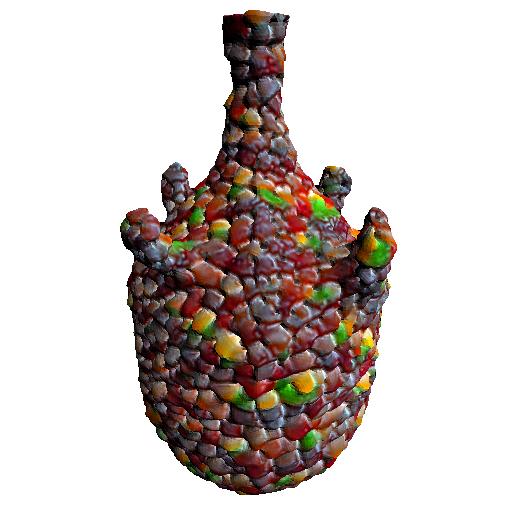} \\
\end{tabular}
\captionof{figure}{Results of direct training on target meshes in \textbf{Text-guided} 3D texture synthesis.}
\label{fig:text-train-direct-mesh-result-supp2}
\end{table*}

\begin{table*}[]
\resizebox{\textwidth}{!}{%
\begin{tabular}{c||cccc}
 \multirow{2}{*}{\textbf{Target}}  &  \multicolumn{4}{c}{\textbf{Methods}}  \\
 
 & Ours & \rev{\small{\citet{text2mesh}}} & \rev{\small{\citet{diff-program-rdsystem}}} & \rev{\small{\citet{on-demand-solid-texture}}} \\
 \midrule
\rowdircomp{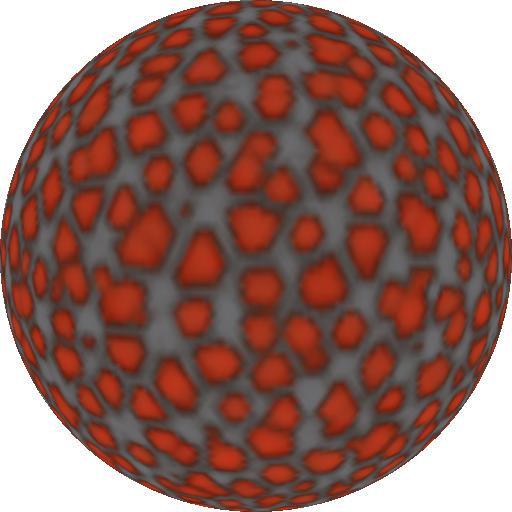} \\ 
\rowdircomp{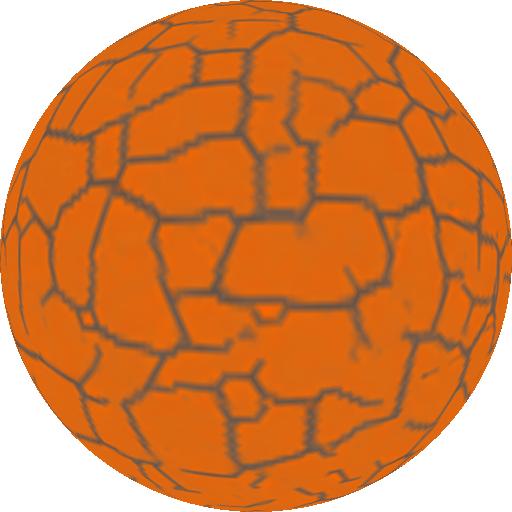} \\ 
\rowdircomp{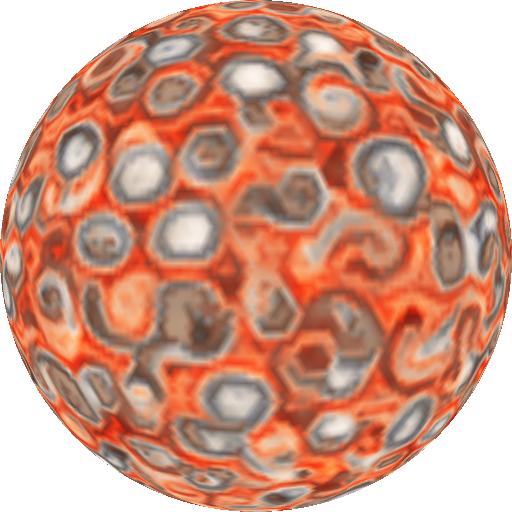} \\ 
\rowdircomp{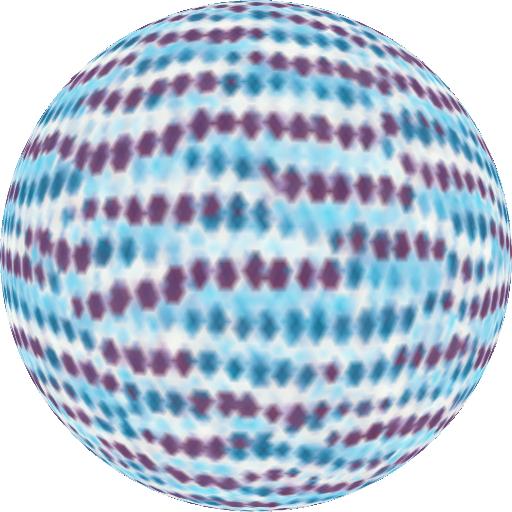} \\ 
\rowdircomp{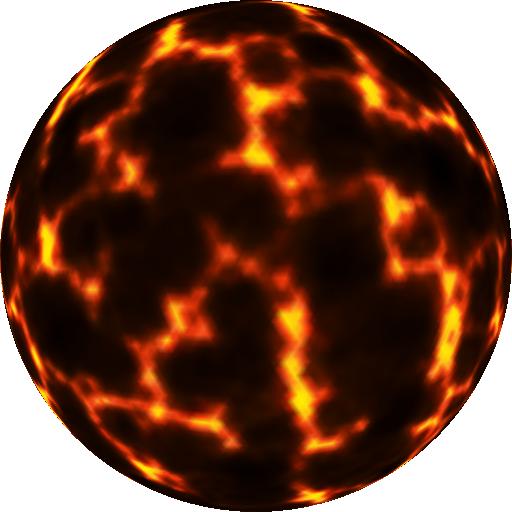} \\ 
\rowdircomp{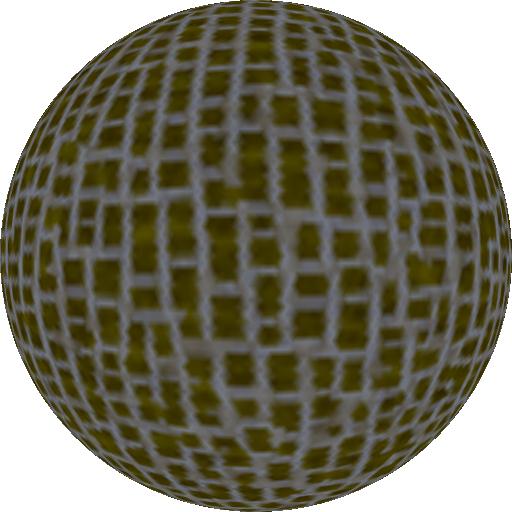} \\ 
\end{tabular}
}
\captionof{figure}{\rev{Additional comparison results for image-guided texture synthesis on the training mesh.}}
\label{fig:image-comparison-train-supp1}
\end{table*}

\begin{table*}[]
\resizebox{\textwidth}{!}{%
\begin{tabular}{c||cccc}
 \multirow{2}{*}{\textbf{Target}}  &  \multicolumn{4}{c}{\textbf{Methods}}  \\
 
 & Ours & \rev{\small{\citet{text2mesh}}} & \rev{\small{\citet{diff-program-rdsystem}}} & \rev{\small{\citet{on-demand-solid-texture}}} \\
 \midrule
\rowdircomp{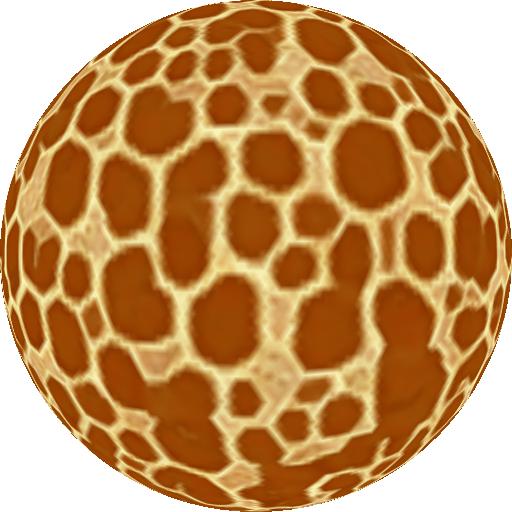} \\ 
\rowdircomp{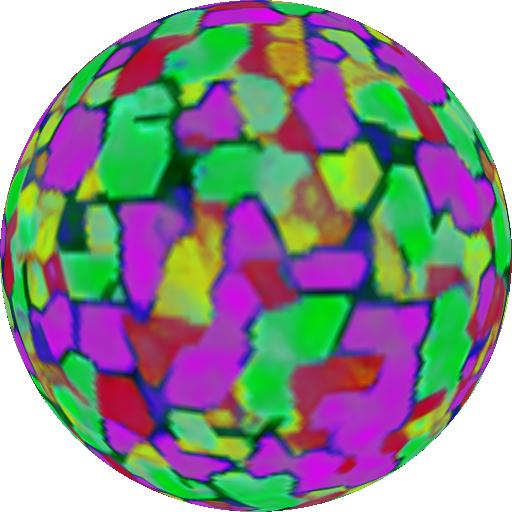} \\ 
\rowdircomp{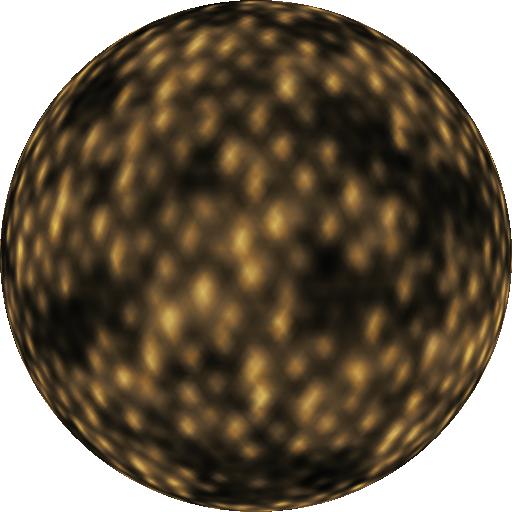} \\ 
\rowdircomp{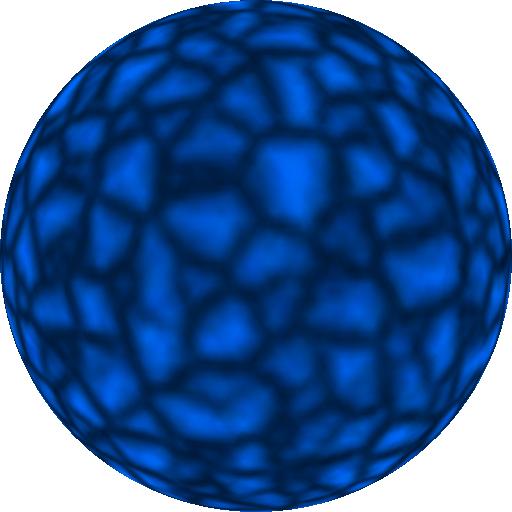} \\ 
\rowdircomp{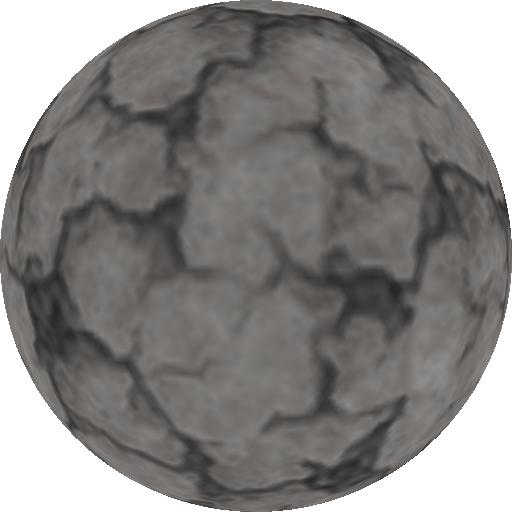} \\ 
\rowdircomp{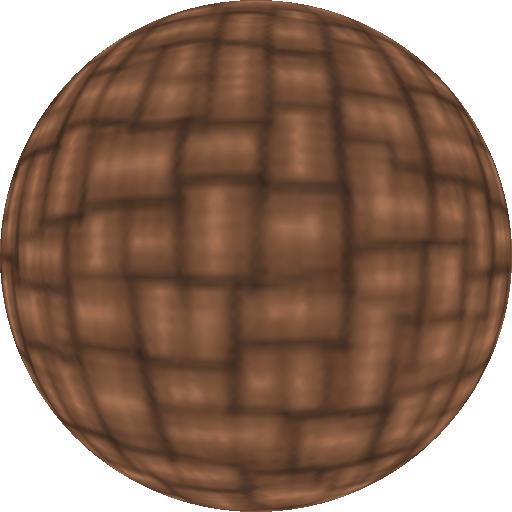} \\ 
\end{tabular}
}
\captionof{figure}{\rev{Additional comparison results for image-guided texture synthesis on the training mesh.}}
\label{fig:image-comparison-train-supp2}
\end{table*}

\begin{table*}[]
\resizebox{\textwidth}{!}{%
\begin{tabular}{c||cccc}
 \multirow{2}{*}{\textbf{Target}}  &  \multicolumn{4}{c}{\textbf{Methods}}  \\
 
 & Ours & \rev{\small{\citet{text2mesh}}} & \rev{\small{\citet{diff-program-rdsystem}}} & \rev{\small{\citet{on-demand-solid-texture}}} \\
 \midrule
\rowdircomp{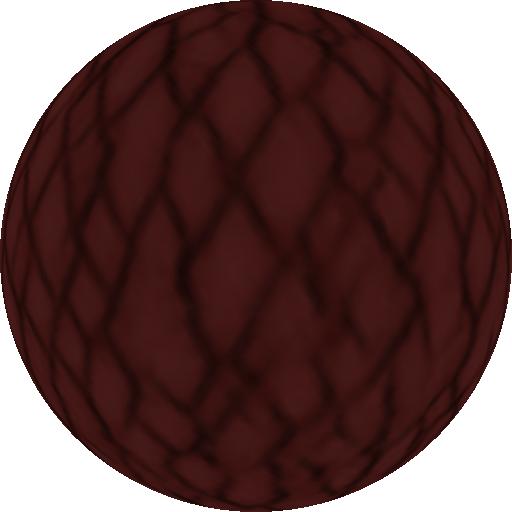} \\ 
\rowdircomp{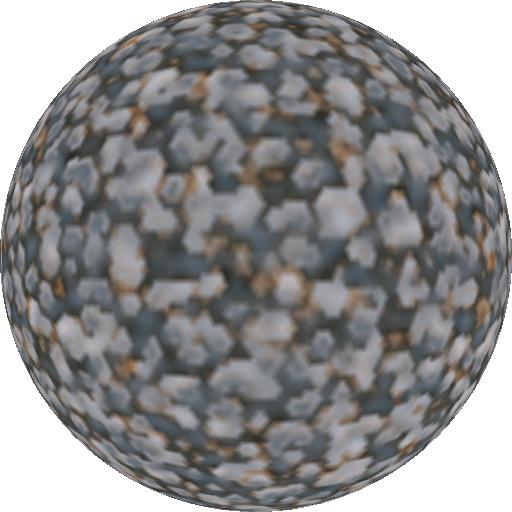} \\ 
\rowdircomp{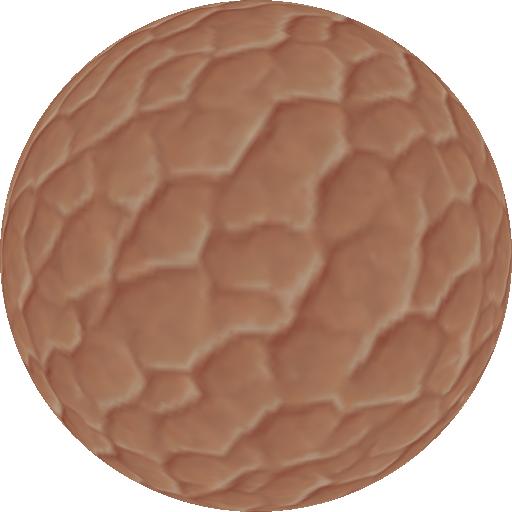} \\ 
\rowdircomp{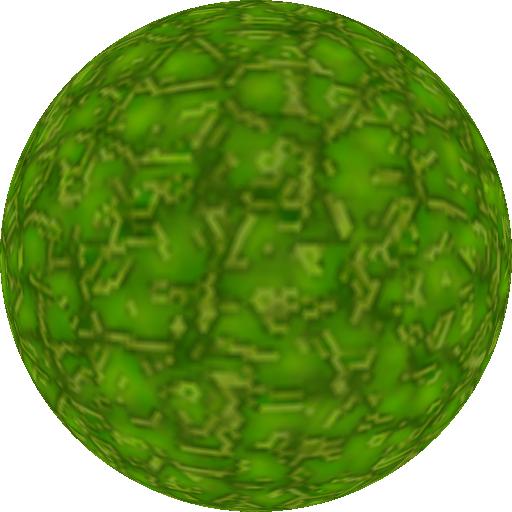} \\ 
\rowdircomp{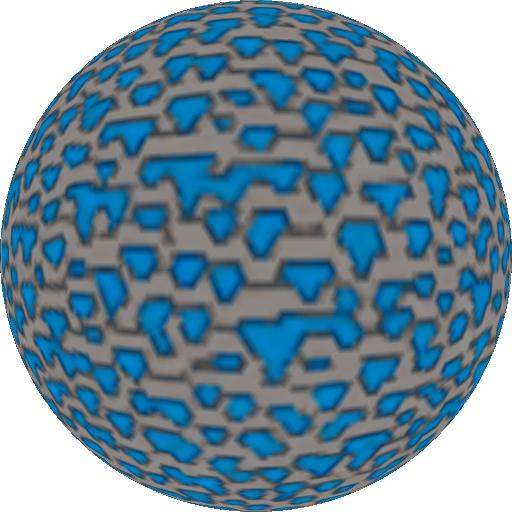} \\ 
\rowdircomp{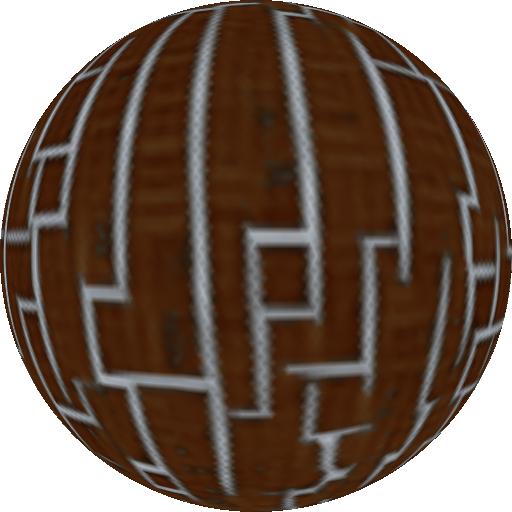} \\ 
\end{tabular}
}
\captionof{figure}{\rev{Additional comparison results for image-guided texture synthesis on the training mesh.}}
\label{fig:image-comparison-train-supp3}
\end{table*}

\begin{table*}[]
\resizebox{\textwidth}{!}{%
\begin{tabular}{c||cccc}
 \multirow{2}{*}{\textbf{Target}}  &  \multicolumn{4}{c}{\textbf{Methods}}  \\
 
 & Ours & \rev{\small{\citet{text2mesh}}} & \rev{\small{\citet{diff-program-rdsystem}}} & \rev{\small{\citet{on-demand-solid-texture}}} \\
 \midrule
\rowgencomp{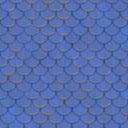}{bunny} \\
\rowgencomp{p1}{dragon} \\
\rowgencomp{p2}{bunny} \\
\rowgencomp{p2}{mobius} \\
\rowgencomp{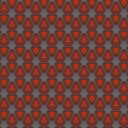}{armor} \\
\rowgencomp{p3}{armadillo} \\
\end{tabular}
}
\captionof{figure}{\rev{Additional comparison results for image-guided texture synthesis on unseen meshes.}}
\label{fig:image-comparison-test-supp1}
\end{table*}

\begin{table*}[]
\resizebox{\textwidth}{!}{%
\begin{tabular}{c||cccc}
 \multirow{2}{*}{\textbf{Target}}  &  \multicolumn{4}{c}{\textbf{Methods}}  \\
 
 & Ours & \rev{\small{\citet{text2mesh}}} & \rev{\small{\citet{diff-program-rdsystem}}} & \rev{\small{\citet{on-demand-solid-texture}}} \\
 \midrule
\rowgencomp{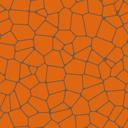}{mobius} \\
\rowgencomp{p4}{dragon} \\
\rowgencomp{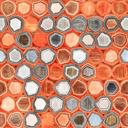}{springer} \\
\rowgencomp{p5}{bunny} \\
\rowgencomp{p6}{dragon} \\
\rowgencomp{p6}{springer} \\
\end{tabular}
}
\captionof{figure}{\rev{Additional comparison results for image-guided texture synthesis on unseen meshes.}}
\label{fig:image-comparison-test-supp2}
\end{table*}

\begin{table*}[]
\resizebox{\textwidth}{!}{%
\begin{tabular}{c||cccc}
 \multirow{2}{*}{\textbf{Target}}  &  \multicolumn{4}{c}{\textbf{Methods}}  \\
 
 & Ours & \rev{\small{\citet{text2mesh}}} & \rev{\small{\citet{diff-program-rdsystem}}} & \rev{\small{\citet{on-demand-solid-texture}}} \\
 \midrule
\rowgencomp{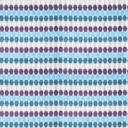}{bunny} \\
\rowgencomp{p7}{mobius} \\
\rowgencomp{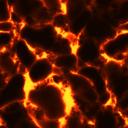}{armadillo} \\
\rowgencomp{p8}{mobius} \\
\rowgencomp{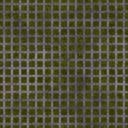}{armor} \\
\rowgencomp{p9}{dragon} \\
\end{tabular}
}
\captionof{figure}{\rev{Additional comparison results for image-guided texture synthesis on unseen meshes.}}
\label{fig:image-comparison-test-supp33}
\end{table*}

\begin{table*}[]
\resizebox{\textwidth}{!}{%
\begin{tabular}{c||cccc}
 \multirow{2}{*}{\textbf{Target}}  &  \multicolumn{4}{c}{\textbf{Methods}}  \\
 
 & Ours & \rev{\small{\citet{text2mesh}}} & \rev{\small{\citet{diff-program-rdsystem}}} & \rev{\small{\citet{on-demand-solid-texture}}} \\
 \midrule
\rowgencomp{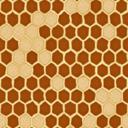}{dragon} \\
\rowgencomp{p10}{bunny} \\
\rowgencomp{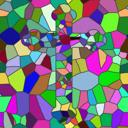}{armor} \\
\rowgencomp{p11}{dragon} \\
\rowgencomp{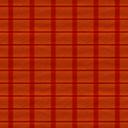}{dragon} \\
\rowgencomp{p12}{springer} \\
\end{tabular}
}
\captionof{figure}{\rev{Additional comparison results for image-guided texture synthesis on unseen meshes.}}
\label{fig:image-comparison-test-supp3}
\end{table*}

\begin{table*}[]
\resizebox{\textwidth}{!}{%
\begin{tabular}{c||cccc}
 \multirow{2}{*}{\textbf{Target}}  &  \multicolumn{4}{c}{\textbf{Methods}}  \\
 
 & Ours & \rev{\small{\citet{text2mesh}}} & \rev{\small{\citet{diff-program-rdsystem}}} & \rev{\small{\citet{on-demand-solid-texture}}} \\
 \midrule
\rowgencomp{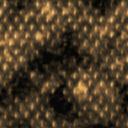}{armadillo} \\
\rowgencomp{p13}{bunny} \\
\rowgencomp{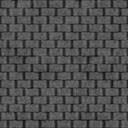}{bunny} \\
\rowgencomp{p14}{mobius} \\
\rowgencomp{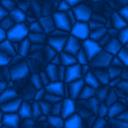}{armor} \\
\rowgencomp{p15}{bunny} \\
\end{tabular}
}
\captionof{figure}{\rev{Additional comparison results for image-guided texture synthesis on unseen meshes.}}
\label{fig:image-comparison-test-supp4}
\end{table*}

\begin{table*}[]
\resizebox{\textwidth}{!}{%
\begin{tabular}{c||cccc}
 \multirow{2}{*}{\textbf{Target}}  &  \multicolumn{4}{c}{\textbf{Methods}}  \\
 
 & Ours & \rev{\small{\citet{text2mesh}}} & \rev{\small{\citet{diff-program-rdsystem}}} & \rev{\small{\citet{on-demand-solid-texture}}} \\
 \midrule
\rowgencomp{p16}{dragon} \\
\rowgencomp{p16}{bunny} \\
\rowgencomp{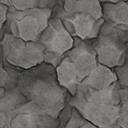}{springer} \\
\rowgencomp{p17}{dragon} \\
\rowgencomp{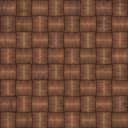}{dragon} \\
\rowgencomp{p18}{armor} \\
\end{tabular}
}
\captionof{figure}{\rev{Additional comparison results for image-guided texture synthesis on unseen meshes.}}
\label{fig:image-comparison-test-supp5}
\end{table*}

\begin{table*}[]
\resizebox{\textwidth}{!}{%
\begin{tabular}{c||cccc}
 \multirow{2}{*}{\textbf{Target}}  &  \multicolumn{4}{c}{\textbf{Methods}}  \\
 
 & Ours & \rev{\small{\citet{text2mesh}}} & \rev{\small{\citet{diff-program-rdsystem}}} & \rev{\small{\citet{on-demand-solid-texture}}} \\
 \midrule
\rowgencomp{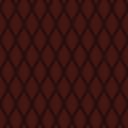}{dragon} \\
\rowgencomp{p19}{bunny} \\
\rowgencomp{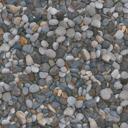}{mobius} \\
\rowgencomp{p20}{bunny} \\
\rowgencomp{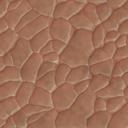}{armadillo} \\
\rowgencomp{p21}{bunny} \\
\end{tabular}
}
\captionof{figure}{\rev{Additional comparison results for image-guided texture synthesis on unseen meshes.}}
\label{fig:image-comparison-test-supp6}
\end{table*}

\begin{table*}[]
\resizebox{\textwidth}{!}{%
\begin{tabular}{c||cccc}
 \multirow{2}{*}{\textbf{Target}}  &  \multicolumn{4}{c}{\textbf{Methods}}  \\
 
 & Ours & \rev{\small{\citet{text2mesh}}} & \rev{\small{\citet{diff-program-rdsystem}}} & \rev{\small{\citet{on-demand-solid-texture}}} \\
 \midrule
\rowgencomp{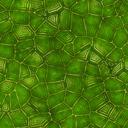}{armor} \\
\rowgencomp{p22}{springer} \\
\rowgencomp{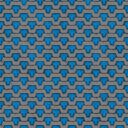}{dragon} \\
\rowgencomp{p23}{armadillo} \\
\rowgencomp{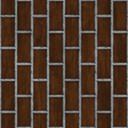}{armor} \\
\rowgencomp{p24}{bunny} \\
\end{tabular}
}
\captionof{figure}{\rev{Additional comparison results for image-guided texture synthesis on unseen meshes.}}
\label{fig:image-comparison-test-supp7}
\end{table*}


\begin{table*}[]
\resizebox{\linewidth}{!}{%
\begin{tabular}{c||ccc|||c||ccc}
\textbf{Prompts} & \textbf{Ours} & \rev{\small{\citet{text2mesh}}} & \rev{\small{\citet{xmesh}}} & \textbf{Prompts} & \textbf{Ours} & \rev{\small{\citet{text2mesh}}} & \rev{\small{\citet{xmesh}}}  \\
 \midrule
\rotatebox{90}{\parbox[c]{2cm}{\centering\hspace{0pt} \textbf{Stained Glass}}}  &  \rowtextgencomp{stained_glass}{vase} &  \rotatebox{90}{\parbox[c]{2cm}{\centering\hspace{0pt} \textbf{Colorful Crochet}}}  &  \rowtextgencomp{colorful_crochet}{bunny} \\
\rotatebox{90}{\parbox[c]{2cm}{\centering\hspace{0pt} \textbf{Stained Glass}}}  &  \rowtextgencomp{stained_glass}{chair} & \rotatebox{90}{\parbox[c]{2cm}{\centering\hspace{0pt} \textbf{Moss}}}  &  \rowtextgencomp{moss}{alien} \\
\rotatebox{90}{\parbox[c]{2cm}{\centering\hspace{0pt} \textbf{Feathers}}}  &  \rowtextgencomp{feathers}{chair} & \rotatebox{90}{\parbox[c]{2cm}{\centering\hspace{0pt} \textbf{Moss}}}  &  \rowtextgencomp{moss}{koala} \\
\rotatebox{90}{\parbox[c]{2cm}{\centering\hspace{0pt} \textbf{Feathers}}}  &  \rowtextgencomp{feathers}{bunny} & \rotatebox{90}{\parbox[c]{2cm}{\centering\hspace{0pt} \textbf{Bark}}}  &  \rowtextgencomp{bark}{chair} \\
\rotatebox{90}{\parbox[c]{2cm}{\centering\hspace{0pt} \textbf{Jelly Beans}}}  &  \rowtextgencomp{jelly_beans}{mug} & \rotatebox{90}{\parbox[c]{2cm}{\centering\hspace{0pt} \textbf{Bark}}}  &  \rowtextgencomp{bark}{alien} \\
\rotatebox{90}{\parbox[c]{2cm}{\centering\hspace{0pt} \textbf{Jelly Beans}}}  &  \rowtextgencomp{jelly_beans}{koala} & \rotatebox{90}{\parbox[c]{2cm}{\centering\hspace{0pt} \textbf{Sandstone}}}  &  \rowtextgencomp{sandstone}{bunny} \\
\rotatebox{90}{\parbox[c]{2cm}{\centering\hspace{0pt} \textbf{Patchwork Leather}}}  &  \rowtextgencomp{patchwork_leather}{chair} & \rotatebox{90}{\parbox[c]{2cm}{\centering\hspace{0pt} \textbf{Sandstone}}}  &  \rowtextgencomp{sandstone}{chair} \\
\rotatebox{90}{\parbox[c]{2cm}{\centering\hspace{0pt} \textbf{Patchwork Leather}}}  &  \rowtextgencomp{patchwork_leather}{alien} & \rotatebox{90}{\parbox[c]{2cm}{\centering\hspace{0pt} \textbf{Marble}}}  &  \rowtextgencomp{marble}{vase} \\
\rotatebox{90}{\parbox[c]{2cm}{\centering\hspace{0pt} \textbf{Colorful Crochet}}}  &  \rowtextgencomp{colorful_crochet}{vase} & \rotatebox{90}{\parbox[c]{2cm}{\centering\hspace{0pt} \textbf{Marble}}}  &  \rowtextgencomp{marble}{mug} \\
\end{tabular}
}
\captionof{figure}{\rev{Additional comparison on generalization ability in \textbf{Text-guided} synthesis.}}
\label{fig:text-comparison-gen-supp1}
\end{table*}

\begin{table*}[]
\resizebox{\linewidth}{!}{%
\begin{tabular}{c||ccc|||c||ccc}
\textbf{Prompts} & \textbf{Ours} & \rev{\small{\citet{text2mesh}}} & \rev{\small{\citet{xmesh}}} & \textbf{Prompts} & \textbf{Ours} & \rev{\small{\citet{text2mesh}}} & \rev{\small{\citet{xmesh}}}  \\
 \midrule
 \rotatebox{90}{\parbox[c]{2cm}{\centering\hspace{0pt} \textbf{Stained Glass}}}  &  \rowtextgencompspec{stained_glass}{vase}  &  \rotatebox{90}{\parbox[c]{2cm}{\centering\hspace{0pt} \textbf{Stained Glass}}}  &  \rowtextgencompspec{stained_glass}{chair} \\
\rotatebox{90}{\parbox[c]{2cm}{\centering\hspace{0pt} \textbf{Feathers}}}  &  \rowtextgencompspec{feathers}{chair}  &  \rotatebox{90}{\parbox[c]{2cm}{\centering\hspace{0pt} \textbf{Feathers}}}  &  \rowtextgencompspec{feathers}{bunny} \\
\rotatebox{90}{\parbox[c]{2cm}{\centering\hspace{0pt} \textbf{Jelly Beans}}}  &  \rowtextgencompspec{jelly_beans}{mug}  &  \rotatebox{90}{\parbox[c]{2cm}{\centering\hspace{0pt} \textbf{Jelly Beans}}}  &  \rowtextgencompspec{jelly_beans}{bunny} \\
\rotatebox{90}{\parbox[c]{2cm}{\centering\hspace{0pt} \textbf{Sandstone}}}  &  \rowtextgencompspec{sandstone}{bunny}  &  \rotatebox{90}{\parbox[c]{2cm}{\centering\hspace{0pt} \textbf{Sandstone}}}  &  \rowtextgencompspec{sandstone}{chair} \\
\rotatebox{90}{\parbox[c]{2cm}{\centering\hspace{0pt} \textbf{Marble}}}  &  \rowtextgencompspec{marble}{vase}  &  \rotatebox{90}{\parbox[c]{2cm}{\centering\hspace{0pt} \textbf{Marble}}}  &  \rowtextgencompspec{marble}{mug} \\
\end{tabular}
}
\captionof{figure}{\rev{Additional comparison on texture quality in \textbf{Text-guided} synthesis. We train the methods of \cite{text2mesh, xmesh} directly for each mesh and prompt pair, while our results are based on direct generalization from the sphere mesh.}}
\label{fig:text-comparison-gen-spec-supp1}
\end{table*}

\begin{table*}[]
\resizebox{\linewidth}{!}{%
\begin{tabular}{c||ccc|||c||ccc}
\textbf{Prompts} & \textbf{Ours} & \rev{\small{\citet{text2mesh}}} & \rev{\small{\citet{xmesh}}} & \textbf{Prompts} & \textbf{Ours} & \rev{\small{\citet{text2mesh}}} & \rev{\small{\citet{xmesh}}}  \\
 \midrule
\rotatebox{90}{\parbox[c]{2cm}{\centering\hspace{0pt} \textbf{Patchwork Leather}}}  &  \rowtextgencompspec{patchwork_leather}{chair}  &  \rotatebox{90}{\parbox[c]{2cm}{\centering\hspace{0pt} \textbf{Patchwork Leather}}}  &  \rowtextgencompspec{patchwork_leather}{vase} \\
\rotatebox{90}{\parbox[c]{2cm}{\centering\hspace{0pt} \textbf{Colorful Crochet}}}  &  \rowtextgencompspec{colorful_crochet}{vase}  &  \rotatebox{90}{\parbox[c]{2cm}{\centering\hspace{0pt} \textbf{Colorful Crochet}}}  &  \rowtextgencompspec{colorful_crochet}{bunny} \\
\rotatebox{90}{\parbox[c]{2cm}{\centering\hspace{0pt} \textbf{Moss}}}  &  \rowtextgencompspec{moss}{mug}  &  \rotatebox{90}{\parbox[c]{2cm}{\centering\hspace{0pt} \textbf{Moss}}}  &  \rowtextgencompspec{moss}{alien} \\
\rotatebox{90}{\parbox[c]{2cm}{\centering\hspace{0pt} \textbf{Bark}}}  &  \rowtextgencompspec{bark}{chair}  &  \rotatebox{90}{\parbox[c]{2cm}{\centering\hspace{0pt} \textbf{Bark}}}  &  \rowtextgencompspec{bark}{alien} \\
\rotatebox{90}{\parbox[c]{2cm}{\centering\hspace{0pt} \textbf{Cactus}}}  &  \rowtextgencompspec{cactus}{alien}  &  \rotatebox{90}{\parbox[c]{2cm}{\centering\hspace{0pt} \textbf{Cactus}}}  &  \rowtextgencompspec{cactus}{chair} \\
\end{tabular}
}
\captionof{figure}{\rev{Additional comparison on texture quality in \textbf{Text-guided} synthesis. We train the methods of \cite{text2mesh, xmesh} directly for each mesh and prompt pair, while our results are based on direct generalization from the sphere mesh.}}
\label{fig:text-comparison-gen-spec-supp2}
\end{table*}







\clearpage
\clearpage
\bibliographystyle{ACM-Reference-Format}
\bibliography{main}


\end{document}